\PassOptionsToPackage{bookmarksnumbered,unicode}{hyperref}
\PassOptionsToPackage{hyphens}{url}
\documentclass[format=acmsmall, review=false, screen=true]{acmart}
\pdfoutput=1
\usepackage{kg-macros}
\usepackage{cube}
\usepackage[utf8]{inputenc}
\usepackage{booktabs} 
\usepackage{xspace} 
\usepackage{paralist}
\usepackage{subcaption}
\usepackage{multirow}
\usepackage{ifsym}
\usepackage{xcolor}
\usepackage{pgfplots}

\pgfplotsset{compat=1.14}

\usepackage[ruled]{algorithm2e} 

\SetAlFnt{\small}
\SetAlCapFnt{\small}
\SetAlCapNameFnt{\small}
\SetAlCapHSkip{0pt}
\IncMargin{-\parindent}

\newcommand{\scaledeftabs}{0.94}

\acmJournal{CSUR}
\acmVolume{X}
\acmNumber{X}
\acmArticle{XX}
\acmYear{2019}
\acmMonth{3}
\copyrightyear{2019}

\hypersetup{
	pdfkeywords={knowledge graph, property graph, ontology, context, knowledge graph embedding, graph neural network, quality, knowledge graph refinement}, 
	pdfsubject={axVix 2020},   
}

\acmDOI{0000001.0000001}

\setcopyright{none}
\renewcommand\footnotetextcopyrightpermission[1]{}
\makeatletter
\renewcommand\@formatdoi[1]{\ignorespaces}
\makeatother
\makeatletter
\let\@authorsaddresses\@empty
\makeatother

\begin{document}
\title[Knowledge Graphs]{Knowledge Graphs}


\author{Aidan Hogan}
\affiliation{%
  \institution{IMFD, DCC, Universidad de Chile}
  \country{Chile}
}
\email{ahogan@dcc.uchile.cl}

\author{Eva Blomqvist}
\affiliation{%
	\institution{Linköping University}
	\country{Sweden}
}

\author{Michael Cochez}
\affiliation{%
	\institution{Vrije Universiteit and Discovery Lab, Elsevier}
	\country{The Netherlands}
}

\author{Claudia d'Amato}
\affiliation{%
	\institution{University of Bari} 
	\country{Italy}
}

\author{Gerard de Melo}
\affiliation{%
	\institution{HPI, Germany and Rutgers University}
	\country{USA}
}

\author{Claudio Gutierrez}
\affiliation{%
	\institution{IMFD, DCC, Universidad de Chile}
	\country{Chile}
}

\author{José Emilio Labra Gayo}
\affiliation{%
	\institution{Universidad de Oviedo}
	\country{Spain}
}

\author{Sabrina Kirrane}
\author{Sebastian Neumaier}
\author{Axel Polleres}
\affiliation{%
	\institution{WU Vienna}
	\country{Austria}
}

\author{Roberto Navigli}
\affiliation{%
	\institution{Sapienza University of Rome}
	\country{Italy}
}

\author{Axel-Cyrille Ngonga Ngomo}
\affiliation{%
	\institution{DICE, Universität Paderborn}
	\country{Germany}
}

\author{Sabbir M. Rashid}
\affiliation{%
	\institution{Tetherless World Constellation, Rensselaer Polytechnic Institute}
	\country{USA}
}

\author{Anisa Rula}
\affiliation{%
	\institution{University of Milano--Bicocca}
	\country{Italy}
}
\affiliation{%
	\institution{University of Bonn}
	\country{Germany}
}

\author{Lukas Schmelzeisen}
\affiliation{%
	\institution{Universität Stuttgart}
	\country{Germany}
}

\author{Juan Sequeda}
\affiliation{%
	\institution{data.world}
	\country{USA}
}

\author{Steffen Staab}
\affiliation{%
	\institution{Universität Stuttgart}
	\country{Germany}
}
\affiliation{%
	\institution{University of Southampton}
	\country{UK}
}

\author{Antoine Zimmermann}
\affiliation{%
  \institution{\'Ecole des mines de Saint-\'Etienne}
	\country{France}}

\renewcommand{\shortauthors}{Hogan et al.}

\begin{abstract}
In this paper we provide a comprehensive introduction to knowledge graphs, which have recently garnered significant attention from both industry and academia in scenarios that require exploiting diverse, dynamic, large-scale collections of data. After some opening remarks, we motivate and contrast various graph-based data models and query languages that are used for knowledge graphs. We discuss the roles of schema, identity, and context in knowledge graphs. We explain how knowledge can be represented and extracted using a combination of deductive and inductive techniques. We summarise methods for the creation, enrichment, quality assessment, refinement, and publication of knowledge graphs. We provide an overview of prominent open knowledge graphs and enterprise knowledge graphs, their applications, and how they use the aforementioned techniques. We conclude with high-level future research directions for knowledge graphs.
\end{abstract}

%
%

%
%


\newcommand{\comm}[1]{#1}

\newcommand{\ah}[1]{{\comm{\color{blue}\textsc{ah:} #1}}} 
\newcommand{\js}[1]{{\comm{\color{green}\textsc{js:} #1}}} 
\newcommand{\ls}[1]{{\comm{\color{orange}\textsc{ls:} #1}}} 
\newcommand{\az}[1]{{\comm{\color{magenta}\textsc{az:} #1}}} 
\newcommand{\cda}[1]{{\comm{\color{red}\textsc{cda:} #1}}} 
\newcommand{\sst}[1]{{\comm{\color{purple}\textsc{SSt:} #1}}} 
\newcommand{\ap}[1]{{\comm{\color{teal}\textsc{ap:} #1}}} 
\newcommand{\mc}[1]{{\comm{\color{cyan}\textsc{MC:} #1}}}
\newcommand{\cg}[1]{{\comm{\color{blue!50!pink}\textsc{cg:} #1}}} 
\newcommand{\sr}[1]{{\comm{\color{orange}\textsc{sr:} #1}}} 
\newcommand{\eb}[1]{{\comm{\color{violet}\textsc{EB:} #1}}} 
\newcommand{\jl}[1]{{\comm{\color{olive}\textsc{JL:} #1}}} 
\newcommand{\an}[1]{{\comm{\color{orange!50!red}\textsc{AN:} #1}}} 
\newcommand{\sn}[1]{{\comm{\color{pink}\textsc{sn:} #1}}} 

\newcommand{\GG}{\ensuremath{\mathcal{G}}\xspace}

\maketitle

\thispagestyle{empty}

\section{Introduction}\label{sec:intro}

Though the phrase ``knowledge graph'' has been used in the literature since at least 1972~\cite{Schneider72}, the modern incarnation of the phrase stems from the 2012 announcement of the Google Knowledge Graph~\cite{GoogleKG}, followed by further announcements of the development of knowledge graphs by Airbnb~\cite{AirBnBKG}, Amazon~\cite{AmazonKG}, eBay~\cite{eBayKG}, Facebook~\cite{NoyGJNPT19}, IBM~\cite{IBMKG}, LinkedIn~\cite{LinkedInKG}, Microsoft~\cite{BingKG}, Uber~\cite{UberKG}, and more besides. The growing industrial uptake of the concept proved difficult for academia to ignore: more and more scientific literature is being published on knowledge graphs, which includes books (e.g.~\cite{QiCLWJW19}), as well as papers outlining definitions (e.g.,~\cite{EhrlingerW16}), novel techniques (e.g.,~\cite{PujaraMGC13,wang2014knowledge,lin2015learning}), and surveys of specific aspects of knowledge graphs (e.g.,~\cite{Paulheim17,Wang2017KGEmbedding}).

Underlying all such developments is the core idea of using graphs to represent data, often enhanced with some way to explicitly represent knowledge~\cite{NoyGJNPT19}. The result is most often used in application scenarios that involve integrating, managing and extracting value from diverse sources of data at large scale~\cite{NoyGJNPT19}. Employing a graph-based abstraction of knowledge has numerous benefits in such settings when compared with, for example, a relational model or NoSQL alternatives. Graphs provide a concise and intuitive abstraction for a variety of domains, where edges capture the (potentially cyclical) relations between the entities inherent in social data, biological interactions, bibliographical citations and co-authorships, transport networks, and so forth~\cite{AnglesG08}. Graphs allow maintainers to postpone the definition of a schema, allowing the data -- and its scope -- to evolve in a more flexible manner than typically possible in a relational setting, particularly for capturing incomplete knowledge~\cite{Abiteboul97}. Unlike (other) NoSQL models, specialised graph query languages support not only standard relational operators (joins, unions, projections, etc.), but also navigational operators for recursively finding entities connected through arbitrary-length paths~\cite{AnglesABHRV17}. Standard knowledge representation formalisms -- such as ontologies~\cite{OWL2,RDFS,obof} and rules~\cite{swrl,rif} -- can be employed to define and reason about the semantics of the terms used to label and describe the nodes and edges in the graph. Scalable frameworks for graph analytics~\cite{DBLP:conf/sigmod/MalewiczABDHLC10,DBLP:conf/sigmod/XinGFS13,signalcollect} can be leveraged for computing centrality, clustering, summarisation, etc., in order to gain insights about the domain being described. Various representations have also been developed that support applying machine learning techniques directly over graphs~\cite{Wang2017KGEmbedding,abs-1901-00596}.

In summary, the decision to build and use a knowledge graph opens up a range of techniques that can be brought to bear for integrating and extracting value from diverse sources of data. However, we have yet to see a general unifying summary that describes how knowledge graphs are being used, what techniques they employ, and how they relate to existing data management topics.

The goal of this tutorial paper is to motivate and give a comprehensive introduction to knowledge graphs: to describe their foundational data models and how they can be queried; to discuss representations relating to schema, identity, and context; to discuss deductive and inductive ways to make knowledge explicit; to present a variety of techniques that can be used for the creation and enrichment of graph-structured data; to describe how the quality of knowledge graphs can be discerned and how they can be refined; to discuss standards and best practices by which knowledge graphs can be published; and to provide an overview of existing knowledge graphs found in practice. Our intended audience includes researchers and practitioners who are new to knowledge graphs. As such, we do not assume that readers have specific expertise on knowledge graphs.

\paragraph{Knowledge graph} The definition of a ``\textit{knowledge graph}'' remains contentious~\cite{EhrlingerW16,BonattiDPP18,Bergman19}, where a number of (sometimes conflicting) definitions have emerged, varying from specific technical proposals to more inclusive general proposals; we address these prior definitions in Appendix~\ref{sec:defs}. Herein we adopt an inclusive definition, where we view a knowledge graph as \textit{a graph of data intended to accumulate and convey knowledge of the real world, whose nodes represent entities of interest and whose edges represent relations between these entities}. The graph of data (aka \textit{data graph}) conforms to a graph-based data model, which may be a \textit{directed edge-labelled graph}, a \textit{property graph}, etc. (we discuss concrete alternatives in Section~\ref{sec:graph}). By \textit{knowledge}, we refer to something that is \textit{known}\footnote{A number of specific definitions for knowledge have been proposed in the literature on epistemology.}. Such knowledge may be accumulated from external sources, or extracted from the knowledge graph itself. Knowledge may be composed of simple statements, such as ``\textit{Santiago is the capital of Chile}'', or quantified statements, such as ``\textit{all capitals are cities}''.  Simple statements can be accumulated as edges in the data graph. If the knowledge graph intends to accumulate quantified statements, a more expressive way to represent knowledge -- such as \textit{ontologies} or \textit{rules} -- is required. \textit{Deductive methods} can then be used to entail and accumulate further knowledge (e.g., ``\textit{Santiago is a city}''). Additional knowledge -- based on simple or quantified statements -- can also be extracted from and accumulated by the knowledge graph using \textit{inductive methods}.

Knowledge graphs are often assembled from numerous sources, and as a result, can be highly diverse in terms of structure and granularity. To address this diversity, representations of \textit{schema}, \textit{identity}, and \textit{context} often play a key role, where a \textit{schema} defines a high-level structure for the knowledge graph, \textit{identity} denotes which nodes in the graph (or in external sources) refer to the same real-world entity, while \textit{context} may indicate a specific setting in which some unit of knowledge is held true. As aforementioned, effective methods for \textit{extraction}, \textit{enrichment}, \textit{quality assessment}, and \textit{refinement} are required for a knowledge graph to grow and improve over time. 

\paragraph{In practice} Knowledge graphs aim to serve as an ever-evolving shared substrate of knowledge within an organisation or community~\cite{NoyGJNPT19}. We distinguish two types of knowledge graphs in practice: \textit{open knowledge graphs} and \textit{enterprise knowledge graphs}. Open knowledge graphs are published online, making their content accessible for the public good. The most prominent examples -- DBpedia~\cite{LehmannIJJKMHMK15}, Freebase~\cite{bollacker2007platform}, Wikidata~\cite{VrandecicK14}, YAGO~\cite{YAGO}, etc. -- cover many domains and are either extracted from Wikipedia~\cite{LehmannIJJKMHMK15,YAGO}, or built by communities of volunteers~\cite{bollacker2007platform,VrandecicK14}. Open knowledge graphs have also been published within specific domains, such as media~\cite{RaimondFSA14}, government~\cite{HendlerHMT12,ShadboltO13},  
geography~\cite{StadlerLHA12}, tourism~\cite{LuLS16,abs-1805-05744,MaturanaALMH18,ZhangCHYAL19}, life sciences~\cite{CallahanCAD13}, and more besides. Enterprise knowledge graphs are typically internal to a company and applied for commercial use-cases~\cite{NoyGJNPT19}. Prominent industries using enterprise knowledge graphs include Web search (e.g., Bing~\cite{BingKG}, Google~\cite{GoogleKG}), commerce (e.g., Airbnb~\cite{AirBnBKG}, Amazon~\cite{AmazonKG,dong2019building}, eBay~\cite{eBayKG}, Uber~\cite{UberKG}), social networks (e.g., Facebook~\cite{NoyGJNPT19}, LinkedIn~\cite{LinkedInKG}), finance (e.g., Accenture~\cite{AccentureKG}, Banca d'Italia~\cite{BellomariniFGS19}, Bloomberg~\cite{BloombergKG}, Capital One~\cite{CapitalOneKG}, Wells Fargo~\cite{WellsFargoKG}), among others. Applications include search~\cite{BingKG,GoogleKG}, recommendations~\cite{AirBnBKG,UberKG,LinkedInKG,NoyGJNPT19}, personal agents~\cite{eBayKG}, advertising~\cite{LinkedInKG}, business analytics~\cite{LinkedInKG}, risk assessment~\cite{ThompsonReutersKG,MaanaKG}, automation~\cite{HensonSTK19}, and more besides. We will provide more details on the use of knowledge graphs in practice in Section~\ref{sec:kgs}.

\paragraph{Running example} To keep the discussion accessible, throughout the paper, we present concrete examples in the context of a hypothetical knowledge graph relating to tourism in Chile (loosely inspired by, e.g.,~\cite{abs-1805-05744,LuLS16}). 
The knowledge graph is managed by a tourism board that aims to increase tourism in the country and promote new attractions in strategic areas. The knowledge graph itself will eventually describe tourist attractions, cultural events, services, and businesses, as well as cities and inter-city travel routes. Some applications the organisation envisages are to:

\begin{itemize}
\item create a tourism portal that allows visitors to search for attractions, upcoming events, and other related services (in multiple languages);
\item gain insights into tourism demographics in terms of season, nationalities, etc.;
\item analyse sentiment about various attractions and events, including positive reviews, summaries of complaints about events and services, reports of crime, etc.;
\item understand tourism trajectories: the sequence of attractions, events, etc., that tourists visit;
\item cross-reference trajectories with available flights/buses to suggest new strategic routes;
\item offer personalised recommendations of places to visit;
\item and so forth. 
\end{itemize}

\begin{table}
\setlength{\tabcolsep}{2.15pt}
\caption{Related tertiary literature on knowledge graphs; \protect\id\ denotes in-depth discussion, \protect\ib\ denotes brief discussion, * denotes informal publication (arXiv) \label{tab:related}}

\centering
\footnotesize
\begin{tabular}{lrlcccccccccccccccccccccc}
\textbf{Publication} & \textbf{Year} & \textbf{Type} & \rot{\textbf{Models}} & \rot{\textbf{Querying}} & \rot{\textbf{Shapes}} & \rot{\textbf{Identity}} & \rot{\textbf{Context}} & \rot{\textbf{Ontologies}} & \rot{\textbf{Entailment}} & \rot{\textbf{Rules}} & \rot{\textbf{DLs}} & \rot{\textbf{Analytics}} & \rot{\textbf{Embeddings}} & \rot{\textbf{GNNs}} & \rot{\textbf{Symbolic Learning}} & \rot{\textbf{Construction}} & \rot{\textbf{Quality}} & \rot{\textbf{Refinement}} & \rot{\textbf{Publication}}  & \rot{\textbf{Enterprise KGs}} & \rot{\textbf{Open KGs}} & \rot{\textbf{Applications}} & \rot{\textbf{History}} & \rot{\textbf{Prior Definitions}} \\
\toprule

\citet{PVGW2017} & \citeyear{PVGW2017} & Book & \noop{Models}\ib & \noop{Querying}\id & \noop{Shapes} & \noop{Identity}\id & \noop{Context} & \noop{Ontologies}\ib & \noop{Entailment}\ib & \noop{Rules} & \noop{DLs} & \noop{Analytics}\ib & \noop{Embeddings} & \noop{GNNs} & \noop{Symbolic Learning} & \noop{Construction}\id & \noop{Quality} & \noop{Refinement} & \noop{Publication} & \noop{Enterprise KGs}\id & \noop{Open KGs} & \noop{Applications}\id & \noop{History}\id & \noop{Prior Definitions} \\

\citet{Paulheim17} & \citeyear{Paulheim17} & Survey & \noop{Models} & \noop{Querying} & \noop{Shapes} & \noop{Identity} & \noop{Context} & \noop{Ontologies} & \noop{Entailment} & \noop{Rules} & \noop{DLs} & \noop{Analytics} & \noop{Embeddings}\ib & \noop{GNNs} & \noop{Symbolic Learning}\ib & \noop{Construction}\ib & \noop{Quality}\ib & \noop{Refinement}\id & \noop{Publication} & \noop{Enterprise KGs}\ib & \noop{Open KGs}\ib & \noop{Applications} & \noop{History} & \noop{Prior Definitions} \\

\citet{Wang2017KGEmbedding} & \citeyear{Wang2017KGEmbedding} & Survey & \noop{Models} & \noop{Querying} & \noop{Shapes} & \noop{Identity} & \noop{Context} & \noop{Ontologies} & \noop{Entailment} & \noop{Rules} & \noop{DLs} & \noop{Analytics} & \noop{Embeddings}\id & \noop{GNNs} & \noop{Symbolic Learning}\ib & \noop{Construction}\ib & \noop{Quality}\ib & \noop{Refinement}\ib & \noop{Publication} & \noop{Enterprise KGs} & \noop{Open KGs}\ib & \noop{Applications}\ib & \noop{History} & \noop{Prior Definitions} \\

\citet{YanWCGZ18} & \citeyear{YanWCGZ18} & Survey & \noop{Models}\ib & \noop{Querying}\ib & \noop{Shapes} & \noop{Identity} & \noop{Context} & \noop{Ontologies} & \noop{Entailment}\ib & \noop{Rules}\id & \noop{DLs} & \noop{Analytics}\ib & \noop{Embeddings}\id & \noop{GNNs} & \noop{Symbolic Learning}\ib & \noop{Construction}\id & \noop{Quality} & \noop{Refinement} & \noop{Publication} & \noop{Enterprise KGs}\id & \noop{Open KGs}\id & \noop{Applications} & \noop{History} & \noop{Prior Definitions} \\

\citet{GeseseBS19} & \citeyear{GeseseBS19} & Survey & \noop{Models} & \noop{Querying} & \noop{Shapes} & \noop{Identity}\ib & \noop{Context} & \noop{Ontologies} & \noop{Entailment} & \noop{Rules} & \noop{DLs} & \noop{Analytics} & \noop{Embeddings}\id & \noop{GNNs} & \noop{Symbolic Learning} & \noop{Construction} & \noop{Quality} & \noop{Refinement} & \noop{Publication} & \noop{Enterprise KGs} & \noop{Open KGs} & \noop{Applications}\ib & \noop{History} & \noop{Prior Definitions} \\

\citet{KazemiGJKSFP19} & \citeyear{KazemiGJKSFP19} & Survey* & \noop{Models} & \noop{Querying} & \noop{Shapes} & \noop{Identity} & \noop{Context}\ib & \noop{Ontologies} & \noop{Entailment} & \noop{Rules}\ib & \noop{DLs} & \noop{Analytics} & \noop{Embeddings}\id & \noop{GNNs}\id & \noop{Symbolic Learning}\id & \noop{Construction}\ib & \noop{Quality} & \noop{Refinement}\ib & \noop{Publication} & \noop{Enterprise KGs} & \noop{Open KGs} & \noop{Applications} & \noop{History} & \noop{Prior Definitions} \\

\citet{Kejriwal19} & \citeyear{Kejriwal19} & Book & \noop{Models} & \noop{Querying} & \noop{Shapes} & \noop{Identity} & \noop{Context} & \noop{Ontologies} & \noop{Entailment} & \noop{Rules} & \noop{DLs} & \noop{Analytics} & \noop{Embeddings} & \noop{GNNs} & \noop{Symbolic Learning} & \noop{Construction}\id & \noop{Quality} & \noop{Refinement} & \noop{Publication} & \noop{Enterprise KGs} & \noop{Open KGs} & \noop{Applications} & \noop{History} & \noop{Prior Definitions} \\

\citet{XiaoDCC19} & \citeyear{XiaoDCC19} & Survey & \noop{Models}\ib & \noop{Querying}\ib & \noop{Shapes} & \noop{Identity} & \noop{Context}\ib & \noop{Ontologies}\ib & \noop{Entailment}\ib & \noop{Rules} & \noop{DLs} & \noop{Analytics} & \noop{Embeddings} & \noop{GNNs} & \noop{Symbolic Learning} & \noop{Construction}\ib & \noop{Quality} & \noop{Refinement} & \noop{Publication} & \noop{Enterprise KGs} & \noop{Open KGs} & \noop{Applications}\id & \noop{History} & \noop{Prior Definitions} \\

\citet{WangY19} & \citeyear{WangY19} & Survey & \noop{Models}\ib & \noop{Querying}\ib & \noop{Shapes} & \noop{Identity} & \noop{Context} & \noop{Ontologies}\ib & \noop{Entailment}\ib & \noop{Rules}\ib & \noop{DLs} & \noop{Analytics} & \noop{Embeddings}\id & \noop{GNNs}\ib & \noop{Symbolic Learning}\ib & \noop{Construction}\id & \noop{Quality} & \noop{Refinement} & \noop{Publication} & \noop{Enterprise KGs} & \noop{Open KGs} & \noop{Applications}\ib & \noop{History} & \noop{Prior Definitions} \\

\citet{Al-MoslmiOOV20} & \citeyear{Al-MoslmiOOV20} & Survey & \noop{Models} & \noop{Querying} & \noop{Shapes} & \noop{Identity} & \noop{Context} & \noop{Ontologies} & \noop{Entailment} & \noop{Rules} & \noop{DLs} & \noop{Analytics} & \noop{Embeddings} & \noop{GNNs} & \noop{Symbolic Learning} & \noop{Construction}\id & \noop{Quality} & \noop{Refinement} & \noop{Publication} & \noop{Enterprise KGs} & \noop{Open KGs} & \noop{Applications} & \noop{History} & \noop{Prior Definitions} \\

\citet{FenselSAHKPTUW20} & \citeyear{FenselSAHKPTUW20} & Book & \noop{Models} & \noop{Querying} & \noop{Shapes} & \noop{Identity} & \noop{Context} & \noop{Ontologies} & \noop{Entailment} & \noop{Rules} & \noop{DLs} & \noop{Analytics} & \noop{Embeddings} & \noop{GNNs} & \noop{Symbolic Learning} & \noop{Construction}\id & \noop{Quality} & \noop{Refinement} & \noop{Publication} & \noop{Enterprise KGs}\ib  & \noop{Open KGs}\ib & \noop{Applications}\id & \noop{History} & \noop{Prior Definitions}\ib \\

\citet{HeistHRP20} & \citeyear{HeistHRP20} & Survey* &  \noop{Models} & \noop{Querying} & \noop{Shapes} & \noop{Identity} & \noop{Context} & \noop{Ontologies}\ib & \noop{Entailment} & \noop{Rules} & \noop{DLs} & \noop{Analytics} & \noop{Embeddings} & \noop{GNNs} & \noop{Symbolic Learning} & \noop{Construction}\ib & \noop{Quality}\ib & \noop{Refinement} & \noop{Publication} & \noop{Enterprise KGs}  & \noop{Open KGs}\id & \noop{Applications} & \noop{History} & \noop{Prior Definitions} \\

\citet{JiPCMY20} & \citeyear{JiPCMY20} & Survey* & \noop{Models} & \noop{Querying} & \noop{Shapes} & \noop{Identity} & \noop{Context} & \noop{Ontologies} & \noop{Entailment} & \noop{Rules}\id & \noop{DLs} & \noop{Analytics} & \noop{Embeddings}\id & \noop{GNNs}\ib & \noop{Symbolic Learning} & \noop{Construction}\id & \noop{Quality} & \noop{Refinement} & \noop{Publication} & \noop{Enterprise KGs} & \noop{Open KGs}\ib & \noop{Applications}\id & \noop{History}\ib & \noop{Prior Definitions}\ib \\
\midrule

Hogan et al. & 2020 & Tutorial* & \id & \id & \id & \id & \id & \id & \id & \id & \id & \id & \id & \id & \id & \id & \id & \id & \id & \id & \id & \id & \id & \id \\

\bottomrule

\end{tabular}
\end{table}

\paragraph{Related Literature}

A number of related surveys, books, etc., have been published relating to knowledge graphs. In Table~\ref{tab:related}, we provide an overview of the tertiary literature -- surveys, books, tutorials, etc. -- relating to knowledge graphs, comparing the topics covered to those covered herein. We see that the existing literature tends to focus on specific aspects of knowledge graphs. Unlike the various surveys that have been published, our goal as a tutorial paper is to provide a broad and accessible introduction to knowledge graphs. Some of the surveys (in particular) provide more in-depth technical details on their chosen topic than this paper; throughout the discussion, where appropriate, we will refer to these surveys for further reading.

\paragraph{Structure} The remainder of the paper is structured as follows:

\begin{description}
\item[Section~\ref{sec:graph}] outlines graph data models and the languages that can be used to query them.
\item[Section~\ref{sec:knowledge}] describes representations of schema, identity, and context in knowledge graphs.
\item[Section~\ref{sec:deductive}] presents deductive formalisms by which knowledge can be represented and entailed.
\item[Section~\ref{sec:inductive}] describes inductive techniques by which additional knowledge can be extracted.
\item[Section~\ref{sec:create}] discusses the creation and enrichment of knowledge graphs from external sources.
\item[Section~\ref{sec:quality}] enumerates quality dimensions by which a knowledge graph can be assessed.
\item[Section~\ref{sec:refine}] discusses various techniques for knowledge graph refinement.
\item[Section~\ref{sec:publish}] discusses principles and protocols for publishing knowledge graphs.
\item[Section~\ref{sec:kgs}] surveys some prominent knowledge graphs and their applications.
\item[Section~\ref{sec:conclude}] concludes with a summary and future research directions for knowledge graphs.
\item[Appendix~\ref{sec:defs}] provides historical background and previous definitions for knowledge graphs.
\item[Appendix~\ref{sec:formal}] enumerates formal definitions that will be referred to from the body of the paper.
\end{description}

\section{Data Graphs}\label{sec:graph}

At the foundation of any knowledge graph is the principle of first applying a graph abstraction to data, resulting in an initial data graph. We now discuss a selection of graph-structured data models that are commonly used in practice to represent data graphs. We then discuss the primitives that form the basis of graph query languages used to interrogate such data graphs. 

\subsection{Models}\label{ssec:graphModels}

Leaving aside graphs, let us assume that the tourism board from our running example has not yet decided how to model relevant data about attractions, events, services, etc. The board first considers using a tabular structure -- in particular, relational databases -- to represent the required data, and though they do not know precisely what data they will need to capture, they start to design an initial relational schema. They begin with an \textsf{Event} table with five columns:
\[ \textsf{Event}(\underline{\textsf{name}}, \textsf{venue}, \textsf{type}, \underline{\textsf{start}}, \textsf{end}) \]
where \underline{\textsf{name}} and \underline{\textsf{start}} together form the primary key of the table in order to uniquely identify recurring events. But as they start to populate the data, they encounter various issues: events may have multiple names (e.g., in different languages), events may have multiple venues, they may not yet know the start and end date-times for future events, events may have multiple types, and so forth. Incrementally addressing these modelling issues as the data become more diverse, they generate internal identifiers for events and adapt their relational schema until they have:

\begin{align}
\textsf{EventName}(\underline{\textsf{id}},\underline{\textsf{name}}),
\textsf{EventStart}(\underline{\textsf{id}},\textsf{start}),
   \textsf{EventEnd}(\underline{\textsf{id}},\textsf{end}),\\
   \textsf{EventVenue}(\underline{\textsf{id}},\underline{\textsf{venue}}),
   \textsf{EventType}(\underline{\textsf{id}},\underline{\textsf{type}})\nonumber
\end{align}
\noindent With the above schema, the organisation can now model events with 0--$n$ names, venues, and types, and 0--1 start dates and end dates (without needing relational nulls/blank cells in tables).

Along the way, the board has to incrementally change the schema several times in order to support new sources of data. Each such change requires a costly remodelling, reloading, and reindexing of data; here we only considered one table. The tourism board struggles with the relational model because they do not know, \textit{a priori}, what data will need to be modelled or what sources they will use. But once they reach the latter relational schema, the board finds that they can integrate further sources without more changes: with minimal assumptions on \textit{multiplicities} (1--1, 1--$n$, etc.) this schema offers a lot of flexibility for integrating incomplete and diverse data. 

In fact, the refined, flexible schema that the board ends up with -- shown in (1) -- is modelling a set of binary relations between entities, which indeed can be viewed as modelling a graph.
By instead adopting a graph data model from the outset, the board could forgo 
the need for an upfront schema, and could define any (binary) relation between any pair of entities at any time.

We now introduce graph data models commonly used in practice~\cite{AnglesABHRV17}.

\subsubsection{Directed edge-labelled graphs} 
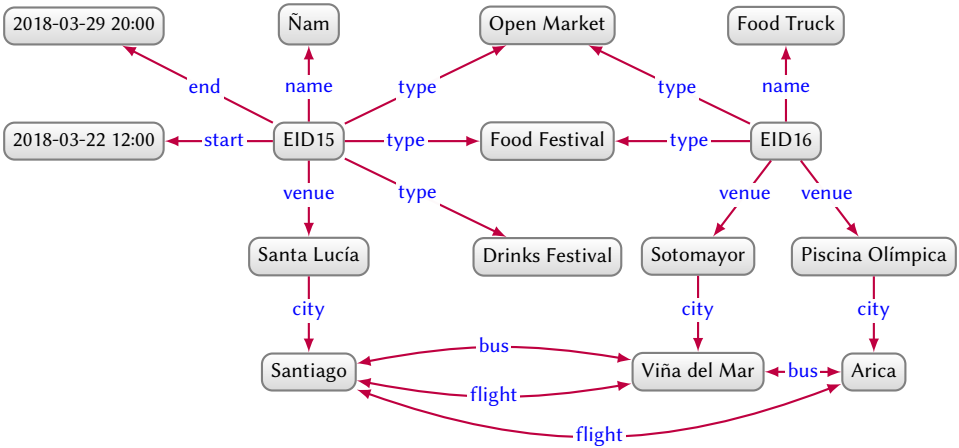
\begin{figure}
\setlength{\vgap}{1cm}
\setlength{\hgap}{1.8cm}

\begin{tikzpicture}
\node[iri,anchor=center] (nId) {EID15};

\node[iri,anchor=center,above=\vgap of nId] (n) {Ñam}
  edge[arrin] node[lab] {name} (nId);
  
\node[iri,anchor=center,right=\hgap of nId] (ff) {Food Festival}
  edge[arrin] node[lab] {type} (nId);
  
\node[iri,anchor=center,below=\vgap of ff] (df) {Drinks Festival}
  edge[arrin] node[lab] {type} (nId);
  
\node[iri,anchor=center,above=\vgap of ff] (om) {Open Market}
  edge[arrin] node[lab] {type} (nId);  
  
\node[iri,anchor=center,below=\vgap of nId] (sl) {Santa Lucía}
  edge[arrin] node[lab] {venue} (nId);  
  
\node[iri,anchor=center,below=\vgap of sl] (san) {Santiago}
  edge[arrin] node[lab] {city} (sl);  
  
\node[iri,anchor=center,right=\hgap of ff] (fId) {EID16}
  edge[arrout] node[lab] {type} (ff)
  edge[arrout] node[lab] {type} (om);  

\node[iri,anchor=center,above=\vgap of fId] (f) {Food Truck}
  edge[arrin] node[lab] {name} (fId);
  
\node[iri,anchor=center,left=0.8\hgap of nId] (fS) {2018-03-22 12:00}
  edge[arrin] node[lab] {start} (nId);  
  
\node[iri,anchor=center,above=\vgap of fS] (fE) {2018-03-29 20:00}
  edge[arrin] node[lab] {end} (nId);    
  
\node[iri,anchor=center,below=\vgap of fId,xshift=0.65\hgap] (po) {Piscina Olímpica}
  edge[arrin] node[lab] {venue} (fId);  
  
\node[iri,anchor=center,below=\vgap of po] (ar) {Arica}
  edge[arrin] node[lab] {city} (po)
  edge[arroutin,bend left=20] node[lab] {flight} (san);   

\node[iri,anchor=center,below=\vgap of fId,xshift=-0.65\hgap] (sm) {Sotomayor}
  edge[arrin] node[lab] {venue} (fId);  
  
\node[iri,anchor=center,below=\vgap of sm] (vdm) {Viña del Mar}
  edge[arrin] node[lab] {city} (sm)
  edge[arroutin,bend left=10] node[lab] {flight} (san)
  edge[arroutin,bend right=10] node[lab] {bus} (san)
  edge[arroutin] node[lab] {bus} (ar);   

%
%
%
\end{tikzpicture}

\caption{Directed edge-labelled graph describing events and their venues. \label{fig:delg}}
\end{figure}

A directed edge-labelled graph (also known as a \textit{multi-relational graph}~\cite{nickel2013tensor,bordes2013translating,BalazevicAH19}) is defined as a set of nodes -- like \gnode{Santiago}, \gnode{Arica}, \gnode{EID16}, \gnode{2018-03-22 12:00} -- and a set of directed labelled edges between those nodes, like \gedge{Santa Lucía}{city}{Santiago}. In the case of knowledge graphs, nodes are used to represent entities and edges are used to represent (binary) relations between those entities. Figure~\ref{fig:delg} provides an example of how the tourism board could model some relevant event data as a directed edge-labelled graph (for a formal definition of a directed edge-labelled graph see Definition~\ref{def:delg} in Appendix~\ref{sec:formal}). The graph includes data about the names, types, start and end date-times, and venues for events.\footnote{We represent bidirectional edges as \gedge[arroutin]{Viña del Mar}{bus}{Arica}, which more concisely depicts two directed edges: \gedge{Viña del Mar}{bus}{Arica} and \gedge[arrout]{Viña del Mar}{bus}{Arica}. Also while some naming conventions recommend more complete edge labels that include a verb, such as \gelab{has venue} or \gelab{is valid from}, in this paper, for presentation purposes, we will omit the ``\texttt{has}'' and ``\texttt{is}'' verbs from such labels, using simply \gelab{venue} or \gelab{valid from}.} Adding information to such a graph typically involves adding new nodes and edges (with some exceptions discussed later). Representing incomplete information requires simply omitting a particular edge; for example, the graph does not yet define a start/end date-time for the Food Truck festival. 

Modelling data as a graph in this way offers more flexibility for integrating new sources of data, compared to the standard relational model, where a schema must be defined upfront and followed at each step. While other structured data models such as trees (XML, JSON, etc.) would offer similar flexibility, graphs do not require organising the data hierarchically (should \texttt{venue} be a parent, child, or sibling of \texttt{type} for example?). They also allow cycles to be represented and queried (e.g., note the directed cycle in the routes between Santiago, Arica, and Viña del Mar). 

A standardised data model based on directed edge-labelled graphs is the Resource Description Framework (RDF)~\cite{rdf11}, which has been recommended by the W3C. The RDF model defines different types of nodes, including \textit{Internationalized Resource Identifiers} (IRIs)~\cite{rfc3987} which allow for global identification of entities on the Web; \textit{literals}, which allow for representing strings (with or without language tags) and other datatype values (integers, dates, etc.); and \textit{blank nodes}, which are anonymous nodes that are not assigned an identifier (for example, rather than create internal identifiers like \texttt{EID15}, \texttt{EID16}, in RDF, we have the option to use blank nodes). We will discuss these different types of nodes further in Section~\ref{sec:identity} when we speak about issues relating to identity.

\subsubsection{Heterogeneous graphs}
\label{subsub:heterograph}

\begin{figure}
 	\begin{subfigure}[b]{.4\textwidth}
 		\setlength{\vgap}{0.7cm}
 		\setlength{\hgap}{1.5cm}
 		\centering
 		\begin{tikzpicture}
 		\node[iri,anchor=center] (san) {Santiago};

 		\node[iri,above=\vgap of san] (city) {City}
 		  edge[arrin] node[lab,pos=0.62] {type} (san);

 		\node[iri,anchor=center,right=\hgap of san] (chile) {Chile}
 		  edge[arrin] node[lab,xshift=-0.03cm] {capital} (san);

  		\node[iri,anchor=center,right=\hgap of chile] (peru) {Perú}
  		  edge[arrin,bend left=15] node[lab] {borders} (chile)
  		  edge[arrout,bend right=15] node[lab] {borders} (chile);

  		\node[between=chile and peru] (m) {};

  		\node[iri] (country) at (m|-city) (country) {Country}
  		  edge[arrin,bend right=15] node[lab,pos=0.62] {type} (chile)
  		  edge[arrin,bend left=15] node[lab,pos=0.62] {type} (peru);
 		\end{tikzpicture}
 		\caption{Del graph}
 		\label{fig:cap}
 	\end{subfigure}

 	\begin{subfigure}[b]{.58\textwidth}
 	 	\setlength{\vgap}{0.9cm}
 	 	\setlength{\hgap}{1.4cm}
 	 	\centering
 		\begin{tikzpicture}
 		\node[iri,anchor=center] (san) {Santiago : City};

 		\node[iri,anchor=center,right=\hgap of san] (chile) {Chile : Country}
 		  edge[arrin] node[lab,xshift=-0.03cm] {capital} (san);

  		\node[iri,anchor=center,right=\hgap of chile] (peru) {Perú : Country}
  		  edge[arrin,bend left=15] node[lab] {borders} (chile)
  		  edge[arrout,bend right=15] node[lab] {borders} (chile);
 		\end{tikzpicture}
 	 	\caption{Heterogeneous graph}
 	 	\label{fig:hg}
 	\end{subfigure}
 	\caption{Data about capitals and countries in a directed edge-labelled graph and a heterogeneous graph \label{fig:capital}}
 \end{figure}
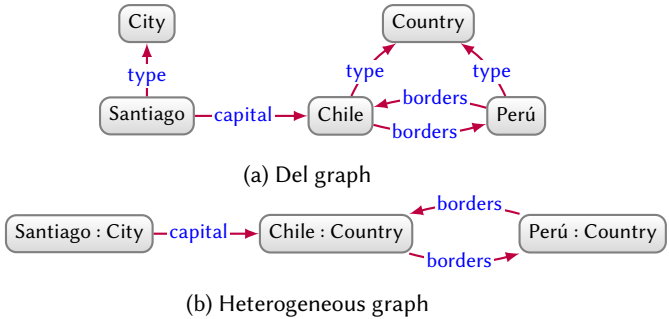

A heterogeneous graph~\cite{HusseinYC18,WangJSWYCY19,YangXJWHW20} (or \textit{heterogeneous information network}~\cite{sun2011pathsim,2012Sun}) is a graph where each node and edge is assigned one type. Heterogeneous graphs are thus akin to del graphs -- with edge labels corresponding to edge types -- but where the type of node forms part of the graph model itself, rather than being expressed as a special relation, as illustrated in Figure~\ref{fig:capital}. An edge is called \textit{homogeneous} if it is between two nodes of the same type (e.g., \gelab{borders}); otherwise it is called \textit{heterogeneous} (e.g., \gelab{capital}). A benefit of heterogeneous graphs is that they allow for partitioning nodes according to their type, for example, for the purposes of machine learning tasks~\cite{HusseinYC18,WangJSWYCY19,YangXJWHW20}. Conversely, they typically only support a one-to-one relation between nodes and types, which is not the case for del graphs (see, for example, the node \gnode{Santiago} with zero types and \gnode{EID15} with multiple types in Figure~\ref{fig:delg}).

\subsubsection{Property graphs}

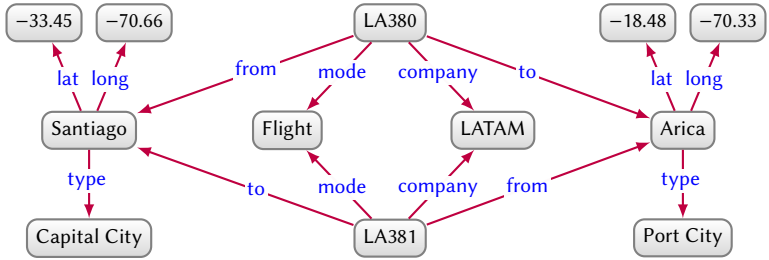
\begin{figure}
\setlength{\vgap}{0.9cm}
\setlength{\hgap}{1.7cm}

\begin{tikzpicture}
\node[iri,anchor=center] (san) {Santiago};  

\node[iri,anchor=center,below=\vgap of san] (sanp) {Capital City}
  edge[arrin] node[lab,xshift=-0.03cm] {type} (san);  
  
\node[iri,anchor=center,above=1\vgap of san,xshift=-0.6cm] (sanlat) {$-$33.45}
  edge[arrin] node[lab] {lat} (san);  
 
\node[iri,anchor=center,above=1\vgap of san,xshift=0.6cm] (sanlong) {$-$70.66}
  edge[arrin] node[lab] {long} (san);  

  
\node[iri,anchor=center,right=0.9\hgap of san] (fl) {Flight};  

\node[iri,anchor=center,right=\hgap of fl] (latam) {LATAM};  

\node[iri,anchor=center,right=0.9\hgap of latam] (ar) {Arica};

\node[iri,anchor=center,above=1\vgap of ar,xshift=-0.6cm] (arlat) {$-$18.48}
  edge[arrin] node[lab] {lat} (ar);  
 
\node[iri,anchor=center,above=1\vgap of ar,xshift=0.6cm] (arlong) {$-$70.33}
  edge[arrin] node[lab] {long} (ar);  

\node[iri,anchor=center,below=\vgap of ar] (arp) {Port City}
  edge[arrin] node[lab,xshift=-0.03cm] {type} (ar);
  
\node[between=fl and latam,anchor=center] (mid) {};  

\node[iri,anchor=center] (fsa) at (mid|-sanp) {LA381}
  edge[arrout] node[lab] {to} (san)
  edge[arrout] node[lab] {from} (ar)
  edge[arrout] node[lab] {company} (latam)
  edge[arrout] node[lab] {mode} (fl); 

\node[iri,anchor=center] (fas) at (mid|-sanlat) {LA380}
  edge[arrout] node[lab] {from} (san)
  edge[arrout] node[lab] {to} (ar)
  edge[arrout] node[lab] {company} (latam)
  edge[arrout] node[lab] {mode} (fl); 
  
\end{tikzpicture}

\caption{Directed edge-labelled graph with companies offering flights between Santiago and Arica \label{fig:fsa}}
\end{figure}

Property graphs were introduced to provide additional flexibility when modelling more complex relations. Consider integrating incoming data that provides information on which companies offer fares on which flights, allowing the board to better understand available routes between cities (for example, on national airlines). In the case of directed-edge labelled graphs, we cannot directly annotate an edge like \gedge{Santiago}{flight}{Arica} with the company (or companies) offering that route. But we could add a new node denoting a flight, connect it with the source, destination, companies, and mode, as shown in Figure~\ref{fig:fsa}. Applying this modelling to all routes in Figure~\ref{fig:delg} would, however, involve a significant change to the graph. Another option might be to put the flights of different companies in different named graphs, but if named graphs are already being used to track the source of graphs (for example), this could become cumbersome.

\begin{figure}
\setlength{\vgap}{0.8cm}
\setlength{\hgap}{2.9cm}

\begin{tikzpicture}
  \node[nrect] (n1) {
   \alt{
       \uri{lat} & =\,\uri{$-$33.45}\\[-1ex]
       \uri{long} & =\,\uri{$-$70.66} } };
  
  \node[rt] (ln1) at (n1.north) {\uri{Santiago : Capital City}};
  
  \node[nrect, right=4.5cm of n1] (n2) {
   \alt{
      \uri{lat} & =\,\uri{$-$18.48}\\[-1ex]
      \uri{long} & =\,\uri{$-$70.33} } };
  
  \node[rt] (ln2) at (n2.north) {\uri{Arica : Port City}};
  
  \draw[arrout,pos=0.5,bend left=15] (n1) to 
   node[erect] (e1) {
   \alt{ \uri{company} & =\,\uri{LATAM} \\ } }
  (n2);
  
  \node[rte] (le1) at (e1.north) {\uri{LA380 $:$ flight}}; 
  
  \draw[arrout,pos=0.5,bend left=15] (n2) to 
   node[erect] (e2) {
   \alt{ \uri{company} & =\,\uri{LATAM}\\ } }
  (n1);
  
  \node[rte] (le2) at (e2.north) {\uri{LA381 $:$ flight}};
\end{tikzpicture}

\caption{Property graph with companies offering flights between Santiago and Arica \label{fig:pg}}
\end{figure}
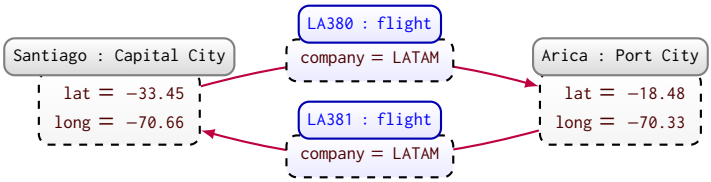

The property graph model was thus proposed to offer additional flexibility when modelling data as a graph~\cite{Miller13,AnglesABHRV17}. A property graph allows a set of \textit{property--value} pairs and a \textit{label} to be associated with both nodes and edges. Figure~\ref{fig:pg} shows a concise example of a property graph with data analogous to Figure~\ref{fig:fsa} (for a formal definition of a property graph, we refer to Definition~\ref{def:pg} in Appendix~\ref{sec:formal}). This time we use property--value pairs on edges to model the companies\footnote{In practical implementations of property graphs, properties with multiple values may be expressed, for example, as a single array value. Such issues do not, however, affect expressivity, nor our discussion.}. The type of relation is captured by the label \texttt{flight}. We further use node labels to indicate the types of the two nodes, and use property--value pairs to indicate their latitude and longitude.

Property graphs are most prominently used in popular graph databases, such as Neo4j~\cite{Miller13,AnglesABHRV17}. In choosing between graph models, it is important to note that property graphs can be translated to/from directed edge-labelled graphs without loss of information~\cite{HernandezHK15,AnglesTT19} (per, e.g., Figure~\ref{fig:pg}). In summary, directed-edge labelled graphs offer a more minimal model, while property graphs offer a more flexible one. Often the choice of model will be secondary to other practical factors, such as the implementations available for different models, etc.

\subsubsection{Graph dataset}
\label{subsub:graphdataset}

\begin{figure}
\setlength{\vgap}{1cm}
\setlength{\hgap}{1.8cm}

\colorlet{ng}{black!2}

\tikzset{lab/.append style={fill=ng}}

\begin{tikzpicture}
\node[block, minimum width=13cm,minimum height=5.6cm, fill=ng, anchor=south west] (eg) at (0,0)  {};

\node[above=0.1cm of eg.south west,anchor=south west,xshift=0.1cm] (egt) {\texttt{Events}};  

\node[iri,anchor=mid,below right=1cm of eg.north west,yshift=0.4cm,xshift=3.1cm] (n) {Ñam};
  
\node[iri,anchor=center,below=\vgap of n] (nId) {EID15}
  edge[arrout] node[lab] {name} (n);

\node[iri,anchor=center,right=\hgap of nId] (ff) {Food Festival}
  edge[arrin] node[lab] {type} (nId);
  
\node[iri,anchor=center,below=\vgap of ff] (df) {Drinks Festival}
  edge[arrin] node[lab] {type} (nId);
  
\node[iri,anchor=center,above=\vgap of ff] (om) {Open Market}
  edge[arrin] node[lab] {type} (nId);  
  
\node[iri,anchor=center,below=\vgap of nId] (sl) {Santa Lucía}
  edge[arrin] node[lab] {venue} (nId);  
  
\node[iri,anchor=center,below=\vgap of sl] (san) {Santiago}
  edge[arrin] node[lab] {city} (sl);  
  
\node[iri,anchor=center,right=\hgap of ff] (fId) {EID16}
  edge[arrout] node[lab] {type} (ff)
  edge[arrout] node[lab] {type} (om);  

\node[iri,anchor=center,above=\vgap of fId] (f) {Food Truck}
  edge[arrin] node[lab] {name} (fId);
  
\node[iri,anchor=center,left=0.8\hgap of nId] (fS) {2018-03-22 12:00}
  edge[arrin] node[lab] {start} (nId);  
  
\node[iri,anchor=center,above=\vgap of fS] (fE) {2018-03-29 20:00}
  edge[arrin] node[lab] {end} (nId);   
  
\node[iri,anchor=center,below=\vgap of fId,xshift=0.65\hgap] (po) {Piscina Olímpica}
  edge[arrin] node[lab] {venue} (fId);  
  
\node[iri,anchor=center,below=\vgap of po] (ar) {Arica}
  edge[arrin] node[lab] {city} (po);   

\node[iri,anchor=center,below=\vgap of fId,xshift=-0.65\hgap] (sm) {Sotomayor}
  edge[arrin] node[lab] {venue} (fId);  
  
\node[iri,anchor=center,below=\vgap of sm] (vdm) {Viña del Mar}
  edge[arrin] node[lab] {city} (sm);   
  
\node[block, minimum width=13cm,minimum height=2cm, fill=ng,below=0.3cm of eg.south west,anchor=north west] (rg) {};  

\node[above=0.1cm of rg.south west,anchor=south west,xshift=0.1cm] (rgt) {\texttt{Routes}};  
  
\node[iri,anchor=center,below=\vgap of san] (sanL) {Santiago};  

\node[iri,anchor=center,below=\vgap of ar] (arL) {Arica}
  edge[arroutin,bend left=20] node[lab] {flight} (sanL);    
  
\node[iri,anchor=center,below=\vgap of vdm] (vdmL) {Viña del Mar}
  edge[arroutin,bend left=10] node[lab] {flight} (sanL)
  edge[arroutin,bend right=10] node[lab] {bus} (sanL)
  edge[arroutin] node[lab] {bus} (arL);      
  
\node[block,dashed,minimum width=13cm,minimum height=1.1cm,below=0.3cm of rg.south west,anchor=north west] (dg) {};  

\tikzset{lab/.append style={fill=white}}

\node[above=0.1cm of dg.south west,anchor=south west,xshift=0.1cm] (dgt) {\textit{Default}};   
  
\node[iri,anchor=center,below=1.6\vgap of sanL,xshift=-1.4cm] (routes) {Routes};     

\node[iri,anchor=center,right=\hgap of routes] (routesU) {2018-04-03}
  edge[arrin] node[lab] {updated} (routes); 
  
\node[iri,anchor=center,right=0.9\hgap of routesU] (events) {Events}; 

\node[iri,anchor=center,right=\hgap of events] (eventsU) {2018-06-14}
  edge[arrin] node[lab] {updated} (events); 
\end{tikzpicture}

\caption{Graph dataset with two named graphs and a default graph describing events and routes \label{fig:gd}}
\end{figure}
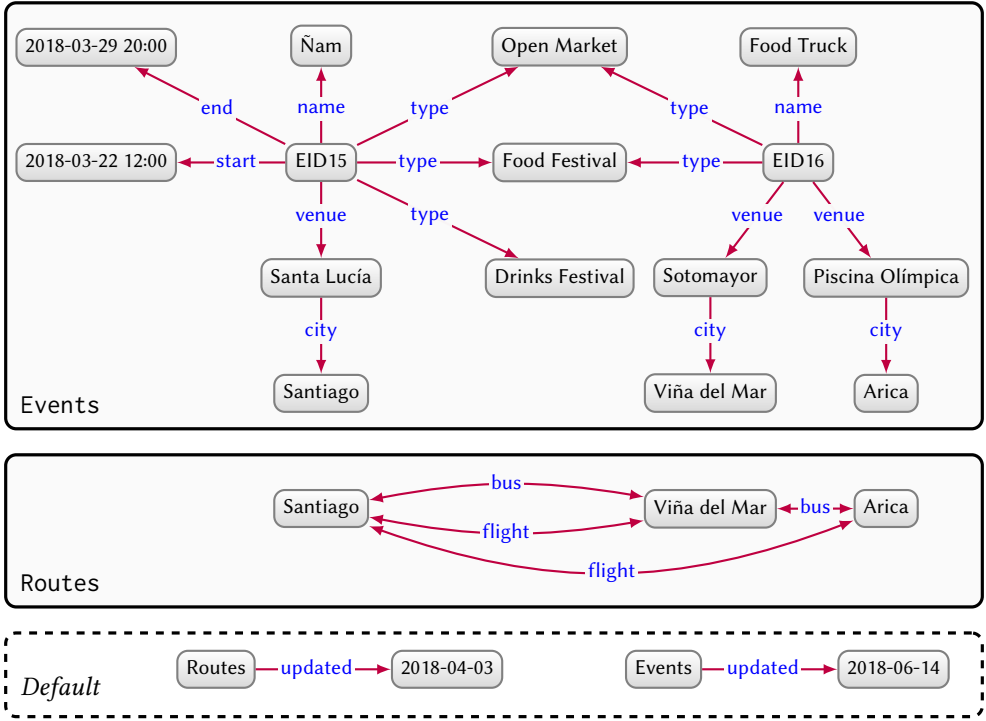

Although multiple directed edge-labelled graphs can be merged by taking their union, it is often desirable to manage several graphs rather than one monolithic graph; for example, it may be beneficial to manage multiple graphs from different sources, making it possible to update or refine data from one source, to distinguish untrustworthy sources from more trustworthy ones, and so forth. A graph dataset then consists of a set of \textit{named graphs} and a \textit{default graph}. Each named graph is a pair of a graph ID and a graph. The default graph is a graph without an ID, and is referenced ``by default'' if a graph ID is not specified. Figure~\ref{fig:gd} provides an example where events and routes are stored in two named graphs, and the default graph manages meta-data about the named graphs (for a formal definition of a graph dataset, see Definition~\ref{def:gd} in Appendix~\ref{sec:formal}). Graph names can also be used as nodes in a graph. Furthermore, nodes and edges can be repeated across graphs, where the same node in different graphs will typically refer to the same entity, allowing data on that entity to be integrated when merging graphs. Though the example illustrates a dataset of directed edge-labelled graphs, the concept generalises straightforwardly to other types of graphs.

A prominent use-case for graph datasets is to manage and query \textit{Linked Data} composed of interlinked documents of RDF graphs spanning the Web. When dealing with Web data, tracking the source of data becomes of key importance~\cite{Dividino09,BonattiHPS11,zimm-etal-2012-JWS}. We will discuss Linked Data later in Section~\ref{sec:identity} and further discuss provenance in Section~\ref{ssec:knowledgeContext}.

\subsubsection{Other graph data models} The previous models are popular examples of graph representations. Other graph data models exist with \textit{complex nodes} that may contain individual edges~\cite{AnglesG08,Hartig14} or nested graphs~\cite{AnglesG08,n3} (sometimes called \textit{hypernodes}~\cite{LeveneP89}). Likewise the mathematical notion of a \textit{hypergraph} defines \textit{complex edges} that connect sets rather than pairs of nodes. In our view, a knowledge graph can adopt any such graph data model based on nodes and edges: often data can be converted from one model to another (see Figure~\ref{fig:fsa} vs.\ Figure~\ref{fig:pg}). In the rest of the paper, we prefer discussing directed-edge labelled graphs given their relative succinctness, but most discussion extends naturally to other models.

\subsubsection{Graph stores} A variety of techniques have been proposed for storing and indexing graphs, facilitating the efficient evaluation of queries (as discussed next). Directed-edge labelled graphs can be stored in relational databases either as a single relation of arity three (\textit{triple table}), as a binary relation for each property (\textit{vertical partitioning}), or as $n$-ary relations for entities of a given type (\textit{property tables})~\cite{WylotHCS18}. Custom storage techniques have also been developed for a variety of graph models, providing efficient access for finding nodes, edges and their adjacent elements~\cite{AnglesG08,Miller13,WylotHCS18}. A number of systems further allow for distributing graphs over multiple machines based on popular NoSQL stores or custom partitioning schemes~\cite{WylotHCS18,JankeS18}. For further details we refer to the book chapter by \citet{JankeS18} and the survey by \citet{WylotHCS18} dedicated to this topic.

\subsection{Querying}\label{ssec:querying}

A number of practical languages have been proposed for querying graphs~\cite{AnglesABHRV17}, including the SPARQL query language for RDF graphs~\cite{sparql11}; and Cypher~\cite{FrancisGGLLMPRS18}, Gremlin~\cite{Rodriguez15}, and G-CORE~\cite{AnglesABBFGLPPS18} for querying property graphs.\footnote{The popularity of these languages is investigated by \citet{seifer19}.} Underlying these query languages are some common primitives, including (basic) graph patterns, relational operators, path expressions, and more besides~\cite{AnglesABHRV17}. We now describe these core features for querying graphs in turn, starting with graph patterns.

\subsubsection{Graph patterns} At the core of every structured query language for graphs are (\textit{basic}) \textit{graph patterns}~\cite{ConsensM90,AnglesABHRV17}, which follow the same model as the data graph being queried (see Section~\ref{ssec:graphModels}), additionally allowing variables as terms.\footnote{The terms of a directed edge-labelled graph are its nodes and edge-labels. The terms of a property graph are its ids, labels, properties, and values (as used on either edges or nodes).} Terms in graph patterns are thus divided into constants, such as \gnode{Arica} or \gelab{venue}, and variables, which we prefix with question marks, such as \gvar{?event} or \gelab{\color{black} ?rel}. A graph pattern is then evaluated against the data graph by generating mappings from the variables of the graph pattern to constants in the data graph such that the image of the graph pattern under the mapping (replacing variables with the assigned constants) is contained within the data graph.

\begin{figure}[t]
	\setlength{\vgap}{1.2cm}
	\setlength{\hgap}{1.2cm}
	
	\begin{tikzpicture}
	\node[iri,anchor=center] (ff) {Food Festival};
	
	\node[var,anchor=center,right=\hgap of ff] (e) {\textbf{?ev}}
	edge[arrout] node[lab] {type} (ff);
	
	\node[var,anchor=center,right=\hgap of e,yshift=0.8\vgap] (l1) {\textbf{?vn1}}
	edge[arrin] node[lab,yshift=-0.2ex] {venue} (e);  
	
	\node[var,anchor=center,right=\hgap of e,yshift=-0.8\vgap] (l2) {\textbf{?vn2}}
	edge[arrin] node[lab,yshift=0.2ex] {venue} (e); 
	\end{tikzpicture}\qquad %
	\tt\footnotesize %
	\begin{tabular}[b]{lll}
		\toprule
		\textbf{?ev} & \textbf{?vn1} & \textbf{?vn2} \\
		\midrule
		EID16 & Piscina Olímpica & Sotomayor \\
		EID16 & Sotomayor & Piscina Olímpica  \\
		EID16 & Piscina Olímpica & Piscina Olímpica  \\
		EID16 & Sotomayor & Sotomayor  \\
		EID15 & Santa Lucía & Santa Lucía \\
		\bottomrule
	\end{tabular}
	\caption{Graph pattern (left) with mappings generated over the graph of Figure~\ref{fig:delg} (right) \label{fig:gp}}
\end{figure}
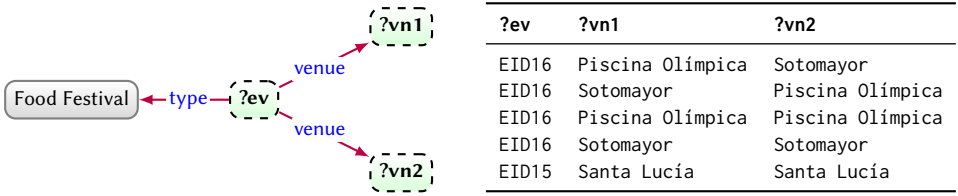

In Figure~\ref{fig:gp}, we provide an example of a graph pattern looking for the venues of Food Festivals, along with the possible mappings generated by the graph pattern against the data graph of Figure~\ref{fig:delg}. In some of the presented mappings (the last two listed), multiple variables are mapped to the same term, which may or may not be desirable depending on the application. Hence a number of semantics have been proposed for evaluating graph patterns~\cite{AnglesABHRV17}, amongst which the most important are: \textit{homomorphism-based semantics}, which allows multiple variables to be mapped to the same term such that all mappings shown in Figure~\ref{fig:gp} would be considered results; and \textit{isomorphism-based semantics}, which requires variables on nodes and/or edges to be mapped to unique terms, thus excluding the latter three mappings of Figure~\ref{fig:gp} from the results. Different practical languages adopt different semantics for evaluating graph patterns where, for example, SPARQL adopts a homomorphism-based semantics, while Cypher adopts an isomorphism-based semantics on edges.

As we will see in later examples (particularly Figure~\ref{fig:cgp}), graph patterns may also form cycles (be they directed or undirected), and may replace edge labels with variables. Graph patterns in the context of other models -- such as property graphs -- can be defined analogously by allowing variables to replace terms in any position of the model. We provide a formalisation of graph patterns and their evaluation for both directed edge-labelled graphs and property graphs in Appendix~\ref{app:gps}.

\subsubsection{Complex graph patterns} A graph pattern transforms an input graph into a table of results (as shown in Figure~\ref{fig:gp}). We may then consider using the relational algebra to combine and/or transform such tables, thus forming more complex queries from one or more graph patterns. Recall that the relational algebra consists of unary operators that accept one input table, and binary operators that accept two input tables. Unary operators include projection ($\pi$) to output a subset of columns, selection ($\sigma$) to output a subset of rows matching a given condition, and renaming of columns ($\rho$). Binary operators include union ($\cup$) to merge the rows of two tables into one table, difference ($-$) to remove the rows from the first table present in the second table, and joins ($\Join$) to extend the rows of one table with rows from the other table that satisfy a join condition. Selection and join conditions typically include equalities ($=$), inequalities ($\leq$), negation ($\neg$), disjunction ($\vee$), etc. From these operators, we can further define other (syntactic) operators, such as intersection ($\cap$) to output rows in both tables, anti-join ($\rhd$, aka \textit{not exists}) to output rows from the first table for which there are no join-compatible rows in the second table, left-join ($\LeftJoin$, aka \textit{optional}) to perform a join but keeping rows from the first table without a compatible row in the second table, etc. 

Graph patterns can then be expressed in a subset of relational algebra (namely $\pi$, $\sigma$, $\rho$, $\Join$). Assuming, for example, a single ternary relation $G(s,p,o)$ representing a graph -- i.e., a table $G$ with three columns $s$, $p$, $o$ -- the query of Figure~\ref{fig:gp} can be expressed in relational algebra as:
\[ \pi_{ev,vn1,vn2}(\sigma_{p=\texttt{type} \wedge o=\texttt{Food Festival} \wedge p_1=p_2=\texttt{venue}}(\rho_{s/ev}(G \bowtie \rho_{p/p_1,o/vn1}(G) \bowtie \rho_{p/p_2,o/vn2}(G)))) \]
where $\Join$ denotes a \textit{natural join}, meaning that equality is checked across pairs of columns with the same name in both tables (here, the join is thus performed on the subject column $s$). The result of this query is a table with a column for each variable: $ev,vn1,vn2$. However, not all queries using $\pi$, $\sigma$, $\rho$ and $\Join$ on $G$ can be expressed as graph patterns; for example, we cannot choose which variables to project in a graph pattern, but rather must project all variables not fixed to a constant.

\begin{figure}
	\setlength{\vgap}{1cm}
	\setlength{\hgap}{1.7cm}
	
	\begin{tikzpicture}
	\node[iri,anchor=center] (ff) {Food Festival};
	
	\node[var,anchor=center,above=\vgap of ff,xshift=-1.5\hgap] (e1) {?event1}
	edge[arrout] node[lab] {type} (ff);  
	
	\node[var,anchor=center,right=\hgap of e1] (l1) {?ven1}
	edge[arrin] node[lab,xshift=-0.3ex] {venue} (e1); 
	
	\node[var,anchor=center,below=0.65\vgap of e1] (n1) {\textbf{?name1}}
	edge[arrin] node[lab,yshift=0.3ex] {name} (e1); 
	
	\node[var,anchor=center,right=0.7\hgap of l1] (c1) {?city1}
	edge[arrin] node[lab,xshift=-0.3ex] {city} (l1); 
	
	\node[var,anchor=center,below=\vgap of ff,xshift=-1.5\hgap] (e2) {?event2}
	edge[arrout] node[lab] {type} (ff); 
	
	\node[var,anchor=center,right=\hgap of e2] (l2) {?ven2}
	edge[arrin] node[lab,xshift=-0.3ex] {venue} (e2);
	
	\node[var,anchor=center,above=0.65\vgap of e2] (n2) {\textbf{?name2}}
	edge[arrin] node[lab,yshift=-0.3ex] {name} (e2);  
	
	\node[var,anchor=center,right=0.7\hgap of l2] (c2) {?city2}
	edge[arrin] node[lab,xshift=-0.3ex] {city} (l2)
	edge[arroutin] node[vlab] {\textbf{?con}} (c1);  
	
	\end{tikzpicture}\qquad %
	\tt\footnotesize %
	\begin{tabular}[b]{lll}
		\toprule
		\textbf{?name1} & \textbf{?con} & \textbf{?name2} \\
		\midrule
		Food Truck & bus & Food Truck \\
		Food Truck & bus & Food Truck \\
		Food Truck & bus & Ñam \\
		Food Truck & flight & Ñam \\
		Food Truck & flight & Ñam \\
		Ñam & bus & Food Truck \\
		Ñam & flight & Food Truck \\
		Ñam & flight & Food Truck \\
		\bottomrule
	\end{tabular}
	
	\caption{Conjunctive query (left) with mappings generated over the graph of Figure~\ref{fig:delg} (right) \label{fig:cq}}
\end{figure}
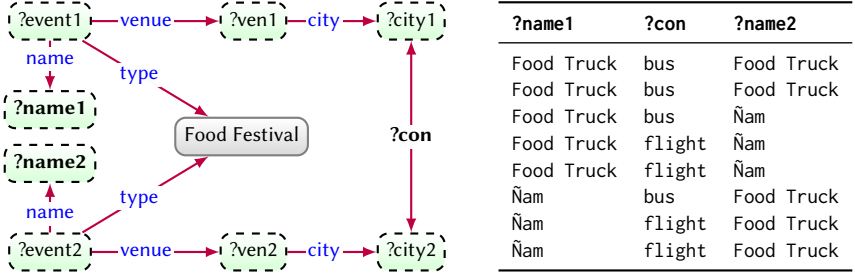

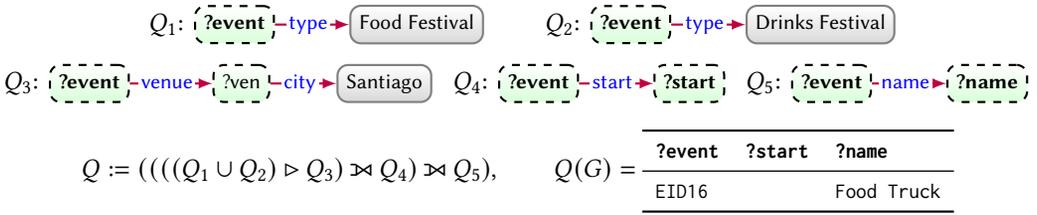
\begin{figure}[t]
	\setlength{\vgap}{1.2cm}
	\setlength{\hgap}{1.1cm}
	
	\centering
	\begin{tikzpicture}
	\node[var,anchor=center] (e) {\textbf{?event}};
	
	\node[iri,anchor=center,right=0.9\hgap of e] (l1) {Food Festival}
	edge[arrin] node[lab] {type} (e);
	
	\node[left=0.01cm of e] {$Q_1$:};  
	\end{tikzpicture}\qquad %
	\begin{tikzpicture}
	\node[var,anchor=center] (e) {\textbf{?event}};
	
	\node[iri,anchor=center,right=0.9\hgap of e] (l1) {Drinks Festival}
	edge[arrin] node[lab] {type} (e);
	
	\node[left=0.01cm of e] {$Q_2$:};   
	\end{tikzpicture}
	
	\vspace{0.2cm} 
	\begin{tikzpicture}
	\node[var,anchor=center] (e) {\textbf{?event}};
	
	\node[var,anchor=center,right=\hgap of e] (l) {?ven}
	edge[arrin] node[lab] {venue} (e);
	
	\node[iri,anchor=center,right=0.8\hgap of l] (s) {Santiago}
	edge[arrin] node[lab] {city} (l);  
	
	\node[left=0.01cm of e] {$Q_3$:};
	\end{tikzpicture}
	\hfill	
	\begin{tikzpicture}
	\node[var,anchor=center] (e) {\textbf{?event}};
	
	\node[var,anchor=center,right=0.9\hgap of e] (l1) {\textbf{?start}}
	edge[arrin] node[lab] {start} (e);
	
	\node[left=0.01cm of e] {$Q_4$:}; 
	\end{tikzpicture}
	\hfill	
	\begin{tikzpicture}
	\node[var,anchor=center] (e) {\textbf{?event}};
	
	\node[var,anchor=center,right=0.9\hgap of e] (l1) {\textbf{?name}}
	edge[arrin] node[lab] {name} (e);
	
	\node[left=0.01cm of e] {$Q_5$:};
	\end{tikzpicture}
	\hfill %
	\vspace{0.3cm}
	
	$Q := ((((Q_1 \cup Q_2) \rhd Q_3) \LeftJoin Q_4 ) \LeftJoin Q_5),\qquad Q(G) =$
	\footnotesize\tt
	\begin{tabular}[m]{lll}
		\toprule
		\textbf{?event} & \textbf{?start} & \textbf{?name}  \\
		\midrule
		EID16 &  & Food Truck \\
		\bottomrule
	\end{tabular}
	
	\caption{Complex graph pattern ($Q$) with mappings generated ($Q(G)$) over the graph of Figure~\ref{fig:delg} ($G$)\label{fig:cgp}}
\end{figure}

Graph query languages such as SPARQL~\cite{sparql11} and Cypher~\cite{FrancisGGLLMPRS18} allow the full use of relational operators over the results of graph patterns, giving rise to \textit{complex graph patterns}~\cite{AnglesABHRV17}. Figure~\ref{fig:cq} presents an example of a complex graph pattern with projected variables in bold, choosing particular variables to appear in the final results. In terms of expressivity, graph patterns with (unrestricted) projection of this form equate to \textit{conjunctive queries} on graphs.
In Figure~\ref{fig:cgp}, we give another example of a complex graph pattern looking for food festivals or drinks festivals not held in Santiago, optionally returning their start date and name (where available). Such queries -- allowing the full use of relational operators on top of graph patterns -- equate to \textit{first-order queries} on graphs. In Appendix~\ref{app:cgps}, we formalise complex graph patterns and their evaluation over data graphs.

Complex graph patterns can give rise to duplicate results; for example, the first result in Figure~\ref{fig:cq} appears twice since \texttt{?city1} matches \texttt{Arica} and \texttt{?city2} matches \texttt{Viña del Mar} in one result, and vice-versa in the other. Query languages then offer two semantics: \textit{bag semantics} preserves duplicates according to the multiplicity of the underlying mappings, while \textit{set semantics} (typically invoked with a \texttt{DISTINCT} keyword) removes duplicates from the results.

\subsubsection{Navigational graph patterns} A key feature that distinguishes graph query languages is the ability to include \textit{path expressions} in queries. A path expression $r$ is a regular expression that allows matching arbitrary-length paths between two nodes, which is expressed as a \textit{regular path query} $(x,r,y)$, where $x$ and $y$ can be variables or constants (or even the same term). The base path expression is where $r$ is a constant (an edge label). Furthermore if $r$ is a path expression, then $r^-$ (\textit{inverse})\footnote{Some authors distinguish \textit{2-way regular path queries} from regular path queries, where only the former supports inverses.} and $r^*$ (\textit{Kleene star}: zero-or-more) are also path expressions. Finally, if $r_1$ and $r_2$ are path expressions, then $r_1 \mid r_2$ (\textit{disjunction}) and $r_1 \cdot r_2$ (\textit{concatenation}) are also path expressions.

\begin{figure}
	\setlength{\hgap}{0.9cm}
	\begin{tabular}{l}
		\begin{tikzpicture}
		\node[iri,anchor=center] (a) {Arica};
		
		\node[numcirc,left=0.2cm of a] {1};
		\end{tikzpicture}\qquad
		\begin{tikzpicture}
		\node[iri,anchor=center] (a) {Arica};  
		\node[iri,anchor=center,right=\hgap of a] (l1) {Viña del Mar}
		edge[arrin] node[lab,xshift=-0.2ex] {bus} (a);
		
		\node[numcirc,left=0.2cm of a] {2};  
		\end{tikzpicture}\qquad
		\begin{tikzpicture}
		\node[iri,anchor=center] (a) {Arica};  
		\node[iri,anchor=center,right=\hgap of a] (v) {Viña del Mar}
		edge[arrin] node[lab] {bus} (a);
		\node[iri,anchor=center,right=\hgap of v] (a2) {Arica}
		edge[arrin] node[lab] {bus} (v);
		
		\node[numcirc,left=0.2cm of a] {3};  
		\end{tikzpicture} \\
		\begin{tikzpicture}
		\node[iri,anchor=center] (a) {Arica};  
		\node[iri,anchor=center,right=\hgap of a] (v) {Viña del Mar}
		edge[arrin] node[lab] {bus} (a);
		\node[iri,anchor=center,right=\hgap of v] (a2) {Arica}
		edge[arrin] node[lab] {bus} (v);
		\node[iri,anchor=center,right=\hgap of a2] (v2) {Viña del Mar}
		edge[arrin] node[lab] {bus} (a2);  
		
		\node[numcirc,left=0.2cm of a] {4}; 
		\end{tikzpicture}\qquad \raisebox{1ex}{...}
	\end{tabular}
	
	\caption{Some possible paths matching $(\texttt{Arica},\texttt{bus*},\texttt{?city})$ over the graph of Figure~\ref{fig:delg}\label{fig:path}}
\end{figure}
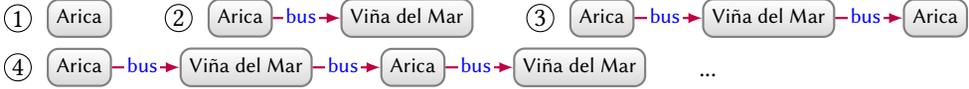

Regular path queries can then be evaluated under a number of different semantics. For example, $(\texttt{Arica},\texttt{bus*},\texttt{?city})$ evaluated against the graph of Figure~\ref{fig:delg} may match the paths in Figure~\ref{fig:path}. In fact, since a cycle is present, an infinite number of paths are potentially matched. For this reason, restricted semantics are often applied, returning only the shortest paths, or paths without repeated nodes or edges (as in the case of Cypher).\footnote{Mapping variables to paths requires special treatment~\cite{AnglesABHRV17}. Cypher~\cite{FrancisGGLLMPRS18} returns a string that encodes a path, upon which certain functions such as \texttt{length($\cdot$)} can be applied. G-CORE~\cite{AnglesABBFGLPPS18}, on the other hand, allows for returning paths, and supports additional operators on them, including projecting them as graphs, applying cost functions, and more besides.} Rather than returning paths, another option is to instead return the (finite) set of pairs of nodes connected by a matching path (as in the case of SPARQL 1.1).

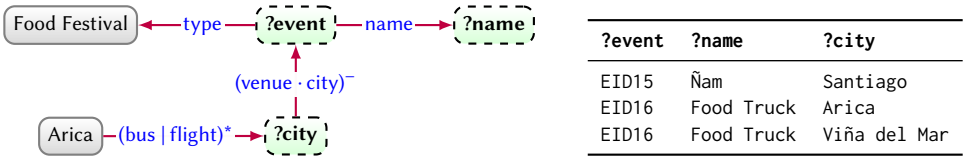
\begin{figure}[t]
	\setlength{\vgap}{1.2cm}
	\setlength{\hgap}{2.1cm}
	
	\begin{tikzpicture}
	\node[iri,anchor=center] (ff) {Food Festival};
	
	\node[var,right=\hgap of ff,anchor=center] (e) {\textbf{?event}}
	edge[arrout] node[lab] {type} (ff);
	
	\node[var,right=\hgap of e,anchor=center] (n) {\textbf{?name}}
	edge[arrin] node[lab] {name} (e);  
	
	\node[var,below=\vgap of e,anchor=center] (c) {\textbf{?city}}
	edge[arrout] node[lab] {(venue$\,\cdot\,$city)$^-$} (e);
	
	\node[iri,below=\vgap of ff,anchor=center] (a) {Arica}
	edge[arrout] node[lab] {(bus$\,\mid\,$flight)*} (c);
	\end{tikzpicture}\qquad %
	\tt\footnotesize %
	\begin{tabular}[b]{lll}
		\toprule
		\textbf{?event} & \textbf{?name} & \textbf{?city} \\
		\midrule
		EID15 & Ñam & Santiago \\
		EID16 & Food Truck & Arica \\
		EID16 & Food Truck & Viña del Mar \\
		\bottomrule
	\end{tabular}
	\caption{Navigational graph pattern (left) with mappings generated over the graph of Figure~\ref{fig:delg} (right) \label{fig:ngp}}
\end{figure}

Regular path queries can then be used in graph patterns to express \textit{navigational graph patterns}~\cite{AnglesABHRV17}, as shown in Figure~\ref{fig:ngp}, which illustrates a query searching for food festivals in cities reachable (recursively) from Arica by bus or flight. Furthermore, when regular path queries and graph patterns are combined with operators such as projection, selection, union, difference, and optional, the result is known as \textit{complex navigational graph patterns}~\cite{AnglesABHRV17}. Appendix~\ref{app:ngps} provides definitions for (complex) navigational graph patterns and their evaluation.

\subsubsection{Other features}\label{app:qother} 

Thus far we have discussed features that form the practical and theoretical foundation of any query language for graphs~\cite{AnglesABHRV17}. However, specific query languages for graphs may support other practical features, such as aggregation (\texttt{GROUP BY}, \texttt{COUNT}, etc.), more complex filters and datatype operators (e.g., range queries on years extracted from a date), federation for querying remotely hosted graphs over the Web, languages for updating graphs, support for semantic entailment regimes, etc. For more information, we refer to the documentation of the respective query languages (e.g.,~\cite{sparql11,AnglesABBFGLPPS18}) and to the survey by~\citet{AnglesABHRV17}.

\section{Schema, Identity, Context}\label{sec:knowledge}

In this section we describe various enhancements and extensions of the data graph -- relating to schema, identity and context -- that provide additional structures for accumulating knowledge. Henceforth, we refer to a \textit{data graph} as a collection of data represented as nodes and edges using one of the models discussed in Section~\ref{sec:graph}. We refer to a \textit{knowledge graph} as a data graph potentially enhanced with representations of schema, identity, context, ontologies and/or rules. These additional representations may be embedded in the data graph, or layered above it. Representations for schema, identity and context are discussed herein, while ontologies and rules will be discussed in Section~\ref{sec:deductive}.

\subsection{Schema}\label{sec:schema}

One of the benefits of modelling data as graphs -- versus, for example, the relational model -- is the option to forgo or postpone the definition of a schema. However, when modelling data as graphs, schemata \textit{can} be used to prescribe a high-level structure and/or semantics that the graph follows or should follow. We discuss three types of graph schemata: \textit{semantic}, \textit{validating}, and \textit{emergent}. 

\subsubsection{Semantic schema}\label{sec:semSchema}

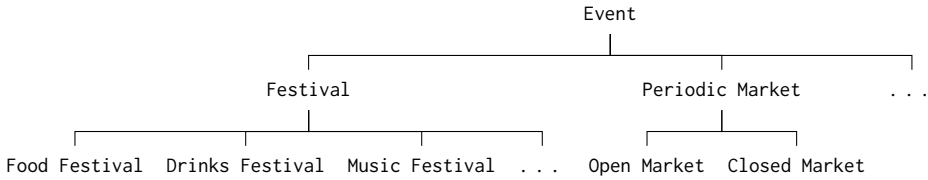
\begin{figure}
	\centering
	\begin{tikzpicture}
	\tikzset{edge from parent/.style=
		{draw,
			edge from parent path={(\tikzparentnode.south)
				-- +(0pt,-8pt)
				-| (\tikzchildnode)}}}
	\tikzset{every tree node/.style={font=\tt\hsp\footnotesize}}
	\tikzset{level distance=1cm}
	\Tree [.Event
	[.Festival Food\ Festival Drinks\ Festival Music\ Festival {\ldots} ] [.Periodic\ Market Open\ Market Closed\ Market ] 
	{\ldots} ]
	\end{tikzpicture}
	\caption{Example class hierarchy for \texttt{Event} \label{fig:classhier}}
\end{figure}

A semantic schema allows for defining the meaning of high-level terms (aka \textit{vocabulary} or \textit{terminology}) used in the graph, which facilitates reasoning over graphs using those terms. Looking at Figure~\ref{fig:delg}, for example, we may notice some natural groupings of nodes based on the types of entities to which they refer. We may thus decide to define \textit{classes} to denote these groupings, such as \texttt{Event}, \texttt{City}, etc. In fact, Figure~\ref{fig:delg} already illustrates three low-level classes -- \texttt{Open Market}, \texttt{Food Market}, \texttt{Drinks Festival} -- grouping similar entities with an edge labelled \gelab{type}. We may subsequently observe some natural relations between some of these classes that we would like to capture. In Figure~\ref{fig:classhier}, we present a class hierarchy for events where children are defined to be \textit{subclasses} of their parents such that if we find an edge \gedge{EID15}{type}{Food Festival} in our graph, we may also \textit{infer} that \gedge{EID15}{type}{Festival} and \gedge{EID15}{type}{Event}.

Aside from classes, we may also wish to define the semantics of edge labels, aka \textit{properties}. Returning to Figure~\ref{fig:delg}, we may consider that the properties \gelab{city} and \gelab{venue} are \textit{sub-properties} of a more general property \gelab{location}, such that given an edge \gedge{Santa Lucía}{city}{Santiago}, for example, we may also infer that \gedge{Santa Lucía}{location}{Santiago}. We may also consider, for example, that \gelab{bus} and \gelab{flight} are both sub-properties of a more general property \gelab{connects to}. As such, properties may also form a hierarchy. We may further define the \textit{domain} of properties, indicating the class(es) of entities for nodes from which edges with that property extend; for example, we may define that the domain of \gelab{connects to} is a class \texttt{Place}, such that given the previous sub-property relations, we could conclude that \gedge{Arica}{type}{Place}. Conversely, we may define the \textit{range} of properties, indicating the class(es) of entities for nodes to which edges with that property extend; for example, we may define that the range of \gelab{city} is a class \texttt{City}, inferring that \gedge{Arica}{type}{City}.

\begin{figure}
	\setlength{\vgap}{0.9cm}
	\setlength{\hgap}{1.9cm}
	
	\begin{tikzpicture}
	\node[iri,anchor=center] (lp) {location};
	
	\node[iri,anchor=center,below=\vgap of lp,xshift=0.85\hgap] (cp) {city}
	edge[arrout] node[lab,xshift=0.2cm] {subp. of} (lp);
	
	\node[iri,anchor=center,below=\vgap of lp,xshift=-0.85\hgap] (vp) {venue}
	edge[arrout] node[lab,xshift=-0.2cm] {subp. of} (lp);

	\node[iri,anchor=center,left=2.4\hgap] (ec) {Event}
	edge[arrin,bend left=10] node[lab,xshift=0.3cm] {domain} (vp);
	
	\node[iri,anchor=center,below=\vgap of ec,xshift=-0.65\hgap] (fc) {Festival}
	edge[arrout] node[lab,xshift=-0.1cm] {subc. of} (ec);
	
	\node[iri,anchor=center,below=\vgap of ec,xshift=0.65\hgap] (pmc) {Periodic Market}
	edge[arrout] node[lab,xshift=0.1cm] {subc. of} (ec);

	\node[iri,anchor=center,right=2.4\hgap] (pc) {Place}
	edge[arrin,bend right=10] node[lab,xshift=-0.3cm] {domain} (cp)
	edge[arrin] node[lab] {range} (lp);
	
	\node[iri,anchor=center,below=\vgap of pc,xshift=-0.65\hgap] (fc) {City}
	edge[arrout] node[lab,xshift=-0.2cm] {subc. of} (pc)
	edge[arrin] node[lab] {range} (cp);
	
	\node[iri,anchor=center,below=\vgap of pc,xshift=0.65\hgap] (vc) {Venue}
	edge[arrout] node[lab,xshift=0.2cm] {subc. of} (pc)
	edge[arrin,bend left=12] node[lab] {range} (vp);
	\end{tikzpicture}
	
	\caption{Example schema graph describing sub-classes, sub-properties, domains, and ranges \label{fig:sg}}
\end{figure}
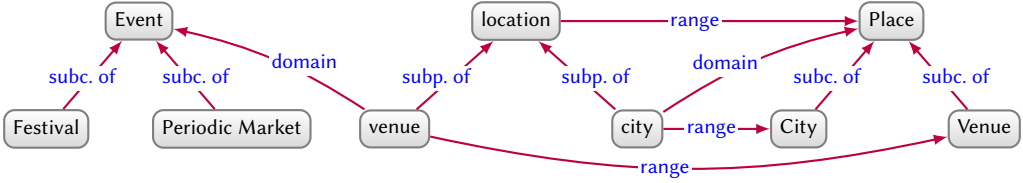

A prominent standard for defining a semantic schema for (RDF) graphs is the \textit{RDF Schema} (\textit{RDFS}) standard~\cite{RDFS}, which allows for defining subclasses, subproperties, domains, and ranges amongst the classes and properties used in an RDF graph, where such definitions can be serialised as a graph. We illustrate the semantics of these features in Table~\ref{tab:semSchema} and provide a concrete example of definitions in Figure~\ref{fig:sg} for a sample of terms used in the running example. These definitions can then be embedded into a data graph. More generally, the semantics of terms used in a graph can be defined in much more depth than seen here, as is supported by the \textit{Web Ontology Language} (\textit{OWL}) standard~\cite{OWL2} for RDF graphs. We will return to such semantics later in Section~\ref{sec:deductive}.

Semantic schema are typically defined for incomplete graph data, where the absence of an edge between two nodes, such as \gedge{Vi\~na del Mar}{flight}{Arica}, does not mean that the relation does not hold in the real world. Therefore, from the graph of Figure~\ref{fig:delg}, we cannot assume that there is no flight between Vi\~na del Mar and Arica. In contrast, if the \emph{Closed World Assumption} (\emph{CWA}) were adopted -- as is the case in many classical database systems -- it would be assumed that the data graph is a complete description of the world, thus allowing to assert with certainty that no flight exists between the two cities. Systems that do not adopt the CWA are said to adopt the \emph{Open World Assumption} (\emph{OWA}). A consequence of CWA is that the addition of an edge to the data graph may contradict what was previously assumed to be false (due to missing information), whereas with OWA, a statement that is proven false continues to be false with the addition of more edges.

Considering our running example, it would be unreasonable to assume that the tourism organisation has complete knowledge of everything describable in its knowledge graph. However, it is inconvenient if a system is unable to definitely answer ``\textit{yes}'' or ``\textit{no}'' to questions such as ``\textit{is there a flight between Arica and Vi\~na del Mar?}'', especially when the organisation is certain that it has complete knowledge of the flights. A compromise between OWA and CWA is the \emph{Local Closed World Assumption} (\textit{LCWA}), where portions of the data graph are assumed to be complete.

\begin{table}
	\caption{Definitions for sub-class, sub-property, domain and range features in semantic schemata \label{tab:semSchema}}
	\scalebox{\scaledeftabs}{
		\begin{tabular}{llll}
			\toprule
			\textbf{Feature} & \textbf{Definition} & \textbf{Condition} & \textbf{Example} \\ 
			\midrule
			
			\footnotesize\textsc{Subclass} & 	\gedge[arrin][1.5cm]{$c$}{subc. of}{$d$} & \footnotesize\gedge[arrin][1cm]{$x$}{type}{$c$} implies \gedge[arrin][1cm]{$x$}{type}{$d$} & \gedge[arrin][1.5cm]{City}{subc. of}{Place} \\
			\midrule
			
			\footnotesize\textsc{Subproperty} & 	\gedge[arrin][1.5cm]{$p$}{subp. of}{$q$} & \footnotesize\gedge[arrin][0.6cm]{$x$}{$p$}{$y$} implies \gedge[arrin][0.6cm]{$x$}{$q$}{$y$} & \gedge[arrin][1.5cm]{venue}{subp. of}{location} \\
			\midrule
			
			\footnotesize\textsc{Domain} & 	\gedge[arrin][1.5cm]{$p$}{domain}{$c$} & \footnotesize\gedge[arrin][0.6cm]{$x$}{$p$}{$y$} implies \gedge[arrin][1cm]{$x$}{type}{$c$} & \gedge[arrin][1.5cm]{venue}{domain}{Event} \\
			\midrule
			
			\footnotesize\textsc{Range} & 	\gedge[arrin][1.5cm]{$p$}{range}{$c$} & \footnotesize\gedge[arrin][0.6cm]{$x$}{$p$}{$y$} implies \gedge[arrin][1cm]{$y$}{type}{$c$} & \gedge[arrin][1.5cm]{venue}{range}{Venue} \\
			\bottomrule
		\end{tabular}
	}
\end{table}

\subsubsection{Validating schema}
\label{sssec:validating-schema}

When graphs are used to represent diverse, incomplete data at large-scale, the OWA is the most appropriate choice for a \textit{default} semantics. But in some scenarios, we may wish to guarantee that our data graph -- or specific parts thereof -- are in some sense ``complete''. Returning to Figure~\ref{fig:delg}, for example, we may wish to ensure that all events have at least a name, a venue, a start date, and an end date, such that applications using the data -- e.g., one that sends event notifications to users -- can ensure that they have the minimal information required. Furthermore, we may wish to ensure that the city of an event is \textit{stated to be} a city (rather than \textit{inferring} that it is a city). We can define such constraints in a validating schema and validate the data graph with respect to the resulting schema, listing constraint violations (if any). Thus while semantic schemata allow for inferring new graph data, validating schemata allow for validating existing graph data. 

A standard way to define a validating schema for graphs is using \textit{shapes}~\cite{SHACLSpec,Prudhommeaux2014,Labra2017}. A shape \textit{targets} a set of nodes in a data graph and specifies \textit{constraints} on those nodes. The shape's target can be defined in many ways, such as targetting all instances of a class, the domain or range of a property, the result of a query, nodes connected to the target of another shape by a given property, etc. Constraints can then be defined on the targetted nodes, such as to restrict the number or types of values taken on a given property. A \textit{shapes graph} is formed from a set of interrelated shapes. 

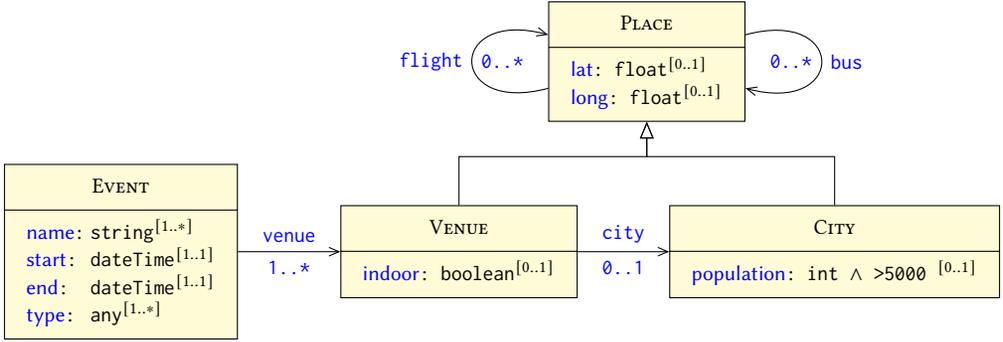
\begin{figure}
	\begin{tikzpicture}
	\umlclass[font=\scshape\footnotesize,rectangle split parts=2]{Event}{
	\begin{tabular}{@{}l@{~~}l@{}}
		\tt	\gelab{name}: & \tt string$^{[1..*]}$ \\
		\tt	\gelab{start}: & \tt dateTime$^{[1..1]}$ \\
		\tt	\gelab{end}: & \tt dateTime$^{[1..1]}$ \\
		\tt	\gelab{type}: & \tt	any$^{[1..*]}$ \\
	\end{tabular}
	}{}
	\umlclass[x = 4.5,font=\scshape\footnotesize,rectangle split parts=2]{Venue}{
		\tt	\gelab{indoor}: boolean$^{[0..1]}$ \\
	}{}
	\umlclass[x = 9.5,font=\scshape\footnotesize,rectangle split parts=2]{City}{
		\tt	\gelab{population}: int $\wedge$ >5000 $^{[0..1]}$\\
	}{}
	\umlclass[x = 7,y=2.5,font=\scshape\footnotesize,rectangle split parts=2]{Place}{
		\tt	\gelab{lat}: float$^{[0..1]}$ \\
		\tt	\gelab{long}: float$^{[0..1]}$ \\
	}{}
	\umluniassoc[arg=venue, mult=1..*,pos=0.5, font=\ttfamily\footnotesize, text=blue]{Event}{Venue}
	\umluniassoc[arg=city, mult=0..1,pos=0.5, font=\ttfamily\footnotesize, text=blue]{Venue}{City}
	\umluniassoc[arg=bus, mult=0..*,pos=0.5, recursive=15|-15|1.4cm, font=\ttfamily\footnotesize, text=blue]{Place}{Place}
	\umluniassoc[arg=flight, mult=0..*, pos=0.5, recursive=195|165|1.4cm, font=\ttfamily\footnotesize, text=blue]{Place}{Place}
	\umlVHVinherit{City}{Place}
	\umlVHVinherit{Venue}{Place}
	\end{tikzpicture}
	\caption{Example shapes graph depicted as a UML-like diagram\label{fig:shapeExample}}
\end{figure}

Shapes graphs can be depicted as UML-like class diagrams, where Figure~\ref{fig:shapeExample} illustrates an example of a shapes graph based on Figure~\ref{fig:delg}, defining constraints on four interrelated shapes. Each shape -- denoted with a box like \shap{Place}, \shap{Event}, etc. -- is associated with a set of constraints. Nodes conform to a shape if and only if they satisfy all constraints defined on the shape. Inside each shape box are placed constraints on the number (e.g., \texttt{[1..*]} denotes one-to-many, \texttt{[1..1]} denotes precisely one, etc.) and types (e.g., \texttt{string}, \texttt{dateTime}, etc.) of nodes that conforming nodes can relate to with a property (e.g., \gelab{name}, \gelab{start}, etc.). Another option is to place constraints on the number of nodes conforming to a particular shape that the conforming node can relate to with a property (thus generating edges between shapes); for example, \sedge{Event}{venue}{1..*}{Venue} denotes that conforming nodes for \shap{Event} must relate to at least one node with the property \gelab{venue} that conforms to the \shap{Venue} shape. Shapes can inherit the constraints of parent shapes -- denoted with an $\triangle$ connector -- as in the case of \shap{City} and \shap{Venue}, whose conforming nodes must also conform to the \shap{Place} shape.

Given a shape and a targetted node, it is possible to check if the node conforms to that shape or not, which may require checking conformance of other nodes; for example, the node \gnode{EID15} conforms to the \shap{Event} shape not only based on its local properties, but also based on conformance of \gnode{Santa Lucía} to \shap{Venue} and \gnode{Santiago} to \shap{City}. Conformance dependencies may also be recursive, where the conformance of \gnode{Santiago} to \shap{City} requires that it conforms to \shap{Place}, which requires that \gnode{Viña del Mar} and \gnode{Arica} conform to \shap{Place}, and so on. Conversely, \gnode{EID16} does not conform to \shap{Event}, as it does not have the \gelab{start} and \gelab{end} properties required by the example shapes graph.

When declaring shapes, the data modeller may not know in advance the entire set of properties that some nodes can have. An \textit{open shape} allows the node to have additional properties not specified by the shape, while a \textit{closed shape} does not. For example, if we add the edge \gedge{Santiago}{founder}{Pedro de Valdivia} to the graph represented in Figure~\ref{fig:delg}, then \gnode{Santiago} only conforms to the \shap{City} shape if that shape is defined as open (since the shape does not mention \gelab{founder}). 

Practical languages for shapes often support additional boolean features, such as conjunction (\textit{\textsc{and}}), disjunction (\textit{\textsc{or}}), and negation (\textit{\textsc{not}}) of shapes; for example, we may say that all the values of \gelab{venue} should conform to the shape \shap{\textsc{Venue} \textit{and} (\textit{not} \textsc{City})}, making explicit that venues in the data graph should not be directly given as cities. However, shapes languages that freely combine recursion and negation may lead to semantic problems, depending on how their semantics are defined. To illustrate, consider the following case inspired by the barber paradox~\cite{Labra2017}, involving a shape \shap{Barber} whose conforming nodes \gelab{shave} at least one node conforming to \shap{\textsc{Person} \textit{and} (\textit{not} \textsc{Barber})}. Now, given (only) \gedge{Bob}{shave}{Bob} with \gnode{Bob} conforming to \shap{Person}, does \gnode{Bob} conform to \shap{Barber}? If \textit{yes} -- if \gnode{Bob} conforms to \shap{Barber} -- then \gnode{Bob} violates the constraint by not shaving at least one node conforming to \shap{\textsc{Person} \textit{and} (\textit{not} \textsc{Barber})}. If \textit{no} -- if \gnode{Bob} does not conform to \shap{Barber} -- then \gnode{Bob} satisfies the \shap{Barber} constraint by shaving such a node. Semantics to avoid such paradoxical situations have been proposed based on stratification~\cite{Boneva2017}, partial assignments~\cite{Corman2018b}, and stable models~\cite{Gelfond88}.

Although validating schemata and semantic schemata serve different purposes, they can complement each other. In particular, a validating schema can take into consideration a semantic schema, such that, for example, validation is applied on the data graph including inferences. Taking the class hierarchy of Figure~\ref{fig:classhier} and the shapes graph of Figure~\ref{fig:shapeExample}, for example, we may define the target of the \shap{Event} shape as the nodes that are of type \texttt{Event} (the class). If we first apply inferencing with respect to the class hierarchy of the semantic schema, the \shap{Event} shape would now target \gnode{EID15} and \gnode{EID16}. The presence of a semantic schema may, however, require adapting the validating schema. Taking into account, for example, the aforementioned class hierarchy would require defining a relaxed cardinality on the \gelab{type} property. Open shapes may also be preferred in such cases rather than enumerating constraints on all possible properties that may be inferred on a node.

We provide high-level definitions for shapes and related concepts in Appendix~\ref{app:shapes}.
Two shapes languages have recently emerged for RDF graphs: \textit{Shape Expressions} (\textit{ShEx}), published as a W3C Community Group Report~\cite{Prudhommeaux2014}; and \textit{SHACL} (\textit{Shapes Constraint Language}), published as a W3C Recommendation~\cite{SHACLSpec}.
These languages support the discussed features (and more) and have been adopted for validating graphs in a number of domains relating to health-care~\cite{ThorntonSSGMPW19}, scientific literature~\cite{HammondPT17}, spatial data~\cite{Car2019}, amongst others. More details about ShEx and SHACL can be found in the book by~\citet{Labra2017}. A recently proposed language that can be used as a common basis for both ShEx and SHACL reveals their similarities and differences~\cite{Labra-Gayo2019}. A similar notion of schema has been proposed by \citet{Angles18} for property graphs.

\subsubsection{Emergent schema}\label{ssec:emergentSchema}

Both semantic and validating schemata require a domain expert to explicitly specify definitions and constraints. However, a data graph will often exhibit latent structures that can be automatically extracted as an \textit{emergent schema}~\cite{PhamPEB15} (aka \textit{graph summary}~\cite{LiuSDK18,CebiricGKKMTZ19,SpahiuPPRM16a}). 

\begin{figure}
	\begin{tikzpicture}
	\setlength{\vgap}{0.9cm}
	\setlength{\hgap}{1.2cm}
	
	\node[iri,anchor=center] (nId) {\begin{tabular}{c}EID15\\EID16\end{tabular}};
	
	\node[iri,anchor=center,above=\vgap of nId] (n) {\begin{tabular}{c}Ñam\\Food Truck\end{tabular}}
	edge[arrin] node[lab] {name} (nId);
	
	\node[iri,anchor=center,left=\hgap of n] (ff) {\begin{tabular}{c}Food Festival\\Open Market\\Drinks Festival\end{tabular}}
	edge[arrin] node[lab] {type} (nId);
	
	\node[iri,anchor=center,right=\hgap of n] (sl) {\begin{tabular}{c}Santa Lucía\\Sotomayor\\Piscina Olímpica\end{tabular}}
	edge[arrin] node[lab] {venue} (nId);  
	
	\node[iri,anchor=center,right=\hgap of sl] (san) {\begin{tabular}{c}Santiago\\Viña de Mar\\Arica\end{tabular}}
	edge[arrin] node[lab] {city} (sl)
	edge[arrin,out=340,in=310,looseness=6] node[lab,xshift=0.3cm] {flight} (san)
	edge[arrin,out=200,in=230,looseness=6] node[lab,xshift=-0.3cm] {bus} (san);  
	
	\node[iri,anchor=center] (fS) at (ff|-nId) {\begin{tabular}{c}2020-03-22 12:00\\2020-03-29 20:00\end{tabular}}
	edge[arrin,bend left=8] node[lab] {start} (nId)
	edge[arrin,bend right=8] node[lab] {end} (nId);  
	\end{tikzpicture}
	\caption{Example quotient graph simulating the data graph in Figure~\ref{fig:delg}\label{fig:emergentSchema}}
\end{figure}
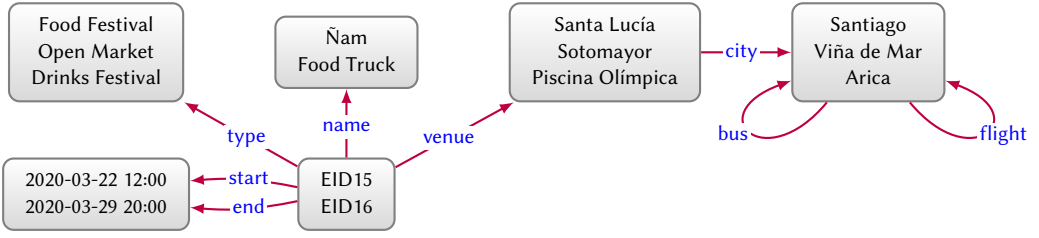

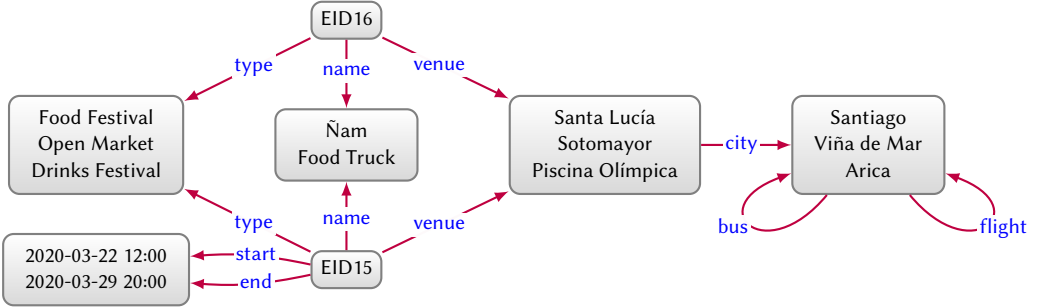
\begin{figure}
	\begin{tikzpicture}
	\setlength{\vgap}{0.9cm}
	\setlength{\hgap}{1.2cm}
	
	\node[iri,anchor=center] (nId) {EID15};
	
	\node[iri,anchor=center,above=\vgap of nId] (n) {\begin{tabular}{c}Ñam\\Food Truck\end{tabular}}
	edge[arrin] node[lab] {name} (nId);
	
	\node[iri,anchor=center,above=\vgap of n] (nId2) {EID16}
	edge[arrout] node[lab] {name} (n);
	
	\node[iri,anchor=center,left=\hgap of n] (ff) {\begin{tabular}{c}Food Festival\\Open Market\\Drinks Festival\end{tabular}}
	edge[arrin] node[lab] {type} (nId)
	edge[arrin] node[lab] {type} (nId2);
	
	\node[iri,anchor=center,right=\hgap of n] (sl) {\begin{tabular}{c}Santa Lucía\\Sotomayor\\Piscina Olímpica\end{tabular}}
	edge[arrin] node[lab] {venue} (nId)
	edge[arrin] node[lab] {venue} (nId2);  
	
	\node[iri,anchor=center,right=\hgap of sl] (san) {\begin{tabular}{c}Santiago\\Viña de Mar\\Arica\end{tabular}}
	edge[arrin] node[lab] {city} (sl)
	edge[arrin,out=340,in=310,looseness=6] node[lab,xshift=0.3cm] {flight} (san)
	edge[arrin,out=200,in=230,looseness=6] node[lab,xshift=-0.3cm] {bus} (san);  
	
	\node[iri,anchor=center] (fS) at (ff|-nId) {\begin{tabular}{c}2020-03-22 12:00\\2020-03-29 20:00\end{tabular}}
	edge[arrin,bend left=8] node[lab] {start} (nId)
	edge[arrin,bend right=8] node[lab] {end} (nId);  
	\end{tikzpicture}
	\caption{Example quotient graph bisimilar with the data graph in Figure~\ref{fig:delg}\label{fig:emergentSchema2}}
\end{figure}

A framework often used for defining emergent schema is that of \textit{quotient graphs}, which partition groups of nodes in the data graph according to some equivalence relation while preserving some structural properties of the graph. Taking Figure~\ref{fig:delg}, we can intuitively distinguish different \textit{types} of nodes based on their context, such as event nodes, which link to venue nodes, which in turn link to city nodes, and so forth. In order to describe the structure of the graph, we could consider six partitions of nodes: \textit{event}, \textit{name}, \textit{venue}, \textit{class}, \textit{date-time}, \textit{city}. In practice, these partitions may be computed based on the class or shape of the node. Merging the nodes of each partition into one node while preserving edges leads to the quotient graph shown in Figure~\ref{fig:emergentSchema}: the nodes of this quotient graph are the partitions of nodes from the data graph and the edge \gedge[arrin][0.7cm]{$X$}{$y$}{$Z$} is in the quotient graph if and only if there exists $x \in X$ and $z \in Z$ such that \gedge[arrin][0.7cm]{$x$}{$y$}{$z$} is in the data graph.

There are many ways in which quotient graphs may be defined, depending not only on how nodes are partitioned, but also how the edges are defined. Different quotient graphs may provide different guarantees with respect to the structure they preserve. Formally, we can say that every quotient graph \textit{simulates} its input graph (based on the \textit{simulation relation} of set membership between data nodes and quotient nodes), meaning that for all $x \in X$ with $x$ an input node and $X$ a quotient node, if \gedge[arrin][0.7cm]{$x$}{$y$}{$z$} is an edge in the data graph, then there must exist an edge \gedge[arrin][0.7cm]{$X$}{$y$}{$Z$} in the quotient graph such that $z \in Z$; for example, the quotient graph of Figure~\ref{fig:emergentSchema} simulates the data graph of Figure~\ref{fig:delg}. However, this quotient graph seems to suggest (for instance) that \gnode{EID16} would have a start and end date in the data graph when this is not the case.  
A stronger notion of structural preservation is given by \textit{bisimilarity}, which in this case would further require that if \gedge[arrin][0.7cm]{$X$}{$y$}{$Z$} is an edge in the quotient graph, then for all $x \in X$, there must exist a $z \in Z$ such that \gedge[arrin][0.7cm]{$x$}{$y$}{$z$} is in the data graph; this is not satisfied by \gnode{EID16} in the quotient graph of Figure~\ref{fig:emergentSchema}, which does not have an outgoing edge labelled \gelab{start} or \gelab{end} in the original data graph. Figure~\ref{fig:emergentSchema2} illustrates a bisimilar version of the quotient graph, splitting the \textit{event} partition into two nodes reflecting their different outgoing edges.  
An interesting property of bisimilarity is that it preserves forward-directed paths: given a path expression $r$ without inverses and two bisimilar graphs, $r$ will match a path in one graph if and only if it matches a corresponding path in the other bisimilar graph. One can verify, for example, that a path matches \gedge[arrin][2.4cm]{$x$}{city$\cdot$(flight|bus)*}{$z$} in Figure~\ref{fig:delg} if and only if there is a path matching \gedge[arrin][2.4cm]{$X$}{city$\cdot$(flight|bus)*}{$Z$} in Figure~\ref{fig:emergentSchema2} such that $x \in X$ and $z \in Z$.

There are many ways in which quotient graphs may be defined, depending on the equivalence relation that partitions nodes. Furthermore, there are many ways in which other similar or bisimilar graphs can be defined, depending on the (bi)simulation relation that preserves the data graph's structure~\cite{CebiricGKKMTZ19}. We provide formal definitions for the notions of \textit{quotient graphs}, \textit{simulation} and \textit{bisimulation} in Appendix~\ref{sec:feschema}. Such techniques aim to \textit{summarise} the data graph into a higher-level topology. In order to reduce the memory overhead of the quotient graph, in practice, nodes may rather be labelled with the cardinality of the partition and/or a high-level label (e.g., \textit{event}, \textit{city}) for the partition rather than storing the labels of all nodes in the partition.

Various other forms of emergent schema not based on a quotient graph framework have also been proposed; examples include emergent schemata based on relational tables~\cite{PhamPEB15}, formal concept analysis~\cite{GonzalezH18}, and so forth. Emergent schemata may be used to provide a human-understandable overview of the data graph, to aid with the definition of a semantic or validating schema, to optimise the indexing and querying of the graph, to guide the integration of data graphs, and so forth. We refer to the survey by ~\citet{CebiricGKKMTZ19} for further details.

\subsection{Identity}
\label{sec:identity}

In Figure~\ref{fig:delg}, we use nodes like \gnode{Santiago}, but to which Santiago does this node refer? Do we refer to Santiago de Chile, Santiago de Cuba, Santiago de Compostela, or do we perhaps refer to the indie rock band Santiago? Based on edges such as \gedge{Santa Lucía}{city}{Santiago}, we may deduce that it is one of the three cities mentioned (not the rock band), and based on the fact that the graph describes tourist attractions in Chile, we may further deduce that it refers to Santiago de Chile. Without further details, however, \textit{disambiguating} nodes of this form may rely on heuristics prone to error in more difficult cases. To help avoid such ambiguity, first we may use globally-unique identifiers to avoid naming clashes when the knowledge graph is extended with external data, and second we may add external identity links to disambiguate a node with respect to an external source. 

\subsubsection{Persistent identifiers}
\label{subsec:globalIdentifiers}
Assume we wished to compare tourism in Chile and Cuba, and we have acquired an appropriate knowledge graph for Cuba. Part of the benefit of using graphs to model data is that we can merge two graphs by taking their union. However, as shown in Figure~\ref{fig:globalIds}, using an ambiguous node like \gnode{Santiago} may result in a \textit{naming clash}: the node is referring to two different real-world cities in both graphs, where the merged graph indicates that Santiago is a city in both Chile and Cuba (rather than two different cities).\footnote{Such a naming clash is not unique to graphs, but could also occur if merging tables, trees, etc.} To avoid such clashes, long-lasting \textit{persistent identifiers} (\textit{PIDs})~\cite{pids} can be created in order to uniquely identify an entity. Prominent examples of PID schemes include \textit{Digital Object Identifiers} (\textit{DOIs}) for papers, \textit{ORCID iDs} for authors, \textit{International Standard Book Numbers} (\textit{ISBNs}) for books, \textit{Alpha-2 codes} for counties, and more besides.

In the context of the Semantic Web, the RDF data model goes one step further and recommends that global Web identifiers be used for nodes and edge labels. However, rather than adopt the \textit{Uniform Resource Locators (URLs)} used to identify the location of \textit{information resources} such as webpages, RDF 1.1 proposes to use \textit{Internationalised Resource Identifiers (IRIs)} to identify \textit{non-information resources} such as cities or events.\footnote{Uniform Resource Identifiers (URIs) can be Uniform Resource Locators (URLs), used to locate information resources, and Uniform Resource Names (URNs), used to name non-information resources. Internationalised Resource Identifiers (IRIs) are URIs that allow Unicode. For example, \texttt{http://example.com/\~Nam} is an IRI, but not a URI, due to the use of ``\~N''. Percentage encoding -- \texttt{http://example.com/\%C3\%91am} -- can encode an IRI as a URI (but reduces readability).} Hence, for example, in the RDF representation of the Wikidata~\cite{VrandecicK14} -- a knowledge graph proposed to complement Wikipedia, discussed in more detail in Section~\ref{sec:kgs} -- while the URL \gnode{\url{https://www.wikidata.org/wiki/Q2887}} refers to a webpage that can be loaded in a browser providing human-readable meta-data about Santiago, the IRI \gnode{\url{http://www.wikidata.org/entity/Q2887}} refers to the city itself. Distinguishing the identifiers for both resources (the webpage and the city itself) avoids naming clashes; for example, if we use the URL to identify both the webpage and the city, we may end up with an edge in our graph, such as (with readable labels below the edge): 

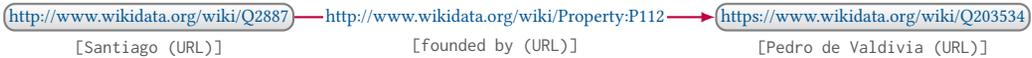
\begin{center}
\begin{tikzpicture}[baseline=-3pt]
    \setlength{\hgap}{5.6cm}

	\node[iri,compact,anchor=center] (s) {\url{http://www.wikidata.org/wiki/Q2887}};
	
	\node[iri,compact,anchor=center,right=\hgap of s] (o) {\url{https://www.wikidata.org/wiki/Q203534}}
	edge[arrin] node[lab,font=\scriptsize,xshift=1ex] (p) {\url{http://www.wikidata.org/wiki/Property:P112}} (s);  
	
	\node[below=0cm of s,font=\ttfamily\scriptsize,color=gray!70!black] {[Santiago (URL)]};
	\node[below=0cm of p,font=\ttfamily\scriptsize,color=gray!70!black] {[founded by (URL)]};
	\node[below=0cm of o,font=\ttfamily\scriptsize,color=gray!70!black] {[Pedro de Valdivia (URL)]};
\end{tikzpicture}
\end{center}

\noindent  Such an edge leaves ambiguity: was Pedro de Valdivia the founder of the webpage, or the city? Using IRIs for entities distinct from the URLs for the webpages that describe them avoids such ambiguous cases, where Wikidata thus rather defines the previous edge as follows:

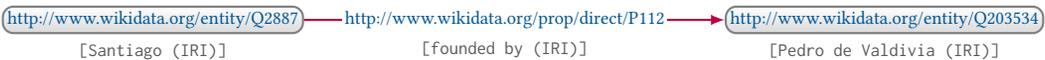
\begin{center}
\begin{tikzpicture}[baseline=-3pt]
    \setlength{\hgap}{5.6cm}

	\node[iri,compact,anchor=center] (s) {\url{http://www.wikidata.org/entity/Q2887}};
	
	\node[iri,compact,anchor=center,right=\hgap of s] (o) {\url{http://www.wikidata.org/entity/Q203534}}
	edge[arrin] node[lab,font=\scriptsize,xshift=1ex] (p) {\url{http://www.wikidata.org/prop/direct/P112}} (s); 
	
	\node[below=0cm of s,font=\ttfamily\scriptsize,color=gray!70!black] {[Santiago (IRI)]};
	\node[below=0cm of p,font=\ttfamily\scriptsize,color=gray!70!black] {[founded by (IRI)]};
	\node[below=0cm of o,font=\ttfamily\scriptsize,color=gray!70!black] {[Pedro de Valdivia (IRI)]};	 
\end{tikzpicture}
\end{center}

\noindent using IRIs for the city, person, and founder of, distinct from the webpages describing them. These Wikidata identifiers use the prefix \url{http://www.wikidata.org/entity/} for entities and the prefix \url{http://www.wikidata.org/prop/direct/} for relations. Such prefixes are known as \textit{namespaces}, and are often abbreviated with prefix strings, such as \texttt{wd:} or \texttt{wdt:}, where the latter triple can then be written more concisely using such abbreviations as \gedge[arrin][1.6cm]{wd:Q2887}{wdt:P112}{wd:Q203534}.

If HTTP IRIs are used to identify the graph's entities, when the IRI is looked up (via HTTP), the web-server can return (or redirect to) a description of that entity in formats such as RDF. This further enables RDF graphs to link to related entities described in external RDF graphs over the Web, giving rise to \textit{Linked Data}~\cite{ldprinciples,ldbook} (discussed in Section~\ref{sec:publish}). Though HTTP IRIs offer a flexible and powerful mechanism for issuing global identifiers on the Web, they are not necessarily persistent: websites may go offline, the resources described at a given location may change, etc. In order to enhance the persistence of such identifiers, \textit{Persistent URL} (\textit{PURL}) services offer redirects from a central server to a particular location, where the PURL can be redirected to a new location if necessary, changing the address of a document without changing its identifier. The persistence of HTTP IRIs can then be improved by using namespaces defined through PURL services.

\begin{figure}
	\setlength{\vgap}{0.8cm}
	\setlength{\hgap}{1.8cm}
	
	\colorlet{ng}{black!2}
	
	\tikzset{lab/.append style={fill=ng}}
	
	\begin{tikzpicture}
	\node[block, minimum width=3.2cm,minimum height=4cm, fill=ng, anchor=south west] (chile) at (0,0)  {};
	
	\node[above=0.1cm of chile.south west,anchor=south west,xshift=0.1cm] (chileL) {\texttt{Chile}};  
	
	\node[iri,anchor=mid,below=0.2cm of chile.north] (sl) {Santa Lucía};  
	
	\node[iri,anchor=center,below=\vgap of sl] (san1) {Santiago}
	edge[arrin] node[lab] {city} (sl);  
	
	\node[iri,anchor=center,below=\vgap of san1,xshift=-0.4\hgap] (ch) {Chile}
	edge[arrin] node[lab] {country} (san1);
	
	\node[iri,anchor=center,below=\vgap of san1,xshift=0.4\hgap] (elip1) {...}
	edge[arrin] node[lab] {...} (san1);

	\node[block, minimum width=3.2cm,minimum height=4cm, fill=ng, right=0.3cm of chile.south east, anchor=south west] (cuba) {};
	
	\node[above=0.1cm of cuba.south west,anchor=south west,xshift=0.1cm] (cubaL) {\texttt{Cuba}};  
	
	\node[iri,anchor=mid,below=0.2cm of cuba.north] (si) {Santa Ifigenia};  
	
	\node[iri,anchor=center,below=\vgap of si] (san2) {Santiago}
	edge[arrin] node[lab] {city} (si);  
	
	\node[iri,anchor=center,below=\vgap of san2,xshift=-0.4\hgap] (cu) {Cuba}
	edge[arrin] node[lab] {country} (san2);
	
	\node[iri,anchor=center,below=\vgap of san2,xshift=0.4\hgap] (elip2) {...}
	edge[arrin] node[lab] {...} (san2); 
	
	\tikzset{lab/.append style={fill=white}}
	
	\node[block, dashed, minimum width=5cm,minimum height=4cm, right=0.3cm of cuba.south east, anchor=south west] (chileCuba) {};
	
	\node[above=0.1cm of chileCuba.south west,anchor=south west,xshift=0.1cm] (chileL) {$\texttt{Chile} \cup \texttt{Cuba}$};
	
	\node[iri] (san3) at (san2.mid-|chileCuba.north) {Santiago};
	
	\node[iri,anchor=mid,above=\vgap of san3,xshift=-0.7\hgap] (sl2) {Santa Lucía}
	edge[arrout] node[lab] {city} (san3);
	
	\node[iri,anchor=mid,above=\vgap of san3,xshift=0.7\hgap] (si2) {Santa Ifigenia}
	edge[arrout] node[lab] {city} (san3);  
	
	\node[iri,anchor=mid,below=\vgap of san3,xshift=-\hgap] (ch2) {Chile}
	edge[arrin] node[lab] {country} (san3);
	
	\node[iri,anchor=mid,below=\vgap of san3,xshift=\hgap] (cu2) {Cuba}
	edge[arrin] node[lab] {country} (san3);
	
	\node[iri,anchor=mid,below=\vgap of san3] (elip3) {...}
	edge[arrin] node[lab] {...} (san3);
	\end{tikzpicture}
	
	\caption{Result of merging two graphs with ambiguous local identifiers \label{fig:globalIds}}
\end{figure}
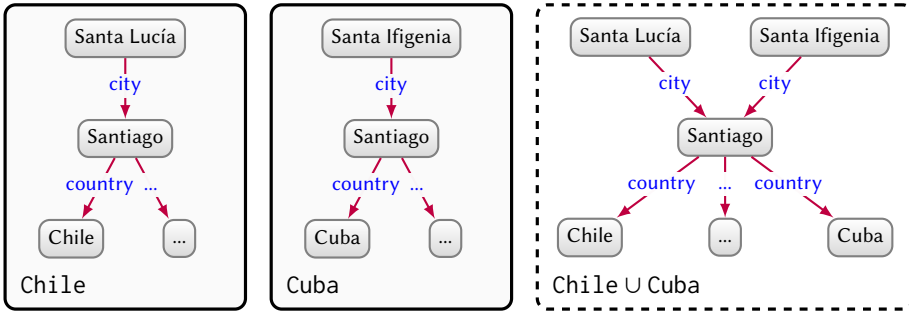

\subsubsection{External identity links}\label{sssec:external_identy} Assume that the tourist board opts to define the \texttt{chile:} namespace with an IRI such as \texttt{http://turismo.cl/entity/} on a web-server that they control, allowing nodes such as \gnode{chile:Santiago} -- a shortcut for the IRI \gnode{http://turismo.cl/entity/Santiago} -- to be looked up over the Web. While using such a naming scheme helps to avoid naming clashes, the use of IRIs does not necessarily help ground the identity of a resource. For example, an external geographic knowledge graph may assign the same city the IRI \gnode{geo:SantiagoDeChile} in their own namespace, where we have no direct way of knowing that the two identifiers refer to the same city. If we merge the two knowledge graphs, we will end up with two distinct nodes for the same city.

There are a number of ways to ground the identity of an entity. The first is to associate the entity with uniquely-identifying information in the graph, such as its geo-coordinates, its postal code, the year it was founded, etc. Each additional piece of information removes ambiguity as to which city is being referred, providing (for example) more options for matching the city with its analogue in external sources. A second option is to use \textit{identity links} to state that a local entity has the same identity as another \textit{coreferent} entity found in an external source; an instantiation of this concept can be found in the OWL standard, which defines the \texttt{owl:sameAs} property relating coreferent entities. Using this property, we could state the edge \gedge[arrin][1.8cm]{chile:Santiago}{owl:sameAs}{geo:SantiagoDeChile} in our RDF graph, thus establishing an identity link between the corresponding nodes in both graphs. The semantics of \texttt{owl:sameAs} defined by the OWL standard then allow us to combine the data for both nodes. Such semantics will be discussed later in Section~\ref{sec:deductive}. Ways in which identity links can be computed will also be discussed later in Section~\ref{sec:refine}.

\subsubsection{Datatypes}

Consider the two date-times on the left of Figure~\ref{fig:delg}: how should we assign these nodes persistent/global identifiers? Intuitively it would not make sense, for example, to assign IRIs to these nodes since their syntactic form tells us what they refer to: specific dates and times in March 2020. This syntactic form is further recognisable by machine, meaning that with appropriate software, we could order such values in ascending or descending order, extract the year, etc.

Most practical data models for graphs allow for defining nodes that are datatype values. RDF utilises \textit{XML Schema Datatypes} (\textit{XSD})~\cite{XSD}, amongst others, where a datatype node is given as a pair $(l,d)$ where $l$ is a lexical string, such as "\texttt{2020-03-29T20:00:00}", and $d$ is an IRI denoting the datatype, such as \texttt{xsd:dateTime}. The node is then denoted \gnode{"\texttt{2020-03-29T20:00:00}"\dtsep xsd:dateTime}. Datatype nodes in RDF are called \textit{literals} and are not allowed to have outgoing edges. Other datatypes commonly used in RDF data include \texttt{xsd:string}, \texttt{xsd:integer}, \texttt{xsd:decimal}, \texttt{xsd:boolean}, etc. In case that the datatype is omitted, the value is assumed to be of type \texttt{xsd:string}. Applications built on top of RDF can then recognise these datatypes, parse them into datatype objects, and apply equality checks, normalisation, ordering, transformations, casting, according to their standard definition. In the context of property graphs, Neo4j~\cite{Miller13} also defines a set of internal datatypes on property values that includes numbers, strings, booleans, spatial points, and temporal values.

\subsubsection{Lexicalisation} Global identifiers for entities will sometimes have a human-interpretable form, such as \gnode{chile:Santiago}, but the identifier strings themselves do not carry any formal semantic significance. In other cases, the identifiers used may not be human-interpretable by design. In Wikidata, for instance, Santiago de Chile is identified as \gnode{wd:Q2887}, where such a scheme has the advantage of providing better persistence and of not being biased to a particular human language. For example, the Wikidata identifier for Eswatini (\gnode{wd:Q1050}) was not affected when the country changed its name from Swaziland, and does not necessitate choosing between languages for creating (more readable) IRIs such as \gnode{wd:Eswatini} (English), \gnode{wd:eSwatini} (Swazi), \gnode{wd:Esuatini} (Spanish), etc. 

Since identifiers can be arbitrary, it is common to add edges that provide a human-interpretable label for nodes, such as \gedge[arrin][1.5cm]{wd:Q2887}{rdfs:label}{``Santiago''}, indicating how people may refer to the subject node linguistically. Linguistic information of this form plays an important role in grounding knowledge such that users can more clearly identify which real-world entity a particular node in a knowledge graph actually references~\cite{Lexvo}; it further permits cross-referencing entity labels with text corpora to find, for example, documents that potentially speak of a given entity~\cite{IESW}. Labels can be complemented with aliases (e.g., \gedge[arrin][2cm]{wd:Q2887}{skos:altLabel}{``Santiago de Chile''}) or comments (e.g. \gedge[arrin][2cm]{wd:Q2887}{rdfs:comment}{``Santiago is the capital of Chile''}) to further help ground the node's identity. 

Nodes such as \gnode{``Santiago''} denote string literals, rather than an identifier. Depending on the specific graph model, such literal nodes may also be defined as a pair $(s,l)$, where $s$ denotes the string and $l$ a language code; in RDF, for example we may state \gedge[arrin][1.5cm]{chile:City}{rdfs:label}{"City"@en}, \gedge[arrin][1.5cm]{chile:City}{rdfs:label}{"Ciudad"@es}, etc., indicating labels for the node in different languages. In other models, the pertinent language can rather be specified, e.g., via metadata on the edge. Knowledge graphs with human-interpretable labels, aliases, comments, etc., (in various languages) are sometimes called (\textit{multilingual}) \textit{lexicalised knowledge graphs}~\cite{BonattiDPP18}.

\subsubsection{Existential nodes} When modelling incomplete information, we may in some cases know that there must exist a particular node in the graph with particular relationships to other nodes, but without being able to identify the node in question.  For example, we may have two co-located events \gnode{chile:EID42} and \gnode{chile:EID43} whose venue has yet to be announced. One option is to simply omit the venue edges, in which case we lose the information that these events have a venue and that both events have the same venue. Another option might be to create a fresh IRI representing the venue, but semantically this becomes indistinguishable from there being a known venue. Hence some graph models permit the use of existential nodes, represented here as a blank circle:

\begin{center}
	\setlength{\vgap}{1cm}
	\setlength{\hgap}{2cm}
	
	\begin{tikzpicture}
	\node[iri,anchor=center] (e42) {chile:EID42};
	
	\node[bnode,anchor=center,right=\hgap of e42] (v) {}
	edge[arrin] node[lab] {chile:venue} (e42);
	
	\node[iri,anchor=center,right=\hgap of v] (e43) {chile:EID43}
	edge[arrout] node[lab] {chile:venue} (v);
	\end{tikzpicture}
\end{center}

\begin{figure}
\centering
	\setlength{\vgap}{1cm}
	\setlength{\hgap}{2.7cm}

\begin{tikzpicture}
\node[iri,anchor=center] (c) {chile:Chile};

\node[bnode,anchor=center,right=0.7\hgap of c] (l1) {}
edge[arrin] node[lab] {chile:peaks} (c);

\node[iri,anchor=center,below=\vgap of l1] (l1e) {chile:OjosDelSalado}
edge[arrin] node[lab] {rdf:first} (l1);

\node[bnode,anchor=center,right=\hgap of l1] (l2) {}
edge[arrin] node[lab] {rdf:rest} (l1);

\node[iri,anchor=center,below=\vgap of l2] (l2e) {chile:NevadoTresCruces}
edge[arrin] node[lab] {rdf:first} (l2);

\node[bnode,anchor=center,right=\hgap of l2] (l3) {}
edge[arrin] node[lab] {rdf:rest} (l2);

\node[iri,anchor=center,below=\vgap of l3] (l3e) {chile:Llullaillaco}
edge[arrin] node[lab] {rdf:first} (l3);

\node[iri,anchor=center,right=0.7\hgap of l3] (nil) {rdf:nil}
edge[arrin] node[lab] {rdf:rest} (l3);
\end{tikzpicture}

\caption{RDF list representing the three largest peaks of Chile, in order \label{fig:list}}
\end{figure}
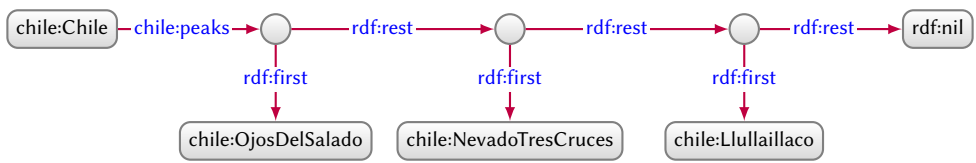

\noindent These edges denote that there exists a common venue for \gnode{chile:EID42} and \gnode{chile:EID42} without identifying it. Existential nodes are supported in RDF as blank nodes~\cite{rdf11}, which are also commonly used to support modelling complex elements in graphs, such as \textit{RDF lists}~\cite{rdf11,HoganAMP14}. Figure~\ref{fig:list} exemplifies an RDF list, which uses blank nodes in a linked-list structure to encode order. Though existential nodes can be convenient, their presence can complicate operations on graphs, such as deciding if two data graphs have the same structure modulo existential nodes~\cite{rdf11,Hogan17}. Hence methods for \textit{skolemising} existential nodes in graphs -- replacing them with canonical labels -- have been proposed~\cite{canon,Hogan17}. Other authors rather call to minimise the use of such nodes in graph data~\cite{ldbook}.

\subsection{Context}\label{ssec:knowledgeContext}

Many (arguably \textit{all}) facts presented in the data graph of Figure~\ref{fig:delg} can be considered true with respect to a certain \textit{context}. With respect to \textit{temporal context}, \gnode{Santiago} has only existed as a city since 1541, flights from \gnode{Arica} to \gnode{Santiago} began in 1956, etc. With respect to \textit{geographic context}, the graph describes events in Chile. With respect to \textit{provenance}, data relating to \gnode{EID15} were taken from -- and are thus said to be true with respect to -- the Ñam webpage on January 4\textsuperscript{th}, 2020. Other forms of context may also be used. We may further combine contexts, such as to indicate that \gnode{Arica} is a Chilean city (\textit{geographic}) since 1883 (\textit{temporal}) according to the Treaty of Ancón (\textit{provenance}).

By context we herein refer to the \textit{scope of truth}, and thus talk about the context in which some data are held to be true~\cite{McCarthy93,GuhaMF04}. The graph of Figure~\ref{fig:delg} leaves much of its context implicit. However, making context explicit can allow for interpreting the data from different perspectives, such as to understand what held true in 2016, what holds true excluding webpages later found to have spurious data, etc. As seen in the previous examples, context for graph data may be considered at different levels: on individual nodes, individual edges, or sets of edges (sub-graphs). We now discuss various representations by which context can be made explicit at different levels.

\subsubsection{Direct representation}

The first way to represent context is to consider it as data no different from other data. For example, the dates for the event \gnode{EID15} in Figure~\ref{fig:delg} can be seen as representing a form of temporal context, indicating the temporal scope within which edges such as \gedge{EID15}{venue}{Santa Lucía} are held true. Another option is to change a relation represented as an edge, such as \gedge{Santiago}{flight}{Arica}, into a node, such as seen in Figure~\ref{fig:fsa}, allowing to assign additional context to the relation. While in these examples context is represented in an ad hoc manner, a number of specifications have been proposed to represent context as data in a more standard way. One example is the \textit{Time Ontology}~\cite{timeOnt}, which specifies how temporal entities, intervals, time instants, etc. -- and relations between them such as before, overlaps, etc. -- can be described in RDF graphs in an interoperable manner. Another example is the \textit{PROV Data Model}~\cite{prov13}, which specifies how provenance can be described in RDF graphs, where entities (e.g., graphs, nodes, physical document) are derived from other entities, are generated and/or used by activities (e.g., extraction, authorship), and are attributed to agents (e.g., people, software, organisations).

\subsubsection{Reification}\label{sec:reify} 
\begin{figure}
	\setlength{\vgap}{1cm}
	%
	\begin{subfigure}[b]{.36\textwidth}
		\setlength{\hgap}{2.5cm}
		\centering
		\begin{tikzpicture}		
		\node[iri,anchor=center] (santiago) {Santiago}; 
		
		\node[iri,anchor=center,right=\hgap of santiago] (arica) {Arica};
		
		\node[iri,anchor=center,between=santiago and arica] (flight) {flight};
		
		\node[iri,anchor=center,below=\vgap of santiago] (e) {$e$}
		edge[arrout] node[lab,yshift=0.1cm] {subject} (santiago)
		edge[arrout,bend right=12] node[lab] {predicate} (flight)
		edge[arrout,bend right=18] node[lab] {object} (arica);
		
		\node[iri,anchor=center] (year) at (e-|arica) {1956}
		edge[arrin] node[lab,xshift=0.5cm] {valid from} (e); 
		\end{tikzpicture}
		\caption{RDF Reification}
		\label{fig:reif}
	\end{subfigure}
	%
	\begin{subfigure}[b]{.31\textwidth}
		\setlength{\hgap}{2cm}
		\centering
		\begin{tikzpicture}		
		\node[iri,anchor=center] (santiago) {Santiago}; 
		
		\node[iri,anchor=center,right=\hgap of santiago] (e) {$e$}
		edge[arrin] node[lab] {flight} (santiago);
		
		\node[iri,anchor=center,below=\vgap of santiago] (year) {1956}
		edge[arrin] node[lab] {valid from} (e); 
		
		\node[iri,anchor=center] (arica) at (year-|e) {Arica}
		edge[arrin] node[lab] {value} (e);
		\end{tikzpicture}
		\caption{$n$-ary Relations}
		\label{fig:nary}
	\end{subfigure}
	%
	\begin{subfigure}[b]{.31\textwidth}
		\setlength{\hgap}{2cm}
		\centering
		\begin{tikzpicture}		
		\node[iri,anchor=center] (santiago) {Santiago}; 
		
		\node[iri,anchor=center,right=\hgap of santiago] (arica) {Arica}
		edge[arrin] node[lab,draw,dotted] (e) {$e$} (santiago);
		
		\node[iri,anchor=center,below=\vgap of santiago] (year) {1956}
		edge[arrin] node[lab,xshift=0.2cm] {valid from} (e); 
		
		\node[iri,anchor=center] (flight) at (year-|arica) {flight}
		edge[arrin] node[lab,xshift=0.2cm] {singleton} (e);
		\end{tikzpicture}
		\caption{Singleton properties}
		\label{fig:singprop}
	\end{subfigure}
	\caption{Three representations of temporal context on an edge in a directed-edge labelled graph \label{fig:temporal}}
\end{figure}
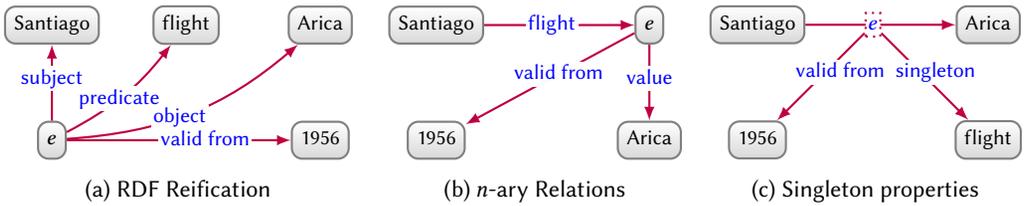

Often we may wish to directly define the context of edges themselves; for example, we may wish to state that the edge \gedge{Santiago}{flight}{Arica} is valid from 1956. While we could use the pattern of turning the edge into a node -- as illustrated in Figure~\ref{fig:fsa} -- to directly represent such context, another option is to use \textit{reification}, which allows for making statements about statements in a generic manner (or in the case of a graph, for defining edges about edges). In Figure~\ref{fig:temporal} we present three forms of reification that can be used for modelling temporal context on the aforementioned edge within a directed edge-labelled graph~\cite{HernandezHK15}. We use $e$ to denote an arbitrary identifier representing the edge itself to which the contextual information can be associated. Unlike in a direct representation, $e$ represents an edge, not a flight. RDF reification~\cite{rdf11} (Figure~\ref{fig:reif}) defines a new node \gnode{$e$} to represent the edge and connects it to the source node (via \gelab{subject}), target node (via \gelab{object}), and edge label (via \gelab{predicate}) of the edge. In contrast, $n$-ary relations~\cite{rdf11} (Figure~\ref{fig:nary}) connect the source node of the edge directly to the edge node \gnode{$e$}  with the label of the edge; the target node of the edge is then connected to \gnode{$e$} (via \gelab{value}). Finally, singleton properties~\cite{Nguyen14} (Figure~\ref{fig:singprop}) rather use \gelab{$e$} as an edge label, connecting it to a node indicating the original edge label (via \gelab{singleton}). Other forms of reification have been proposed in the literature, including, for example, NdFluents~\cite{Gimenez-GarciaZ17}. In general, a reified edge does not assert the edge it reifies; for example, we may reify an edge to state that it is no longer valid. We refer to the work of~\citeT{HernandezHK15} for further comparison of reification alternatives and their relative strengths and weaknesses.
 
\subsubsection{Higher-arity representation} 
\begin{figure}
	\begin{subfigure}[b]{.31\textwidth}
		\setlength{\hgap}{1.6cm}
		\setlength{\vgap}{0.7cm}
		\centering
		\begin{tikzpicture}		
		\node[draw,thick,rounded corners,anchor=center,minimum width=2.45\hgap,minimum height=1.2cm] (e) {};
		
		\node[iri,right=0.1cm of e.north west,yshift=-0.1cm,anchor=north west] (santiago) {Santiago}; 
		
		\node[iri,anchor=center,right=\hgap of santiago] (arica) {Arica};
		
		\node[lab,anchor=center,between=santiago and arica,arrout,opacity=0] (flight) {flight}
		edge[arrin,-] (santiago)
		edge[arrout] (arica);
		
		\node[draw,inner sep=0.4ex,dashed,thick,font=\sf\footnotesize\hsp,anchor=south,above=0.1cm of e.south east,xshift=-0.4cm] (saflight) {$e$};
		
		\node[iri,anchor=center,below=1.1\vgap of saflight] (year) {1956}
		edge[arrin] node[lab] {valid from} (saflight);
		\end{tikzpicture}
		\caption{Named graph}
		\label{fig:ngraph}
	\end{subfigure}
	%
	\begin{subfigure}[b]{.33\textwidth}
		\setlength{\hgap}{3cm}
		\centering
		\begin{tikzpicture}
		\node[iri] (ln1) {Santiago};
		
		\node[iri,right=0.7\hgap of ln1] (ln2) {Arica};
		
		\draw[arrout, bend right=30] (ln1) to 
		node[below, erect] (e1) {
			\alt{
				\uri{valid from} & =\,\uri{1956}\\} }
		(ln2);
		%
		\node[rte] (le1) at (e1.north) {\uri{$e$} : \uri{flight}}; 
		\end{tikzpicture}
		\caption{Property graph}
		\label{fig:pgc}
	\end{subfigure}
	\setlength{\vgap}{1cm}
	\begin{subfigure}[b]{.31\textwidth}
		\setlength{\hgap}{1.6cm}
		\centering
		\begin{tikzpicture}		
		\node[iri,anchor=center,minimum width=2.45\hgap,minimum height=0.8cm] (e) {};
		
		\node[iri,anchor=west,right=0.1cm of e.west] (santiago) {Santiago}; 
		
		\node[iri,anchor=center,right=\hgap of santiago] (arica) {Arica};
		
		\node[lab,anchor=center,between=santiago and arica,arrout,opacity=0] (flight) {flight}
		edge[arrin,-] (santiago)
		edge[arrout] (arica);
		
		\node[iri,anchor=center,below=\vgap of e] (year) {1956}
		edge[arrin] node[lab] {valid from} (e);
		\end{tikzpicture}
		\caption{RDF*}
		\label{fig:rdfstar}
	\end{subfigure}
	\caption{Three higher-arity representations of temporal context on an edge \label{fig:temporal2}}
\end{figure}
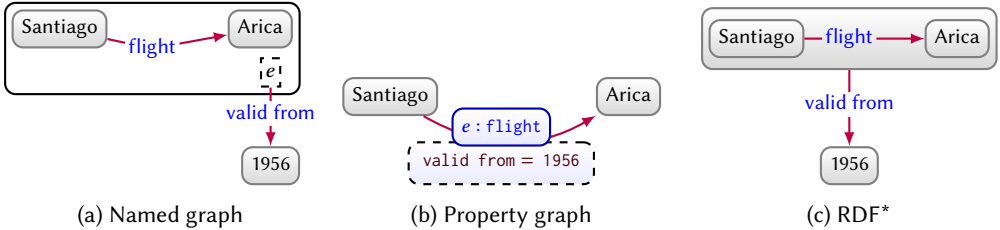

As an alternative to reification, we can rather use higher-arity representations for modelling context. Taking again the edge \gedge{Santiago}{flight}{Arica}, Figure~\ref{fig:temporal2} illustrates three higher-arity representations of temporal context. First, we can use a named graph (Figure~\ref{fig:ngraph}) to contain the edge and then define the temporal context on the graph name. Second, we can use a property graph (Figure~\ref{fig:pgc}) where the temporal context is defined as an attribute on the edge. Third, we can use \textit{RDF*}~\cite{Hartig17} (Figure~\ref{fig:rdfstar}): an extension of RDF that allows edges to be defined as nodes. Amongst these options, the most flexible is the named graph representation, where we can assign context to multiple edges at once by placing them in one named graph; for example, we can add more edges to the named graph of Figure~\ref{fig:ngraph} that are also valid from 1956. The least flexible option is RDF*, which, in the absence of an edge id, does not permit different groups of contextual values to be assigned to an edge; for example, considering the edge \gedge[arrin][1.8cm]{Chile}{president}{M. Bachelet}, if we add four contextual values to this edge to state that it was valid from 2006 until 2010 and valid from 2014 until 2018, we cannot pair the values, but may rather have to create a node to represent different presidencies (in the other models, we could have used two named graphs or edge ids).

\subsubsection{Annotations} Thus far we have discussed representing context in a graph, but we have not spoken about automated mechanisms for reasoning about context; for example, if there are only seasonal summer flights from \gnode{Santiago} to \gnode{Arica}, we may wish to find other routes from Santiago for winter events taking place in \gnode{Arica}. While the dates for buses, flights, etc., can be represented directly in the graph, or using reification, writing a query to manually intersect the corresponding temporal contexts will be tedious -- or may not even be possible at all. Another alternative is to consider \textit{annotations} that provide mathematical definitions of a contextual domain and key operations possible within that domain that can then be applied automatically. 

Some annotations model a particular contextual domain; for example, \textit{Temporal RDF}~\cite{GutierrezHV07} allows for annotating edges with time intervals, such as \gedge[arrin][2cm]{Chile}{\begin{tabular}{@{}c@{}}president\\[-0.5ex]$[2006,2010]$\end{tabular}}{M. Bachelet}, while \textit{Fuzzy RDF}~\cite{Straccia09} allows for annotating edges with a degree of truth such as \gedge[arrin][1.7cm]{Santiago}{\begin{tabular}{@{}c@{}}climate\\[-0.5ex]$0.8$\end{tabular}}{Semi-Arid}, indicating that it is more-or-less true -- with a degree of 0.8 -- that Santiago has a semi-arid climate. 

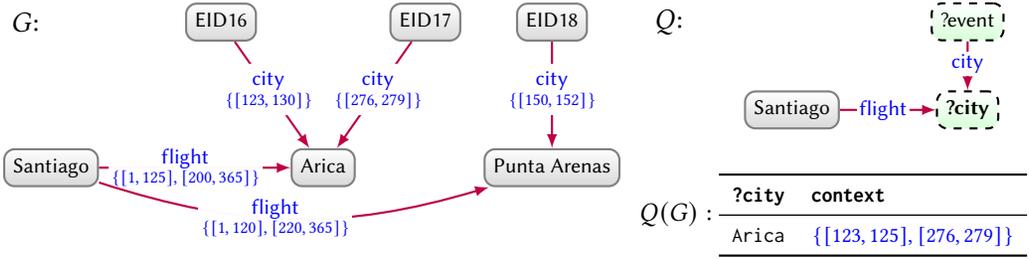
\begin{figure}
	\setlength{\vgap}{1.4cm}
	\setlength{\hgap}{1.7cm}
	\centering
	
	\begin{tikzpicture}
	\node[iri,anchor=center] (s) {Santiago};
	 
	\node[iri,anchor=center,right=1.5\hgap of s] (a) {Arica}
	   edge[arrin] node[lab] {\aelab{flight}{$\{ [1,125],[200,365] \}$}} (s);
	   
	\node[iri,anchor=center,above=\vgap of a,xshift=-0.8\hgap] (e16) {EID16}
		edge[arrout] node[lab] {\aelab{city}{$\{ [123,130] \}$}} (a);
		
	\node[iri,anchor=center,above=\vgap of a,xshift=0.8\hgap] (e17) {EID17}
	edge[arrout] node[lab] {\aelab{city}{$\{ [276,279] \}$}} (a);
	
	\node[iri,anchor=center,right=\hgap of a] (p) {Punta Arenas}
	   edge[arrin,bend left=17] node[lab] {\aelab{flight}{$\{ [1,120],[220,365] \}$}} (s);
	
	\node[iri,anchor=center,above=\vgap of p] (e17) {EID18}
	edge[arrout] node[lab] {\aelab{city}{$\{ [150,152] \}$}} (p);
	
	\node[anchor=west] at (s.west|-e16.west) {\large $G$:};
	\end{tikzpicture}
	\raisebox{2ex}{\begin{tabular}[b]{l}
	\setlength{\vgap}{0.9cm}
	\begin{tikzpicture}
	\node[iri,anchor=center] (s) {Santiago};
	
	\node[var,right=\hgap of s,anchor=center] (c) {\textbf{?city}}
	edge[arrin] node[lab] {flight} (s);
	
	\node[var,above=\vgap of c,anchor=center] (e) {?event}
	edge[arrout] node[lab] {city} (c);  
	
	\node[anchor=west,xshift=-1.3cm] at (s.west|-e.west) {\large $Q$:};
	\end{tikzpicture}
	\\[3ex]
	$Q(G):$~~ 
	\tt\footnotesize %
	\begin{tabular}{ll}
		\toprule
		\textbf{?city} & \textbf{context} \\
		\midrule
		Arica & \color{blue}$\{ [123,125], [276,279] \}$\\
		\bottomrule
	\end{tabular}
	\end{tabular}
	}

	\caption{Example query on a temporally annotated graph \label{fig:time}}
\end{figure}

Other forms of annotation are domain-independent; for example, \textit{Annotated RDF}~\cite{Dividino09,UdreaRS10,zimm-etal-2012-JWS} allows for representing various forms of context modelled as \textit{semi-rings}: algebraic structures consisting of domain values (e.g., temporal intervals, fuzzy values, etc.) and two main operators to combine domain values: \textit{meet} and \textit{join}.\footnote{The \textit{join} operator for annotations is different from the join operator for relational algebra.}  We provide an example in Figure~\ref{fig:time}, where $G$ is annotated with values from a simplified temporal domain consisting of sets of integers (1--365) representing days of the year. For brevity we use an interval notation, where, for example, $\{[150,152]\}$ indicates the set $\{150,151,152\}$. Query $Q$ then asks for flights from Santiago to cities with events; this query will check and return an annotation reflecting the temporal validity of each answer. To derive these answers, we first require applying a conjunction of annotations on compatible \gelab{flight} and \gelab{city} edges, applying the \textit{meet operator} to compute the annotation for which both edges hold. The natural way to define meet in our scenario is as the intersection of sets of days, where, for example, applying meet on the event annotation $\color{blue}\{[150,152]\}$ and the flight annotation $\color{blue}\{[1,120],[220,365]\}$ for \gnode{Punta Arenas} leads to the empty time interval $\color{blue}\{\}$, which may thus lead to the city being filtered from the results (depending on the query evaluation semantics). However, for \gnode{Arica}, we find two different non-empty intersections: $\color{blue}\{[123,125]\}$ for \gnode{EID16} and $\color{blue}\{[276,279]\}$ for \gnode{EID17}. Given that we are interested in the city (a projected variable), rather than the event, we can thus combine these two annotations for \gnode{Arica} using the \textit{join operator}, returning the annotation in which either result holds true. In our scenario, the natural way to define join is as the union of the sets of days, giving $\color{blue}\{[123,125],[276,279]\}$.  We provide formal definitions in Appendix~\ref{sec:annotationDomain} based on the general framework proposed by Zimmermann et al.~\cite{zimm-etal-2012-JWS} for annotations on graphs. 

\subsubsection{Other contextual frameworks} Other frameworks have been proposed for modelling and reasoning about context in graphs. A notable example is that of \textit{contextual knowledge repositories}~\cite{SerafiniH12}, which allow for assigning individual (sub-)graphs to their own context. Unlike in the case of named graphs, context is explicitly modelled along one or more dimensions, where each (sub-)graph must take a value for each dimension. Each dimension is further associated with a partial order over its values -- e.g., \gnode{2020-03-22} $\preceq$ \gnode{2020-03} $\preceq$ \gnode{2020} -- allowing to select and combine sub-graphs that are valid within contexts at different levels of granularity. \citet{SchuetzBNSS20} similarly propose a form of contextual OnLine Analytic Processing (OLAP), based on a data cube formed by dimensions where individual cells contain knowledge graphs. Operations such as ``\textit{slice-and-dice}'' (selecting knowledge according to given dimensions), as well as ``\textit{roll-up}'' (aggregating knowledge at a higher level) can then be supported. We refer the reader to the respective papers for more details~\cite{SerafiniH12,SchuetzBNSS20}.
\section{Deductive Knowledge}\label{sec:deductive}

As humans, we can \textit{deduce} more from the data graph of Figure~\ref{fig:delg} than what the edges explicitly indicate. We may deduce, for example, that the Ñam festival (\gnode{EID15}) will be located in Santiago, even though the graph does not contain an edge \gedge[arrin][1.7cm]{EID15}{location}{Santiago}. We may further deduce that the cities connected by flights must have some airport nearby, even though the graph does not contain nodes referring to these airports. In these cases, given the data as premises, and some general rules about the world that we may know \textit{a priori}, we can use a deductive process to derive new data, allowing us to know more than what is explicitly given by the data. These types of general premises and rules, when shared by many people, form part of ``\textit{commonsense knowledge}''~\cite{Commonsense}; conversely, when rather shared by a few experts in an area, they form part of ``\textit{domain knowledge}'', where, for example, an expert in biology may know that \textit{hemocyanin} is a protein containing copper that carries oxygen in the blood of some species of \textit{Mollusca} and \textit{Arthropoda}. 

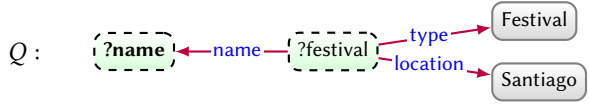
\begin{figure}
	\setlength{\hgap}{1.5cm}
	\setlength{\vgap}{1cm}
		
		$Q$ :\qquad
		\begin{tikzpicture}[baseline=-3pt]
		\node[var,anchor=center] (fv) {?festival};
		
		\node[iri,anchor=center,right=\hgap of fv,yshift=0.4\vgap] (f) {Festival}
		edge[arrin] node[lab] {type} (fv);
		
		\node[iri,anchor=center,right=\hgap of fv,yshift=-0.4\vgap] (s) {Santiago}
		edge[arrin] node[lab] {location} (fv);
		
		\node[var,anchor=center,left=\hgap of fv] (fn) {\textbf{?name}}
		edge[arrin] node[lab] {name} (fv);
		\end{tikzpicture}
		
	\caption{Graph pattern querying for names of festivals in Santiago \label{fig:bgpFS}}
\end{figure}

Machines, in contrast, do not have \textit{a priori} access to such deductive faculties; rather they need to be given formal instructions, in terms of premises and \textit{entailment regimes}, in order to make similar deductions to what a human can make. These entailment regimes formalise the conclusions that logically follow as a consequence of a given set of premises. Once instructed in this manner, machines can (often) apply deductions with a precision, efficiency, and scale beyond human performance. These deductions may serve a range of applications, such as improving query answering, (deductive) classification, finding inconsistencies, etc. As a concrete example involving query answering, assume we are interested in knowing \textit{the festivals located in Santiago}; we may straightforwardly express such a query as per the graph pattern shown in Figure~\ref{fig:bgpFS}. This query returns no results for the graph in Figure~\ref{fig:delg}: there is no node named \gnode{Festival}, and nothing has (directly) the \gelab{location} \gnode{Santiago}. However, an answer (\gnode{Ñam}) could be automatically entailed were we to state that $x$ being a Food Festival \textit{entails} that $x$ is a Festival, or that $x$ having venue $y$ in city $z$ \textit{entails} that $x$ has location $z$. How, then, should such entailments be captured? In Section~\ref{sec:semSchema} we already discussed how the former entailment can be captured with sub-class relations in a semantic schema; the second entailment, however, requires a more expressive entailment regime than seen thus far. 

In this section, we discuss ways in which more complex entailments can be expressed and automated. Though we could leverage a number of logical frameworks for these purposes -- such as First-Order Logic, Datalog, Prolog, Answer Set Programming, etc. -- we focus on \textit{ontologies}, which constitute a formal representation of knowledge that, importantly for us, can be represented as a graph. We then discuss how these ontologies can be formally defined, how they relate to existing logical frameworks, and how reasoning can be conducted with respect to such ontologies. 

\subsection{Ontologies}

To enable entailment, we must be precise about what the terms we use mean. Returning to Figure~\ref{fig:delg}, for example, and examining the node \gnode{EID16} more closely, we may begin to question how it is modelled, particularly in comparison with \gnode{EID15}. Both nodes -- according to the class hierarchy of Figure~\ref{fig:classhier} -- are considered to be events. But what if, for example, we wish to define two pairs of start and end dates for \gnode{EID16} corresponding to the different venues? Should we rather consider what takes place in each venue as a different event? What then if an event has various start and end dates in a single venue: would these also be considered as one (recurring) event, or many events? These questions are facets of a more general question: \textit{what precisely do we mean by an ``event''}? Does it happen in one contiguous time interval or can it happen many times? Does it happen in one place or can it happen in multiple? There are no ``correct'' answers to such questions -- we may understand the term ``event'' in a variety of ways, and thus the answers are a matter of \textit{convention}. 

In the context of computing, an \textit{ontology}\footnote{The term stems from the philosophical study of \textit{ontology}, concerned with the different kinds of entities that exist, the nature of their existence, what kinds of properties they have, and how they may be identified and categorised.} is then a concrete, formal representation of what terms mean within the scope in which they are used (e.g., a given domain). For example, one event ontology may formally define that if an entity is an ``event'', then it has precisely one venue and precisely one time instant in which it begins. Conversely, a different event ontology may define that an ``event'' can have multiple venues and multiple start times, etc. Each such ontology formally captures a particular perspective -- a particular \textit{convention}. Under the first ontology, for example, we could not call the Olympics an ``event'', while under the second ontology we could. Likewise ontologies can guide how graph data are modelled. Under the first ontology we may split \gnode{EID16} into two events. Under the second, we may elect to keep \gnode{EID16} as one event with two venues. Ultimately, given that ontologies are formal representations, they can be used to automate entailment.

Like all conventions, the usefulness of an ontology depends on the level of agreement on what that ontology defines, how detailed it is, and how broadly and consistently it is adopted. Adoption of an ontology by the parties involved in one knowledge graph may lead to a consistent use of terms and consistent modelling in that knowledge graph. Agreement over multiple knowledge graphs will, in turn, enhance the interoperability of those knowledge graphs. 

Amongst the most popular ontology languages used in practice are the \textit{Web Ontology Language} (\textit{OWL})~\cite{OWL2}\footnote{We could include RDF Schema (RDFS) in this list, but it is largely subsumed by OWL, which builds upon its core.}, recommended by the W3C and compatible with RDF graphs; and the \textit{Open Biomedical Ontologies Format} (\textit{OBOF})~\cite{obof}, used mostly in the biomedical domain. Since OWL is the more widely adopted, we focus on its features, though many similar features are found in both~\cite{obof}. Before introducing such features, however, we must discuss how graphs are to be \textit{interpreted}.

\subsubsection{Interpretations} 

We as humans may \textit{interpret} the node \gnode{Santiago} in the data graph of Figure~\ref{fig:delg} as referring to the real-world city that is the capital of Chile. We may further \textit{interpret} an edge \gedge[arrin][1.2cm]{Arica}{flight}{Santiago} as stating that there are flights from the city of Arica to this city. We thus interpret the data graph as another graph -- what we here call the \textit{domain graph} -- composed of real-world entities connected by real-world relations. The process of interpretation, here, involves \textit{mapping} the nodes and edges in the data graph to nodes and edges of the domain graph. 

Along these lines, we can abstractly define an \textit{interpretation} of a data graph as being composed of two elements: a domain graph, and a mapping from the \textit{terms} (nodes and edge-labels) of the data graph to those of the domain graph. The domain graph follows the same model as the data graph; for example, if the data graph is a directed edge-labelled graph, then so too will be the domain graph. For simplicity, we will speak of directed edge-labelled graphs and refer to the nodes of the domain graph as \textit{entities}, and the edges of the domain graph as \textit{relations}. Given a data graph and an interpretation, while we denote nodes in the data graph by \gnode{Santiago}, we will denote the entity it refers to in the domain graph by \ginode{Santiago} (per the mapping of the given interpretation). Likewise, while we denote an edge by \gedge{Arica}{flight}{Santiago}, we will denote the relation by \giedge{Arica}{flight}{Santiago} (again, per the mapping of the given interpretation). In this abstract notion of an interpretation, we do not require that \ginode{Santiago} nor \ginode{Arica} be the real-world cities, nor even that the domain graph contain real-world entities and relations: an interpretation can have any domain graph and mapping. 

Why is such an abstract notion of interpretation useful? The distinction between nodes/edges and entities/relations becomes important when we define the meaning of ontology features and entailment. 
To illustrate this distinction, if we ask whether there is an edge labelled \gelab{flight} between \gnode{Arica} and \gnode{Vi\~na del Mar} for the data graph in Figure~\ref{fig:delg}, the answer is \emph{no}. However, if we ask if the entities \ginode{Arica} and \ginode{Vi\~na del Mar} are connected by the relation \gielab{flight}, then the answer depends on what assumptions we make when interpreting the graph. Under the Closed World Assumption (CWA), if we do not have additional knowledge, then the answer is a definite \emph{no} -- since what is not known is assumed to be false. Conversely, under the Open World Assumption (OWA), we cannot be certain that this relation does not exist as this could be part of some knowledge not (yet) described by the graph. Likewise under the Unique Name Assumption (UNA), the data graph describes \textit{at least two} flights to \gienode{Santiago} (since \ginode{Viña del Mar} and \ginode{Arica} are assumed to be different entities and therefore, \giedge{Arica}{flight}{Santiago} and \giedge{Vi\~na del Mar}{flight}{Santiago} must be different edges). Conversely, under No Unique Name Assumption (NUNA), we can only say that there is \textit{at least one} such flight since \ginode{Viña del Mar} and \ginode{Arica} may be the same entity  with two ``names''.

These assumptions (or lack thereof) define which interpretations are valid, and which interpretations \textit{satisfy} which data graphs. The UNA forbids interpretations that map two data terms to the same domain term. The NUNA allows such interpretations. Under CWA, an interpretation that contains an edge \giedge[arrin][0.7cm]{x}{p}{y} in its domain graph can only satisfy a data graph from which we can entail \gedge[arrin][0.7cm]{x}{p}{y}. Under OWA, an interpretation containing the edge \giedge[arrin][0.7cm]{x}{p}{y} can satisfy a data graph not entailing \gedge[arrin][0.7cm]{x}{p}{y} so long it does not contradict that edge.\footnote{Variations of the CWA can provide a middle ground between a completely open world that makes no assumption about completeness, falsehood of unknown statements, or unicity of names. One example of such variation is Local Closed World Assumption, already mentioned in Section~\ref{sec:semSchema}.} In the case of OWL, the NUNA and OWA are adopted, thus representing the most general case, whereby multiple nodes/edge-labels in the graph may refer to the same entity/relation-type (NUNA), and where anything not entailed by the data graph is \textit{not} assumed to be false as a consequence (OWA).
 
Beyond our base assumptions, we can associate certain patterns in the data graph with \textit{semantic conditions} that define which interpretations satisfy it; for example, we can add a semantic condition to enforce that if our data graph contains the edge \gedge[arrin][1.3cm]{p}{subp. of}{q}, then any edge \giedge[arrin][0.7cm]{x}{p}{y} in the domain graph of the interpretation must also have a corresponding edge \giedge[arrin][0.7cm]{x}{q}{y} to satisfy the data graph. These semantic conditions then form the features of an ontology language. In what follows, to aid readability, we will introduce the features of OWL using an abstract graphical notation with abbreviated terms. For details of concrete syntaxes, we rather refer to the OWL and OBOF standards~\cite{OWL2,obof}. Likewise we present semantic conditions for interpretations associated with each feature in the same graphical format;\footnote{We use ``iff'' as an abbreviation for ``if and only if'' whereby ``$\phi$ iff $\psi$'' can be read as ``if $\phi$ then $\psi$'' and ``if $\psi$ then $\phi$''.} further details of these conditions will be described later in Section~\ref{sec:ontSemantics}, with formal definitions rather provided in Appendix~\ref{app:deductive}.

\subsubsection{Individuals} In Table~\ref{tab:ontEqIneq}, we list the main features supported by OWL for describing \textit{individuals} (e.g., \textsf{Santiago}, \textsf{EID16}), sometimes distinguished from classes and properties. First, we can \textit{assert} (binary) relations between individuals using edges such as \gedge[arrin][0.9cm]{Santa Lucía}{city}{Santiago}. In the condition column, when we write \giedge[arrin][0.8cm]{$x$}{$y$}{$z$}, for example, we refer to the condition that the given relation holds in the interpretation; if so, the interpretation \textit{satisfies} the axiom. OWL further allows for defining relations to explicitly state that two terms refer to the \textit{same} entity, where, e.g., \gedge[arrin][1.5cm]{Región V}{same as}{Región de Valparaíso} states that both refer to the same region (per Section~\ref{sec:identity}); or that two terms refer to \textit{different} entities, where, e.g., \gedge[arrin][1.6cm]{Valparaíso}{diff. from}{Región de Valparaíso} distinguishes the city from the region of the same name. We may also state that a relation does not hold using \textit{negation}, which can be serialised as a graph using a form of reification (see Figure~\ref{fig:reif}). 

\begin{table}
	\caption{Ontology features for individuals \label{tab:ontEqIneq}}
	\scalebox{\scaledeftabs}{
		\begin{tabular}{llll}
			\toprule
			\textbf{Feature} & \textbf{Axiom} & \textbf{Condition} & \textbf{Example} \\ 
			\midrule
			
			\footnotesize\textsc{Assertion} & \gedge[arrin][\mhgap]{$x$}{$y$}{$z$} & \footnotesize \giedge[arrin][\mhgap]{$x$}{$y$}{$z$} 
			& \gedge[arrin][1.6cm]{Chile}{capital}{Santiago}\\
			\midrule
			
			\footnotesize\textsc{Negation} & \begin{tikzpicture}[baseline=-3pt]
			\node[iri,compact](n){$n$};
			
			\node[iri,compact,right=1.6cm of n,yshift=0.3cm](s){$x$}
			edge[arrin] node[lab] {sub} (n);
			\node[iri,compact,right=1.6cm of n,yshift=-0.3cm](p){$y$}
			edge[arrin] node[lab] {pre} (n);
			\node[iri,compact,right=1.6cm of n,yshift=-0.9cm](o){$z$}
			edge[arrin] node[lab] {obj} (n);
			\node[iri,compact,right=1.6cm of n,yshift=0.9cm](o){Neg}
			edge[arrin] node[lab] {type} (n);
			\end{tikzpicture} & \footnotesize not \giedge[arrin][\mhgap]{$x$}{$y$}{$z$} 
			& \begin{tikzpicture}[baseline=-3pt]
			\node[iri,compact](n){$n$};
			
			\node[iri,compact,right=1.6cm of n,yshift=0.3cm](s){Chile}
			edge[arrin] node[lab] {sub} (n);
			\node[iri,compact,right=1.6cm of n,yshift=-0.3cm](p){capital}
			edge[arrin] node[lab] {pre} (n);
			\node[iri,compact,right=1.6cm of n,yshift=-0.9cm](o){Arica}
			edge[arrin] node[lab] {obj} (n);
			\node[iri,compact,right=1.6cm of n,yshift=0.9cm](o){Neg}
			edge[arrin] node[lab] {type} (n);
			\end{tikzpicture}\\\midrule
			
			\footnotesize\textsc{Same As} & 	\gedge[arrin][1.6cm]{$x_1$}{same as}{$x_2$} & \footnotesize 
			$\ginode{$x_1$} = \ginode{$x_2$}$
			& \gedge[arrin][1.4cm]{Región V}{same as}{Región de Valparaíso} \\
			\midrule
			
			\footnotesize\textsc{Different From} & 	\gedge[arrin][1.6cm]{$x_1$}{diff. from}{$x_2$} & \footnotesize $\ginode{$x_1$} \neq \ginode{$x_2$}$ &
			\gedge[arrin][1.6cm]{Valparaíso}{diff. from}{Región de Valparaíso}\\
			 \bottomrule
		\end{tabular}
	}
\end{table}

\subsubsection{Properties} In Section~\ref{sec:semSchema}, we already discussed how \textit{subproperties}, \textit{domains} and \textit{ranges} may be defined for properties. OWL allows such definitions, and further includes other features, as listed in Table~\ref{tab:ontProp}. We may define a pair of properties to be \textit{equivalent}, \textit{inverses}, or \textit{disjoint}. We can further define a particular property to denote a \textit{transitive}, \textit{symmetric}, \textit{asymmetric}, \textit{reflexive}, or \textit{irreflexive} relation. We can also define the multiplicity of the relation denoted by properties, based on being \textit{functional} (many-to-one) or \textit{inverse-functional} (one-to-many). We may further define a \textit{key} for a class, denoting the set of properties whose values uniquely identify the entities of that class. Without adopting a Unique Name Assumption (UNA), from these latter three features we may conclude that two or more terms refer to the same entity. Finally, we can relate a property to a \textit{chain} (a path expression only allowing concatenation of properties) such that pairs of entities related by the chain are also related by the given property. Note that for the latter two features in Table~\ref{tab:ontProp} we require representing a list, denoted with a vertical notation \gnode{\tiny\raisebox{1ex}{$\vdots$}}; while such a list may be serialised as a graph in a number of concrete ways, OWL uses RDF lists (see Figure~\ref{fig:list}).

\begin{table}
\caption{Ontology features for property axioms \label{tab:ontProp}}
\scalebox{\scaledeftabs}{
\begin{tabular}{llll}
\toprule
\textbf{Feature} & \textbf{Axiom} & \textbf{Condition} (for all $x_{*}$, $y_{*}$, $z_{*}$) & \textbf{Example} \\ 
\midrule

\footnotesize
\textsc{Subproperty} & 	\gedge[arrin][\fhgap]{$p$}{subp. of}{$q$} & \footnotesize\giedge[arrin][\mhgap]{$x$}{$p$}{$y$} implies \giedge[arrin][\mhgap]{$x$}{$q$}{$y$} & \gedge[arrin][\fhgap]{venue}{subp. of}{location} \\
\midrule

\footnotesize
\textsc{Domain} & 	\gedge[arrin][\fhgap]{$p$}{domain}{$c$} & \footnotesize\giedge[arrin][\mhgap]{$x$}{$p$}{$y$} implies \giedge[arrin][\thgap]{$x$}{type}{$c$} & \gedge[arrin][\fhgap]{venue}{domain}{Event} \\
\midrule

\footnotesize
\textsc{Range} & 	\gedge[arrin][\fhgap]{$p$}{range}{$c$} & \footnotesize\giedge[arrin][\mhgap]{$x$}{$p$}{$y$} implies \giedge[arrin][\thgap]{$y$}{type}{$c$} & \gedge[arrin][\fhgap]{venue}{range}{Venue} \\
\midrule

\footnotesize
\textsc{Equivalence} & 	\gedge[arrin][\fhgap]{$p$}{equiv. p.}{$q$} & \footnotesize\giedge[arrin][\mhgap]{$x$}{$p$}{$y$} iff \giedge[arrin][\mhgap]{$x$}{$q$}{$y$} & \gedge[arrin][\fhgap]{start}{equiv. p.}{begins} \\
\midrule

\footnotesize
\textsc{Inverse} & 	\gedge[arrin][\fhgap]{$p$}{inv. of}{$q$} & \footnotesize\giedge[arrin][\mhgap]{$x$}{$p$}{$y$} iff \giedge[arrin][\mhgap]{$y$}{$q$}{$x$} & \gedge[arrin][\fhgap]{venue}{inv. of}{hosts} \\
\midrule

\footnotesize
\textsc{Disjoint} & 	\gedge[arrin][\fhgap]{$p$}{disj. p.}{$q$} & \footnotesize not \begin{tikzpicture}[baseline=-3pt]
\node[iri,intn,compact](x){$x$};

\node[iri,intn,compact,right=\mhgap of x](y){$y$}
edge[arrin,inte,bend left=25] node[lab] {$p$} (x)
edge[arrin,inte,bend right=25] node[lab] {$q$} (x);
\end{tikzpicture} & \gedge[arrin][\fhgap]{venue}{disj. p.}{hosts} \\
\midrule

\footnotesize
\textsc{Transitive} & \gedge[arrin][\thgap]{$p$}{type}{Transitive} & \footnotesize \begin{tikzpicture}[baseline=-3pt]
\node[iri,intn,compact](x){$x$};

\node[iri,intn,compact,right=\mhgap of x](y){$y$}
edge[arrin,inte] node[lab] {$p$} (x);

\node[iri,intn,compact,right=\mhgap of y](z){$z$}
edge[arrin,inte] node[lab] {$p$} (y);
\end{tikzpicture} implies \giedge[arrin][\mhgap]{$x$}{$p$}{$z$} & \gedge[arrin][\thgap]{part of}{type}{Transitive} \\
\midrule

\footnotesize
\textsc{Symmetric} & \gedge[arrin][\thgap]{$p$}{type}{Symmetric} & \footnotesize \giedge[arrin][\mhgap]{$x$}{$p$}{$y$} iff \giedge[arrin][\mhgap]{$y$}{$p$}{$x$} &  \gedge[arrin][\thgap]{nearby}{type}{Symmetric} \\
\midrule

\footnotesize
\textsc{Asymmetric} & \gedge[arrin][\thgap]{$p$}{type}{Asymmetric} & \footnotesize not \begin{tikzpicture}[baseline=-3pt]
\node[iri,intn,compact](x){$x$};

\node[iri,intn,compact,right=\mhgap of x](y){$y$}
edge[arrin,inte,bend left=25] node[lab] {$p$} (x)
edge[arrout,inte,bend right=25] node[lab] {$p$} (x);
\end{tikzpicture} &  \gedge[arrin][\thgap]{capital}{type}{Asymmetric} \\
\midrule

\footnotesize
\textsc{Reflexive} & \gedge[arrin][\thgap]{$p$}{type}{Reflexive} & \footnotesize  \giloop{$x$}{$p$} &  \gedge[arrin][\thgap]{part of}{type}{Reflexive} \\
\midrule

\footnotesize
\textsc{Irreflexive} & \gedge[arrin][\thgap]{$p$}{type}{Irreflexive} & \footnotesize  not \giloop{$x$}{$p$} &  \gedge[arrin][\thgap]{flight}{type}{Irreflexive} \\
\midrule

\footnotesize
\textsc{Functional} & \gedge[arrin][\thgap]{$p$}{type}{Functional} & \footnotesize \begin{tikzpicture}[baseline=-3pt]
\node[iri,intn,compact](y1){$y_1$};

\node[iri,intn,compact,right=\mhgap of y1](x){$x$}
edge[arrout,inte] node[lab] {$p$} (y1);

\node[iri,intn,compact,right=\mhgap of x](y2){$y_2$}
edge[arrin,inte] node[lab] {$p$} (x);
\end{tikzpicture} implies $\ginode{$y_1$} = \ginode{$y_2$}$ &  \gedge[arrin][\thgap]{population}{type}{Functional} \\\midrule

\footnotesize
\textsc{Inv. Functional}& \gedge[arrin][\thgap]{$p$}{type}{Inv. Functional} & \footnotesize \begin{tikzpicture}[baseline=-3pt]
\node[iri,intn,compact](x1){$x_1$};

\node[iri,intn,compact,right=\mhgap of x1](y){$y$}
edge[arrin,inte] node[lab] {$p$} (x1);

\node[iri,intn,compact,right=\mhgap of y](x2){$x_2$}
edge[arrout,inte] node[lab] {$p$} (y);
\end{tikzpicture} implies $\ginode{$x_1$} = \ginode{$x_2$}$ &  \gedge[arrin][\thgap]{capital}{type}{Inv. Functional} \\
\midrule

\footnotesize
\textsc{Key} & \gedge[arrin][\fhgap]{$c$}{key}{\begin{tabular}{@{}c@{}}$p_1$\\[-0.8ex]$\vdots$\\$p_n$\end{tabular}} &  \footnotesize
\begin{tikzpicture}[baseline=-3pt]
\node[iri,intn,compact](x1){$x_1$};

\node[iri,intn,compact,right=0.8cm of x1,yshift=1cm](c){$c$}
  edge[arrin,inte] node[lab] {type} (x1);

\node[iri,intn,compact,right=0.8cm of c,yshift=-1cm](x2){$x_2$}
  edge[arrout,inte] node[lab] {type} (c);
  
\node[iri,intn,compact,below=0.3cm of c](y1){$y_1$}
  edge[arrin,inte] node[lab] {$p_1$} (x1)
  edge[arrin,inte] node[lab] {$p_1$} (x2);
	
\node[iri,intn,compact,below=0.3cm of y1](yd){$...$}
  edge[arrin,inte] node[lab] {$...$} (x1)
  edge[arrin,inte] node[lab] {$...$} (x2);
  
\node[iri,intn,compact,below=0.3cm of yd](yn){$y_n$}
 edge[arrin,inte] node[lab] {$p_n$} (x1)
 edge[arrin,inte] node[lab] {$p_n$} (x2);
\end{tikzpicture} implies $\ginode{$x_1$} = \ginode{$x_2$}$
  &  \gedge[arrin][\fhgap]{City}{key}{\begin{tabular}{@{}c@{}}lat\\long\end{tabular}} \\
\midrule

\footnotesize
\textsc{Chain} & \gedge[arrin][\fhgap]{$p$}{chain}{\begin{tabular}{@{}c@{}}$q_1$\\[-0.8ex]$\vdots$\\$q_n$\end{tabular}} &  \footnotesize \begin{tabular}{@{}l@{}}\begin{tikzpicture}[baseline=-3pt]
\node[iri,intn,compact](x){$x$};

\node[iri,intn,compact,right=\mhgap of x](y1){$y_1$}
edge[arrin,inte] node[lab] {$q_1$} (x);

\node[iri,intn,compact,right=\mhgap of y1](yn1){$y_{n-1}$}
edge[arrin,inte] node[lab] {...} (y1);

\node[iri,intn,compact,right=\mhgap of yn1](z){$z$}
edge[arrin,inte] node[lab] {$q_{n}$} (yn1);
\end{tikzpicture} \\{\color{white}.\hspace{1cm}}implies \giedge[arrin][\mhgap]{$x$}{$p$}{$z$}\end{tabular}  &  \gedge[arrin][\fhgap]{location}{chain}{\begin{tabular}{@{}c@{}}location\\part of\end{tabular}} \\
\bottomrule

\end{tabular}
}
\end{table}

\subsubsection{Classes} In Section~\ref{sec:semSchema}, we discussed how class hierarchies can be modelled using a \textit{sub-class} relation. OWL supports sub-classes, and many additional features, for defining and making claims about classes; these additional features are summarised in Table~\ref{tab:ontClass}. Given a pair of classes, OWL allows for defining that they are \textit{equivalent}, or \textit{disjoint}. Thereafter, OWL provides a variety of features for defining novel classes by applying set operators on other classes, or based on conditions that the properties of its instances satisfy. First, using set operators, one can define a novel class as the \textit{complement} of another class, the \textit{union} or \textit{intersection} of a list (of arbitrary length) of other classes, or as an \textit{enumeration} of all of its instances. Second, by placing restrictions on a particular property $p$, one can define classes whose instances are all of the entities that have: \textit{some value} from a given class on $p$; \textit{all values} from a given class on $p$;\footnote{While something like \begin{tikzpicture}[baseline=-3pt]
	\node[iri,compact](c){DomesticAirport};
	\node[iri,compact,right=0.8cm of c](d){NationalFlight}
	edge[arrin] node[lab] {all} (c);
	\node[iri,compact,left=1cm of c](p){flight}
	edge[arrin] node[lab] {prop} (c);
\end{tikzpicture} might appear to be a more natural example for \textsc{All Values}, this would be a modelling mistake, as the corresponding \textit{for all} condition is satisfied when no such node exists. In other words, with this example definition, we could infer anything known not to have any flights to be a domestic airport. (We could, however, define the intersection of this class and airport as being a domestic airport.)} have a specific individual as a value on $p$ (\textit{has value}); have themselves as a reflexive value on $p$ (\textit{has self}); have at least, at most or exactly some number of values on $p$ (\textit{cardinality}); and have at least, at most or exactly some number of values on $p$ from a given class (\textit{qualified cardinality}). For the latter two cases, in Table~\ref{tab:ontClass}, we use the notation ``$\#\{ \ginode{a} \mid \phi \}$'' to count distinct entities satisfying $\phi$ in the interpretation. 
These features can then be combined to create more complex classes, where combining the examples for \textsc{Intersection} and \textsc{Has Self} in Table~\ref{tab:ontClass} gives the definition: \textit{self-driving taxis are taxis having themselves as a driver}.

\begin{table}
	\caption{Ontology features for class axioms and definitions \label{tab:ontClass}}
	\scalebox{\scaledeftabs}{
		\begin{tabular}{llll}
			\toprule
			\textbf{Feature} & \textbf{Axiom} & \textbf{Condition} (for all $x_{*}$, $y_{*}$, $z_{*}$) & \textbf{Example} \\ 
			\midrule
			
			\footnotesize\textsc{Subclass} & \gedge[arrin][\fhgap]{$c$}{subc. of}{$d$} & \footnotesize\giedge[arrin][\thgap]{$x$}{type}{$c$} implies \giedge[arrin][\thgap]{$x$}{type}{$d$} & \gedge[arrin][\fhgap]{City}{subc. of}{Place} \\
			\midrule
			
			\footnotesize\textsc{Equivalence} & 	\gedge[arrin][\fhgap]{$c$}{equiv. c.}{$d$} & \footnotesize\giedge[arrin][\thgap]{$x$}{type}{$c$} iff \giedge[arrin][\thgap]{$x$}{type}{$d$} & \gedge[arrin][\fhgap]{Human}{equiv. c.}{Person} \\
			\midrule
			
			\footnotesize\textsc{Disjoint} & 	\gedge[arrin][\fhgap]{$c$}{disj. c.}{$d$} & \footnotesize not \begin{tikzpicture}[baseline=-3pt]
			\node[iri,intn,compact](c){$c$};
			\node[iri,intn,compact,right=\thgap of c](x){$x$}
			edge[arrout,inte] node[lab] {type} (c);
			\node[iri,intn,compact,right=\thgap of x](d){$d$}
			edge[arrin,inte] node[lab] {type} (x);
			\end{tikzpicture} & \gedge[arrin][1.6cm]{City}{disj. c.}{Region} \\
			\midrule
			
			\footnotesize\textsc{Complement} & 	\gedge[arrin][\fhgap]{$c$}{comp.}{$d$} & \footnotesize\giedge[arrin][\thgap]{$x$}{type}{$c$} iff not \giedge[arrin][\thgap]{$x$}{type}{$d$}  & \gedge[arrin][\fhgap]{Dead}{comp.}{Alive} \\
			\midrule			
			
			\footnotesize\textsc{Union} & 	\gedge[arrin][\fhgap]{$c$}{union}{\begin{tabular}{@{}c@{}}$d_1$\\[-1ex]$\vdots$\\$d_n$\end{tabular}} & \footnotesize\giedge[arrin][\thgap]{$x$}{type}{$c$} iff~ \begin{tabular}{@{}l@{~}l@{}}
				\giedge[arrin][\thgap]{$x$}{type}{$d_1$} & or\\
				\giedge[arrin][\thgap]{$x$}{type}{$...$} & or\\
				\giedge[arrin][\thgap]{$x$}{type}{$d_n$} 	
			\end{tabular} & \gedge[arrin][1.1cm]{Flight}{union}{\begin{tabular}{@{}c@{}}DomesticFlight\\InternationalFlight\end{tabular}} \\
			\midrule	
			
			\footnotesize\textsc{Intersection} & 	\gedge[arrin][\fhgap]{$c$}{inter.}{\begin{tabular}{@{}c@{}}$d_1$\\[-1ex]$\vdots$\\$d_n$\end{tabular}} & \footnotesize\giedge[arrin][\thgap]{$x$}{type}{$c$} iff~ \begin{tikzpicture}[baseline=-3pt]
			\node[iri,intn,compact](x){$x$};
			\node[iri,intn,compact,right=1.1\thgap of c](dd){...}
			edge[arrin,inte] node[lab] {type} (x);
			\node[iri,intn,compact,above=0.2cm of dd](d1){$d_1$}
			edge[arrin,inte,bend right=20] node[lab] {type} (x);
			\node[iri,intn,compact,below=0.2cm of dd](dn){$d_n$}
			edge[arrin,inte,bend left=20] node[lab] {type} (x);
			\end{tikzpicture} & \gedge[arrin][1.1cm]{SelfDrivingTaxi}{inter.}{\begin{tabular}{@{}c@{}}Taxi\\ SelfDriving\end{tabular}} \\
			\midrule
			
			\footnotesize\textsc{Enumeration} & 	\gedge[arrin][\fhgap]{$c$}{one of}{\begin{tabular}{@{}c@{}}$x_1$\\[-1ex]$\vdots$\\$x_n$\end{tabular}} & \footnotesize\giedge[arrin][\thgap]{$x$}{type}{$c$} iff $\ginode{$x$} \in \{ \ginode{$x_1$}, \ldots, \ginode{$x_n$}\}$  
			& 
			\gedge[arrin][\fhgap]{EUState}{one of}{\begin{tabular}{@{}c@{}}Austria\\[-0.8ex]$\vdots$\\Sweden\end{tabular}} \\
			\midrule
			
			\footnotesize\textsc{Some Values} & 	
			\begin{tikzpicture}[baseline=-3pt]
			\node[iri,compact](c){$c$};
			\node[iri,compact,right=\fhgap of c,yshift=-0.3cm](d){$d$}
			edge[arrin] node[lab] {some} (c);
			\node[iri,compact,right=\fhgap of c,yshift=0.3cm](p){$p$}
			edge[arrin] node[lab] {prop} (c);
			\end{tikzpicture}
			& \footnotesize\giedge[arrin][\thgap]{$x$}{type}{$c$} iff~ 
			\begin{tabular}{@{~}l@{}}
				there exists \ginode{$a$} such that\\ \begin{tikzpicture}[baseline=-3pt]
				\node[iri,intn,compact](x){$x$};
				\node[iri,intn,compact,right=\mhgap of c](a){$a$}
				edge[arrin,inte] node[lab] {$p$} (x);
				\node[iri,intn,compact,right=\thgap of a](d){$d$}
				edge[arrin,inte] node[lab] {type} (a);
				\end{tikzpicture}
			\end{tabular} & 
			\begin{tikzpicture}[baseline=-3pt]
			\node[iri,compact](c){EUCitizen};
			\node[iri,compact,right=\fhgap of c,yshift=-0.3cm](d){EUState}
			edge[arrin] node[lab] {some} (c);
			\node[iri,compact,right=\fhgap of c,yshift=0.3cm](p){nationality}
			edge[arrin] node[lab] {prop} (c);
			\end{tikzpicture} \\
			\midrule
			
			\footnotesize\textsc{All Values} & 	
			\begin{tikzpicture}[baseline=-3pt]
			\node[iri,compact](c){$c$};
			\node[iri,compact,right=\fhgap of c,yshift=-0.3cm](d){$d$}
			edge[arrin] node[lab] {all} (c);
			\node[iri,compact,right=\fhgap of c,yshift=0.3cm](p){$p$}
			edge[arrin] node[lab] {prop} (c);
			\end{tikzpicture}
			& \footnotesize\giedge[arrin][\thgap]{$x$}{type}{$c$} iff~ 
			\begin{tabular}{@{~}l@{}}
				for all \ginode{$a$} with \giedge[arrin][\mhgap]{$x$}{$p$}{$a$}\\
				it holds that \giedge[arrin][\thgap]{$a$}{type}{$d$}
			\end{tabular} & 
			\begin{tikzpicture}[baseline=-3pt]
			\node[iri,compact](c){Weightless};
			\node[iri,compact,right=\fhgap of c,yshift=-0.3cm](d){Weightless}
			edge[arrin] node[lab] {all} (c);
			\node[iri,compact,right=\fhgap of c,yshift=0.3cm](p){has part}
			edge[arrin] node[lab] {prop} (c);
			\end{tikzpicture} \\
			\midrule
			
			\footnotesize\textsc{Has Value} & 	
			\begin{tikzpicture}[baseline=-3pt]
			\node[iri,compact](c){$c$};
			\node[iri,compact,right=\fhgap of c,yshift=-0.3cm](y){$y$}
			edge[arrin] node[lab] {value} (c);
			\node[iri,compact,right=\fhgap of c,yshift=0.3cm](p){$p$}
			edge[arrin] node[lab] {prop} (c);
			\end{tikzpicture}
			& \footnotesize\giedge[arrin][\thgap]{$x$}{type}{$c$} iff \giedge[arrin][\mhgap]{$x$}{$p$}{$y$} & 
			\begin{tikzpicture}[baseline=-3pt]
			\node[iri,compact](c){ChileanCitizen};
			\node[iri,compact,right=1.1cm of c,yshift=-0.3cm](y){Chile}
			edge[arrin] node[lab] {value} (c);
			\node[iri,compact,right=1.1cm of c,yshift=0.3cm](p){nationality}
			edge[arrin] node[lab] {prop} (c);
			\end{tikzpicture} \\
			\midrule
			
			\footnotesize\textsc{Has Self} & 	
			\begin{tikzpicture}[baseline=-3pt]
			\node[iri,compact](c){$c$};
			\node[iri,compact,right=\fhgap of c,yshift=-0.3cm,xshift=-0.1cm](t){true}
			edge[arrin] node[lab] {self} (c);
			\node[iri,compact,right=\fhgap of c,yshift=0.3cm](p){$p$}
			edge[arrin] node[lab] {prop} (c);
			\end{tikzpicture}
			& \footnotesize\giedge[arrin][\thgap]{$x$}{type}{$c$} iff \giedge[arrin][\mhgap]{$x$}{$p$}{$x$} & 
			\begin{tikzpicture}[baseline=-3pt]
			\node[iri,compact](c){SelfDriving};
			\node[iri,compact,right=\fhgap of c,yshift=-0.3cm](t){true}
			edge[arrin] node[lab] {self} (c);
			\node[iri,compact,right=\fhgap of c,yshift=0.3cm](p){driver}
			edge[arrin] node[lab] {prop} (c);
			\end{tikzpicture} \\
			\midrule
			
			\footnotesize
			\begin{tabular}{@{}l@{}}
				\textsc{Cardinality}\\ 
				$\star \in \{ =, \leq, \geq \}$
			\end{tabular} & 	
			\begin{tikzpicture}[baseline=-3pt]
			\node[iri,compact](c){$c$};
			\node[iri,compact,right=\fhgap of c,yshift=-0.3cm](n){$n$}
			edge[arrin] node[lab] {$\star$} (c);
			\node[iri,compact,right=\fhgap of c,yshift=0.3cm](p){$p$}
			edge[arrin] node[lab] {prop} (c);
			\end{tikzpicture}
			& \footnotesize\begin{tabular}{@{}l@{}}\giedge[arrin][\thgap]{$x$}{type}{$c$} iff \\{\color{white}.\hspace{0.5cm}}$\# \{ \ginode{$a$} \mid \giedge[arrin][\mhgap]{$x$}{$p$}{$a$} \} \star n$\end{tabular}
			 & 
			\begin{tikzpicture}[baseline=-3pt]
			\node[iri,compact](c){Polyglot};
			\node[iri,compact,right=\fhgap of c,yshift=-0.3cm](t){$2$}
			edge[arrin] node[lab] {$\geq$} (c);
			\node[iri,compact,right=\fhgap of c,yshift=0.3cm](p){fluent}
			edge[arrin] node[lab] {prop} (c);
			\end{tikzpicture} \\
			\midrule
			
			\footnotesize
			\begin{tabular}{@{}l@{}}
				\textsc{Qualified}\\
				\textsc{Cardinality}\\ 
				$\star \in \{ =, \leq, \geq \}$
			\end{tabular} & 	
			\begin{tikzpicture}[baseline=-3pt]
			\node[iri,compact](c){$c$};
			\node[iri,compact,right=\fhgap of c](d){$d$}
			edge[arrin] node[lab] {class} (c);
			\node[iri,compact,right=\fhgap of c,yshift=-0.6cm](n){$n$}
			edge[arrin] node[lab] {$\star$} (c);
			\node[iri,compact,right=\fhgap of c,yshift=0.6cm](p){$p$}
			edge[arrin] node[lab] {prop} (c);
			\end{tikzpicture}
			& \footnotesize\begin{tabular}{@{}l@{}}\giedge[arrin][\thgap]{$x$}{type}{$c$} iff \\{\color{white}.\hspace{0.5cm}}$\# \{ \ginode{$a$} \mid \begin{tikzpicture}[baseline=-3pt]
					\node[iri,intn,compact](x){$x$};
					\node[iri,intn,compact,right=\mhgap of c](a){$a$}
					edge[arrin,inte] node[lab,xshift=-0.2ex] {$p$} (x);
					\node[iri,intn,compact,right=\thgap of a](d){$d$}
					edge[arrin,inte] node[lab,xshift=-0.2ex] {type} (a);
					\end{tikzpicture} \} \star n$\end{tabular} & 
			\begin{tikzpicture}[baseline=-3pt]
			\node[iri,compact](c){BinaryStarSystem};
			\node[iri,compact,right=\fhgap of c](d){Star}
			edge[arrin] node[lab] {class} (c);
			\node[iri,compact,right=\fhgap of c,yshift=-0.6cm](n){$2$}
			edge[arrin] node[lab] {$=$} (c);
			\node[iri,compact,right=\fhgap of c,yshift=0.6cm](p){body}
			edge[arrin] node[lab] {prop} (c);
			\end{tikzpicture} \\
			
			\bottomrule
		\end{tabular}
	}
\end{table}

\subsubsection{Other features} OWL supports other language features not previously discussed, including: \textit{annotation properties}, which provide metadata about ontologies, such as versioning info; \textit{datatype vs. object properties}, which distinguish properties that take datatype values from those that do not; and \textit{datatype facets}, which allow for defining new datatypes by applying restrictions to existing datatypes, such as to define that places in Chile must have a \textit{float between -66.0 and -110.0} as their value for the (datatype) property \gelab{latitude}. For more details we refer to the OWL 2 standard~\cite{OWL2}. We will further discuss methodologies for the creation of ontologies in Section~\ref{ssec:knowledgeConceptual}.

\subsection{Semantics and Entailment}\label{sec:ontSemantics}

The conditions listed in the previous tables indicate how each feature should be interpreted. These conditions give rise to \textit{entailments}, where, for example, in reference to the \textsc{Symmetric} feature of Table~\ref{tab:ontProp}, the definition \gedge{nearby}{type}{Symmetric} and edge \gedge{Santiago}{nearby}{Santiago Airport} entail the edge \gedge{Santiago Airport}{nearby}{Santiago} according to the condition given for that feature. We now describe how these conditions lead to entailments.

\subsubsection{Model-theoretic semantics} Each axiom described by the previous tables, when added to a graph, enforces some condition(s) on the interpretations that \textit{satisfy} the graph. The interpretations that satisfy a graph are called \textit{models} of the graph. Were we to consider only the base condition of the \textsc{Assertion} feature in Table~\ref{tab:ontEqIneq}, for example, then the models of a graph would be any interpretation such that for every edge \gedge[arrin][\mhgap]{x}{y}{z} in the graph, there exists a relation \giedge[arrin][\mhgap]{x}{y}{z} in the model. Given that there may be other relations in the model (under the OWA), the number of models of any such graph is infinite. Furthermore, given that we can map multiple nodes in the graph to one entity in the model (under the NUNA), any interpretation with (for example) the relation \giedge[arrin][\mhgap]{a}{a}{a} is a model of any graph so long as for every edge \gedge[arrin][\mhgap]{x}{y}{z} in the graph, it holds that \ginode{x} = \ginode{y} = \ginode{z} = \ginode{a} in the interpretation (in other words, the interpretation maps everything to \ginode{a}). As we add axioms with their associated conditions to the graph, we restrict models for the graph; for example, considering a graph with two edges -- \gedge[arrin][\mhgap]{x}{y}{z} and \gedge[arrin][\thgap]{y}{type}{Irreflexive} -- the interpretation with \giedge[arrin][\mhgap]{a}{a}{a}, \ginode{x} = \ginode{y} = ... = \ginode{a} is no longer a model as it breaks the condition for the irreflexive axiom.

\subsubsection{Entailment} We say that one graph \textit{entails} another if and only if any model of the former graph is also a model of the latter graph. Intuitively this means that the latter graph says nothing new over the former graph and thus holds as a logical consequence of the former graph. For example, consider the graph \begin{tikzpicture}[baseline=-3pt]
\node[iri,compact](x){Santiago};

\node[iri,compact,right=\thgap of x](y){City}
edge[arrin] node[lab] {type} (x);

\node[iri,compact,right=\fhgap of y](z){Place}
edge[arrin] node[lab] {subc. of} (y);
\end{tikzpicture} and the graph \begin{tikzpicture}[baseline=-3pt]
\node[iri,compact](x){Santiago};

\node[iri,compact,right=\thgap of x](y){Place}
edge[arrin] node[lab] {type} (x);
\end{tikzpicture}. All models of the latter must have that \giedge[arrin][\thgap]{Santiago}{type}{Place}, but so must all models of the former, which must have \begin{tikzpicture}[baseline=-3pt]
\node[iri,intn,compact](x){Santiago};

\node[iri,intn,compact,right=\thgap of x](y){City}
edge[arrin,inte] node[lab] {type} (x);

\node[iri,intn,compact,right=\fhgap of y](z){Place}
edge[arrin,inte] node[lab] {subc. of} (y);
\end{tikzpicture} and further must satisfy the condition for \textsc{Subclass}, which requires that \giedge[arrin][\thgap]{Santiago}{type}{Place} also hold. Hence we conclude that any model of the former graph must be a model of the latter graph, or, in other words, the former graph entails the latter graph.

\subsubsection{If--then vs. if-and-only-if semantics}  Consider the graph \gedge[arrin][\thgap]{nearby}{type}{Symmetric} and the graph \gedge[arrin][\fhgap]{nearby}{inv. of}{nearby}. They result in the same semantic conditions being applied in the domain graph, but does one entail the other? The answer depends on the semantics applied. Considering the axioms and conditions of Tables~\ref{tab:ontEqIneq}, we can consider two semantics. Under if--then semantics -- \textsc{if} \textbf{Axiom} matches data graph \textsc{then} \textbf{Condition} holds in domain graph -- the graphs do not entail each other: though both graphs give rise to the same condition, this condition is not translated back into the axioms that describe it.\footnote{Observe that $\giedge[arrin][\thgap]{nearby}{type}{Symmetric}$ is a model of the first graph but not the second, while $\giedge[arrin][\fhgap]{nearby}{inv. of}{nearby}$ is a model of the second graph but not the first. Hence neither graph entails the other.} Conversely, under if-and-only-if semantics -- \textbf{Axiom} matches data graph \textsc{if-and-only-if} \textbf{Condition} holds in domain graph -- the graphs entail each other: both graphs give rise to the same condition, which is translated back into all possible axioms that describe it. Hence if-and-only-if semantics allows for entailing more axioms in the ontology language than if--then semantics. OWL generally applies an if-and-only-if semantics~\cite{OWL2}.

\subsection{Reasoning}\label{ssec:reasoning}

Unfortunately, given two graphs, deciding if the first entails the second -- per the notion of entailment we have defined and for all of the ontological features listed in Tables~\ref{tab:ontEqIneq}--\ref{tab:ontClass} -- is \textit{undecidable}: no (finite) algorithm for such entailment can exist that halts on all inputs with the correct \texttt{true}/\texttt{false} answer~\cite{Hitzler2010}. However, we can provide practical reasoning algorithms for ontologies that (1) halt on any input ontology but may miss entailments, returning \texttt{false} instead of \texttt{true}, (2) always halt with the correct answer but only accept input ontologies with restricted features, or (3) only return correct answers for any input ontology but may never halt on certain inputs. Though option (3) has been explored using, e.g., theorem provers for First Order Logic~\cite{SchneiderS11}, options (1) and (2) are more commonly pursued using rules and/or Description Logics. Option (1) generally allows for more efficient and scalable reasoning algorithms and is useful where data are incomplete and having some entailments is valuable. Option (2) may be a better choice in domains -- such as medical ontologies -- where missing entailments may have undesirable outcomes.

\subsubsection{Rules}\label{sec:rules}

One of the most straightforward ways to provide automated access to deductive knowledge is through \textit{inference rules} (or simply \textit{rules}) encoding \textsc{if}--\textsc{then}-style consequences. A rule is composed of a \emph{body} (\textsc{if}) and a \emph{head} (\textsc{then}). Both the body and head are given as graph patterns. A rule indicates that if we can replace the variables of the body with terms from the data graph and form a subgraph of a given data graph, then using the same replacement of variables in the head will yield a valid entailment. The head must typically use a subset of the variables appearing in the body to ensure that the conclusion leaves no variables unreplaced. 
Rules of this form correspond to (positive) Datalog~\cite{CeriGT89} in databases, Horn clauses~\cite{lloyd2012foundations} in logic programming, etc. 

Rules can be used to capture entailments under ontological conditions. In Table~\ref{tab:rulesRdfs}, we list some example rules for sub-class, sub-property, domain and range features~\cite{MunozPG09}; these rules may be considered incomplete, not capturing, for example, that every class is a sub-class of itself, that every property is a sub-property of itself, etc. A more comprehensive set of rules for the OWL features of Tables~\ref{tab:ontEqIneq}--\ref{tab:ontClass} have been defined as OWL 2 RL/RDF~\cite{key:owl2profiles}; these rules are likewise incomplete as such rules cannot fully capture negation (e.g., \textsc{Complement}), existentials (e.g., \textsc{Some Values}), universals (e.g., \textsc{All Values}), or counting (e.g., \textsc{Cardinality} and \textsc{Qualified Cardinality}). Other rule languages have, however, been proposed to support additional such features, including existentials (see, e.g., Datalog$^\pm$~\cite{BellomariniSG18}), disjunction (see, e.g., Disjunctive Datalog~\cite{RudolphKH08}), etc.

Rules can be leveraged for reasoning in a number of ways. \textit{Materialisation} refers to the idea of applying rules recursively to a graph, adding the conclusions generated back to the graph until a fixpoint is reached and nothing more can be added. The materialised graph can then be treated as any other graph. Although the efficiency and scalability of materialisation can be enhanced through optimisations like Rete networks~\cite{Forgy82}, or using distributed frameworks like MapReduce~\cite{UrbaniKMHB12}, depending on the rules and the data, the materialised graph may become unfeasibly large to manage. Another strategy is to use rules for \textit{query rewriting}, which given a query, will automatically extend the query in order to find solutions entailed by a set of rules; for example, taking the schema graph in Figure~\ref{fig:sg} and the rules in Table~\ref{tab:rulesRdfs}, the (sub-)pattern \begin{tikzpicture}[baseline=-3pt]
\node[var,compact](x){?x};

\node[iri,compact,right=\thgap of x](e){Event}
edge[arrin] node[lab] {type} (x);
\end{tikzpicture} in a given input query would be rewritten to the following disjunctive pattern evaluated on the original graph:

\begin{center}
\begin{tikzpicture}[baseline=-3pt]
\node[var,compact](x){?x};

\node[iri,compact,right=\thgap of x](e){Event}
edge[arrin] node[lab] {type} (x);
\end{tikzpicture} $\cup$ 
\begin{tikzpicture}[baseline=-3pt]
\node[var,compact](x){?x};

\node[iri,compact,right=\thgap of x](e){Festival}
edge[arrin] node[lab] {type} (x);
\end{tikzpicture} $\cup$ 
\begin{tikzpicture}[baseline=-3pt]
\node[var,compact](x){?x};

\node[iri,compact,right=\thgap of x](e){Periodic Market}
edge[arrin] node[lab] {type} (x);
\end{tikzpicture} $\cup$ 
\begin{tikzpicture}[baseline=-3pt]
\node[var,compact](x){?x};

\node[var,compact,right=\fhgap of x](y){?y}
edge[arrin] node[lab] {venue} (x);
\end{tikzpicture}
\end{center}

\begin{figure}
	\centering
	\begin{tabular}{r@{}l}			
		$O$: 
		&
		\setlength{\hgap}{1.4cm}
		\setlength{\vgap}{0.5cm}
		\begin{tikzpicture}[baseline=-3pt]
			
			\node[iri,anchor=center] (l) {location};
				
			\node[iri,anchor=center,right=\hgap of l] (lv) {\begin{tabular}{@{}c@{}}venue\\city\end{tabular}}
			  edge[arrin] node[lab] (k) {chain} (l);

			\node[iri,anchor=center,right=2.7\hgap of lv] (f) {Festival};
			  
			\node[iri,anchor=center,left=\hgap of f] (ff) {Food Festival}
			  edge[arrout] node[lab] {subc. of} (f);
			
			\node[iri,anchor=center,right=\hgap of f] (df) {Drinks Festival}
			  edge[arrout] node[lab] {subc. of} (f);
		\end{tikzpicture}
		\\[4ex]
		
		$O(Q):$
		& 
		\setlength{\hgap}{0.93cm}
		\setlength{\vgap}{1cm}	
		(
		\begin{tikzpicture}[baseline=-3pt]
			\node[var,anchor=center] (fv) {?festival};
		
			\node[iri,anchor=center,right=\hgap of fv] (fn) {Festival}
			  edge[arrin] node[lab] {type} (fv);
		\end{tikzpicture} $\cup$
		\begin{tikzpicture}[baseline=-3pt]
			\node[var,anchor=center] (fv) {?festival};
		
			\node[iri,anchor=center,right=\hgap of fv] (fn) {Food Festival}
			  edge[arrin] node[lab] {type} (fv);
		\end{tikzpicture} $\cup$
		\begin{tikzpicture}[baseline=-3pt]
			\node[var,anchor=center] (fv) {?festival};
	
			\node[iri,anchor=center,right=\hgap of fv] (fn) {Drinks Festival}
			  edge[arrin] node[lab] {type} (fv);
		\end{tikzpicture}
		)
		\\[1ex]
		
		&
		\setlength{\hgap}{1.2cm}
		\setlength{\vgap}{1cm}
		$\Join$
		(
		\begin{tikzpicture}[baseline=-3pt]
			\node[var,anchor=center] (fv) {?festival};
		
			\node[iri,anchor=center,right=1.3\hgap of fv] (fn) {Santiago}
			  edge[arrin] node[lab] {location} (fv);
		\end{tikzpicture} $\cup$		
		\begin{tikzpicture}[baseline=-3pt]
			\node[var,anchor=center] (fv) {?festival};
		
			\node[var,anchor=center,right=\hgap of fv] (fn) {?x}
			  edge[arrin] node[lab] {venue} (fv);
			  
			\node[iri,anchor=center,right=\hgap of fn] (s) {Santiago}
			  edge[arrin] node[lab] {city} (fn);			  
		\end{tikzpicture}
		)
		
		\\[1ex]
		
		&
		\setlength{\hgap}{1.2cm}
		\setlength{\vgap}{1cm}
		$\Join$
		\begin{tikzpicture}[baseline=-3pt]
			\node[var,anchor=center] (fv) {?festival};
		
			\node[var,anchor=center,right=\hgap of fv] (fn) {\textbf{?name}}
			  edge[arrin] node[lab] {name} (fv);
		\end{tikzpicture}
	\end{tabular}				

	\caption{Query rewriting example for the query $Q$ of Figure~\ref{fig:bgpFS} \label{fig:qrew}}
\end{figure}
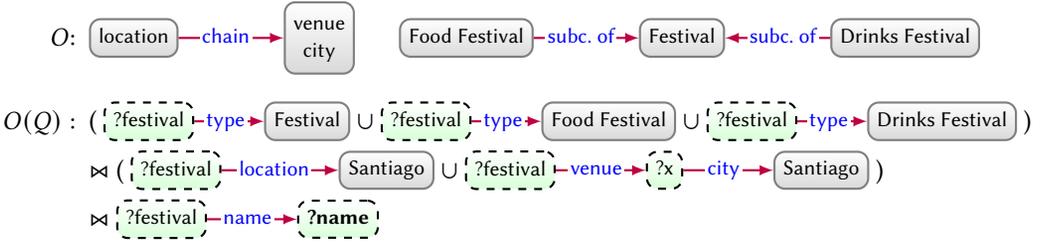

\noindent Figure~\ref{fig:qrew} provides a more complete example of an ontology that is used to rewrite the query of Figure~\ref{fig:bgpFS}; if evaluated over the graph of Figure~\ref{fig:delg}, \gnode{Ñam} will be returned as a solution. However, not all of the aforementioned features of OWL can be supported in this manner. The OWL 2 QL profile~\cite{key:owl2profiles} is a subset of OWL designed specifically for query rewriting of this form~\cite{ArtaleCKZ09}.

While rules can be used to (partially) capture ontological entailments, they can also be defined independently of an ontology language, capturing entailments for a given domain. In fact, some rules -- such as the following -- cannot be captured by the ontology features previously seen, as they do not support ways to infer relations from cyclical graph patterns (for computability reasons):

\begin{center}
\begin{tikzpicture}[baseline=-3pt]
\node[var,compact](x){?x};

\node[var,compact,right=\fhgap of x](y){?y}
edge[arrin] node[lab] {flight} (x);

\node[var,compact,right=\fhgap of y](z){?z}
  edge[arrin,bend left=20] node[lab] {country} (x)
  edge[arrin] node[lab] {country} (y);
\end{tikzpicture} $\Rightarrow$ 
\begin{tikzpicture}[baseline=-3pt]
\node[var,compact](x){?x};

\node[var,compact,right=1.6\fhgap of x](e){?y}
edge[arrin] node[lab] {domestic flight} (x);
\end{tikzpicture}
\end{center}

\noindent
Various languages allow for expressing rules over graphs -- independently or alongside of an ontology language -- including: Notation3 (N3)~\cite{n3}, Rule Interchange Format (RIF)~\cite{rif}, Semantic Web Rule Language (SWRL)~\cite{swrl}, and SPARQL Inferencing Notation (SPIN)~\cite{spin}.

\begin{table}
	\caption{Example rules for sub-class, sub-property, domain, and range features \label{tab:rulesRdfs}}
	\scalebox{\scaledeftabs}{
		\begin{tabular}{llcl}
			\toprule
			\textbf{Feature} & \textbf{Body} & $\Rightarrow$ & \textbf{Head} \\ 
			\midrule
			
			\footnotesize\textsc{Subclass} (I) & 	
			\begin{tikzpicture}[baseline=-3pt]
			\node[var,compact](x){?x};
			
			\node[var,compact,right=\thgap of x](c){?c}
			edge[arrin] node[lab] {type} (x);
			
			\node[var,compact,right=\fhgap of c](d){?d}
			edge[arrin] node[lab] {subc. of} (c);
			\end{tikzpicture}	& $\Rightarrow$		
			 & \begin{tikzpicture}[baseline=-3pt]
			 \node[var,compact](x){?x};
			 
			 \node[var,compact,right=\thgap of x](d){?d}
			 edge[arrin] node[lab] {type} (x);
			 \end{tikzpicture} \\
			 \midrule
			 
			 \footnotesize\textsc{Subclass} (II) & 	
			 \begin{tikzpicture}[baseline=-3pt]
			 \node[var,compact](c){?c};
			 
			 \node[var,compact,right=\fhgap of c](d){?d}
			 edge[arrin] node[lab] {subc. of} (c);
			 
			 \node[var,compact,right=\fhgap of d](e){?e}
			 edge[arrin] node[lab] {subc. of} (d);
			 \end{tikzpicture} & $\Rightarrow$		
			 & \begin{tikzpicture}[baseline=-3pt]
			 \node[var,compact](c){?c};
			 
			 \node[var,compact,right=\fhgap of c](e){?e}
			 edge[arrin] node[lab] {subc. of} (c);
			 \end{tikzpicture} \\
			 \midrule
			 
			 \footnotesize\textsc{Subproperty} (I) & 	
			 \begin{tikzpicture}[baseline=4pt]
			 \node[var,compact](x){?x};
			 
			 \node[var,compact,right=\mhgap of x](y){?y}
			 edge[arrin] node[vlab,draw,dotted] (p) {?p} (x);
			 
			 \node[var,compact,right=\mhgap of y](q){?q}
			 edge[arrin,bend right=40] node[lab] {subp. of} (p);
			 \end{tikzpicture} & $\Rightarrow$		
			 & \begin{tikzpicture}[baseline=-3pt]
			 \node[var,compact](x){?x};
			 
			 \node[var,compact,right=\thgap of x](y){?y}
			 edge[arrin] node[lab] {?q} (x);
			 \end{tikzpicture} \\
			 \midrule
			 
			 \footnotesize\textsc{Subproperty} (II) & 	
			 \begin{tikzpicture}[baseline=-3pt]
			 \node[var,compact](p){?p};
			 
			 \node[var,compact,right=\fhgap of p](q){?q}
			 edge[arrin] node[lab] {subp. of} (p);
			 
			 \node[var,compact,right=\fhgap of q](r){?r}
			 edge[arrin] node[lab] {subp. of} (q);
			 \end{tikzpicture} & $\Rightarrow$
			 & \begin{tikzpicture}[baseline=-3pt]
			 \node[var,compact](p){?p};
			 
			 \node[var,compact,right=\fhgap of p](r){?r}
			 edge[arrin] node[lab] {subp. of} (p);
			 \end{tikzpicture} \\
			 \midrule
			 
			 \footnotesize\textsc{Domain} & 	
			\begin{tikzpicture}[baseline=4pt]
			\node[var,compact](x){?x};
			
			\node[var,compact,right=\mhgap of x](y){?y}
			edge[arrin] node[vlab,draw,dotted] (p) {?p} (x);
			
			\node[var,compact,right=\mhgap of y](c){?c}
			edge[arrin,bend right=40] node[lab] {domain} (p);
			\end{tikzpicture} & $\Rightarrow$		
			& \begin{tikzpicture}[baseline=-3pt]
			\node[var,compact](x){?x};
			
			\node[var,compact,right=\thgap of x](c){?c}
			edge[arrin] node[lab] {type} (x);
			\end{tikzpicture} \\
			\midrule
			
			 \footnotesize\textsc{Range} & 	
			\begin{tikzpicture}[baseline=4pt]
			\node[var,compact](x){?x};
			
			\node[var,compact,right=\mhgap of x](y){?y}
			edge[arrin] node[vlab,draw,dotted] (p) {?p} (x);
			
			\node[var,compact,right=\mhgap of y](c){?c}
			edge[arrin,bend right=40] node[lab] {range} (p);
			\end{tikzpicture} & $\Rightarrow$	
			& \begin{tikzpicture}[baseline=-3pt]
			\node[var,compact](y){?y};
			
			\node[var,compact,right=\thgap of y](c){?c}
			edge[arrin] node[lab] {type} (y);
			\end{tikzpicture} \\
			\bottomrule
		\end{tabular}
	}
\end{table}

\subsubsection{Description Logics}\label{sssec:dls} Description Logics (DLs) were initially introduced as a way to formalise the meaning of \emph{frames} \cite{minsky} and \emph{semantic networks} \cite{quillian}. Considering that semantic networks are an early version of knowledge graphs, and the fact that DLs have heavily influenced the Web Ontology Language, DLs thus hold an important place in the logical formalisation of knowledge graphs. DLs form a family of logics rather than a particular logic. Initially, DLs were restricted fragments of First Order Logic (FOL) that permit decidable reasoning tasks, such as entailment checking~\cite{BaaderHLS17}. Different DLs strike different balances between expressive power and computational complexity of reasoning. DLs would later be extended with features that go beyond FOL but are useful in the context of modelling graph data, such as transitive closure, datatypes, etc.

Description Logics are based on three types of elements: \textit{individuals}, such as \texttt{Santiago}; \textit{classes} (aka \textit{concepts}) such as \texttt{City}; and \textit{properties} (aka \textit{roles}) such as \texttt{flight}. DLs then allow for making claims, known as \textit{axioms}, about these elements. \textit{Assertional axioms} can be either unary class relations on individuals, such as \texttt{City(Santiago)}, or binary property relations on individuals, such as \texttt{flight(Santiago,Arica)}. Such axioms form the \textit{Assertional Box} (\textit{A-Box}). DLs further introduce logical symbols to allow for defining \textit{class axioms} (forming the \textit{Terminology Box}, or \textit{T-Box} for short), and \textit{property axioms} (forming the \textit{Role Box}, \textit{R-Box}); for example, the class axiom $\texttt{City} \sqsubseteq \texttt{Place}$ states that the former class is a subclass of the latter one, while the property axiom $\texttt{flight} \sqsubseteq \texttt{connectsTo}$ states that the former property is a subproperty of the latter one. DLs may then introduce a rich set of logical symbols, not only for defining class and property axioms, but also defining new classes based on existing terms; as an example of the latter, we can define a class $\exists \texttt{nearby}.\texttt{Airport}$ as the class of individuals that have some airport nearby. Noting that the symbol $\top$ is used in DLs to denote the class of all individuals, we can then add a class axiom $\exists \texttt{flight}.\top \sqsubseteq \exists \texttt{nearby}.\texttt{Airport}$ to state that individuals with an outgoing flight must have some airport nearby. Noting that the symbol $\sqcup$ can be used in DL to define that a class is the union of other classes, we can further define that  $\texttt{Airport} \sqsubseteq \texttt{DomesticAirport} \sqcup  \texttt{InternationalAirport}$, i.e., that an airport is either a domestic airport or an international airport (or both).

The similarities between these DL features and the OWL features previously outlined in Tables~\ref{tab:ontEqIneq}--\ref{tab:ontClass} are not coincidental: the OWL standard was heavily influenced by DLs, where, for example, the OWL 2 DL language is a fragment of OWL restricted so that entailment becomes decidable. As an example of a restriction, with $\texttt{DomesticAirport} \sqsubseteq  ~=1~\texttt{destination} \circ \texttt{country}.\top$, we can define in DL syntax that domestic airports have flights destined to precisely one country (where $\texttt{p} \circ \texttt{q}$ denotes a chain of properties). However, counting chains is often disallowed in DLs to ensure decidability. In Appendix~\ref{sec:dlformal}, we present formal definitions for DL syntax and semantics, as well as notions of entailment. For further reading, we also refer to the textbook by~\citet{BaaderHLS17}.

Expressive DLs support complex entailments involving existentials, universals, counting, etc. A common strategy for deciding such entailments is to reduce entailment to \textit{satisfiability}, which decides if an ontology is consistent or not~\cite{HorrocksP04}.\footnote{$G$ entails $G'$ if and only if $G \cup \text{not}(G')$ is not satisfiable.} Thereafter methods such as \textit{tableau} can be used to check satisfiability, cautiously constructing models by completing them along similar lines to the materialisation strategy previously described, but additionally branching models in the case of disjunction, introducing new elements to represent existentials, etc. If any model is successfully ``completed'', the process concludes that the original definitions are satisfiable (see, e.g.,~\cite{MotikSH09}). Due to their prohibitive computational complexity~\cite{key:owl2profiles} -- where for example, disjunction may lead to an exponential number of branching possibilities -- such reasoning strategies are not typically applied in the case of large-scale data, though they may be useful when modelling complex domains.
\section{Inductive Knowledge}\label{sec:inductive}

While deductive knowledge is characterised by precise logical consequences, inductively acquiring knowledge involves generalising patterns from a given set of input observations, which can then be used to generate novel but potentially imprecise predictions. For example, from a large data graph with geographical and flight information, we may observe the pattern that almost all capital cities of countries have international airports serving them, and hence predict that if Santiago is a capital city, it \textit{likely} has an international airport serving it; however, the predictions drawn from this pattern do not hold for certain, where (e.g.) Vaduz, the capital city of Liechtenstein, has no (international) airport serving it. Hence predictions will often be associated with a level of confidence; for example, we may say that a capital has an international airport in $\frac{187}{195}$ of cases, offering a confidence of $0.959$ for predictions made with that pattern. We then refer to knowledge acquired inductively as \textit{inductive knowledge}, which includes both the models used to encode patterns, as well as the predictions made by those models. Though fallible, inductive knowledge can be highly valuable.

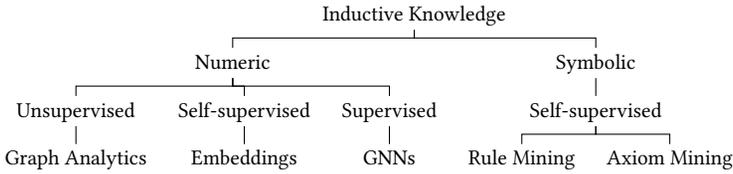
\begin{figure}
\centering
	\begin{tikzpicture}
	\tikzset{edge from parent/.style=
		{draw,
			edge from parent path={(\tikzparentnode.south)
				-- +(0,-3pt)
				-| (\tikzchildnode)}}}
	\tikzset{every tree node/.style={font=\hsp\footnotesize,inner sep=1pt},,sibling
	distance=10pt,level distance=18pt}
	\Tree [.Inductive~Knowledge 
	  [.Numeric [.Unsupervised [.Graph~Analytics ] ] [.Self-supervised [.Embeddings ] ] [.Supervised [.GNNs ] ] ]
	  [.Symbolic [.Self-supervised [.Rule~Mining ] [.Axiom~Mining ] ] ] 
	 ]
	\end{tikzpicture}
\caption{Conceptual overview of popular inductive techniques for knowledge graphs in terms of type of representation generated (Numeric/Symbolic) and type of paradigm used (Unsupervised/Self-supervised/Supervised). \label{fig:ind}}
\end{figure}

In Figure~\ref{fig:ind} we provide an overview of the inductive techniques typically applied to knowledge graphs. In the case of unsupervised methods, there is a rich body of work on \textit{graph analytics}, which uses well-known functions/algorithms to detect communities or clusters, find central nodes and edges, etc., in a graph. Alternatively, \textit{knowledge graph embeddings} can use self-supervision to learn a low-dimensional numeric model of a knowledge graph that (typically) maps input edges to an output \textit{plausibility score} indicating the likelihood of the edge being true. The structure of graphs can also be directly leveraged for supervised learning, as explored in the context of \textit{graph neural networks}. Finally, while the aforementioned techniques learn numerical models, \textit{symbolic learning} can learn symbolic models -- i.e., logical formulae in the form of rules or axioms -- from a graph in a self-supervised manner. We now discuss each of the aforementioned techniques in turn.

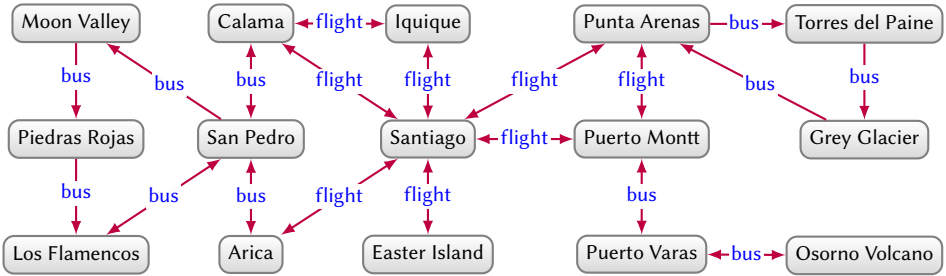
\begin{figure}
	\setlength{\vgap}{1cm}
	\setlength{\hgap}{1cm}
	\setlength{\sgap}{0.3cm}
	\setlength{\mgap}{0.1cm}

	\centering
	\begin{tikzpicture}
	\node[iri,anchor=mid] (ca) {Calama};

	\node[iri,anchor=mid,below=\vgap of ca] (sp) {San Pedro}
	edge[arroutin] node[lab] {bus} (ca);

	\node[iri,anchor=mid,left=0.7\hgap of sp] (pr) {Piedras Rojas};

	\node[iri,anchor=mid,above=\vgap of pr] (vl) {Moon Valley}
	edge[arrin] node[lab] {bus} (sp)
	edge[arrout] node[lab] {bus} (pr);

	\node[iri,anchor=mid,below=\vgap of pr] (lf) {Los Flamencos}
	edge[arrin] node[lab] {bus} (pr)
	edge[arroutin] node[lab] {bus} (sp);

	\node[iri,anchor=mid,right=\hgap of sp] (sa) {Santiago}
	edge[arroutin] node[lab] {flight}  (ca);

	\node[iri,anchor=mid,below=\vgap of sp] (ar) {Arica}
	edge[arroutin] node[lab] {flight} (sa)
	edge[arroutin] node[lab] {bus} (sp);

	\node[iri,anchor=mid,above=\vgap of sa] (iq) {Iquique}
	edge[arroutin] node[lab] {flight} (sa)
	edge[arroutin] node[lab] {flight} (ca);

	\node[iri,anchor=mid,below=\hgap of sa] (ei) {Easter Island}
	edge[arroutin] node[lab] {flight} (sa);

	\node[iri,anchor=mid,right=1.3\hgap of sa] (pm) {Puerto Montt}
	edge[arroutin] node[lab] {flight} (sa);

	\node[iri,anchor=mid,above=\vgap of pm] (pa) {Punta Arenas}
	edge[arroutin] node[lab] {flight} (pm)
	edge[arroutin] node[lab] {flight} (sa);

	\node[iri,anchor=mid,below=\vgap of pm] (pv) {Puerto Varas}
	edge[arroutin] node[lab] {bus} (pm);

	\node[iri,anchor=mid,right=\hgap of pa] (tdp) {Torres del Paine}
	edge[arrin] node[lab] {bus} (pa);

	\node[iri,anchor=mid,below=\vgap of tdp] (gg) {Grey Glacier}
	edge[arrin] node[lab] {bus} (tdp)
	edge[arrout] node[lab] {bus} (pa);

	\node[iri,anchor=mid,below=\vgap of gg] (os) {Osorno Volcano}
	edge[arroutin] node[lab] {bus} (pv);
	\end{tikzpicture}
	\caption{Data graph representing transport routes in Chile \label{fig:chileTransport}}
\end{figure}

\subsection{Graph Analytics}\label{sec:gAnalytics}

Analytics is the process of discovering, interpreting, and communicating meaningful patterns inherent to (typically large) data collections. Graph analytics is then the application of analytical processes to (typically large) graph data. The nature of graphs naturally lends itself to certain types of analytics that derive conclusions about nodes and edges based on the \textit{topology} of the graph, i.e., how the nodes of the graph are connected. Graph analytics hence draws many of its techniques from related areas such as graph theory and network analysis, which have been used to study graphs that represent social networks, the Web, internet routing, transportation networks, ecosystems, protein--protein interactions, linguistic cooccurrences, and more besides~\cite{Estrada2011}.

Returning to the domain of our running example, the tourism board could use graph analytics to extract knowledge about, for instance: key transport hubs that serve many tourist attractions (centrality); groupings of attractions visited by the same tourists (community detection); attractions that may become unreachable in the event of strikes or other route failures (connectivity), or pairs of attractions that are similar to each other (node similarity). Given that such analytics will require a complex, large-scale graph, for the purposes of illustration, in Figure~\ref{fig:chileTransport} we present a more concise example of some transportation connections in Chile directed towards popular touristic destinations. We first introduce a selection of key techniques that can be applied for graph analytics. We then discuss frameworks and languages that can be used to compute such analytics in practice. Given that many traditional graph algorithms are defined for unlabelled graphs, we then describe ways in which analytics can be applied over directed edge-labelled graphs. Finally we discuss the potential connections between graph analytics and querying and reasoning.

\subsubsection{Techniques}\label{sssec:graph-analytics-tasks} A wide variety of techniques can be applied for graph analytics. In the following we will enumerate
some of the main techniques -- as recognised, for example, by the survey of~\citet{IosupHNHPMCCSAT16} -- that can be invoked in this setting. 

\begin{itemize}
    \item{\it Centrality:} aims to identify the most important (aka \textit{central}) nodes or edges of a graph. Specific node centrality measures include \textit{degree}, \textit{betweenness}, \textit{closeness}, \textit{Eigenvector}, \textit{PageRank}, \textit{HITS}, \textit{Katz}, among others. Betweenness centrality can also be applied to edges. A node centrality measure would allow, e.g., to predict the transport hubs in Figure~\ref{fig:chileTransport}, while edge centrality would allow us to find the edges on which many shortest routes depend for predicting traffic.
    \item{\it Community detection:} aims to identify \textit{communities} in a graph, i.e., sub-graphs that are more densely connected internally than to the rest of the graph. Community detection algorithms, such as \textit{minimum-cut algorithms}, \textit{label propagation}, \textit{Louvain modularity},  etc.\ enable discovering such communities. Community detection applied to Figure~\ref{fig:chileTransport} may, for example, detect a community to the left (referring to the north of Chile), to the right (referring to the south of Chile), and perhaps also the centre (referring to cities with airports).
    \item{\it Connectivity:} aims to estimate how well-connected the graph is, revealing, for instance, the resilience and (un)reachability of elements of the graph. Specific techniques include measuring \textit{graph density} or \textit{$k$-connectivity}, detecting \textit{strongly connected components} and \textit{weakly connected components}, computing \textit{spanning trees} or \textit{minimum cuts}, etc. In the context of Figure~\ref{fig:chileTransport}, such analysis may tell us that routes to \gnode{Grey Glacier}, \gnode{Osorno Volcano} and \gnode{Piedras Rojas} are the most ``brittle'', becoming disconnected if one of two \gelab{bus} routes fail.
    \item{\it Node similarity:} aims to find nodes that are similar to other nodes by virtue of how they are connected within their neighbourhood. Node similarity metrics may be computed using \textit{structural equivalence}, \textit{random walks}, \textit{diffusion kernels}, etc. These methods provide an understanding of what connects nodes, and, thereafter, in what ways they are similar.  In the context of Figure~\ref{fig:chileTransport}, such analysis may tell us that \gnode{Calama} and \gnode{Arica} are similar nodes based on both having return flights to \gnode{Santiago} and return buses to \gnode{San Pedro}.
\end{itemize}

\noindent While the previous techniques accept a graph alone as input,\footnote{Node similarity can be run over an entire graph to find the $k$ most similar nodes for each node, or can also be run for a specific node to find its most similar nodes. There are also measures for graph similarity (based on, e.g., frequent itemsets~\cite{MaillotB18}) that accept multiple graphs as input.} other forms of graph analytics may further accept a node, a pair of nodes, etc., along with the graph.

\begin{itemize}
    \item{\it Path finding:} aims to find paths in a graph, typically between pairs of nodes given as input. Various technical definitions exist that restrict the set of valid paths between such nodes, including simple paths that do not visit the same node twice, shortest paths that visit the fewest number of edges, or -- as previously discussed in Section~\ref{ssec:querying} -- regular path queries that restrict the labels of edges that can be traversed by the path~\cite{AnglesABHRV17}. We could use such algorithms to find, for example, the shortest path(s) in Figure~\ref{fig:chileTransport} from \gnode{Torres del Paine} to \gnode{Moon Valley}.
\end{itemize}

\noindent Most such techniques have been proposed and studied for simple graphs or directed graphs without edge labels. We will discuss their application to more complex graph models -- and how they can be combined with other techniques such as reasoning and querying -- later in Section~\ref{sssec:query-languages}.

\subsubsection{Frameworks}\label{sssec:technologies-graph-analytics}

Various frameworks have been proposed for large-scale graph analytics, often in a distributed (cluster) setting. Amongst these we can mention Apache Spark (GraphX)~\cite{DBLP:conf/sigmod/XinGFS13,DBLP:conf/grades/DaveJLXGZ16}, GraphLab~\cite{LowGKBGH12}, Pregel~\cite{DBLP:conf/sigmod/MalewiczABDHLC10}, Signal--Collect~\cite{signalcollect}, Shark~\cite{DBLP:conf/sigmod/XinRZFSS13}, etc. These \textit{graph parallel frameworks} apply a \textit{systolic abstraction}~\cite{Kung82} based on a directed graph, where nodes are processors that can send messages to other nodes along edges. Computation is then iterative, where in each iteration, each node reads messages received through inward edges (and possibly its own previous state), performs a computation, and then sends messages through outward edges based on the result. These frameworks then define the systolic computational abstraction on top of the data graph being processed: nodes and edges in the data graph become nodes and edges in the systolic graph. We refer to Appendix~\ref{app:gpfs} for more formal details on graph parallel frameworks.

\begin{figure}
\centering
\setlength{\vgap}{1cm}
\setlength{\hgap}{0.8cm}
\setlength{\sgap}{0.1cm}
\setlength{\mgap}{0.1cm}

\begin{tikzpicture}
\node[block, minimum width=4.9cm,minimum height=4.4cm,fill=blue!5!white] (msg)  {};

\node[iri,anchor=mid,right=0.6cm of msg.west,yshift=0.25cm] (pr) {Piedras Rojas};

\node[iri,anchor=mid,below=\vgap of pr] (lf) {Los Flamencos}
edge[arrin] node[msg] {$\frac{d}{6 \cdot 1}$} (pr);

\node[iri,anchor=mid,right=0.7\hgap of pr] (sp) {San Pedro}
edge[arrin,bend left=13] node[msg] {$\frac{d}{6 \cdot 1}$} (lf)
edge[arrout,bend right=13] node[msg] {$\frac{d}{6 \cdot 4}$} (lf);

\node[iri,anchor=mid,above=\vgap of sp] (c)  {Calama}
edge[arrin,bend left=25] node[msg] {$\frac{d}{6 \cdot 4}$} (sp)
edge[arrout,bend right=25] node[msg] {$\frac{d}{6 \cdot 1}$} (sp);

\node[iri,anchor=mid,below=\vgap of sp] (a) {Arica}
edge[arrin,bend left=25] node[msg] {$\frac{d}{6 \cdot 4}$} (sp)
edge[arrout,bend right=25] node[msg] {$\frac{d}{6 \cdot 1}$} (sp);

\node[iri,anchor=mid,above=\vgap of pr] (mv) {Moon Valley}
edge[arrin] node[msg] {$\frac{d}{6 \cdot 4}$} (sp)
edge[arrout] node[msg] {$\frac{d}{6 \cdot 1}$} (pr);

\node[left=0.1cm of msg.south,anchor=south] {\textsc{Msg} \scriptsize(\texttt{iter = 1})};

\node[state,right=\sgap of a,anchor=west,faded] (sa) { $\frac{1}{6}$ };

\node[state,left=\sgap of pr,anchor=east,faded] (spr) { $\frac{1}{6}$ };

\node[state,left=\sgap of lf,anchor=east,faded] (slf) { $\frac{1}{6}$ };

\node[state,right=\sgap of sp,anchor=west,faded] (ssp) { $\frac{1}{6}$ };

\node[state,right=\sgap of c,anchor=west,faded] (sc) { $\frac{1}{6}$ };

\node[state,left=\sgap of mv,anchor=east,faded] (smvf) { $\frac{1}{6}$ };

\node[block,minimum width=8.4cm,minimum height=4.4cm,fill=green!5!white,right=0.4cm of msg.east,anchor=west] (com)  {};

\node[iri,anchor=mid,right=2.6cm of ssp] (pr) {Piedras Rojas};

\node[iri,anchor=mid,below=\vgap of pr] (lf) {Los Flamencos}
edge[arrin] node[msg,faded] {$\frac{d}{6 \cdot 1}$} (pr);

\node[iri,anchor=mid,right=0.7\hgap of pr] (sp) {San Pedro}
edge[arrin,bend left=13] node[msg,faded] {$\frac{d}{6 \cdot 1}$} (lf)
edge[arrout,bend right=13] node[msg,faded] {$\frac{d}{6 \cdot 4}$} (lf);

\node[iri,anchor=mid,above=\vgap of sp] (c)  {Calama}
edge[arrin,bend left=25] node[msg,faded] {$\frac{d}{6 \cdot 4}$} (sp)
edge[arrout,bend right=25] node[msg,faded] {$\frac{d}{6 \cdot 1}$} (sp);

\node[iri,anchor=mid,below=\vgap of sp] (a) {Arica}
edge[arrin,bend left=25] node[msg,faded] {$\frac{d}{6 \cdot 4}$} (sp)
edge[arrout,bend right=25] node[msg,faded] {$\frac{d}{6 \cdot 1}$} (sp);

\node[iri,anchor=mid,above=\vgap of pr] (mv) {Moon Valley}
edge[arrin] node[msg,faded] {$\frac{d}{6 \cdot 4}$} (sp)
edge[arrout] node[msg,faded] {$\frac{d}{6 \cdot 1}$} (pr);

\node[left=0.1cm of com.south,anchor=south] {\textsc{Agg} \scriptsize(\texttt{iter = 1})};

\node[state,right=\sgap of a,anchor=west] (sa) { $\frac{d}{6\cdot4} + \frac{1-d}{6}$ };

\node[state,left=\sgap of pr,anchor=east] (spr) { $\frac{d}{6\cdot1} + \frac{1-d}{6}$ };

\node[state,left=\sgap of lf,anchor=east] (slf) { $\frac{d}{6\cdot4} + \frac{d}{6\cdot1} + \frac{1-d}{6}$ };

\node[state,right=\sgap of sp,anchor=west] (ssp) { $\frac{d}{6\cdot1} + \frac{d}{6\cdot1} + \frac{d}{6\cdot1} + \frac{1-d}{6}$ };

\node[state,right=\sgap of c,anchor=west] (sc) {  $\frac{d}{6\cdot4} + \frac{1-d}{6}$ };

\node[state,left=\sgap of mv,anchor=east] (smv) {  $\frac{d}{6\cdot4} + \frac{1-d}{6}$ };
\end{tikzpicture}
\caption{Example of a systolic iteration of PageRank for a sample sub-graph of Figure~\ref{fig:chileTransport} \label{fig:pagerank}}
\end{figure}
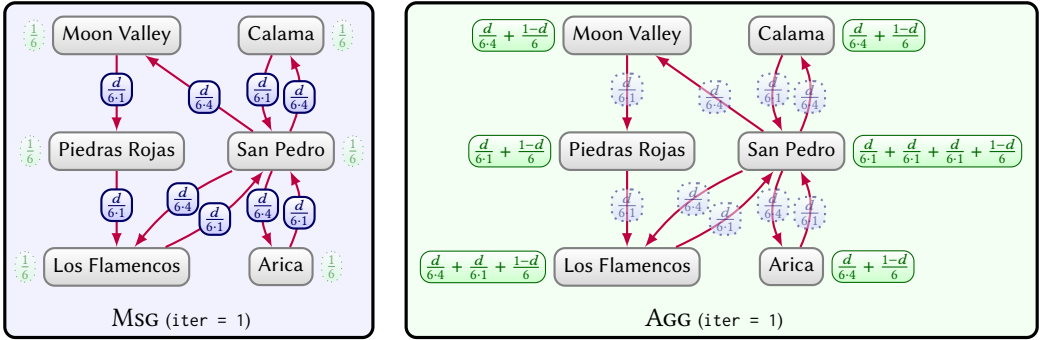

To take an example, assume we wish to compute the places that are most (or least) easily reached by the routes shown in the graph of Figure~\ref{fig:chileTransport}. A good way to measure this is using centrality, where we choose PageRank~\cite{page1999pagerank}, which computes the probability of a tourist randomly following the routes shown in the graph being at a particular place after a given number of ``hops''. We can implement PageRank on large graphs using a graph parallel framework. In Figure~\ref{fig:pagerank}, we provide an example of an iteration of PageRank for an illustrative sub-graph of Figure~\ref{fig:chileTransport}. The nodes are initialised with a score of $\frac{1}{|V|} = \frac{1}{6}$, where we assume the tourist to have an equal chance of starting at any point. In the \textit{message phase} (\textsc{Msg}), each node $v$ passes a score of $\frac{d \textrm{R}_i(v)}{|E(v)|}$ on each of its outgoing edges, where we denote by $d$ a constant damping factor used to ensure convergence (typically $d = 0.85$, indicating the probability that a tourist randomly ``jumps'' to any place), by $\textrm{R}_i(v)$ the score of node $v$ in iteration $i$ (the probability of the tourist being at node $v$ after $i$ hops), and by $|E(v)|$ the number of outgoing edges of $v$. The aggregation phase (\textsc{Agg}) for $v$ then sums all incoming messages received along with its constant share of the damping factor ($\frac{1-d}{|V|}$) to compute  $\textrm{R}_{i+1}(v)$. We then proceed to the message phase of the next iteration, continuing until some termination criterion is reached (e.g., iteration count or residual threshold, etc.) and final scores are output.

While the given example is for PageRank, the systolic abstraction is general enough to support a wide variety of graph analytics, including those previously mentioned. An algorithm in this framework consists of the functions to compute message values in the \textit{message phase} (\textsc{Msg}), and to accumulate the messages in the aggregation phase (\textsc{Agg}). The framework will take care of distribution, message passing, fault tolerance, etc. However, such frameworks -- based on message passing between neighbours -- have limitations: not all types of analytics can be expressed in such frameworks~\cite{XuHLJ19}.\footnote{Formally \citet{XuHLJ19} have shown that such frameworks are as powerful as the (incomplete) Weisfeiler--Lehman (WL) graph isomorphism test -- based on recursively hashing neighbouring hashes -- for distinguishing graph structures.} Hence frameworks may allow additional features, such as a \textit{global step} that performs a global computation on all nodes, making the result available to each node~\cite{DBLP:conf/sigmod/MalewiczABDHLC10}; or a \textit{mutation step} that allows for adding or removing nodes and edges during processing~\cite{DBLP:conf/sigmod/MalewiczABDHLC10}. 

\subsubsection{Analytics on data graphs}\label{sssec:query-languages}

\begin{figure}
	\setlength{\vgap}{2.6cm}
	\setlength{\hgap}{2.6cm}
	\begin{subfigure}[b]{.32\textwidth}
		\centering
		\begin{tikzpicture}
		\node[iri,anchor=mid](y){Santiago};

		\node[iri,right=\hgap of y,anchor=mid](z){Arica}
		edge[arrin] node[lab] {flight} (y);

		\node[between=y and z] (yz) {};

		\node[iri,above=\vgap of yz,anchor=mid](x){Iquique}
		edge[arrout] node[lab] {flight} (y)
		edge[arrout] node[lab] {bus} (z);
		\end{tikzpicture}
		\caption{Original graph}
	\end{subfigure}
	\begin{subfigure}[b]{.32\textwidth}
		\centering
		\begin{tikzpicture}
		\node[iri,anchor=mid](y){Santiago};

		\node[iri,right=\hgap of y,anchor=mid](z){Arica};

		\node[between=y and z] (yz) {};

		\node[iri,above=\vgap of yz,anchor=mid](x){Iquique};

		\node[iri,above=0.35\vgap of yz,anchor=mid](v){flight}
		edge[arrin] (x)
		edge[arroutin] (y)
		edge[arrout] (z);

		\node[iri,between=x and z,anchor=mid](w){bus}
		edge[arrin] (x)
		edge[arrout] (z);
		\end{tikzpicture}
		\caption{Lossy transformation}
	\end{subfigure}
	\begin{subfigure}[b]{.32\textwidth}
		\centering
		\begin{tikzpicture}
		\node[iri,anchor=mid](y){Santiago};

		\node[iri,right=\hgap of y,anchor=mid](z){Arica};

		\node[iri,above=\vgap of yz,anchor=mid](x){Iquique};

		\node[iri,above=0.3\vgap of yz,anchor=mid](v){flight};

		\node[iri,circle,compact,between=y and z] (yz) {}
		edge[arrout] (v)
		edge[arrin] (y);

		\node[iri,between=x and z,anchor=mid] (w) {bus};

		\node[iri,circle,compact,anchor=mid] (xz) at (z|-w) {}
		edge[arrout] (w)
		edge[arrin] (x);

		\node[iri,circle,compact,between=x and y] (xy) {}
		edge[arrout] (v)
		edge[arrin] (x);

		\node[iri,circle,compact,between=yz and z,anchor=mid](zo){}
		edge[arrout] (z)
		edge[arrin] (yz)
		edge[arrin] (xz);

		\node[iri,circle,compact,between=xy and y,anchor=mid](yo){}
		edge[arrout] (y)
		edge[arrin] (xy);
		\end{tikzpicture}
		\caption{Lossless transformation}
	\end{subfigure}
	\caption{Transformations from a directed edge-labelled graph to a directed graph \label{fig:transform}}
\end{figure}
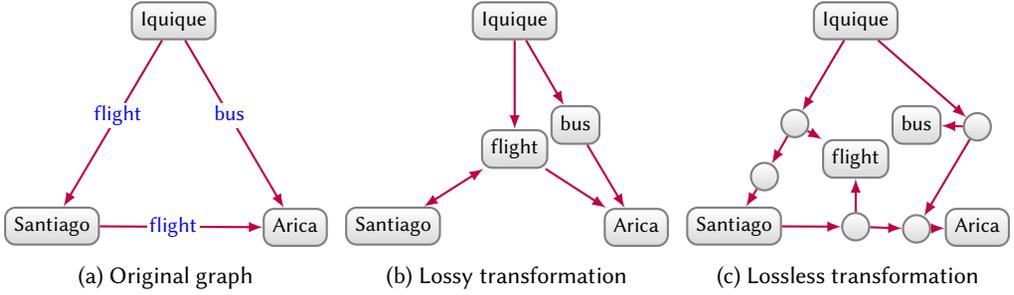

As aforementioned, most analytics presented thus far are, in their ``native'' form, applicable for undirected or directed graphs without the \textit{edge meta-data} -- i.e., edge labels or property--value pairs -- typical of graph data models.\footnote{We remark that in the case of property graphs, property--value pairs on nodes can be converted by mapping values to nodes and properties to edges with the corresponding label.} A number of strategies can be applied to make data graphs subject to analytics of this form:

\begin{itemize}
	\item \textit{Projection} involves simply ``projecting'' an undirected or directed graph by optionally selecting a sub-graph from the data graph from which all edge meta-data are dropped; for example, Figure~\ref{fig:pagerank} may be the result of extracting the sub-graph induced by the edge labels \gelab{bus} and \gelab{flight} from a larger data graph, where the labels are then dropped to create a directed graph.
	\item \textit{Weighting} involves converting edge meta-data into numerical values according to some function. Many of the aforementioned techniques are easily adapted to the case of weighted (directed) graphs; for example, we could consider weights on the graph of Figure~\ref{fig:pagerank} denoting trip duration (or price, traffic, etc.), and then compute the shortest paths adding the duration of each leg.\footnote{Other forms of analytics are possible if we assume the graph is weighted; for example, if we annotated the graph of Figure~\ref{fig:pagerank} with probabilities of tourists moving from one place to the next, we could leverage \textit{Markov processes} to understand features such as reducibility, periodicity, transience, recurrence, ergodicity, steady-states, etc., of the routes~\cite{markov}.} In the absence of external weights, we may rather map edge labels to weights, assigning the same weight to all \gelab{flight} edges, to all \gelab{bus} edges, etc., based on some criteria.
	\item \textit{Transformation} involves transforming the graph to a lower arity model. A transformation may be \textit{lossy}, meaning that the original graph cannot be recovered; or \textit{lossless}, meaning that the original graph can be recovered. Figure~\ref{fig:transform} provides an example of a lossy and lossless transformation from a directed edge-labelled graph to directed graphs. In the lossy transformation, we cannot tell, for example, if the original graph contained the edge \gedge[arrin][0.8\fhgap]{Iquique}{flight}{Santiago} or \gedge[arrin][0.8\fhgap]{Iquique}{flight}{Arica}, etc. The lossless transformation must introduce new nodes (similar to reification) to maintain information about directed labelled edges. Both transformed graphs further attempt to preserve the directionality of the original graph.
	\item \textit{Customisation} involves changing the analytical procedure to incorporate edge meta-data, such as was the case for path finding based on path expressions. Other examples might include structural measures for node similarity that not only consider common neighbours, but also common neighbours connected by edges with the same label, or aggregate centrality measures that capture the importance of edges grouped by label, etc.
\end{itemize}

\noindent The results of an analytical process may change drastically depending on which of the previous strategies are chosen to prepare the data for analysis. This choice may be a non-trivial one to make \textit{a priori} and may require empirical validation. More study is required to more generally understand the effects of such strategies on the results of different analytical techniques.

\subsubsection{Analytics with queries}\label{sssec:analyticsQ} As discussed in Section~\ref{ssec:querying}, various languages for querying graphs have been proposed~\cite{AnglesABHRV17}. One may consider a variety of ways in which query languages and analytics can complement each other. First, we may consider using query languages to project or transform a graph suitable for a particular analytical task, such as to extract the graph of Figure~\ref{fig:chileTransport} from a larger data graph. Query languages such as SPARQL~\cite{sparql11}, Cypher~\cite{FrancisGGLLMPRS18}, and G-CORE~\cite{AnglesABBFGLPPS18} allow for outputting graphs, where such queries can be used to select sub-graphs for analysis. These languages can also express some limited (non-recursive) analytics, where aggregations can be used to compute degree centrality, for example; they may also have some built-in analytical support, where, for example, Cypher~\cite{FrancisGGLLMPRS18} allows for finding shortest paths. In the other direction, analytics can contribute to the querying process in terms of \textit{optimisations}, where, for example, analysis of connectivity may suggest how to better distribute a large data graph over multiple machines for querying using, e.g., \textit{minimum cuts}~\cite{AkhterNS18,JankeST18}. Analytics have also been used to \textit{rank} query results over large graphs~\cite{WagnerTLHS12,FanWW13}, selecting the most important results for presentation to the user.

In some use-cases we may further wish to interleave querying and analytical processes. For example, from the full data graph collected by the tourist board, consider an upcoming airline strike where the board wishes to find \textit{the events during the strike with venues in cities unreachable from Santiago by public transport due to the strike}. Hypothetically, we could use a query to extract the transport network excluding the airline's routes (assuming, per Figure~\ref{fig:fsa} that the airline information is available), use analytics to extract the strongly connected component containing Santiago, and finally use a query to find events in cities not in the Santiago component on the given dates.\footnote{Such a task could not be solved in a single query using regular path queries as such expressions would not be capable of filtering edges representing flights of a particular airline.} While one could solve this task using an imperative language such as Gremlin~\cite{Rodriguez15}, GraphX~\cite{DBLP:conf/sigmod/XinGFS13}, or R~\cite{R}, more declarative languages are also being explored to more easily express such tasks, with proposals including the extension of graph query languages with recursive capabilities~\cite{BishofDKLP12,ReutterSV15},\footnote{Recursive query languages become Turing complete assuming one can also express operations on binary arrays.} combining linear algebra with relational (query) algebra~\cite{HutchisonHS17}, and so forth.

\subsubsection{Analytics with entailment}\label{sssec:analyticsE}

Knowledge graphs are often associated with a semantic schema or ontology that defines the semantics of domain terms, giving rise to entailments (per Section~\ref{sec:deductive}). Applying analytics with or without such entailments -- e.g., before or after materialisation -- may yield radically different results. For example, observe that an edge  \gedge[arrin][1.1cm]{Santa Lucía}{hosts}{EID15} is semantically equivalent to an edge \gedge[arrin][1.1cm]{EID15}{venue}{Santa Lucía} once the inverse axiom \gedge[arrin][1.1cm]{hosts}{inv. of}{venue} is invoked; however, these edges are far from equivalent from the perspective of analytical techniques that consider edge direction, for which including one type of edge, or the other, or both, may have a major bearing on the final results. To the best of our knowledge, the combination of analytics and entailment has not been well-explored, leaving open interesting research questions. Along these lines, it may be of interest to explore \textit{semantically-invariant analytics} that yield the same results over semantically-equivalent graphs (i.e., graphs that entail one another), thus analysing the semantic content of the knowledge graph rather than simply the topological features of the data graph; for example, semantically-invariant analytics would yield the same results over a graph containing the inverse axiom \gedge[arrin][1.1cm]{hosts}{inv. of}{venue} and a number of \gelab{hosts} edges, the same graph but where every \gelab{hosts} edge is replaced by an inverse \gelab{venue} edge, and the union of both graphs.

\subsection{Knowledge Graph Embeddings}\label{ssec:embeddings}

Methods for machine learning have gained significant attention in recent years. In the context of knowledge graphs, machine learning can either be used for directly \textit{refining} a knowledge graph~\cite{Paulheim17} (discussed further in Section~\ref{sec:refine}); or for \textit{downstream tasks} using the knowledge graph, such as recommendation~\cite{zhang2016collaborative}, information extraction~\cite{VashishthJT18}, question answering~\cite{HuangZLL19}, query relaxation~\cite{WangWLCZQ18}, query approximation~\cite{HamiltonBZJL18}, etc.\ (discussed further in Section~\ref{sec:kgs}). However, many traditional machine learning techniques assume dense numeric input representations in the form of vectors, which is quite distinct from how graphs are usually expressed. So how can graphs -- or nodes, edges, etc., thereof -- be encoded as numeric vectors?

A first attempt to represent a graph using vectors would be to use a \textit{one-hot encoding}, generating a vector for each node of length $|L|  \cdot |V|$ -- with $|V|$ the number of nodes in the input graph and $|L|$ the number of edge labels -- placing a one at the corresponding index to indicate the existence of the respective edge in the graph, or zero otherwise. Such a representation will, however, typically result in large and sparse vectors, which will be detrimental for most machine learning models. 

The main goal of knowledge graph embedding techniques is to create a dense representation of the graph (i.e., \textit{embed} the graph) in a continuous, low-dimensional vector space that can then be used for machine learning tasks. The dimensionality $d$ of the embedding is fixed and typically low (often, e.g., $50 \geq d \geq 1000$). Typically the graph embedding is composed of an \textit{entity embedding} for each node: a vector with $d$ dimensions that we denote by $\mathbf{e}$; and a \textit{relation embedding} for each edge label: (typically) a vector with $d$ dimensions that we denote by $\mathbf{r}$. The overall goal of these vectors is to abstract and preserve latent structures in the graph. There are many ways in which this notion of an embedding can be instantiated. Most commonly, given an edge $\gedge[arrin][0.5cm]{s}{p}{o}$, a specific embedding approach defines a \textit{scoring function} that accepts $\mathbf{e}_\texttt{s}$ (the entity embedding of node \gnode{s}), $\mathbf{r}_\gelab{p}$ (the entity embedding of edge label \gelab{p}) and $\mathbf{e}_\texttt{o}$ (the entity embedding of node \gnode{o}) and computes the \textit{plausibility} of the edge: how likely it is to be true. Given a data graph, the goal is then to compute the embeddings of dimension $d$ that maximise the plausibility of positive edges (typically edges in the graph) and minimise the plausibility of negative examples (typically edges in the graph with a node or edge label changed such that they are no longer in the graph) according to the given scoring function. The resulting embeddings can then be seen as models learnt through self-supervision that encode (latent) features of the graph, mapping input edges to output plausibility scores.

Embeddings can then be used for a number of low-level tasks involving the nodes and edge-labels of the graph from which they were computed. First, we can use the plausibility scoring function to assign a confidence to edges that may, for example, have been extracted from an external source (discussed later in Section~\ref{sec:create}). Second, the plausibility scoring function can be used to complete edges with missing nodes/edge labels for the purposes of link prediction (discussed later in Section~\ref{sec:refine}); for example, in Figure~\ref{fig:chileTransport}, we might ask which nodes in the graph are likely to complete the edge \gedge[arrin][0.8cm]{Grey Glacier}{bus}{?}, where -- aside from \gnode{Punta Arenas}, which is already given -- we might intuitively expect \gnode{Torres del Paine} to be a plausible candidate. Third, embedding models will typically assign similar vectors to similar nodes and similar edge-labels, and thus they can be used as the basis of similarity measures, which may be useful for finding duplicate nodes that refer to the same entity, or for the purposes of providing recommendations (discussed later in Section~\ref{sec:kgs}).

A wide range of knowledge graph embedding techniques have been proposed~\cite{Wang2017KGEmbedding}. Our goal here is to provide a high-level introduction to some of the most popular techniques proposed thus far. First we discuss \textit{translational models} that adopt a geometric perspective whereby relation embeddings translate subject entities to object entities in the low-dimensional space. We then describe \textit{tensor decomposition models} that extract latent factors approximating the graph's structure. Thereafter we discuss \textit{neural models} that use neural networks to train embeddings that provide accurate plausibility scores. Finally, we discuss \textit{language models} that leverage existing word embedding techniques, proposing ways of generating graph-like analogues for their expected (textual) inputs. A more formal treatment of these models is provided in Appendix~\ref{app:gembeddings}.

\subsubsection{Translational models}

\begin{figure}
	\setlength{\vgap}{1cm}
	\setlength{\hgap}{1.3cm}
	\setlength{\sgap}{0.3cm}
	\setlength{\mgap}{0.1cm}
	\newlength{\squ}
	\setlength{\squ}{5.1cm}
	\begin{subfigure}[b]{.3\textwidth}
		\centering
		\begin{tikzpicture}
			\node[iri,anchor=mid] (v) {Valparaíso};
		
			\node[iri,anchor=mid,right=\hgap of v] (sa) {Santiago}
			edge[arrin] node[lab] {west of}  (v);
		
			\node[iri,anchor=mid,below=\vgap of sa] (c) {Curico}
			edge[arrin] node[lab] {north of} (sa);
		
			\node[iri,anchor=mid,below=\vgap of v] (l) {Licantén}
			edge[arrout] node[lab] {west of} (c)
			edge[arrin] node[lab] {north of} (v);
			
			\node[iri,anchor=mid,above=\vgap of sa] (t) {Toconao}
			edge[arrout] node[lab] {north of} (sa);
		
			\node[iri,anchor=mid,above=\vgap of v] (a) {Antofagasta}
			edge[arrout] node[lab] {north of} (v);
		\end{tikzpicture}
		\caption{Original graph \label{fig:distEg}}
	\end{subfigure}
	\begin{subfigure}[b]{.31\textwidth}
		\centering
		\begin{tikzpicture}
		   \begin{axis}[
		        xmin=-1,xmax=1,
		        ymin=-1,ymax=1,
		        scatter/classes={a={mark=o,draw=black}},
		        width=\squ,
		        height=\squ,
		        xticklabels={,,},
		        yticklabels={,,},
		        ticks=none
		       ]		

				\draw[-latex, black] (axis cs:0,0) -- (axis cs:0.9,0) node[midway,lab]{$\mathbf{r}_{\gelab{wo.}}$};
				
				\draw[-latex, black] (axis cs:0,0) -- (axis cs:0,-0.8) node[midway,lab]{$\mathbf{r}_{\gelab{no.}}$};
	        \end{axis}
		\end{tikzpicture}
		\caption{Relation embeddings}
	\end{subfigure}
	\begin{subfigure}[b]{.31\textwidth}
		\centering
		\begin{tikzpicture}
		        \begin{axis}[
		        xmin=-1,xmax=1,
		        ymin=-1,ymax=1,
		        scatter/classes={a={mark=o,draw=black}},
		        width=\squ,
		        height=\squ,
		        xticklabels={,,},
		        yticklabels={,,},
		        ticks=none
		       ]
		
				\draw[-latex, black] (axis cs:0,0) -- (axis cs:-0.45,0.8) node[midway,lab]{$\mathbf{e}_{\texttt{A.}}$};
				
				\draw[-latex, black] (axis cs:0,0) -- (axis cs:0.45,0.8) node[midway,lab]{$\mathbf{e}_{\texttt{T.}}$};
				
				\draw[-latex, black] (axis cs:0,0) -- (axis cs:-0.45,0.0) node[midway,lab]{$\mathbf{e}_{\texttt{V.}}$};
				
				\draw[-latex, black] (axis cs:0,0) -- (axis cs:0.45,0.0) node[midway,lab]{$\mathbf{e}_{\texttt{S.}}$};		
				
				\draw[-latex, black] (axis cs:0,0) -- (axis cs:-0.45,-0.8) node[midway,lab]{$\mathbf{e}_{\texttt{L.}}$};
				
				\draw[-latex, black] (axis cs:0,0) -- (axis cs:0.45,-0.8) node[midway,lab]{$\mathbf{e}_{\texttt{C.}}$};	
				
				\draw[-latex, black, dotted] (axis cs:-0.45,0.8) -- (axis cs:0.45,0.8) node[midway,lab]{$\mathbf{r}_{\gelab{wo.}}$};					
		        \end{axis}
		\end{tikzpicture}
		\caption{Entity embeddings}\label{fig:transeEE}
	\end{subfigure}
	\caption{Toy example of two-dimensional relation and entity embeddings learnt by TransE; the entity embeddings use abbreviations and include an example of vector addition to predict what is west of Antofagasta \label{fig:transE}}
\end{figure}
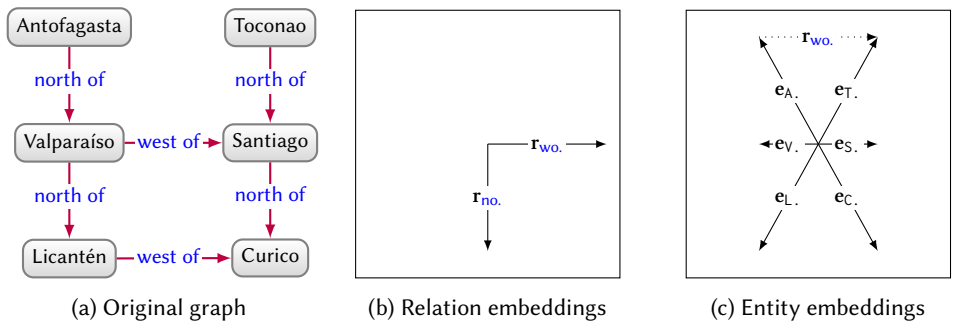

\emph{Translational models} interpret edge labels as transformations from subject nodes (aka the \textit{source} or \textit{head}) to object nodes (aka the \textit{target} or \textit{tail}); for example, in the edge \gedge[arrin][1cm]{San Pedro}{bus}{Moon Valley}, the edge label \gelab{bus} is seen as transforming \gnode{San Pedro} to \gnode{Moon Valley}, and likewise for other \gelab{bus} edges. The most elementary approach in this family is TransE~\cite{bordes2013translating}. Over all positive edges \gedge[arrin][0.6cm]{s}{p}{o}, TransE learns vectors $\ee_\texttt{s}$, $\re_\gelab{p}$, and $\ee_\texttt{o}$ aiming to make $\ee_\texttt{s}+\re_\gelab{p}$ as close as possible to $\ee_\texttt{o}$. Conversely, if the edge is a negative example, TransE attempts to learn a representation that keeps $\ee_\texttt{s} + \re_\gelab{p}$ away from $\ee_\texttt{o}$. To illustrate, Figure~\ref{fig:transE} provides a toy example of two-dimensional ($d = 2$) entity and relation embeddings computed by TransE. We keep the orientation of the vectors similar to the original graph for clarity. For any edge \gedge[arrin][0.6cm]{s}{p}{o} in the original graph, adding the vectors $\ee_\texttt{s} + \re_\gelab{p}$ should approximate $\ee_\texttt{o}$. In this toy example, the vectors correspond precisely where, for instance, adding the vectors for \gnode{Licantén} ($\ee_\texttt{L.}$) and $\gelab{west of}$ ($\re_\gelab{wo.}$) gives a vector corresponding to \gnode{Curico} ($\ee_\texttt{C.}$). We can use these embeddings to predict edges (among other tasks); for example, in order to predict which node in the graph is most likely to be \gelab{west of} \gnode{Antofagasta} (\texttt{A.}), by computing $\ee_\texttt{A.}+\re_\gelab{wo.}$ we find that the resulting vector (dotted in Figure~\ref{fig:transeEE}) is closest to $\ee_\texttt{T.}$, thus predicting \gnode{Toconao} (\texttt{T.}) to be the most \textit{plausible} such node.

Aside from this toy example, TransE can be too simplistic; for example, in Figure~\ref{fig:chileTransport}, \gelab{bus} not only transforms \gnode{San Pedro} to \gnode{Moon Valley}, but also to \gnode{Arica}, \gnode{Calama}, and so forth. TransE will, in this case, aim to give similar vectors to all such target locations, which may not be feasible given other edges. TransE will also tend to assign cyclical relations a zero vector, as the directional components will tend to cancel each other out. To resolve such issues, many variants of TransE have been investigated. Amongst these, for example, TransH~\cite{wang2014knowledge} represents different relations using distinct hyperplanes, where for the edge \gedge[arrin][0.6cm]{s}{p}{o}, \gnode{s} is first projected onto the hyperplane of \gelab{p} before the translation to \gnode{o} is learnt (uninfluenced by edges with other labels for \gnode{s} and for \gnode{o}). TransR~\cite{lin2015learning} generalises this approach by projecting \gnode{s} and \gnode{o} into a vector space specific to \gelab{p}, which involves multiplying the entity embeddings for \gnode{s} and \gnode{o} by a projection matrix specific to \gelab{p}. TransD~\cite{TransD} simplifies TransR by associating entities and relations with a second vector, where these secondary vectors are used to project the entity into a relation-specific vector space. Recently, RotatE~\cite{SunDNT19} proposes translational embeddings in complex space, which allows to capture more characteristics of relations, such as direction, symmetry, inversion, antisymmetry, and composition. Embeddings have also been proposed in non-Euclidean space, e.g., MuRP~\cite{BalazevicAH19} uses relation embeddings that transform entity embeddings in the hyperbolic space of the Poincaré ball mode, whose curvature provides more ``space'' to separate entities with respect to the dimensionality. For discussion of other translational models, we refer to the survey by \citet{Wang2017KGEmbedding}.

\subsubsection{Tensor decomposition models}

A second approach to derive graph embeddings is to apply methods based on \textit{tensor decomposition}. A \textit{tensor} is a multidimensional numeric field that generalises scalars (0-order tensors), vectors (1-order tensors) and matrices (2-order tensors) towards arbitrary dimension/order. Tensors have become a widely used abstraction for machine learning~\cite{RabanserSG17}. Tensor decomposition involves decomposing a tensor into more ``elemental'' tensors (e.g., of lower order) from which the original tensor can be recomposed (or approximated) by a fixed sequence of basic operations. These elemental tensors can be viewed as capturing \textit{latent factors} underlying the information contained in the original tensor. There are many approaches to tensor decomposition, where we will now briefly introduce the main ideas behind \textit{rank decompositions}~\cite{RabanserSG17}.

Leaving aside graphs momentarily, consider an $(a,b)$-matrix (i.e., a 2-order tensor) $\mathbf{C}$, where $a$ is the number of cities in Chile, $b$ is the number of months in a year, and each element $(\mathbf{C})_{ij}$ denotes the average temperature of the $i$\textsuperscript{th} city in the $j$\textsuperscript{th} month. Noting that Chile is a long, thin country -- ranging from subpolar climates in the south, to a desert climate in the north -- we may find a decomposition of $\mathbf{C}$ into two vectors representing latent factors -- specifically $\vect{x}$ (with $a$ elements) giving lower values for cities with lower latitude, and $\vect{y}$ (with $b$ elements), giving lower values for months with lower temperatures -- such that computing the outer product\footnote{The outer product of two (column) vectors $\vect{x}$ of length $a$ and $\vect{y}$ of length $b$, denoted $\vect{x} \otimes \vect{y}$, is defined as $\vect{x}\vect{y}^{\mathrm{T}}$, yielding an $(a,b)$-matrix $\mathbf{M}$ such that $(\mathbf{M})_{ij} = (\vect{x})_i  \cdot (\vect{y})_j$. Analogously, the outer product of $k$ vectors is a $k$-order tensor.} of the two vectors approximates $\mathbf{C}$ reasonably well: $\vect{x} \otimes \vect{y} \approx \mathbf{C}$. In the (unlikely) case that there exist vectors $\vect{x}$ and $\vect{y}$ such that $\mathbf{C}$ is precisely the outer product of two vectors ($\vect{x} \otimes \vect{y} = \mathbf{C}$) we call $\mathbf{C}$ a rank-1 matrix; we can then precisely encode $\mathbf{C}$ using $a + b$ values rather than $a \times b$ values. Most times, however, to get precisely $\mathbf{C}$, we will need to sum multiple rank-1 matrices, where the rank $r$ of $\mathbf{C}$ is the minimum number of rank-1 matrices that need to be summed to derive precisely $\mathbf{C}$, such that $\vect{x}_1 \otimes \vect{y}_1 + \ldots \vect{x}_r \otimes \vect{y}_r = \mathbf{C}$. In the temperature example, $\vect{x}_2 \otimes \vect{y}_2$ might correspond to a correction for altitude, $\vect{x}_3 \otimes \vect{y}_3$ for higher temperature variance further south, etc. A (low) rank decomposition of a matrix then sets a limit $d$ on the rank and computes the vectors $(\vect{x}_1,\vect{y}_1,\ldots,\vect{x}_{d},\vect{y}_{d})$ such that $\vect{x}_1 \otimes \vect{y}_1 + \ldots + \vect{x}_{d} \otimes \vect{y}_{d}$ gives the best $d$-rank approximation of $\mathbf{C}$. Noting that to generate $n$-order tensors we need to compute the outer product of $n$ vectors, we can generalise this idea towards low rank decomposition of tensors; this method is called Canonical Polyadic (CP) decomposition~\cite{Hitchcock27}. For example, we might have a 3-order tensor $\mathcal{C}$ containing monthly temperatures for Chilean cities \textit{at four different times of day}, which could be approximated with $\vect{x}_1 \otimes \vect{y}_1 \otimes \vect{z}_1 + \ldots \vect{x}_{d} \otimes \vect{y}_{d} \otimes \vect{z}_{d}$ (e.g., $\vect{x}_1$ might be a latitude factor, $\vect{y}_1$ a monthly variation factor, and $\vect{z}_1$ a daily variation factor, and so on). Various algorithms then exist to compute (approximate) CP decompositions, including Alternating Least Squares, Jennrich's Algorithm, and the Tensor Power method~\cite{RabanserSG17}.

\begin{figure}
\centering
\begin{tikzpicture}   
\tikzcuboid{
 dimx=6,
 dimy=6,
 dimz=2,
 scalex=0.3,
 scaley=0.3,
 scalez=0.2,
 front/.style={draw=green!75!black,fill=green!10!white},%
 right/.style={draw=red!25!black,fill=red!10!white},%
 top/.style={draw=blue!50!black,fill=blue!10!white}
};

\tikzset{every node/.style={font=\scriptsize}}
\tikzset{
  vi/.style={font=\scriptsize\ttfamily}
}

\node[vi] at (0.15,1.93) {A.};
\node[vi] at (0.45,1.93) {C.};
\node[vi] at (0.75,1.93) {L.};
\node[vi] at (1.05,1.93) {S.};
\node[vi] at (1.35,1.93) {T.};
\node[vi] at (1.65,1.93) {V.};

\node[vi] at (-0.45,1.35) {A.};
\node[vi] at (-0.45,1.05) {C.};
\node[vi] at (-0.45,0.75) {L.};
\node[vi] at (-0.45,0.45) {S.};
\node[vi] at (-0.45,0.15) {T.};
\node[vi] at (-0.45,-0.15) {V.};

\node[vi,anchor=east,xslant=1,inner sep=0] (wo) at (-0.3,1.57) {\color{blue}\tiny west of};

\node[vi,anchor=south east,xslant=1,inner sep=0] at (wo.north east) {\color{blue}\tiny north of};

\node at (-0.15,-0.15) {0}; 
\node at (0.15,-0.15) {0};  
\node at (0.45,-0.15) {0};  
\node at (0.75,-0.15) {1};  
\node at (1.05,-0.15) {0};  
\node at (1.35,-0.15) {0};  

\node at (-0.15,0.15) {0}; 
\node at (0.15,0.15) {0};  
\node at (0.45,0.15) {0};  
\node at (0.75,0.15) {0};  
\node at (1.05,0.15) {0};  
\node at (1.35,0.15) {0};  

\node at (-0.15,0.45) {0}; 
\node at (0.15,0.45) {0};  
\node at (0.45,0.45) {0};  
\node at (0.75,0.45) {0};  
\node at (1.05,0.45) {0};  
\node at (1.35,0.45) {0};  

\node at (-0.15,0.75) {0}; 
\node at (0.15,0.75) {1};  
\node at (0.45,0.75) {0};  
\node at (0.75,0.75) {0};  
\node at (1.05,0.75) {0};  
\node at (1.35,0.75) {0};  

\node at (-0.15,1.05) {0}; 
\node at (0.15,1.05) {0};  
\node at (0.45,1.05) {0};  
\node at (0.75,1.05) {0};  
\node at (1.05,1.05) {0};  
\node at (1.35,1.05) {0};  

\node at (-0.15,1.35) {0}; 
\node at (0.15,1.35) {0};  
\node at (0.45,1.35) {0};  
\node at (0.75,1.35) {0};  
\node at (1.05,1.35) {0};  
\node at (1.35,1.35) {0};  

\begin{scope}[shift={(4.5,1.5)}] 
\begin{scope}[shift={(0,0.155)}]
\tikzcuboid{
 dimx=6,
 dimy=1,
 dimz=1,
 scalex=0.3,
 scaley=0.3,
 scalez=0.2,
 front/.style={draw=green!75!black,fill=green!10!white},%
 right/.style={draw=red!25!black,fill=red!10!white},%
 top/.style={draw=blue!50!black,fill=blue!10!white}
};

\tikzset{every node/.style={font=\scriptsize}}
\tikzset{
  vi/.style={font=\scriptsize\ttfamily}
}

\end{scope}

\begin{scope}[shift={(-0.7,-2.0)}]
\tikzcuboid{
 dimx=1,
 dimy=6,
 dimz=1,
 scalex=0.3,
 scaley=0.3,
 scalez=0.2,
 front/.style={draw=green!75!black,fill=green!10!white},%
 right/.style={draw=red!25!black,fill=red!10!white},%
 top/.style={draw=blue!50!black,fill=blue!10!white}
};

\tikzset{every node/.style={font=\scriptsize}}
\tikzset{
  vi/.style={font=\scriptsize\ttfamily}
}

\end{scope}

\begin{scope}[shift={(-0.55,0.3)}]
\tikzcuboid{
 dimx=1,
 dimy=1,
 dimz=2,
 scalex=0.3,
 scaley=0.3,
 scalez=0.2,
 front/.style={draw=green!75!black,fill=green!10!white},%
 right/.style={draw=red!25!black,fill=red!10!white},%
 top/.style={draw=blue!50!black,fill=blue!10!white}
};

\tikzset{every node/.style={font=\scriptsize}}
\tikzset{
  vi/.style={font=\scriptsize\ttfamily}
}

\end{scope}

\node (x1) at (-1.2,-1.2) {\Large $\vect{x}_1$};
\node (z1) at (0.93,0.8) {\Large $\vect{z}_1$};
\node (y1) at (x1|-z1) {\Large $\vect{y}_1$};
\end{scope}

\begin{scope}[shift={(10.5,1.5)}] 
\begin{scope}[shift={(0,0.155)}]
\tikzcuboid{
 dimx=6,
 dimy=1,
 dimz=1,
 scalex=0.3,
 scaley=0.3,
 scalez=0.2,
 front/.style={draw=green!75!black,fill=green!10!white},%
 right/.style={draw=red!25!black,fill=red!10!white},%
 top/.style={draw=blue!50!black,fill=blue!10!white}
};

\end{scope}

\begin{scope}[shift={(-0.7,-2.0)}]
\tikzcuboid{
 dimx=1,
 dimy=6,
 dimz=1,
 scalex=0.3,
 scaley=0.3,
 scalez=0.2,
 front/.style={draw=green!75!black,fill=green!10!white},%
 right/.style={draw=red!25!black,fill=red!10!white},%
 top/.style={draw=blue!50!black,fill=blue!10!white}
};

\end{scope}

\begin{scope}[shift={(-0.55,0.3)}]
\tikzcuboid{
 dimx=1,
 dimy=1,
 dimz=2,
 scalex=0.3,
 scaley=0.3,
 scalez=0.2,
 front/.style={draw=green!75!black,fill=green!10!white},%
 right/.style={draw=red!25!black,fill=red!10!white},%
 top/.style={draw=blue!50!black,fill=blue!10!white}
};

\end{scope}

\node (xr) at (-1.2,-1.2) {\Large $\vect{x}_d$};
\node (zr) at (0.93,0.8) {\Large $\vect{z}_d$};
\node (yr) at (xr|-zr) {\Large $\vect{y}_d$};
\end{scope}

\node (ap) at (2.5,0.65) {\Huge $\approx$};
\node[right=1.8cm of ap] (ot1) {\Huge $\otimes$};

\node[right=1cm of ot1] (plus) {\Huge $+ \ldots +$};

\node[right=2cm of plus] (otr) {\Huge $\otimes$};

\node (g) at (0.5,-1.1) {\Large $\mathcal{G}$};

\node (aps) at (g-|ap) {\Large $\approx$};

\node (xyz1) at (g-|ot1) {\Large $\vect{x}_1 \oplus \vect{y}_1 \oplus \vect{z}_1$};

\node (pluss) at (g-|plus) {\Large  $+ \ldots +$};

\node (xyzr) at (g-|otr) {\Large $\vect{x}_d \oplus \vect{y}_d \oplus \vect{z}_d$};
\end{tikzpicture}

\caption{Abstract illustration of a CP $d$-rank decomposition of a tensor representing the graph of Figure~\ref{fig:distEg}} \label{fid:cpRank}
\end{figure}
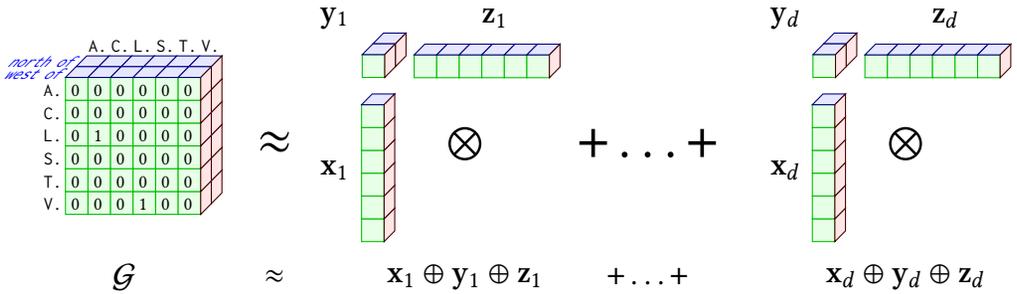

Returning to graphs, similar principles can be used to decompose a graph into vectors, thus yielding embeddings. In particular, a graph can be encoded as a one-hot 3-order tensor $\mathcal{G}$ with $|V| \times |L| \times |V|$ elements, where the element $(\mathcal{G})_{ijk}$ is set to one if the $i$\textsuperscript{th} node links to the $k$\textsuperscript{th} node with an edge having the $j$\textsuperscript{th} label, or zero otherwise. As previously mentioned, such a tensor will typically be very large and sparse, where rank decompositions are thus applicable. A CP decomposition~\cite{Hitchcock27} would compute a sequence of vectors $(\vect{x}_1,\vect{y}_1,\vect{z}_1,\ldots,\vect{x}_d,\vect{y}_d,\vect{z}_d)$ such that $\vect{x}_1 \otimes \vect{y}_1 \otimes \vect{z}_1 + \ldots + \vect{x}_d \otimes \vect{y}_d \otimes \vect{z}_d \approx \mathcal{G}$. We illustrate this scheme in Figure~\ref{fid:cpRank}. Letting $\mathbf{X}, \mathbf{Y}, \textbf{Z}$ denote the matrices formed by $\begin{bmatrix} \vect{x}_1\,\cdots\,\vect{x}_d \end{bmatrix}$, $\begin{bmatrix} \vect{y}_1\,\cdots\,\vect{y}_d \end{bmatrix}$, $\begin{bmatrix} \vect{z}_1\,\cdots\,\vect{z}_d \end{bmatrix}$, respectively, with each vector forming a column of the corresponding matrix, we could then extract the $i$\textsuperscript{th} row of $\mathbf{Y}$ as an embedding for the $i$\textsuperscript{th} relation, and the $j$\textsuperscript{th} rows of $\mathbf{X}$ and $\mathbf{Z}$ as \textit{two} embeddings for the $j$\textsuperscript{th} entity. However, knowledge graph embeddings typically aim to assign \textit{one} vector to each entity.

DistMult~\cite{distmult} is a seminal method for computing knowledge graph embeddings based on rank decompositions, where each entity and relation is associated with a vector of dimension $d$, such that for an edge \gedge[arrin][0.6cm]{s}{p}{o}, a plausibility scoring function $\sum_{i=1}^d (\ee_{\texttt{s}})_i (\re_{\gelab{p}})_i (\ee_{\texttt{o}})_i$ is defined, where $(\ee_{\texttt{s}})_i$, $(\re_{\gelab{p}})_i$ and $(\ee_{\texttt{o}})_i$ denote the $i$\textsuperscript{th} elements of vectors $\ee_\texttt{s}$, $\re_\gelab{p}$, $\ee_\texttt{o}$, respectively. The goal, then, is to learn vectors for each node and edge label that maximise the plausibility of positive edges and minimise the plausibility of negative edges. This approach equates to a CP decomposition of the graph tensor $\mathcal{G}$, but where entities have one vector that is used twice: $\vect{x}_1 \otimes \vect{y}_1 \otimes \vect{x}_1 + \ldots + \vect{x}_d \otimes \vect{y}_d \otimes \vect{x}_d \approx \mathcal{G}$. A weakness of this approach is that per the scoring function, the plausibility of \gedge[arrin][0.6cm]{s}{p}{o} will always be equal to that of \gedge[arrin][0.6cm]{o}{p}{s}; in other words, DistMult does not consider edge direction. 

Rather than use a vector as a relation embedding, RESCAL~\cite{nickel2013tensor} uses a matrix, which allows for combining values from $\ee_\texttt{s}$ and $\ee_\texttt{o}$ across all dimensions, and thus can capture (e.g.) edge direction. However, RESCAL incurs a higher cost in terms of space and time than DistMult. HolE~\cite{NickelRP16} uses vectors for relation and entity embeddings, but proposes to use the \textit{circular
correlation operator} -- which takes sums along the diagonals of the outer product of two vectors -- to combine them. This operator is not commutative, and can thus consider edge direction. ComplEx~\cite{TrouillonWRGB16}, on the other hand, uses a complex vector (i.e., a vector containing complex numbers) as a relational embedding, which similarly allows for breaking the aforementioned symmetry of DistMult's scoring function while keeping the number of parameters low. SimplE~\cite{Kazemi018} rather proposes to compute a standard CP decomposition computing two initial vectors for entities from $\mathbf{X}$ and $\mathbf{Z}$ and then averaging terms across $\mathbf{X}$, $\mathbf{Y}$, $\mathbf{Z}$ to compute the final plausibility scores. TuckER~\cite{BalazevicAH19a} employs a different type of decomposition -- called a Tucker Decomposition~\cite{tucker64extension}, which computes a smaller ``core'' tensor $\mathcal{T}$ and a sequence of three matrices $\mathbf{A}$, $\mathbf{B}$ and $\mathbf{C}$, such that $\mathcal{G} \approx \mathcal{T} \otimes \mathbf{A} \otimes \mathbf{B} \otimes \mathbf{C}$ -- where entity embeddings are taken from $\mathbf{A}$ and $\mathbf{C}$, while relation embeddings are taken from $\mathbf{B}$. Of these approaches, TuckER~\cite{BalazevicAH19a} currently provides state-of-the-art results on standard benchmarks.

\subsubsection{Neural models}

A limitation of the previously discussed approaches is that they assume either linear (preserving addition and scalar multiplication) or bilinear (e.g., matrix multiplication) operations over embeddings to compute plausibility scores. A number of approaches rather use neural networks to learn embeddings with non-linear scoring functions for plausibility.

One of the earliest proposals of a neural model was Semantic Matching Energy (SME)~\cite{GlorotBWB13}, which learns parameters (aka weights: $\mathbf{w}$, $\mathbf{w}'$) for two functions -- $f_{\mathbf{w}}(\ee_\texttt{s},\re_\gelab{p})$ and $g_{\mathbf{w}'}(\ee_\texttt{o},\re_\gelab{p})$ -- such that the dot product of the result of both functions -- $f_{\mathbf{w}}(\ee_\texttt{s},\re_\gelab{p}) \cdot g_{\mathbf{w}'}(\ee_\texttt{o},\re_\gelab{p})$ -- gives the plausibility score. Both linear and bilinear variants of $f_{\mathbf{w}}$ and $g_{\mathbf{w}'}$ are proposed. Another early proposal was Neural Tensor Networks (NTN)~\cite{socher2013reasoning}, which rather proposes to maintain a tensor $\mathcal{W}$ of internal weights, such that the plausibility score is computed by a complex function that combines the outer product $\ee_\texttt{s} \otimes \mathcal{W} \otimes \ee_\texttt{o}$ with a standard neural layer over $\ee_\texttt{s}$ and $\ee_\texttt{o}$, which in turn is combined with $\re_\gelab{p}$, to produce a plausibility score. The use of the tensor $\mathcal{W}$ results in a high number of parameters, which limits scalability~\cite{Wang2017KGEmbedding}. Multi Layer Perceptron (MLP)~\cite{DongGHHLMSSZ14} is a simpler model, where $\ee_\texttt{s}$, $\re_\gelab{p}$ and $\ee_\texttt{o}$ are concatenated and fed into a hidden layer to compute the plausibility score. 

A number of more recent approaches have proposed using convolutional kernels in their models. ConvE~\cite{DettmersMS018} proposes to generate a matrix from $\ee_\texttt{s}$ and $\re_\gelab{p}$ by ``wrapping'' each vector over several rows and concatenating both matrices. The concatenated matrix serves as the input for a set of (2D) convolutional layers, which returns a feature map tensor. The feature map tensor is vectorised and projected into $d$ dimensions using a parametrised linear transformation. The plausibility score is then computed based on the dot product of this vector and $\ee_\texttt{o}$. A disadvantage of ConvE is that by wrapping vectors into matrices, it imposes an artificial two-dimensional structure on the embeddings. HypER~\cite{BalazevicAH19b} is a similar model using convolutions, but avoids the need to wrap vectors into matrices. Instead, a fully connected layer (called the ``hypernetwork'') is applied to $\re_\gelab{p}$ and used to generate a matrix of relation-specific convolutional filters. These filters are applied directly to $\ee_\texttt{s}$ to give a feature map, which is vectorised. The same process is then applied as in ConvE: the resulting vector is projected into $d$ dimensions, and a dot product applied with $\ee_\texttt{o}$ to produce the plausibility score. The resulting model is shown to outperform ConvE on standard benchmarks~\cite{BalazevicAH19b}.

The presented approaches strike different balances in terms of expressivity and the number of parameters that need to be trained. While more expressive models, such as NTN, may better fit more complex plausibility functions over lower dimensional embeddings by using more hidden parameters, simpler models, such as that proposed by Dong et al.~\cite{DongGHHLMSSZ14}, and convolutional networks~\cite{DettmersMS018,BalazevicAH19b} that enable parameter sharing by applying the same (typically small) kernels over different regions of a matrix, require handling fewer parameters overall and are more scalable. 

\subsubsection{Language models}

Embedding techniques were first explored as a way to represent natural language within machine learning frameworks, with word2vec~\cite{mikolov2013efficient} and GloVe~\cite{pennington2014glove} being two seminal approaches. Both approaches compute embeddings for words based on large corpora of text such that words used in similar contexts (e.g., ``\texttt{frog}'', ``\texttt{toad}'') have similar vectors. Word2vec uses neural networks trained either to predict the current word from surrounding words (\textit{continuous bag of words}), or to predict the surrounding words given the current word (\textit{continuous skip-gram}). GloVe rather applies a regression model over a matrix of co-occurrence probabilities of word pairs. Embeddings generated by both approaches are widely used in natural language processing tasks.

Another approach for graph embeddings is thus to leverage proven approaches for language embeddings. However, while a graph consists of an unordered set of sequences of three terms (i.e., a set of edges), text in natural language consists of arbitrary-length sequences of terms (i.e., sentences of words). Along these lines, RDF2Vec~\cite{ristoski2016rdf2vec} performs (biased~\cite{cochez2017biased}) random walks on the graph and records the paths (the sequence of nodes and edge labels traversed) as ``sentences'', which are then fed as input into the word2vec~\cite{mikolov2013efficient} model. An example of such a path extracted from Figure~\ref{fig:chileTransport} might be, for example, \begin{tikzpicture}[baseline=-3pt]
    \setlength{\hgap}{1.1cm}
	\node[iri,compact](sp){San Pedro};

	\node[iri,compact,right=0.8\hgap of sp](c){Calama}
		edge[arrin] node[lab] {bus} (sp);
	
	\node[iri,compact,right=\hgap of c](i){Iquique}
		edge[arrin] node[lab] {flight} (c);
		
	\node[iri,compact,right=\hgap of i](s){Santiago}
		edge[arrin] node[lab] {flight} (i);
\end{tikzpicture}, where the paper experiments with 500 paths of length 8 per entity. RDF2Vec also proposes a second mode where sequences are generated for nodes from canonically-labelled sub-trees of which they are a root node, where the paper experiments with sub-trees of depth 1 and 2. Conversely, KGloVe~\cite{cochez2017global} is based on the GloVe model. Much like how the original GloVe model~\cite{pennington2014glove} considers words that co-occur frequently in windows of text to be more related, KGloVe uses personalised PageRank\footnote{Intuitively speaking, personalised PageRank starts at a given node and then determines the probability of a random walk being at a particular node after a given number of steps. A higher number of steps converges towards standard PageRank emphasising global node centrality, while a lower number emphasises proximity/relatedness to the starting node.} to determine the most related nodes to a given node, whose results are then fed into the GloVe model.

\subsubsection{Entailment-aware models} The embeddings thus far consider the data graph alone. But what if an ontology or set of rules is provided? Such deductive knowledge could be used to improve the embeddings. One approach is to use constraint rules to refine the predictions made by embeddings; for example,~\citet{WangWG15} use functional and inverse-functional definitions as constraints (under UNA) such that, for example, if we define that an event can have at most one value for \gelab{venue}, this is used to lower the plausibility of edges that would assign multiple venues to an event. 

More recent approaches rather propose joint embeddings that consider both the data graph and rules when computing embeddings. KALE~\cite{GuoWWWG16} computes entity and relation embeddings using a translational model (specifically TransE) that is adapted to further consider rules using \textit{t-norm fuzzy logics}. With reference to Figure~\ref{fig:chileTransport}, consider a simple rule \begin{tikzpicture}[baseline=-3pt]
\node[var,compact](x){?x};

\node[var,compact,right=0.7\fhgap of x](y){?y}
edge[arrin] node[lab] {bus} (x);
\end{tikzpicture} $\Rightarrow$ 
\begin{tikzpicture}[baseline=-3pt]
\node[var,compact](x){?x};

\node[var,compact,right=1.3\fhgap of x](e){?y}
edge[arrin] node[lab] {connects to} (x);
\end{tikzpicture}. We can use embeddings to assign plausibility scores to new edges, such as $e_1$: \gedge[arrin][0.7\fhgap]{Piedras Rojas}{bus}{Moon Valley}. We can further apply the previous rule to generate a new edge $e_2$: \gedge[arrin][1.3\fhgap]{Piedras Rojas}{connects to}{Moon Valley} from the predicted edge $e_1$. But what plausibility should we assign to this second edge? Letting $p_1$ and $p_2$ be the current plausibility scores of $e_1$ and $e_2$ (initialised using the standard embedding), then t-norm fuzzy logics suggests that the plausibility be updated as $p_1p_2 - p_1 + 1$. Embeddings are then trained to jointly assign larger plausibility scores to positive examples versus negative examples of both edges and \textit{ground rules}. An example of a positive ground rule based on Figure~\ref{fig:chileTransport} would be \begin{tikzpicture}[baseline=-3pt]
\node[iri,compact](x){Arica};

\node[iri,compact,right=0.7\fhgap of x](y){San Pedro}
edge[arrin] node[lab] {bus} (x);
\end{tikzpicture} $\Rightarrow$ 
\begin{tikzpicture}[baseline=-3pt]
\node[iri,compact](x){Arica};

\node[iri,compact,right=1.3\fhgap of x](e){San Pedro}
edge[arrin] node[lab] {connects to} (x);
\end{tikzpicture}. Negative ground rules randomly replace the relation in the head of the rule; for example, \begin{tikzpicture}[baseline=-3pt]
\node[iri,compact](a){Arica};

\node[iri,compact,right=0.7\fhgap of x](y){San Pedro}
edge[arrin] node[lab] {bus} (x);
\end{tikzpicture} $\not\Rightarrow$ 
\begin{tikzpicture}[baseline=-3pt]
\node[iri,compact](x){Arica};

\node[iri,compact,right=0.9\fhgap of x](e){San Pedro}
edge[arrin] node[lab] {flight} (x);
\end{tikzpicture}. \citet{GuoWWWG18} later propose RUGE, which uses a joint model over ground rules (possibly soft rules with confidence scores) and plausibility scores to align both forms of scoring for unseen edges.

Generating ground rules can be costly. An alternative approach, called FSL~\cite{DemeesterRR16}, observes that in the case of a simple rule, such as \begin{tikzpicture}[baseline=-3pt]
\node[var,compact](x){?x};

\node[var,compact,right=0.7\fhgap of x](y){?y}
edge[arrin] node[lab] {bus} (x);
\end{tikzpicture} $\Rightarrow$ 
\begin{tikzpicture}[baseline=-3pt]
\node[var,compact](x){?x};

\node[var,compact,right=1.3\fhgap of x](e){?y}
edge[arrin] node[lab] {connects to} (x);
\end{tikzpicture}, the relation embedding \gelab{bus} should always return a lower plausibility than \gelab{connects to}. Thus, for all such rules, FSL proposes to train relation embeddings while avoiding violations of such inequalities. While relatively straightforward, FSL only supports simple rules, while KALE also supports more complex rules.

These works are interesting examples of how deductive and inductive forms of knowledge -- in this case rules and embeddings -- can interplay and complement each other.

\subsection{Graph Neural Networks}\label{ssec:gnns}

While embeddings aim to provide a dense numerical representation of graphs suitable for use within existing machine learning models, another approach is to build custom machine learning models adapted for graph-structured data. Most custom learning models for graphs are based on (artificial) neural networks~\cite{abs-1901-00596}, exploiting a natural correspondence between both: a neural network already corresponds to a weighted, directed graph, where nodes serve as artificial neurons, and edges serve as weighted connections (axons). However, the typical topology of a traditional neural network -- more specifically, a fully-connected feed-forward neural network -- is quite homogeneous, being defined in terms of sequential layers of nodes where each node in one layer is connected to all nodes in the next layer. Conversely, the topology of a data graph is quite heterogeneous, being determined by the relations between entities that its edges represent.

A \textit{graph neural network} (GNN)~\cite{ScarselliGTHM09} builds a neural network based on the topology of the data graph; i.e., nodes are connected to their neighbours per the data graph. Typically a model is then learnt to map input features for nodes to output features in a supervised manner; output features for example nodes may be manually labelled, or may be taken from the knowledge graph. Unlike knowledge graphs embeddings, GNNs support end-to-end supervised learning for specific tasks: given a set of labelled examples, GNNs can be used to classify elements of the graph or the graph itself. GNNs have been used to perform classification over graphs encoding compounds, objects in images, documents, etc.; as well as to predict traffic, build recommender systems, verify software, etc.~\cite{abs-1901-00596}. Given labelled examples, GNNs can even replace graph algorithms; for example, GNNs have been used to find central nodes in knowledge graphs in a supervised manner~\cite{ScarselliGTHM09,ParkKDZF19,ParkKDZF20}.

We now discuss the ideas underlying two flavours of GNN -- \textit{recursive GNNs} and \textit{convolutional GNNs} -- where we refer to Appendix~\ref{app:gnns} for more formal definitions relating to GNNs.

\subsubsection{Recursive graph neural networks}

Recursive graph neural networks (RecGNNs) are the seminal approach to graph neural networks~\cite{SperdutiS97,ScarselliGTHM09}. The approach is conceptually similar to the systolic abstraction illustrated in Figure~\ref{fig:pagerank}, where messages are passed between neighbours towards recursively computing some result. However, rather than define the functions used to decide the messages to pass, we rather label the output of a training set of nodes and let the framework learn the functions that generate the expected output, thereafter applying them to label other examples.

In a seminal paper, \citet{ScarselliGTHM09} proposed what they generically call a graph neural network (GNN), which takes as input a directed graph where nodes and edges are associated with \textit{feature vectors} that can capture node and edge labels, weights, etc. These feature vectors remain fixed throughout the process. Each node in the graph is also associated with a \textit{state vector}, which is recursively updated based on information from the node's neighbours -- i.e., the feature and state vectors of the neighbouring nodes and the feature vectors of the edges extending to/from them -- using a parametric function, called the \textit{transition function}. A second parametric function, called the \textit{output function}, is used to compute the final output for a node based on its own feature and state vector. These functions are applied recursively up to a fixpoint. Both parametric functions can be implemented using neural networks where, given a partial set of \textit{supervised nodes} in the graph -- i.e., nodes labelled with their desired output -- parameters for the transition and output functions can be learnt that best approximate the supervised outputs. The result can thus be seen as a recursive neural network architecture.\footnote{Some authors refer to such architectures as \textit{recurrent graph neural networks}, observing that the internal state maintained for nodes can be viewed as a form of recurrence over a sequence of transitions.} To ensure convergence up to a fixpoint, certain restrictions are applied, namely that the transition function be a \textit{contractor}, meaning that upon each application of the function, points in the numeric space are brought closer together (intuitively, in this case, the numeric space ``shrinks'' upon each application, ensuring a unique fixpoint).

To illustrate, consider, for example, that we wish to find priority locations for creating new tourist information offices. A good strategy would be to install them in hubs from which many tourists visit popular destinations. Along these lines, in Figure~\ref{fig:gnn} we illustrate the GNN architecture proposed by \citet{ScarselliGTHM09} for a sub-graph of Figure~\ref{fig:chileTransport}, where we highlight the neighbourhood of \gnode{Punta Arenas}. In this graph, nodes are annotated with feature vectors ($\mathbf{n}_x$) and hidden states at step $t$ ($\mathbf{h}_x^{(t)}$), while edges are annotated with feature vectors ($\mathbf{a}_{xy}$). Feature vectors for nodes may, for example, one-hot encode the type of node (\textit{City}, \textit{Attraction}, etc.), directly encode statistics such as the number of tourists visiting per year, etc. Feature vectors for edges may, for example, one-hot encode the edge label (the type of transport), directly encode statistics such as the distance or number of tickets sold per year, etc. Hidden states can be randomly initialised. The right-hand side of Figure~\ref{fig:gnn} provides the GNN transition and output functions, where $\mathrm{N}(x)$ denotes the neighbouring nodes of $x$,  $f_{\mathbf{w}}(\cdot)$ denotes the transition function with parameters $\mathbf{w}$, and $g_{\mathbf{w}'}(\cdot)$ denotes the output function with parameters $\mathbf{w'}$. An example is also provided for Punta Arenas ($x = 1$). These functions will be recursively applied until a fixpoint is reached. To train the network, we can label examples of places that already have (or should have) tourist offices and places that do (or should) not have tourist offices. These labels may be taken from the knowledge graph, or may be added manually. The GNN can then learn parameters $\mathbf{w}$ and $\mathbf{w'}$ that give the expected output for the labelled examples, which can subsequently be used to label other nodes.

This GNN model is flexible and can be adapted in various ways~\cite{ScarselliGTHM09}: we may define neighbouring nodes differently, for example to include nodes for outgoing edges, or nodes one or two hops away; we may allow pairs of nodes to be connected by multiple edges with different vectors; we may consider transition and output functions with distinct parameters for each node; we may add states and outputs for edges; we may change the sum to another aggregation function; etc.

\begin{figure}
	\setlength{\vgap}{0.9cm}
	\setlength{\hgap}{1.1cm}
	\setlength{\sgap}{0.1cm}
	\setlength{\mgap}{0.1cm}

	\centering
	\begin{tikzpicture}[baseline=-1pt]
	\node[iri,anchor=mid] (pm) {Puerto Montt$_{(3)}$};
	
	\node[feature,anchor=mid,left=\sgap of pm] (pms) {$\mathbf{n}_{3}, \mathbf{h}_{3}^{(t)}$};

	\node[iri,anchor=mid,above=\vgap of pm] (pa) {Punta Arenas$_{(1)}$}
	edge[arrin,bend left=40] node[feature,yshift=0.1cm] {$\mathbf{a}_{31}$} (pm)
	edge[arrout,faded,bend right=40] node[feature,yshift=-0.1cm] {$\mathbf{a}_{13}$} (pm);
	
	\node[feature,anchor=mid] (pas) at (pa-|pms) {$\mathbf{n}_{1}, \mathbf{h}_{1}^{(t)}$};

	\node[iri,faded,anchor=mid,below=\vgap of pm] (pv) {Puerto Varas$_{(5)}$}
	edge[arrin,faded,bend left=40] node[feature,yshift=-0.1cm] {$\mathbf{a}_{35}$} (pm)
	edge[arrout,faded,bend right=40] node[feature,yshift=0.1cm] {$\mathbf{a}_{53}$} (pm);
	
	\node[feature,anchor=mid,faded] at (pv-|pms) (pvs) {$\mathbf{n}_{5}, \mathbf{h}_{5}^{(t)}$};

	\node[iri,anchor=mid,faded,right=\hgap of pa] (tdp) {Torres del Paine$_{(2)}$}
	edge[arrin,faded] node[feature] {$\mathbf{a}_{12}$} (pa);
	
	\node[feature,anchor=mid,right=\sgap of tdp] (tdps) {$\mathbf{n}_{2}, \mathbf{h}_{2}^{(t)}$};

	\node[iri,anchor=mid,below=\vgap of tdp] (gg) {Grey Glacier$_{(4)}$}
	edge[arrin,faded] node[feature] {$\mathbf{a}_{24}$} (tdp)
	edge[arrout] node[feature] {$\mathbf{a}_{41}$} (pa);
	
	\node[feature,anchor=mid] at (gg-|tdps) (ggs) {$\mathbf{n}_{4}, \mathbf{h}_{4}^{(t)}$};

	\node[iri,faded,anchor=mid,below=\vgap of gg] (os) {Osorno Volcano$_{(6)}$}
	edge[arrin,faded,bend left=8] node[feature] {$\mathbf{a}_{56}$} (pv)
	edge[arrout,faded,bend right=8] node[feature] {$\mathbf{a}_{65}$} (pv);
	
	\node[feature,anchor=mid,faded] at (os-|tdps) (oss) {$\mathbf{n}_{6}, \mathbf{h}_{6}^{(t)}$};
	\end{tikzpicture}\hfill
	\begin{tabular}{l@{~}l}
	$\mathbf{h}_x^{(t)} \coloneq$ & $\sum_{y \in \textrm{N}(x)} f_\mathbf{w}(\mathbf{n}_{x},\mathbf{n}_{y},\mathbf{a}_{yx},\mathbf{h}_{y}^{(t-1)})$ \\
	$\mathbf{o}_x^{(t)} \coloneq$ & $g_{\mathbf{w}'}(\mathbf{h}_x^{(t)},\mathbf{n}_x)$\\[3ex]
	
	$\mathbf{h}_1^{(t)} \coloneq$ & $f_\mathbf{w}(\mathbf{n}_{1},\mathbf{n}_{3},\mathbf{a}_{31},\mathbf{h}_{3}^{(t-1)})$ \\ 
	& $+ f_\mathbf{w}(\mathbf{n}_{1},\mathbf{n}_{4},\mathbf{a}_{41},\mathbf{h}_{4}^{(t-1)})$ \\
	$\mathbf{o}_1^{(t)} \coloneq$ & $g_{\mathbf{w}'}(\mathbf{h}_1^{(t)},\mathbf{n}_1)$\\[1ex]
	\ldots \\
	\end{tabular}
		
	\caption{On the left a sub-graph of Figure~\ref{fig:chileTransport} highlighting the neighbourhood of Punta Arenas, where nodes are annotated with feature vectors ($\mathbf{n}_x$) and hidden states at step $t$ ($\mathbf{h}_x^{(t)}$), and edges are annotated with feature vectors ($\mathbf{a}_{xy}$); on the right, the GNN transition and output functions proposed by \citet{ScarselliGTHM09} and an example for Punta Arenas ($x = 1$), where $\mathrm{N}(x)$ denotes the neighbouring nodes of $x$, $f_{\mathbf{w}}(\cdot)$ denotes the transition function with parameters $\mathbf{w}$ and $g_{\mathbf{w}'}(\cdot)$ denotes the output function with parameters $\mathbf{w'}$ \label{fig:gnn}}
\end{figure}

\subsubsection{Convolutional graph neural networks}

Convolutional neural networks (CNNs) have gained a lot of attention, in particular, for machine learning tasks involving images~\cite{KrizhevskySH17}. The core idea in the image setting is to apply small kernels (aka filters) over localised regions of an image using a convolution operator to extract features from that local region. When applied to all local regions, the convolution outputs a feature map of the image. Typically multiple kernels are applied, forming multiple convolutional layers. These kernels can be learnt, given sufficient labelled examples.

One may note an analogy between GNNs as previously discussed, and CNNs as applied to images: in both cases, operators are applied over local regions of the input data. In the case of GNNs, the transition function is applied over a node and its neighbours in the graph. In the case of CNNs, the convolution is applied on a pixel and its neighbours in the image. Following this intuition, a number of \textit{convolutional graph neural networks} (\textit{ConvGNNs})~\cite{BrunaZSL13,KipfW17,abs-1901-00596} have been proposed, where the transition function is implemented by means of convolutions. A key consideration for ConvGNNs is how regions of a graph are defined. Unlike the pixels of an image, nodes in a graph may have varying numbers of neighbours. This creates a challenge: a benefit of CNNs is that the same kernel can be applied over all the regions of an image, but this requires more careful consideration in the case of ConvGNNs since neighbourhoods of different nodes can be diverse. Approaches to address these challenges involve working with spectral (e.g.~\cite{BrunaZSL13,KipfW17}) or spatial (e.g.,~\cite{MontiBMRSB17}) representations of graphs that induce a more regular structure from the graph. An alternative is to use an attention mechanism~\cite{VelickovicCCRLB18} to \textit{learn} the nodes whose features are most important to the current node.

Aside from architectural considerations, there are two main differences between RecGNNs and ConvGNNs. First, RecGNNs aggregate information from neighbours recursively up to a fixpoint, whereas ConvGNNs typically apply a fixed number of convolutional layers. Second, RecGNNs typically use the same function/parameters in uniform steps, while different convolutional layers of a ConvGNN can apply different kernels/weights at each distinct step.

\subsection{Symbolic Learning}\label{ssec:symlearn}

\begin{figure}
	\setlength{\vgap}{0.9cm}
	\setlength{\hgap}{3cm}
	\setlength{\sgap}{0.1cm}
	\setlength{\mgap}{0.1cm}	
	
	\begin{tikzpicture}
		\node[iri,anchor=mid] (sa) {Santiago};
		
		\node[iri,anchor=mid,below=\vgap of sa] (scl) {SCL}
		  edge[arrin] node[lab] {nearby} (sa);
		
		\node[iri,anchor=mid,right=\hgap of sa] (lma) {Lima};
		 
		\node[iri,anchor=mid,below=\vgap of lma] (lim) {LIM}
		  edge[arrin] node[lab] {nearby} (lma)
		  edge[arroutin,bend right=10] node[lab] {flight} (scl)
		  edge[arroutin,bend left=10] node[lab] {international flight} (scl);
		 
		\node[iri,anchor=mid,above=\vgap of lma] (per) {Peru}
          edge[arrout] node[lab] {capital} (lma)
          edge[arrin,bend right=55] node[lab] {country} (lim);
		  
		\node[iri,anchor=mid,left=\hgap of sa] (ar) {Arica};
		  
		\node[iri,anchor=mid,above=\vgap of ar] (cl) {Chile}
		  edge[arrout,bend left=8] node[lab] {capital} (sa)
		  edge[arrin,bend right=8] node[lab] {country} (sa)
		  edge[arrin] node[lab] {country} (ar);
		  
		\node[iri,anchor=mid,above=\vgap of sa] (co) {Country}
		  edge[arrin] node[lab] {type} (per)
		  edge[arrin] node[lab] {type} (cl);
		
	    \node[iri,anchor=mid,below=\vgap of ar] (ari) {ARI}
		  edge[arrin] node[lab] {nearby} (ar)
		  edge[arrout,bend left=10] node[lab] {flight} (scl)
		  edge[arrout,bend right=10] node[lab] {domestic flight} (scl)
		  edge[arrout,bend right=50] node[lab,xshift=0.2cm] {country} (cl);
		  
		\node[iri,anchor=mid,left=\hgap of ar] (iqu) {Iquique}
		  edge[arrout] node[lab] {country} (cl);
				
		\node[iri,anchor=mid,below=\vgap of iqu] (iqq) {IQQ}
		  edge[arrin] node[lab] {nearby} (iqu)
		  edge[arrout] node[lab] {country} (cl)
		  edge[arroutin,bend left=10] node[lab] {flight} (ari)
		  edge[arroutin,bend right=10] node[lab] {domestic flight} (ari);
		  
	    \node[iri,anchor=mid,below=\vgap of ari] (da) {Domestic Airport}
			edge[arrin,bend left=5] node[lab] {type} (iqq)
		    edge[arrin] node[lab] {type} (ari);

	    \node[iri,anchor=mid,below=\vgap of scl] (da) {International Airport}
			edge[arrin] node[lab] {type} (scl)
		    edge[arrin,bend right=5] node[lab] {type} (lim);
		
		\node[iri,anchor=mid,between=iqq and lim,yshift=-2.4\vgap] (a) {Airport}
			edge[arrin,bend left=14] node[lab] {type} (scl)
			edge[arrin,bend right=22,pos=0.6] node[lab] {type} (lim)
			edge[arrin,bend left=22,pos=0.6] node[lab] {type} (iqq)
			edge[arrin,bend right=14] node[lab] {type} (ari);
	\end{tikzpicture}
	
	\caption{An incomplete directed edge-labelled graph describing flights between airports \label{fig:airports}}	
\end{figure}
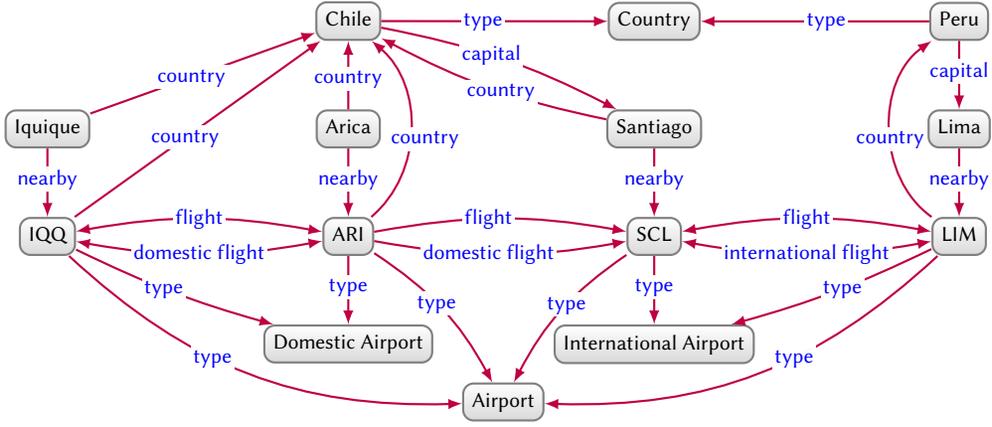

The supervised techniques discussed thus far -- namely knowledge graph embeddings and graph neural networks -- learn numerical models over graphs. However, such models are often difficult to explain or understand. For example, taking the graph of Figure~\ref{fig:airports}, knowledge graph embeddings might predict the edge \gedge[arrin][1.1cm]{SCL}{flight}{ARI} as being highly plausible, but they will not provide an interpretable model to help understand why this is the case: the reason for the result may lie in a matrix of parameters learnt to fit a plausibility score on training data. Such approaches also suffer from the \textit{out-of-vocabulary} problem, where they are unable to provide results for edges involving previously unseen nodes or edges; for example, if we add an edge \gedge[arrin]{SCL}{flight}{CDG}, where \gnode{CDG} is new to the graph, a knowledge graph embedding will not have the entity embedding for \gnode{CDG} and would need to be retrained in order to estimate the plausibility of an edge \gedge[arrin]{CDG}{flight}{SCL}. 

An alternative (sometimes complementary) approach is to adopt \textit{symbolic learning} in order to learn \textit{hypotheses} in a symbolic (logical) language that ``explain'' a given set of positive and negative edges. These edges are typically generated from the knowledge graph in an automatic manner (similar to the case of knowledge graph embeddings). The hypotheses then serve as interpretable models that can be used for further deductive reasoning. Given the graph of Figure~\ref{fig:airports}, we may, for example, learn the rule \begin{tikzpicture}[baseline=-3pt]
\node[var,compact](x){?x};

\node[var,compact,right=1cm of x](e){?y}
edge[arrin] node[lab] {flight} (x);
\end{tikzpicture} $\Rightarrow$
\begin{tikzpicture}[baseline=-3pt]
\node[var,compact](y){?y};

\node[var,compact,right=1cm of y](x){?x}
edge[arrin] node[lab] {flight} (y);
\end{tikzpicture} from observing that \gelab{flight} routes tend to be return routes. Alternatively, we might learn a DL axiom stating that airports are either domestic, international, or both: $\texttt{Airport} \sqsubseteq \texttt{DomesticAirport} \sqcup \texttt{InternationalAirport}$. Such rules and axioms can then be used for deductive reasoning, and offer an interpretable model for new knowledge that is entailed/predicted; for example, from the aforementioned rule for return flights, one can interpret why a novel edge \gedge[arrin][1.1cm]{SCL}{flight}{ARI} is predicted. This further offers domain experts the opportunity to verify the models -- e.g., the rules and axioms -- derived by such processes. Finally, rules/axioms are quantified (\textit{all} flights have a return flight, \textit{all} airports are domestic or international, etc.), so they can be applied to unseen examples (e.g., with the aforementioned rule, we can derive \gedge[arrin]{CDG}{flight}{SCL} from a new edge \gedge[arrin]{SCL}{flight}{CDG} with the unseen node \gnode{CDG}).

In this section, we discuss two forms of symbolic learning: \textit{rule mining}, which learns rules from a knowledge graph, and \textit{axiom mining}, which learns other forms of logical axioms. We refer to Appendix~\ref{app:sym} for a more formal treatment of these two tasks.

\subsubsection{Rule mining}

Rule mining, in the general sense, refers to discovering meaningful patterns in the form of rules from large collections of background knowledge. In the context of knowledge graphs, we assume a set of positive and negative edges as given. Typically positive edges are observed edges (i.e., those given or entailed by a knowledge graph) while negative edges are defined according to a given assumption of completeness (discussed later). The goal of rule mining is to identify new rules that entail a high ratio of positive edges from other positive edges, but entail a low ratio of negative edges from positive edges. The types of rules considered may vary from more simple cases, such as \begin{tikzpicture}[baseline=-3pt]
\node[var,compact](x){?x};

\node[var,compact,right=1cm of x](e){?y}
edge[arrin] node[lab] {flight} (x);
\end{tikzpicture} $\Rightarrow$
\begin{tikzpicture}[baseline=-3pt]
\node[var,compact](y){?y};

\node[var,compact,right=1cm of y](x){?x}
edge[arrin] node[lab] {flight} (y);
\end{tikzpicture} mentioned previously, to more complex rules, such as
\begin{tikzpicture}[baseline=-3pt]
\node[var,compact](x){?x};

\node[var,compact,right=\fhgap of x](y){?y}
edge[arrin] node[lab] {capital} (x);

\node[var,compact,right=\fhgap of y](z){?z}
edge[arrin] node[lab] {nearby} (y);

\node[iri,compact,right=\thgap of z](a){Airport}
edge[arrin] node[lab] {type} (z);
\end{tikzpicture} $\Rightarrow$
\begin{tikzpicture}[baseline=-3pt]
\node[var,compact](y){?z};

\node[iri,compact,right=\thgap of y](i){International Airport}
edge[arrin] node[lab] {type} (y);
\end{tikzpicture}, indicating that airports near capitals tend to be international airports; or \begin{tikzpicture}[baseline=-3pt]
\node[var,compact](x){?x};

\node[var,compact,right=\fhgap of x](y){?y}
edge[arrin] node[lab] {flight} (x);

\node[var,compact,right=\fhgap of y](z){?z}
edge[arrin,bend left=20] node[lab] {country} (x)
edge[arrin] node[lab] {country} (y);
\end{tikzpicture} $\Rightarrow$
\begin{tikzpicture}[baseline=-3pt]
\node[var,compact](x){?x};

\node[var,compact,right=1.6\fhgap of x](e){?y}
edge[arrin] node[lab] {domestic flight} (x);
\end{tikzpicture}, indicating that flights within the same country denote domestic flights (as seen in Section~\ref{sec:rules}). 

Per the international airport example, rules are not assumed to hold in all cases, but rather are associated with measures of how well they conform to the positive and negative edges. In more detail, we call the edges entailed by a rule and the set of positive edges (not including the entailed edge itself), the \textit{positive entailments} of that rule. The number of entailments that are positive is called the \textit{support} for the rule, while the ratio of a rule's entailments that are positive is called the \textit{confidence} for the rule~\cite{SuchanekLBW19}. As such, support and confidence indicate, respectively, the number and ratio of entailments ``confirmed'' to be true for the rule, where the goal is to identify rules that have both high support and high confidence. In fact, techniques for rule mining in relational settings have long been explored in the context of \emph{Inductive Logic Programming} (\textit{ILP})~\cite{DeRaedt08}. However, knowledge graphs present novel challenges due to the scale of the data and the frequent assumption of incomplete data (OWA), where dedicated techniques have been proposed to address these issues~\cite{GalarragaTHS13}. 

When dealing with an incomplete knowledge graph, it is not immediately clear how to define negative edges. A common heuristic -- also used for knowledge graph embeddings -- is to adopt a Partial Completeness Assumption (PCA)~\cite{GalarragaTHS13}, which considers the set of positive edges to be those contained in the data graph, and the set of negative examples to be the set of all edges \gedge[arrin][0.5cm]{$x$}{$p$}{$y'$} not in the graph but where there exists a node \gnode{$y$} such that \gedge[arrin][0.5cm]{$x$}{$p$}{$y$} is in the graph. Taking Figure~\ref{fig:airports}, an example of a negative edge under PCA would be \gedge{SCL}{flight}{ARI} (given the presence of \gedge{SCL}{flight}{LIM}); conversely, \gedge[arrin][2.2cm]{SCL}{domestic flight}{ARI} is neither positive nor negative. The PCA confidence measure is then the ratio of the support to all entailments in the positive or negative set~\cite{GalarragaTHS13}. For example, the support for the rule \begin{tikzpicture}[baseline=-3pt]
\node[var,compact](x){?x};

\node[var,compact,right=2.2cm of x](e){?y}
edge[arrin] node[lab] {domestic flight} (x);
\end{tikzpicture} $\Rightarrow$
\begin{tikzpicture}[baseline=-3pt]
\node[var,compact](y){?y};

\node[var,compact,right=2.2cm of y](x){?x}
edge[arrin] node[lab] {domestic flight} (y);
\end{tikzpicture} is 2 (since it entails \gedge[arrin][2.2cm]{IQQ}{domestic flight}{ARI} and \gedge[arrin][2.2cm]{ARI}{domestic flight}{IQQ} in the graph), while the confidence is $\frac{2}{2} = 1$ (noting that \gedge[arrin][2.2cm]{SCL}{domestic flight}{ARI}, though entailed, is neither positive nor negative, and is thus ignored by the measure). The support for the rule \begin{tikzpicture}[baseline=-3pt]
\node[var,compact](x){?x};

\node[var,compact,right=1cm of x](e){?y}
edge[arrin] node[lab] {flight} (x);
\end{tikzpicture} $\Rightarrow$
\begin{tikzpicture}[baseline=-3pt]
\node[var,compact](y){?y};

\node[var,compact,right=1cm of y](x){?x}
edge[arrin] node[lab] {flight} (y);
\end{tikzpicture} is analogously 4, while the confidence is $\frac{4}{5} = 0.8$ (noting that \gedge{SCL}{flight}{ARI} is negative). 

The goal then, is to find rules satisfying given support and confidence thresholds. An influential rule-mining system for graphs is AMIE~\cite{GalarragaTHS13,GalarragaTHS15}, which adopts the PCA measure of confidence, and builds rules in a top-down fashion~\cite{SuchanekLBW19} starting with rule heads like $\Rightarrow$
\begin{tikzpicture}[baseline=-3pt]
\node[var,compact](x){?x};

\node[var,compact,right=\fhgap of x](y){?y}
edge[arrin] node[lab] {country} (x);
\end{tikzpicture}. For each rule head of this form (one for each edge label), three types of \textit{refinements} are considered, each of which adds a new edge to the body of the rule. This new edge takes an edge label from the graph and may otherwise use \textit{fresh variables} not appearing previously in the rule, \textit{existing variables} that already appear in the rule, or nodes from the graph. The three refinements may then:

\begin{enumerate}
\item add an edge with one existing variable and one fresh variable; for example, refining the aforementioned rule head might give: \begin{tikzpicture}[baseline=-3pt]
\node[var,compact](z){?z};

\node[var,compact,right=\fhgap of z](x){?x}
edge[arrin] node[lab] {flight} (z);
\end{tikzpicture} $\Rightarrow$
\begin{tikzpicture}[baseline=-3pt]
\node[var,compact](x){?x};

\node[var,compact,right=1.1\fhgap of x](y){?y}
edge[arrin] node[lab] {country} (x);
\end{tikzpicture};
\item add an edge with an existing variable and a node from the graph; for example, refining the above rule might give: \begin{tikzpicture}[baseline=-3pt]
\node[var,compact](z){?z};

\node[iri,compact,left=\thgap of z](d){Domestic Airport}
edge[arrin] node[lab] {type} (z);

\node[var,compact,right=\fhgap of z](x){?x}
edge[arrin] node[lab] {flight} (z);
\end{tikzpicture} $\Rightarrow$
\begin{tikzpicture}[baseline=-3pt]
\node[var,compact](x){?x};

\node[var,compact,right=1.1\fhgap of x](y){?y}
edge[arrin] node[lab] {country} (x);
\end{tikzpicture};
\item add an edge with two existing variables; for example, refining the above rule might give: \begin{tikzpicture}[baseline=-3pt]
\node[var,compact](z){?z};

\node[iri,compact,left=\thgap of z](d){Domestic Airport}
edge[arrin] node[lab] {type} (z);

\node[var,compact,right=\fhgap of z](x){?x}
edge[arrin] node[lab] {flight} (z);

\node[var,compact,right=0.3\fhgap of x](y){?y}
edge[arrin,bend left=24] node[lab] {country} (z);
\end{tikzpicture}  $\Rightarrow$
\begin{tikzpicture}[baseline=-3pt]
\node[var,compact](x){?x};

\node[var,compact,right=1.1\fhgap of x](y){?y}
edge[arrin] node[lab] {country} (x);
\end{tikzpicture}.
\end{enumerate}   

\noindent
These refinements can be combined arbitrarily, which gives rise to a potentially exponential search space, where rules meeting given thresholds for support and confidence are maintained. To improve efficiency, the search space can be pruned; for example, these three refinements always decrease support, so if a rule does not meet the support threshold, there is no need to explore its refinements. Further restrictions are imposed on the types of rules generated. First, only rules up to a certain fixed size are considered. Second, a rule must be \textit{closed}, meaning that each variable appears in at least two edges of the rule, which ensures that rules are \textit{safe}, meaning that each variable in the head appears in the body; for example, the rules produced previously by the first and second refinements are neither closed (variable \gvar{y} appears once) nor safe (variable \gvar{y} appears only in the head).\footnote{Safe rules like 
\begin{tikzpicture}[baseline=-3pt]
\node[var,compact](x){?x};

\node[var,compact,right=\fhgap of x](y){?y}
edge[arrin] node[lab] {capital} (x);

\node[var,compact,right=\fhgap of y](z){?z}
edge[arrin] node[lab] {nearby} (y);

\node[iri,compact,right=\thgap of z](a){Airport}
edge[arrin] node[lab] {type} (z);
\end{tikzpicture} $\Rightarrow$
\begin{tikzpicture}[baseline=-3pt]
\node[var,compact](y){?z};

\node[iri,compact,right=\thgap of y](i){International Airport}
edge[arrin] node[lab] {type} (y);
\end{tikzpicture} are not closed as \gvar{?x} appears only in one edge. Hence the condition that rules are closed is strictly stronger than the condition that they are safe.} To ensure closed rules, the third refinement is applied until a rule is closed. For further discussion of possible optimisations based on pruning and indexing, we refer to the paper on AMIE+~\cite{GalarragaTHS15}.

Later works have built on these techniques for mining rules from knowledge graphs. \citet{Gad-ElrabSUW16} propose a method to learn non-monotonic rules -- rules with negated edges in the body -- in order to capture exceptions to base rules; for example, the approach may learn a rule \begin{tikzpicture}[baseline=-3pt]
\node[var,compact](z){?z};

\node[iri,dotted,compact,left=1.4\thgap of z](d){International Airport}
edge[arrin,dotted] node[lab] {$\neg$ type} (z);

\node[var,compact,right=\fhgap of z](x){?x}
edge[arrin] node[lab] {flight} (z);

\node[var,compact,right=0.3\fhgap of x](y){?y}
edge[arrin,bend left=24] node[lab] {country} (z);
\end{tikzpicture}  $\Rightarrow$
\begin{tikzpicture}[baseline=-3pt]
\node[var,compact](x){?x};

\node[var,compact,right=1.1\fhgap of x](y){?y}
edge[arrin] node[lab] {country} (x);
\end{tikzpicture}, indicating that flights are within the same country \textit{except} when the (departure) airport is international, where the exception is shown dotted and we use $\neg$ to negate an edge. The RuLES system~\cite{HoSGKW18} -- which is also capable of learning non-monotonic rules -- proposes to mitigate the limitations of the PCA heuristic by extending the confidence measure to consider the plausibility scores of knowledge graph embeddings for entailed edges not appearing in the graph. Where available, explicit statements about the completeness of the knowledge graph (such as expressed in shapes; see Section~\ref{sssec:validating-schema}) can be used in lieu of PCA for identifying negative edges. Along these lines, CARL~\cite{TanonSRMW17} exploits additional knowledge about the cardinalities of relations to refine the set of negative examples and the confidence measure for candidate rules. Alternatively, where available, ontologies can be used to derive logically-certain negative edges under OWA through, for example, disjointness axioms. The system proposed by~\citet{dAmatoTM16,DBLP:conf/sac/dAmatoSTMG16} leverages ontologically-entailed negative edges for determining the confidence of rules generated through an evolutionary algorithm. 

While the previous works involve discrete expansions of candidate rules for which a fixed confidence scoring function is applied, another line of research is on a technique called \textit{differentiable rule mining}~\cite{Rocktaschel017,YangYC17,SadeghianADW19}, which allows end-to-end learning of rules. The core idea is that the joins in rule bodies can be represented as matrix multiplication. More specifically, we can represent the relations of an edge label $p$ by the adjacency matrix $\mathbf{A}_p$ (of size $|V| \times |V|$) such that the value on the $i$\textsuperscript{th} row of the $j$\textsuperscript{th} column is 1 if there is an edge labelled $p$ from the $i$\textsuperscript{th} entity to the $j$\textsuperscript{th} entity; otherwise the value is 0. Now we can represent a join in a rule body as matrix multiplication; for example, given \begin{tikzpicture}[baseline=-3pt]
\node[var,compact](x){?x};

\node[var,compact,right=2.2\thgap of x](y){?y}
edge[arrin] node[lab] {domestic flight} (x);

\node[var,compact,right=\fhgap of y](z){?z}
edge[arrin] node[lab] {country} (y);
\end{tikzpicture} $\Rightarrow$ \begin{tikzpicture}[baseline=-3pt]
\node[var,compact](x){?x};

\node[var,compact,right=\fhgap of x](z){?z}
edge[arrin] node[lab] {country} (x);
\end{tikzpicture}, we can denote the body by the matrix multiplication $\mathbf{A}_\gelab{df.}\mathbf{A}_\gelab{c.}$, which gives an adjacency matrix representing entailed \gelab{country} edges, where we should expect the 1's in $\mathbf{A}_\gelab{df.}\mathbf{A}_\gelab{c.}$ to be covered by the head's adjacency matrix $\mathbf{A}_\gelab{c.}$. Since we are given adjacency matrices for all edge labels, we are left to learn confidence scores for individual rules, and to learn rules (of varying length) with a threshold confidence. Along these lines, NeuralLP~\cite{YangYC17} uses an \textit{attention mechanism} to select a variable-length sequence of edge labels for path-like rules of the form \begin{tikzpicture}[baseline=-3pt]
\node[var,compact](x){?x};

\node[var,compact,right=0.8\thgap of x](y1){?y$_1$}
edge[arrin] node[lab] {p$_1$} (x);

\node[compact,right=0.8\thgap of y1](d){\ldots}
edge[arrin] node[lab] {p$_2$} (y1);

\node[var,compact,right=0.8\thgap of d](yn){?y$_n$}
edge[arrin] node[lab] {p$_{n}$} (d);

\node[var,compact,right=0.8\fhgap of yn](z){?z}
edge[arrin] node[lab] {p$_{n+1}$} (yn);
\end{tikzpicture} $\Rightarrow$ \begin{tikzpicture}[baseline=-3pt]
\node[var,compact](x){?x};

\node[var,compact,right=0.8\thgap of x](z){?z}
edge[arrin] node[lab] {p} (x);
\end{tikzpicture}, for which confidences are likewise learnt. DRUM~\cite{SadeghianADW19} also learns path-like rules, where, observing that some edge labels are more/less likely to follow others in the rules -- for example, \gelab{flight} will not be followed by \gelab{capital} in the graph of Figure~\ref{fig:chileTransport} as the join will be empty -- the system uses bidirectional recurrent neural networks (a popular technique for learning over sequential data) to learn sequences of relations for rules, and their confidences. These differentiable rule mining techniques are, however, currently limited to learning path-like rules.
 
\subsubsection{Axiom mining}
 
Aside from rules, more general forms of axioms -- expressed in logical languages such as DLs (see Section~\ref{sssec:dls}) -- can be mined from a knowledge graph. We can divide these approaches into two categories: those mining specific axioms and more general axioms.

Among systems mining specific types of axioms, disjointness axioms are a popular target; for example, the disjointness axiom $\texttt{DomesticAirport} \sqcap \texttt{InternationalAirport} \equiv \bot$ states that the intersection of the two classes is equivalent to the empty class, or in simpler terms, no node can be simultaneously of type \gnode{Domestic Airport} and \gnode{International Airport}. The system proposed by~\citet{Volker2015} extracts disjointness axioms based on (negative) \emph{association rule mining}~\cite{Agrawal93}, which finds pairs of classes where each has many instances in the knowledge graph but there are relatively few (or no) instances of both classes. \citet{TopperKS12} rather extract disjointness for pairs of classes that have a cosine similarity below a fixed threshold. For computing this cosine similarity, class vectors are computed using a TF--IDF analogy, where the ``document'' of each class is constructed from all of its instances, and the ``terms'' of this document are the properties used on the class' instances (preserving multiplicities). While the previous two approaches find disjointness constraints between named classes (e.g., \textit{city} is disjoint with \textit{airport}), \citet{Rizzo2017} propose an approach that can capture disjointness constraints between class descriptions (e.g., \textit{city without an airport nearby} is disjoint with \textit{city that is the capital of a country}). The approach first clusters similar nodes of the knowledge base. Next, a \textit{terminological cluster tree} is extracted, where each leaf node indicates a cluster extracted previously, and each internal (non-leaf) node is a class definition (e.g., \textit{cities}) where the left child is either a cluster having all nodes in that class or a sub-class description (e.g., \textit{cities without airports}) and the right child is either a cluster having no nodes in that class or a disjoint-class description (e.g., \textit{non-cities with events}). Finally, candidate disjointness axioms are proposed for pairs of class descriptions in the tree that are not entailed to have a subclass relation.

Other systems propose methods to learn more general axioms. A prominent such system is DL-Learner~\cite{BuhmannLW16}, which is based on algorithms for \textit{class learning} (aka \textit{concept learning}), whereby given a set of positive nodes and negative nodes, the goal is to find a logical class description that divides the positive and negative sets. For example, given $\{\gnode{Iquique}, \gnode{Arica} \}$ as the positive set and $\{\gnode{Santiago} \}$ as the negative set, we may learn a (DL) class description $\exists \texttt{nearby}.\texttt{Airport} \sqcap \neg(\exists \texttt{capital}^-.\top)$, denoting entities near to an airport that are not capitals, of which all positive nodes are instances and no negative nodes are instances. Such class descriptions are learnt in an analogous manner to how aforementioned systems like AMIE learn rules, with a refinement operator used to move from more general classes to more specific classes (and vice-versa), a confidence scoring function, and a search strategy. The system further supports learning more general axioms through a scoring function that uses count queries to determine what ratio of expected edges -- edges that would be entailed were the axiom true -- are indeed found in the graph; for example, to score the axiom $\exists\texttt{flight}^{-}.\texttt{DomesticAirport} \sqsubseteq \texttt{InternationalAirport}$ over Figure~\ref{fig:airports}, we can use a graph query to count how many nodes have incoming flights from a domestic airport (there are 3), and how many nodes have incoming flights from a domestic airport \textit{and} are international airports (there is 1), where the greater the difference between both counts, the weaker the evidence for the axiom.

\section{Creation and Enrichment}\label{sec:create}

In this section, we discuss the principal techniques by which knowledge graphs can be created and subsequently enriched from diverse sources of legacy data that may range from plain text to structured formats (and anything in between). The appropriate methodology to follow when creating a knowledge graph depends on the actors involved, the domain, the envisaged applications, the available data sources, etc. Generally speaking, however, the flexibility of knowledge graphs lends itself to starting with an initial core that can be incrementally enriched from other sources as required (typically following an Agile~\citep{HuntT03a} or ``pay-as-you-go''~\cite{SequedaBMH19} methodology). For our running example, we assume that the tourism board decides to build a knowledge graph from scratch, aiming to initially describe the main tourist attractions -- places, events, etc. -- in Chile in order to help visiting tourists identify those that most interest them. The board decides to postpone adding further data, like transport routes, reports of crime, etc., for a later date.

\subsection{Human Collaboration}\label{sssec:graphCreationHuman}

One approach for creating and enriching knowledge graphs is to solicit direct contributions from human editors. Such editors may be found in-house (e.g., employees of the tourist board), using crowd-sourcing platforms, through feedback mechanisms (e.g., tourists adding comments on attractions), through collaborative-editing platforms (e.g., an attractions wiki open to public edits), etc. Though human involvement incurs high costs~\cite{Paulheim18a}, some prominent knowledge graphs have been primarily based on direct contributions from human editors~\citep{VrandecicK14,LinkedInKG}. Depending on how the contributions are solicited, however, the approach has a number of key drawbacks, due primarily to human error~\cite{pellissier2016freebase}, disagreement~\cite{YasseriSRKK12}, bias~\cite{Janowicz0RZM18}, vandalism~\cite{HeindorfPSE16}, etc. Successful collaborative creation further raises challenges concerning licensing, tooling, and culture~\citep{pellissier2016freebase}. Humans are sometimes rather employed to verify and curate additions to a knowledge graph extracted by other means~\cite{pellissier2016freebase} (through, e.g., video games with a purpose~\cite{JurgensNavigli14}), to define high-quality mappings from other sources~\cite{r2rml}, to define appropriate high-level schema~\cite{onteng,Labra2017},  and so forth. 

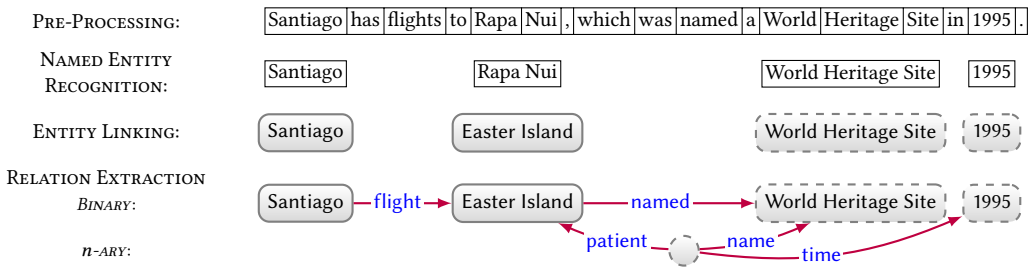
\begin{figure}
\centering
\newlength{\tgap}
\setlength{\tgap}{0.0cm}
\begin{tikzpicture}
\node[task] (tokenL) {Pre-Processing:};

\node[right=1cm of tokenL.east,anchor=west,split,rectangle split parts=17] (token) {\nodepart{one}Santiago \nodepart{two}has \nodepart{three}flights \nodepart{four}to \nodepart{five}Rapa \nodepart{six}Nui \nodepart{seven}, \nodepart{eight}which \nodepart{nine}was \nodepart{ten}named \nodepart{eleven}a \nodepart{twelve}World \nodepart{thirteen}Heritage \nodepart{fourteen}Site \nodepart{fifteen}in \nodepart{sixteen}1995 \nodepart{seventeen}.};

\node[task,below=\tgap of tokenL.south] (nerL) {\begin{tabular}{@{}c@{}}Named Entity\\Recognition:\end{tabular}};

\node[word] at (nerL-|token.one south) (santiagoNER) {Santiago};

\node[word,xshift=-0.7ex] at (nerL-|token.six) (rnNER) {Rapa Nui};

\node[word,xshift=-1ex] at (nerL-|token.thirteen south) (whsNER) {World Heritage Site};

\node[word] at (nerL-|token.sixteen south) (yearNER) {1995};

\node[task,below=0.6\tgap of nerL.south] (elL) {\begin{tabular}{@{}c@{}}Entity Linking:\end{tabular}};

\node[iri] at (elL-|santiagoNER) (santiagoEL) {Santiago};

\node[iri] at (elL-|rnNER) (eiEL) {Easter Island};

\node[iri,dashed] at (elL-|whsNER) (whsEL) {World Heritage Site};

\node[iri,dashed] at (elL-|yearNER) (yearEL) {1995};

\node[task,below=1.5\tgap of elL.south] (reL) {\begin{tabular}{@{}c@{}}Relation Extraction\\\scriptsize \textit{Binary}:\end{tabular}};

\node[iri,yshift=-1ex] at (reL-|santiagoEL) (santiagoRE) {Santiago};

\node[iri] at (santiagoRE-|eiEL) (eiRE) {Easter Island}
  edge[arrin] node[lab] {flight} (santiagoRE);

\node[iri,dashed] at (santiagoRE-|whsEL) (whsRE) {World Heritage Site}
  edge[arrin] node[lab] {named} (eiRE);

\node[iri,dashed] at (santiagoRE-|yearEL) (yearRE) {1995};

\node[task,below=1.5\tgap of reL.south] (neL) {\begin{tabular}{@{}c@{}}\scriptsize \textit{$n$-ary}:\end{tabular}};

\node[between=eiRE and whsRE] (mid) {};

\node[bnode,dashed] at (neL-|mid) (nary) {}
  edge[arrout,bend left=10] node[lab] {patient} (eiRE)
  edge[arrout,bend right=10] node[lab] {name} (whsRE)
  edge[arrout,bend right=15] node[lab] {time} (yearRE);
\end{tikzpicture}

\caption{Text extraction example; nodes new to the knowledge graph are shown dashed \label{fig:textExtract}}
\end{figure}

\subsection{Text Sources}\label{sssec:graphCreationText}

Text corpora -- such as sourced from newspapers, books, scientific articles, social media, emails, web crawls, etc. -- are an abundant source of rich information~\citep{HellmannLAB13,RospocherEVFARS16}. However, extracting such information with high precision and recall for the purposes of creating or enriching a knowledge graph is a non-trivial challenge. To address this, techniques from Natural Language Processing (NLP)~\cite{NLP-SW,JurafskyM18} and Information Extraction (IE)~\cite{WeikumT10,Grishman12,IESW} can be applied. Though processes vary considerably across text extraction frameworks, in Figure~\ref{fig:textExtract} we illustrate four core tasks for text extraction on a sample sentence. We will discuss these tasks in turn.

\subsubsection{Pre-processing} The pre-processing task may involve applying various techniques to the input text, where Figure~\ref{fig:textExtract} illustrates \textit{Tokenisation}, which parses the text into atomic terms and symbols. Other pre-processing tasks applied to a text corpus may include: \textit{Part-of-Speech} (\textit{POS}) \textit{tagging}~\cite{NLP-SW,JurafskyM18} to identify terms representing verbs, nouns, adjectives, etc.; \textit{Dependency Parsing}, which extracts a grammatical tree structure for a sentence where leaf nodes indicate individual words that together form phrases (e.g., noun phrases, verb phrases) and eventually clauses and sentences~\cite{NLP-SW,JurafskyM18}; and \textit{Word Sense Disambiguation} (\textit{WSD})~\cite{Navigli:09} to identify the meaning (aka \textit{sense}) in which a word is used, linking words with a lexicon of senses (e.g., WordNet~\cite{MillerF07} or BabelNet~\cite{NavigliPonzetto:12}), where, for instance, the term \tnode{flights} may be linked with the WordNet sense ``\textsf{an instance of travelling by air}'' rather than ``\textsf{a stairway between one floor and the next}'. The appropriate type of pre-processing to apply often depends on the requirements of later tasks in the pipeline.

\subsubsection{Named Entity Recognition (NER)} The NER task identifies mentions of named entities in a text~\citep{nadeaus07,ratinovr09}, typically targetting mentions of people, organisations, locations, and potentially other types~\citep{LingW12,NakasholeTW13,YogatamaGL15}. A variety of NER techniques exist, with many modern approaches based on learning frameworks that leverage lexical features (e.g., POS tags, dependency parse trees, etc.) and gazetteers (e.g., lists of common first names, last names, countries, prominent businesses, etc.). Supervised methods~\citep{BikelSW99,FinkelGM05,LampleBSKD16} require manually labelling all entity mentions in a training corpus, whereas \textit{bootstrapping}-based approaches~\citep{CollinsS99,EtzioniCDKPSSWY04,NakasholeTW13,GuptaM14} rather require a small set of \textit{seed examples} of entity mentions from which patterns can be learnt and applied to unlabelled text. \textit{Distant supervision}~\citep{LingW12,RenEWTVH15,YogatamaGL15} uses known entities in a knowledge graph as seed examples through which similar entities can be detected. Aside from learning-based frameworks, manually-crafted rules~\cite{KlueglAP09,ChiticariuDLRZ18} are still sometimes used due to their more controllable and predictable behaviour~\citep{ChiticariuLR13}. The named entities identified by NER may be used to generate new candidate nodes for the knowledge graph (known as \textit{emerging entities}, shown dashed in Figure~\ref{fig:textExtract}), or may be linked to existing nodes per the Entity Linking task described in the following.

\subsubsection{Entity Linking (EL)} The EL task associates mentions of entities in a text with the existing nodes of a target knowledge graph, which may be the nucleus of a knowledge graph under creation, or an external knowledge graph~\cite{WuHH18}. In Figure~\ref{fig:textExtract}, we assume that the nodes \gnode{Santiago} and \gnode{Easter Island} already exist in the knowledge graph (possibly extracted from other sources). EL may then link the given mentions to these nodes. The EL task presents two main challenges. First, there may be multiple ways to mention the same entity, as in the case of \tnode{Rapa Nui} and \tnode{Easter Island}; if we created a node \gnode{Rapa Nui} to represent that mention, we would split the information available under both mentions across different nodes, where it is thus important for the target knowledge graph to capture the various aliases and multilingual labels by which one can refer to an entity~\cite{Moroetal:14}. Secondly, the same mention in different contexts can refer to distinct entities; for instance, \tnode{Santiago} can refer to cities in Chile, Cuba, Spain, among others. The EL task thus considers a \textit{disambiguation phase} wherein mentions are associated to candidate nodes in the knowledge graph, the candidates are ranked, and the most likely node being mentioned is chosen~\cite{WuHH18}. Context can be used in this phase; for example, if \gnode{Easter Island} is a likely candidate for the corresponding mention alongside \tnode{Santiago}, we may boost the probability that this mention refers to the Chilean capital as both candidates are located in Chile. Other heuristics for disambiguation consider a prior probability, where for example, \tnode{Santiago} most often refers to the Chilean capital (being, e.g., the largest city with that name); centrality measures on the knowledge graph can be used for such purposes~\cite{WuHH18}.

\subsubsection{Relation Extraction (RE)} The RE task extracts relations between entities in the text~\citep{ZhouSZZ05,BachB07}. The simplest case is that of extracting binary relations in a \textit{closed setting} wherein a fixed set of relation types are considered. While traditional approaches often used manually-crafted patterns~\cite{Hearst92}, modern approaches rather tend to use learning-based frameworks~\citep{RollerKN18}, including supervised methods over manually-labelled examples~\citep{BunescuM05,ZhouSZZ05}. Other learning-based approaches again use bootstrapping~\cite{EtzioniCDKPSSWY04,BunescuM07} and distant supervision~\citep{MintzBSJ09,RiedelYM10,HoffmannZLZW11,SurdeanuTNM12,XuHZG13,SmirnovaC19} to forgo 
the need for manual labelling; the former requires a subset of manually-labelled seed examples, while the latter finds sentences in a large corpus of text mentioning pairs of entities with a known relation/edge, which are used to learn patterns for that relation. Binary RE can also be applied using unsupervised methods in an open setting -- often referred to as \textit{Open Information Extraction} (\textit{OIE})~\citep{BankoCSBE07,EtzioniFCSM11,FaderSE11,MausamSSBE12,Mausam16,MitchellCHTYBCM18} -- whereby the set of target relations is not pre-defined but rather extracted from text based on, for example, dependency parse trees from which relations are taken. 

A variety of RE methods have been proposed to extract $n$-ary relations that capture further context for how entities are related. In Figure~\ref{fig:textExtract}, we see how an $n$-ary relation captures additional temporal context, denoting when Rapa Nui was named a World Heritage site; in this case, an anonymous node is created to represent the higher-arity relation in the directed-labelled graph. Various methods for $n$-ary RE are based on \textit{frame semantics}~\cite{fillmore1976frame}, which, for a given verb (e.g., ``\textit{named}''), captures the entities involved and how they may be interrelated. Resources such as FrameNet~\cite{framenet} then define frames for words, such as to identify that the semantic frame for ``\textit{named}'' includes a \textit{speaker} (the person naming something), an \textit{entity} (the thing named) and a \textit{name}. Optional frame elements are an \textit{explanation}, a \textit{purpose}, a \textit{place}, a \textit{time}, etc., that may add context to the relation. Other RE methods are rather based on \textit{Discourse Representation Theory} (\textit{DRT})~\cite{Kamp1981ATheoryOfTruth}, which considers a logical representation of text based on existential events. Under this theory, for example, the naming of Easter Island as a World Heritage Site is considered to be an (existential) event where Easter Island is the \textit{patient} (the entity affected), leading to the logical (neo-Davidsonian) formula:
\[ \exists e: \big(\text{naming}(e), \text{patient}(e,\tnode{Easter Island}), \text{name}(e,\tnode{World Heritage Site})\big) \]
\noindent Such a formula is analogous to the idea of reification, as discussed previously in Section~\ref{ssec:knowledgeContext}.

Finally, while relations extracted in a closed setting are typically mapped directly to a knowledge graph, relations that are extracted in an open setting may need to be aligned with the knowledge graph; for example, if an OIE process extracts a binary relation \gedge[arrin][2cm]{Santiago}{has flights to}{Easter Island}, it may be the case that the knowledge graph does not have other edges labelled \gelab{has flights to}, where alignment may rather map such a relation to the edge \gedge[arrin][1.2cm]{Santiago}{flight}{Easter Island} assuming \gelab{flight} is used in the knowledge graph. A variety of methods have been applied for such purposes, including mappings~\cite{CorcoglionitiRA16,GangemiPRNDM17} and rules~\cite{rouces2015framebase} for aligning $n$-ary relations; distributional and dependency-based similarities~\cite{MoroNavigli13}, association rule mining~\cite{dutta2014semantifying}, Markov clustering~\cite{Dutta2015ESKwithOI} and linguistic techniques~\cite{Martinez-Rodriguez18} for aligning OIE relations; amongst others.

\subsubsection{Joint tasks}

Having presented the four main tasks for building knowledge graphs from text, it is important to note that frameworks do not always follow this particular sequence of tasks. A common trend, for example, is to combine interdependent tasks, jointly performing WSD and EL~\cite{Moroetal:14}, or NER and EL~\cite{LuoHLN15,NguyenTW16}, or NER and RE~\cite{RenWHQVJAH17,ZhengWBHZX17}, etc., in order to mutually improve the performance of multiple tasks. For further details on extracting knowledge from text we refer to the book by \citet{NLP-SW} and the recent survey by \citet{IESW}.

\subsection{Markup Sources}\label{sssec:graphCreationSemistructured}

\begin{figure}[t]
	\setlength{\vgap}{1.2cm}
	\setlength{\hgap}{1.2cm}
	\begin{minipage}{0.57\textwidth}
\begin{lstlisting}[style=lst,language=xml,basicstyle={\ttfamily\scriptsize}]
<html>
 <head><title>UNESCO World Heritage Sites</title></head>
 <body>
  <h1>World Heritage Sites</h1>
  <h2>Chile</h2>
  <p>Chile has 6 UNESCO World Heritage Sites.</p>
  <table border="1">
   <tr><th>Place</th><th>Year</th><th>Criteria</th></tr>
   <tr><td>Rapa Nui</td><td>1995</td>
       <td rowspan="6">Cultural</td></tr>
   <tr><td>Churches of Chilo`{\color{ctxt}é}`</td><td>2000</td></tr>
   <tr><td>Historical Valpara`{\color{ctxt}í}`so</td><td>2003</td></tr>
   <tr><td>Saltpeter Works</td><td>2005</td></tr>
   <tr><td>Sewell Mining Town</td><td>2006</td></tr>
   <tr><td>Qhapaq `{\color{ctxt}Ñ}`an</td><td>2014</td></tr>
  </table>
 </body>
</html>
\end{lstlisting}
	\end{minipage}
	\quad 
	\begin{tikzpicture}[baseline=-3pt]
	 \node[rectangle,draw,fill=white!93!black] (nl) {
	 	\textsf{\footnotesize
	 	  \begin{tabular}{@{~}p{5cm}@{~}}
	 		\textbf{\large World Heritage Sites}\\[1ex]
	 		\textbf{\normalsize Chile}\\[0.5ex]
	 		Chile has 6 UNESCO World Heritage Sites.\\[1ex]
	 		\begin{tabular}{|l|l|l|}\hline
	 		  \textbf{Place} & \textbf{Year} & \textbf{Criteria}\\\hline\hline
	 	 	  Rapa Nui & 1995 & \multirow{6}{*}{Cultural} \\\cline{1-2}
	 		  Churches of Chiloé & 2000 & \\\cline{1-2}
	 		  Historical Valparaíso & 2003  &  \\\cline{1-2}
	 		  Saltpeter Works & 2005  &  \\\cline{1-2}
	 		  Sewell Mining Town& 2006  &  \\\cline{1-2}
	 		  Qhapaq Ñan & 2014  &  \\\hline
	 	    \end{tabular}
	 	  \end{tabular}
	 	 }
	 };

	 \node[rectangle,fill=white!93!black,anchor=south west] (tab) at (nl.north west) {\scriptsize\Mundus~\texttt{UNESCO World Heritage Sites} $\times$};
	\end{tikzpicture}
	\caption{Example markup document (HTML) with source-code (left) and formatted document (right) \label{fig:html}}
\end{figure}
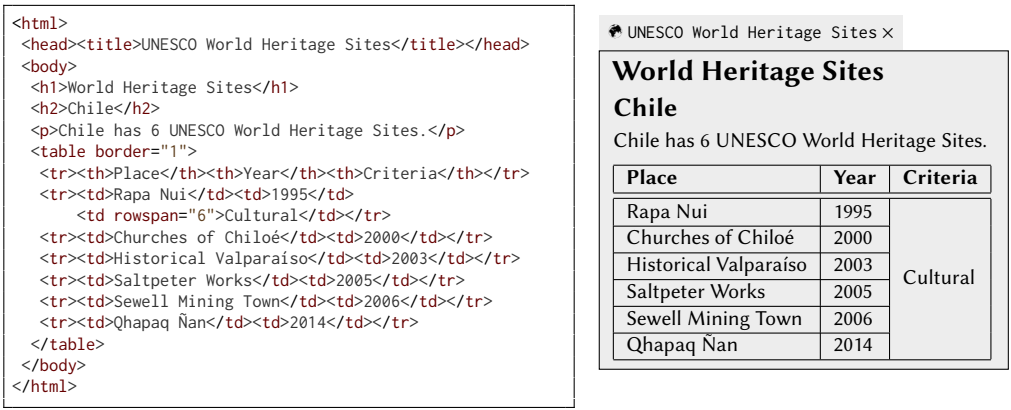

The Web was founded on interlinking \textit{markup documents} wherein markers (aka \textit{tags}) are used to separate elements of the document (typically for formatting purposes). Most documents on the Web use the HyperText Markup Language (HTML). Figure~\ref{fig:html} presents an example HTML webpage about World Heritage Sites in Chile. Other formats of markup include Wikitext used by Wikipedia, TeX for typesetting, Markdown used by Content Management Systems, etc. One approach for extracting information from markup documents -- in order to create and/or enrich a knowledge graph -- is to strip the markers (e.g., HTML tags), leaving only plain text upon which the techniques from the previous section can be applied. However, markup can be useful for extraction purposes, where variations of the aforementioned tasks for text extraction have been adapted to exploit such markup~\citep{LuBLCG13,LockardDSE18,IESW}. We can divide extraction techniques for markup documents into three main categories: general approaches that work independently of the markup used in a particular format, often based on \textit{wrappers} that map elements of the document to the output; focussed approaches that target specific forms of markup in a document, most typically \textit{web tables} (but sometimes also lists, links, etc.); and form-based approaches that extract the data underlying a webpage, per the notion of the \textit{Deep Web}. These approaches can often benefit from the regularities shared by webpages of a given website, be it due to informal conventions on how information is published across webpages, or due to the re-use of templates to automatically generate content across webpages; for example, intuitively speaking, while the webpage of Figure~\ref{fig:html} is about Chile, we will likely find pages for other countries following the same structure on the same website.

\subsubsection{Wrapper-based extraction}

Many general approaches are based on \emph{wrappers} that locate and extract the useful information directly from the markup document. While the traditional approach was to define such wrappers manually -- a task for which a variety of declarative languages and tools have been defined -- such approaches are brittle to changes in a website's layout~\cite{FerraraMFB14}. Hence other approaches allow for (semi-)automatically \textit{inducing} wrappers~\cite{FlescaMM04}. A modern such approach -- used to enrich knowledge graphs in systems such as LODIE~\cite{GentileZC14} -- is to apply distant supervision, whereby EL is used to identify and link entities in the webpage to nodes in the knowledge graph such that paths in the markup that connect pairs of nodes for known edges can be extracted, ranked, and applied to other examples. Taking Figure~\ref{fig:html}, for example, distant supervision may link \tnode{Rapa Nui} and \tnode{\textbf{World Heritage Sites}} to the nodes \gnode{Easter Island} and \gnode{World Heritage Site} in the knowledge graph using EL, and given the edge \gedge{Easter Island}{named}{World Heritage Site} in the knowledge graph (extracted per Figure~\ref{fig:textExtract}), identify the candidate path $(x,{\markupf{td}[1]^{-} \cdot \markupf{tr}^{-} \cdot \markupf{table}^- \cdot \markupf{h1}},y)$ as reflecting edges of the form \gedge{$x$}{named}{$y$}, where $t[n]$ indicates the $n$\textsuperscript{th} child of tag $t$, $t^-$ its inverse, and $t_1 \cdot t_2$ concatenation. Finally, paths with high confidence (e.g., ones ``witnessed'' by many known edges in the knowledge graph) can then be used to extract novel edges, such as \gedge{Qhapaq Ñan}{named}{World Heritage Site}, both on this page and on related pages of the website with similar structure (e.g., for other countries).

\subsubsection{Web table extraction}

Other approaches target specific types of markup, most commonly \textit{web tables}, i.e., tables embedded in HTML webpages. However, web tables are designed to enhance human readability, which often conflicts with machine readability. Many web tables are used for layout and page structure (e.g., navigation bars), while those that do contain data may follow different formats such as relational tables, listings, attribute-value tables, matrices, etc.~\cite{CafarellaHWWZ08,CrestanP11}. Hence a first step is to classify tables to find ones appropriate for the given extraction mechanism(s)~\cite{CrestanP11,EberiusBHTAL15}. Next, web tables may contain column spans, row spans, inner tables, or may be split vertically to improve human aesthetics. Hence a table normalisation phase is required to identify headers, merge split tables, un-nest tables, transpose tables, etc.~\cite{PivkCSGRS07,CafarellaHWWZ08,CrestanP11,DengJLLY13,ErmilovN16,LehmbergRMB16}. Subsequently, approaches may need to identify the \textit{protagonist}~\cite{CrestanP11,MunozHM14} -- the main entity that the table describes -- which is rather found elsewhere in the webpages; for example, though \tnode{World Heritage Sites} is the protagonist of the table of Figure~\ref{fig:textExtract}, it is not mentioned by the table. Finally, extraction processes can be applied, potentially associating cells with entities~\cite{LimayeSC10,MulwadFJ13}, columns with types~\cite{DengJLLY13,LimayeSC10,MulwadFJ13}, and column pairs with relations~\cite{LimayeSC10,MunozHM14}. For the purposes of enriching knowledge graphs, more recent approaches again apply distant supervision, first linking table cells to knowledge graph nodes, which are used to generate candidates for type and relation extraction~\cite{LimayeSC10,MulwadFJ13,MunozHM14}. Statistical distributions can also aid in linking numerical columns~\cite{NeumaierUPP16}. Specialised extraction frameworks have also been designed for tables on specific websites, where prominent knowledge graphs, such as DBpedia~\cite{LehmannIJJKMHMK15} and YAGO~\cite{suchanek2008yago} focus on extraction from info-box tables in Wikipedia.

\subsubsection{Deep Web crawling}

The \textit{Deep Web} presents a rich source of information accessible only through searches on web forms, thus requiring \emph{Deep Web crawling} techniques to access~\citep{MadhavanKKGRH08}. Systems have been proposed to extract knowledge graphs from Deep Web sources~\cite{GellerCA08,LehmannFGNSSUBGHLA12,CollaranaG0GVA16}. Approaches typically attempt to generate sensible form inputs -- which may be based on a user query or generated from reference knowledge -- and then extract data from the generated responses (markup documents) using the aforementioned techniques~\cite{GellerCA08,LehmannFGNSSUBGHLA12,CollaranaG0GVA16}.

\subsection{Structured Sources}\label{sssec:graphCreationStructured}

\begin{figure}
\footnotesize\tt
\begin{tabular}{|l|l|l|l|}
\multicolumn{4}{l}{\textbf{\large Report}}\\
\hline
\textbf{crime} & \textbf{claimant} & \textbf{station} & \textbf{date} \\
\hline \hline
Pickpocketing & XY12SDA & Viña del Mar & 2019-04-12 \\
Assault & AB9123N & Arica & 2019-04-12 \\
Pickpocketing & XY12SDA & Rapa Nui & 2019-04-12 \\
Fraud & FI92HAS & Arica  & 2019-04-13 \\
\hline
\end{tabular}\qquad
\begin{tabular}{|l|l|l|}
\multicolumn{3}{l}{\textbf{\large Claimant}}\\
\hline
\underline{\textbf{id}} & \textbf{name} & \textbf{country}  \\
\hline \hline
XY12SDA & John Smith & U.S. \\
AB9123N & Joan Dubois & France \\
XI92HAS & Jorge Hernández & Chile \\
\hline
\end{tabular}
\caption{Example relational database instance with two tables describing crime data \label{fig:rdbCrime}}
\end{figure}

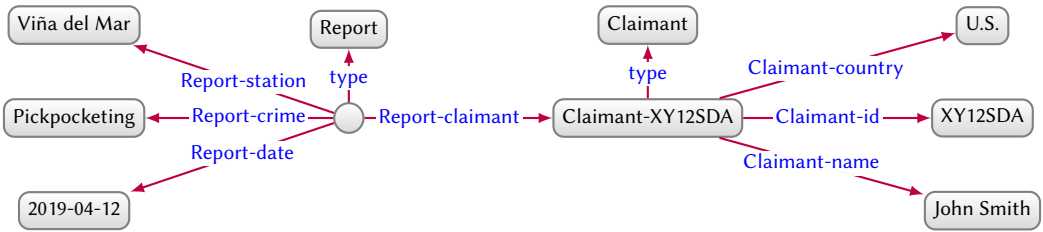
\begin{figure}
\setlength{\hgap}{2.5cm}
\setlength{\vgap}{0.7cm}
\begin{tikzpicture}
\node[iri,anchor=center] (xy) {Claimant-XY12SDA};

\node[iri,anchor=center,right=\hgap of xy] (xyid) {XY12SDA}
  edge[arrin] node[lab] {Claimant-id} (xy);

\node[iri,anchor=center,below=\vgap of xyid] (js) {John Smith}
  edge[arrin] node[lab] {Claimant-name} (xy);

\node[iri,anchor=center,above=\vgap of xyid] (us) {U.S.}
  edge[arrin] node[lab] {Claimant-country} (xy);

\node[iri,anchor=center,above=\vgap of xy] (ctype) {Claimant}
  edge[arrin] node[lab] {type} (xy);

\node[bnode,anchor=center,left=\hgap of xy] (bnode) {}
  edge[arrout] node[lab] {Report-claimant} (xy);

\node[iri,anchor=center,left=\hgap of bnode] (crime) {Pickpocketing}
  edge[arrin] node[lab] {Report-crime} (bnode);

\node[iri,anchor=center,above=\vgap of crime] (station) {Viña del Mar}
  edge[arrin] node[lab] {Report-station} (bnode);

\node[iri,anchor=center,below=\vgap of crime] (date) {2019-04-12}
  edge[arrin] node[lab] {Report-date} (bnode);

\node[iri,anchor=center,above=\vgap of bnode] (crtype) {Report}
  edge[arrin] node[lab] {type} (bnode);
\end{tikzpicture}
\caption{Possible result of applying a direct mapping to the first rows of both tables in Figure~\ref{fig:rdbCrime} \label{fig:direct}}
\end{figure}

Much of the legacy data available within organisations and on the Web is represented in structured formats, primarily tables -- in the form of relational databases, CSV files, etc. -- but also tree-structured formats such as JSON, XML etc. Unlike text and markup documents, structured sources can often be \textit{mapped} to knowledge graphs whereby the structure is (precisely) transformed according to a mapping rather than (imprecisely) extracted. The mapping process involves two steps: 1) create a mapping from the source to a graph, and 2) use the mapping in order to materialise the source data as a graph or to virtualise the source (creating a graph view over the legacy data).

\subsubsection{Mapping from tables} Tabular sources of data are prevalent, where, for example, the structured content underlying many organisations, websites, etc., are housed in relational databases. In Figure~\ref{fig:rdbCrime} we present an example of a relational database instance that we would like to integrate into our knowledge graph under construction. There are then two approaches for mapping content from tables to knowledge graphs: a \textit{direct mapping}, and a \textit{custom mapping}.

A direct mapping automatically generates a graph from a table. We present in Figure~\ref{fig:direct} the result of a standard direct mapping~\cite{dm}, which creates an edge $\gedge[arrin][0.6cm]{x}{y}{z}$ for each (non-header, non-empty, non-\textsc{null}) cell of the table, such that $\gnode{x}$ represents the row of the cell, $\gelab{y}$ the column name of the cell, and $\gnode{z}$ the value of the cell. In particular, $\gnode{x}$ typically encodes the values of the primary key for a row (e.g., \textbf{\texttt{Claimant}}.\underline{\textbf{\texttt{id}}}); otherwise, if no primary key is defined (e.g., per the \textbf{\texttt{Report}} table), $\gnode{x}$ can be an anonymous node or a node based on the row number. The node $\gnode{x}$ and edge label $\gelab{y}$ further encode the name of the table to avoid clashes across tables that have the same column names used with different meanings. For each row $\gnode{x}$, we may add a type edge based on the name of its table. The value $\gnode{z}$ may be mapped to datatype values in the corresponding graph model based on the source domain (e.g., a value in an SQL column of type \texttt{Date} can be mapped to \texttt{xsd:date} in the RDF data model). If the value is \textsc{null} (or empty), typically the corresponding edge will be omitted.\footnote{One might consider representing \textsc{null}s with anonymous nodes. However, \textsc{null}s in SQL can be used to mean that there is no such value, which conflicts with the existential semantics of anonymous nodes in models such as RDF (i.e., blank nodes).} With respect to  Figure~\ref{fig:direct}, we highlight the difference between the nodes \gnode{Claimant-XY12SDA} and \gnode{XY12SDA}, where the former denotes the row (or entity) identified by the latter primary key value. In case of a foreign key between two tables --  such as \textbf{Report.claimant} referencing \textbf{Claimant.\underline{id}} -- we can link, for example, to \gnode{Claimant-XY12SDA} rather than \gnode{XY12SDA}, where the former node also has the name and country of the claimant. A direct mapping along these lines has been standardised for mapping relational databases to RDF~\cite{dm}, where \citet{StoicaFS19} have recently proposed an analogous direct mapping for property graphs. Another direct mapping has been defined for CSV and other tabular data~\cite{csvweb} that further allows for specifying column names, primary/foreign keys, and data types -- which are often missing in such data formats -- as part of the mapping itself.

Although a direct mapping can be applied automatically on tabular sources of data and preserve the information of the original source -- i.e., allowing a deterministic inverse mapping that reconstructs the tabular source from the output graph~\cite{SequedaAM12} -- in many cases it is desirable to customise a mapping, such as to align edge labels or nodes with a knowledge graph under enrichment, etc. Along these lines, declarative mapping languages allow for manually defining custom mappings from tabular sources to graphs. A standard language along these lines is the RDB2RDF Mapping Language (R2RML)~\cite{r2rml}, which allows for mapping from individual rows of a table to one or more custom edges, with nodes and edges defined either as constants, as individual cell values, or using templates that concatenate multiple cell values from a row and static substrings into a single term; for example, a template \texttt{\{id\}-\{country\}} may produce nodes such as \gnode{XY12SDA-U.S.} from the \textbf{Claimant} table. In case that the desired output edges cannot be defined from a single row, R2RML allows for (SQL) queries to generate tables from which edges can be extracted where, for example, edges such as \gedge[arrin][1.2cm]{U.S.}{crimes}{2} can be generated by defining the mapping with respect to a query that joins the \textbf{\texttt{Report}} and \textbf{\texttt{Claimant}} tables on \texttt{\textbf{claimant}=\textbf{id}}, grouping by \texttt{\textbf{country}}, and applying a count. A mapping can then be defined on the results table such that the source node denotes the value of \texttt{\textbf{country}}, the edge label is the constant \gelab{crimes}, and the target node is the count value. An analogous standard also exists for mapping CSV and other tabular data to RDF graphs, again allowing keys, column names, and datatypes to be chosen as part of the mapping~\citep{csvwmeta}.

Once the mappings have been defined, one option is to use them to \textit{materialise} graph data following an \textit{Extract-Transform-Load} (\textit{ETL}) approach, whereby the tabular data are transformed and explicitly serialised as graph data using the mapping. A second option is to use \textit{virtualisation} through a \textit{Query Rewriting} (\textit{QR}) approach, whereby queries on the graph (using, e.g., SPARQL, Cypher, etc.) are translated to queries over the tabular data (typically using SQL). Comparing these two options, ETL allows the graph data to be used as if they were any other data in the knowledge graph. However, ETL requires updates to the underlying tabular data to be explicitly propagated to the knowledge graph, whereas a QR approach only maintains one copy of data to be updated. The area of \textit{Ontology-Based Data Access} (\textit{OBDA})~\cite{XiaoCKLPRZ18} is then concerned with QR approaches that support ontological entailments as discussed in Section~\ref{sec:deductive}. Although most QR approaches only support non-recursive entailments expressible as a single (non-recursive) query, some QR approaches support recursive entailments through rewritings to recursive queries~\cite{SequedaAM14}.

\subsubsection{Mapping from trees} A number of popular data formats are based on trees, including XML and JSON. While one could imagine -- leaving aside issues such as the ordering of children in a tree -- a trivial direct mapping from trees to graphs by simply creating edges of the form \gedge[arrin][1cm]{$x$}{child}{$y$} for each node $y$ that is a child of $x$ in the source tree, such an approach is not typically used, as it represents the literal structure of the source data. Instead, the content of tree-structured data can be more naturally represented as a graph using a custom mapping. Along these lines, the GRDLL standard~\cite{grddl} allows for mapping from XML to (RDF) graphs, while the JSON-LD standard~\cite{jsonld} allows for mapping from JSON to (RDF) graphs. In contrast, hybrid query languages such as XSPARQL~\cite{BishofDKLP12} allow for querying XML and RDF in an integrated fashion, thus supporting both materialisation and virtualisation of graphs over tree-structured sources of legacy data.

\subsubsection{Mapping from other knowledge graphs}\label{sssec:graphCreationKGs}

Another route to construct or enrich knowledge graphs is to leverage existing knowledge graphs as a source. In our scenario, for instance, a large number of points of interest for the Chilean tourist board may be available in existing knowledge graphs such as DBpedia~\cite{LehmannIJJKMHMK15}, LinkedGeoData~\cite{StadlerLHA12}, Wikidata~\cite{VrandecicK14}, YAGO~\cite{YAGO}, BabelNet~\cite{NavigliPonzetto:12},  etc. However, depending on the knowledge graph under construction, not all entities and/or relations may be of interest. A standard option to extract a relevant sub-graph of data is to use SPARQL construct-queries that generate graphs as output~\cite{neumaier2018enabling}. Entity and schema alignment between the knowledge graphs may be further necessary to better integrate (parts of) external knowledge graphs; this may be done using linking tools for graphs~\cite{silk,NgomoA11}, based on the use of external identifiers~\cite{pellissier2016freebase}, or indeed may be done manually~\cite{pellissier2016freebase}. For instance, Wikidata~\cite{VrandecicK14} uses Freebase~\cite{bollacker2007freebase,pellissier2016freebase} as a source; \citet{gottschalk2018eventkg} extract an event-centric knowledge graph from Wikidata, DBpedia and YAGO; while \citet{neumaier2018enabling} construct a spatio-temporal knowledge graph from Geonames, Wikidata, and PeriodO~\cite{GoldenS16} (as well as tabular data).

\subsection{Schema/Ontology Creation}\label{ssec:knowledgeConceptual}

The discussion thus far has focussed on extracting \textit{data} from external sources in order to create and enrich a knowledge graph. In this section, we discuss some of the principal methods for generating a schema based on external sources of data, including human knowledge. For discussion on extracting a schema from the knowledge graph itself, we refer back to Section~\ref{ssec:emergentSchema}. In general, much of the work in this area has focussed on the creation of ontologies using either ontology engineering methodologies, and/or ontology learning. We discuss these two approaches in turn.

\subsubsection{Ontology engineering}

Ontology engineering refers to the development and application of methodologies for building ontologies, proposing principled processes by which better quality ontologies can be constructed and maintained with less effort. Early methodologies~\cite{Gruninger1995,Fernandez1997,Noy2001} were often based on a waterfall-like process, where requirements and conceptualisation were fixed before starting to implement the ontology in a logical language, using, for example, an ontology engineering tool~\cite{gomez2006ontological,onteng,kendall2019ontology}. However, for situations involving large or ever-evolving ontologies, more iterative and agile ways of building and maintaining ontologies have been proposed.

DILIGENT \cite{Pinto2009} was an early example of an agile methodology, proposing a complete process for ontology life-cycle management and knowledge evolution, as well as separating local changes (local views on knowledge) from global updates of the core part of the ontology, using a review process to authorise the propagation of changes from the local to the global level. This methodology is similar to how, for instance, the large clinical reference terminology SNOMED~CT~\cite{snomed2019} (also available as an ontology) is maintained and evolved, where the (international) core terminology is maintained based on global requirements, while national or local extensions to SNOMED CT are maintained based on local requirements. A group of authors then decides which national or local extensions to propagate to the core terminology. More modern agile methodologies include eXtreme Design (XD)~\cite{PresuttiDGB09,Blomqvist2016},  Modular Ontology Modelling (MOM) \cite{hitzler2016modeling,hitzler2018tutorial}, Simplified Agile Methodology for Ontology Development (SAMOD) \cite{peroni2016simplified}, etc. Such methodologies typically include two key elements: \textit{ontology requirements} and (more recently) \textit{ontology design patterns}.

Ontology requirements specify the intended task of the resulting ontology -- or indeed the knowledge graph itself -- based on the ontology as its schema. A common way to express ontology requirements is through Competency Questions (CQ) \cite{gruninger1995role}, which are natural language questions illustrating the typical knowledge that one would require the ontology (or the knowledge graph) to provide. Such CQs can then be complemented with additional restrictions, and reasoning requirements, in case that the ontology should also contain restrictions and general axioms for inferring new knowledge or checking data consistency. A common way of testing ontologies (or knowledge graphs based on them) is then to formalise the CQs as queries over some test set of data, and make sure the expected results are entailed~\cite{blomqvist2012ontology,keet2016test}. We may, for example, consider the CQ ``\textit{What are all the events happening in Santiago?}'', which can be represented as a graph query
\begin{tikzpicture}[baseline=-3pt]
    \setlength{\hgap}{1cm}
	\node[iri,compact,anchor=center] (e) {Event};

	\node[var,compact,anchor=center,right=1.1\hgap of e] (ev) {\textbf{?event}}
	edge[arrout] node[lab] {type} (e);

	\node[iri,compact,anchor=center,right=1.4\hgap of ev] (s) {Santiago}
	edge[arrin] node[lab] {location} (ev);
\end{tikzpicture}. Taking the data graph of Figure~\ref{fig:delg} and the axioms of Figure~\ref{fig:sg}, we can check to see if the expected result \gnode{EID15} is entailed by the ontology and the data, and since it is not, we may consider expanding the axioms to assert that \gedge[arrin][\thgap]{location}{type}{Transitive}.

Ontology Design Patterns (ODPs) are another common feature of modern methodologies~\cite{gangemi2005ontology,blomqvist2005patterns}, specifying generalisable ontology modelling patterns that can be used as inspiration for modelling similar patterns, as modelling templates~\cite{Egana2008,Skjaeveland2018}, or as directly reusable components~\cite{ODPportal,shimizu2019modl}. Several pattern libraries have been made available online, ranging from carefully curated ones~\cite{Aranguren2008,shimizu2019modl} to open and community moderated ones \cite{ODPportal}. As an example, in modelling an ontology for our scenario, we may decide to follow the Core Event ontology pattern proposed by \citet{KrisnadhiH16}, which specifies a spatio-temporal extent, sub-events, and participants of an event, further suggesting competency questions, formal definitions, etc., to support this pattern.

\subsubsection{Ontology learning} The previous methodologies outline methods by which ontologies can be built and maintained manually. Ontology learning, in contrast, can be used to \mbox{(semi-)automatically} extract information from text that is useful for the ontology engineering process~\cite{buitelaar2005ontology,cimiano2006ontology}. Early methods focussed on extracting terminology from text that may represent the relevant domain's classes; for example, from a collection of text documents about tourism, a terminology extraction tool -- using measures of \textit{unithood} that determine how cohesive an $n$-gram is as a unitary phrase, and \textit{termhood} that determine how relevant the phrase is to a domain~\cite{Martinez-Rodriguez18} -- may identify $n$-grams such as ``\textsf{visitor visa}'', ``\textsf{World Heritage Site}'', ``\textsf{off-peak rate}'', etc., as terminology of particular importance to the tourist domain, and that thus may merit inclusion in such an ontology. Axioms may also be extracted from text, where subclass axioms are commonly targetted, based on modifying nouns and adjectives that incrementally specialise concepts (e.g., extracting \gedge{Visitor Visa}{subc. of}{Visa} from the noun phrase ``\textsf{visitor visa}'' and isolated appearances of ``\textsf{visa}'' elsewhere), or using Hearst patterns~\cite{Hearst92} (e.g., extracting \gedge{Off-Peak Rate}{subc. of}{Discount} from ``\textsf{many \underline{discounts, such as off-peak rates}, are available}'' based on the pattern ``\textsf{\underline{X, such as Y}}''). Textual definitions can also be harvested from large texts to extract hypernym relations and induce a taxonomy from scratch~\cite{OntolearnReloaded13}.  More recent works aim to extract more expressive axioms from text, including disjointness axioms~\cite{Volker2015}; and axioms involving the union and intersection of classes, along with existential, universal, and qualified-cardinality restrictions~\cite{petrucci2016ontology}. The results of an ontology learning process can then serve as input to a more general ontology engineering methodology, allowing us to validate the terminological coverage of an ontology, to identify new classes and axioms, etc.

\section{Quality Assessment}\label{sec:quality}

Independently of the (kinds of) source(s) from which a knowledge graph is created, data extracted for the initial knowledge graph will usually be incomplete, and will contain duplicate, contradictory or even incorrect statements -- especially when taken from multiple sources. After the initial creation and enrichment of a knowledge graph from external sources, a crucial step is thus to assess the \textit{quality} of the resulting knowledge graph. By quality, we here refer to \textit{fitness for purpose}. Quality assessment then helps to ascertain for which purposes a knowledge graph can be reliability used.

In the following we discuss \textit{quality dimensions} that capture aspects of multifaceted data quality which evolves from the traditional domain of databases to the domain of knowledge graphs \cite{BatiniRSV15}, some of which are general, others of which are more particular to knowledge graphs~\cite{ZaveriRMPLA16}. While quality dimensions aim to capture qualitative aspects of the data, we also discuss \textit{quality metrics} that provide ways to measure quantitative aspects of these dimensions. We discuss groupings of dimensions and metrics as inspired by \citet{BatiniS16}.

\subsection{Accuracy}

Accuracy refers to the extent to which entities and relations -- encoded by nodes and edges in the graph -- correctly represent real-life phenomena. Accuracy can be further sub-divided into three dimensions: \textit{syntactic accuracy}, \textit{semantic accuracy}, and \textit{timeliness}.

\subsubsectionnd{Syntactic accuracy} is the degree to which the data are accurate with respect to the grammatical rules defined for the domain and/or data model. A prevalent example of syntactic inaccuracies occurs with datatype nodes, which may be incompatible with a defined range or be malformed. For example, assuming that a property \gelab{start} is defined with the range \texttt{xsd:dateTime}, taking a value such as \gnode{"March 29, 2019, 20:00"\dtsep xsd:string} would be incompatible with the defined range, while a value \gnode{"March 29, 2019, 20:00"\dtsep xsd:dateTime} would be malformed (a value such as \gnode{"2019-03-29T20:00:00"\dtsep xsd:dateTime} is rather expected). A corresponding metric for syntactic accuracy is the ratio between the number of incorrect values of a given property and the total number of values for the same property~\cite{ZaveriRMPLA16}. Such forms of syntactic accuracy can typically be assessed using validation tools~\cite{Furber,Hogan}. 

\subsubsectionnd{Semantic accuracy} is the degree to which data values correctly represent real world phenomena, which may be affected by imprecise extraction results, imprecise entailments, vandalism, etc. For instance, given that the National Congress of Chile is located in Valparaíso, this may give rise to the edge \gedge[arrin][1.6cm]{Chile}{capital}{Valparaiso} (through entailment, extraction, completion, etc.), which is in fact semantically inaccurate: the Chilean capital is Santiago. Assessing the level of semantic inaccuracies is challenging. While one option is to apply manual verification, an automatic option may be to check the stated relation against several sources~\cite{Lei,EstevesRRL18}. Another option is to rather validate the quality of individual processes used to generate the knowledge graph, based on measures such as precision, possibly with the help of human experts or gold standards~\cite{IESW}.

\subsubsectionnd{Timeliness} is the degree to which the knowledge graph is currently up-to-date with the real world state~\cite{KaferAUOH13}; in other words, a knowledge graph may be semantically accurate now, but may quickly become inaccurate (outdated) if no procedures are in place to keep it up-to-date in a timely manner. For example, consider a user checking the tourist knowledge graph for flights from one city to another. Suppose that the flight timetable is updated every minute with current flight statuses, but the knowledge graph is only updated every hour. In this case, we see that there is a quality issue regarding timeliness in the knowledge graph. Timeliness can be assessed based on how frequently the knowledge graph is updated with respect to underlying sources~\cite{KaferAUOH13,RulaPPM14}, which can be done using temporal annotations of changes in the knowledge graph~\cite{RulaPHSM12,RulaPRNLME19}, as well as contextual representations that capture the temporal validity of data (see Section~\ref{ssec:knowledgeContext}).

\subsection{Coverage} Coverage refers to avoiding the omission of domain-relevant elements, which otherwise may yield incomplete query results or entailments, biased models, etc.

\subsubsectionnd{Completeness} refers to the degree to which all required information is present in a particular dataset. Completeness comprises the following aspects: (i) \textit{schema completeness} refers to the degree to which the classes and properties of a schema are represented in the data graph, (ii) \textit{property completeness} refers to the ratio of missing values for a specific property, (iii) \textit{population completeness} provides the percentage of all real-world entities of a particular type that are represented in the datasets, and (iv) \textit{linkability completeness} refers to the degree to which instances in the data set are interlinked. Measuring completeness directly is non-trivial as it requires knowledge of a hypothetical \textit{ideal knowledge graph}~\cite{DarariNPR18} that contains all the elements that the knowledge graph in question should ``\textit{ideally}'' represent. Concrete strategies involve comparison with gold standards that provide samples of the ideal knowledge graph (possibly based on \textit{completeness statements}~\cite{DarariNPR18}), or measuring the recall of extraction methods from complete sources~\cite{IESW}, and so forth.

\subsubsectionnd{Representativeness} is a related dimension that, instead of focusing on the ratio of domain-relevant elements that are missing, rather focuses on assessing high-level \textit{biases} in what is included/excluded from the knowledge graph~\cite{Baeza-Yates18}. As such, this dimension assumes that the knowledge graph is incomplete -- i.e., that it is a sample of the ideal knowledge graph -- and asks how biased this sample is. Biases may occur in the data, in the schema, or during reasoning~\cite{Janowicz0RZM18}. Examples of data biases include geographic biases that under-represent entities/relations from certain parts of the world~\cite{Janowicz0RZM18}, linguistic biases that under-represent multilingual resources (e.g., labels and descriptions) for certain languages~\cite{KaffeePVSCP17}, social biases that under-represent people of particular genders or races~\cite{WagnerGGM16}, and so forth. In contrast, schema biases may result from high-level definitions extracted from biased data~\cite{Janowicz0RZM18}, semantic definitions that do not cover uncommon cases, etc. Unrecognised biases may lead to adverse effects; for example, if our tourism knowledge graph has a geographic bias towards events and attractions close to Santiago city -- due perhaps to the sources used for creation, the employment of curators from the city, etc. -- then this may lead to tourism in and around Santiago being disproportionally promoted (potentially compounding future biases). Measures of representativeness involve comparison of known statistical distributions with those of the knowledge graph, for example, comparing geolocated entities with known population densities~\cite{Janowicz0RZM18}, linguistic distributions with known distributions of speakers~\cite{KaffeePVSCP17}, etc. Another option is to compare the knowledge graph with general statistical laws, where \citet{SouletGMS18} use (non-)conformance with Benford's law\footnote{Benford's law states that the leading significant digit in many collections of numbers is more likely to be small.} to measure representativeness in knowledge graphs.

\subsection{Coherency} Coherency refers to how well the knowledge graph conforms to -- or is coherent with -- the formal semantics and constraints defined at the schema-level.

\subsubsectionnd{Consistency} means that a knowledge graph is free of (logical/formal) contradictions with respect to the particular logical entailment considered. For example, in the ontology of our knowledge graph, we may define that \begin{tikzpicture}[baseline=-3pt]
    \setlength{\hgap}{1cm}
	\node[iri,compact,anchor=center] (f) {flight};

	\node[iri,compact,anchor=center,right=1.1\hgap of f] (a) {Airport}
	edge[arrin] node[lab] {range} (f);

	\node[iri,compact,anchor=center,right=1.2\hgap of a] (c) {City}
	edge[arrin] node[lab] {disj. c.} (a);
\end{tikzpicture}, which when combined with the edges \begin{tikzpicture}[baseline=-3pt]
    \setlength{\hgap}{1cm}
	\node[iri,compact,anchor=center] (a) {Arica};

	\node[iri,compact,anchor=center,right=1.1\hgap of a] (s) {Santiago}
	edge[arrin] node[lab] {flight} (a);

	\node[iri,compact,anchor=center,right=1.1\hgap of s] (c) {City}
	edge[arrin] node[lab] {type} (s);
\end{tikzpicture}, gives rise to an inconsistency, entailing that \gnode{Santiago} is a member of the disjoint classes \gnode{City} and \gnode{Airport}. More generally, any semantic feature in Tables~\ref{tab:ontEqIneq}--\ref{tab:ontClass} with a ``not'' condition can give rise to inconsistencies if the negated condition is entailed. A measure of consistency can be the number of inconsistencies found in a knowledge graph, possibly sub-divided into the number of such inconsistencies identified by each semantic feature~\cite{BonattiHPS11}.

\subsubsectionnd{Validity} means that the knowledge graph is free of constraint violations, such as captured by shape expressions~\cite{ThorntonSSGMPW19} (see Section \ref{sssec:validating-schema}). We may, for example, specify a shape \shap{City} whose target nodes have at most one country. Then, given the edges  \begin{tikzpicture}[baseline=-3pt]
    \setlength{\hgap}{1cm}
	\node[iri,compact,anchor=center] (a) {Chile};

	\node[iri,compact,anchor=center,right=1.3\hgap of a] (s) {Santiago}
	edge[arrout] node[lab] {country} (a);

	\node[iri,compact,anchor=center,right=1.3\hgap of s] (c) {Cuba}
	edge[arrin] node[lab] {country} (s);
\end{tikzpicture}, and assuming that \gnode{Santiago} becomes a target of \shap{City}, we have a constraint violation. Conversely, even if we defined analogous cardinality restrictions in an ontology, this would not necessarily cause an inconsistency since, without UNA, we would first infer that \gnode{Chile} and \gnode{Cuba} refer to the same entity. A straightforward measure of validity is to count the number of violations per constraint.

\subsection{Succinctness} Succinctness refers to the inclusion only of relevant content (avoiding ``information overload'') that is represented in a concise and intelligible manner.

\subsubsectionnd{Conciseness} refers to avoiding the inclusion of schema and data elements that are irrelevant to the domain. \citet{MendesMB12} distinguish \textit{intensional conciseness} (schema level), which refers to the case when the data does not contain redundant schema elements (properties, classes, shapes, etc.), and \textit{extensional conciseness} (data level), when the data does not contain redundant entities and relations. For example, including events in \gnode{Santiago de Cuba} in our knowledge graph dedicated to tourism in Chile may affect the extensional conciseness of the knowledge graph, potentially returning irrelevant results for the given domain. In general, conciseness can be measured in terms of the ratio of properties, classes, shapes, entities, relations, etc., of relevance to the domain, which may in turn require a gold standard, or techniques to assess domain-relevance.

\subsubsectionnd{Representational-conciseness} refers to the extent to which content is compactly represented in the knowledge graph, which may again be intensional or extensional~\cite{ZaveriRMPLA16}. For example, having two properties \gelab{flight} and \gelab{flies to} serving the same purpose would negatively affect the intensional form of representational conciseness, while having two nodes \gnode{Santiago} and \gnode{Santiago de Chile} representing the capital of Chile (with neither linked to the other) would affect the extensional form of representational conciseness. Another example of representational conciseness is the unnecessary use of complex modelling constructs, such as using reification unnecessarily, or using linked lists when the order of elements is not important~\cite{HoganUHCPD12}. Though representational conciseness is challenging to assess, measures such as the number of redundant nodes can be used~\cite{Furber}.

\subsubsectionnd{Understandability} refers to the ease with which data can be interpreted without ambiguity by human users, which involves -- at least -- the provision of human-readable labels and descriptions (preferably in different languages~\cite{KaffeePVSCP17}) that allow them to understand what is being spoken about~\cite{HoganUHCPD12}. Referring back to Figure~\ref{fig:delg}, though the nodes \gnode{EID15} and \gnode{EID16} are used to ensure unique identifiers for events, they should also be associated with labels such as \gnode{Ñam} and \gnode{Food Truck}. Ideally the human readable information is sufficient to disambiguate a particular node, such as associating a description \gnode{"Santiago, the capital of Chile"@en} with \gnode{Santiago} to disambiguate the city from synonymous ones. Measures of understandability may include the ratio of nodes with human-readable labels and descriptions, the uniqueness of such labels and descriptions, the languages supported, etc.

\subsection{Other Quality Dimensions} We have discussed some key quality dimensions that have been discussed for -- and apply generally to -- knowledge graphs. Further dimensions may be pertinent in the context of specific domains, specific applications, or specific graph data models. For further details, we refer to the survey by~\citet{ZaveriRMPLA16} and to the book by \citet{BatiniS16}.

\section{Refinement}\label{sec:refine}

Beyond assessing the quality of a knowledge graph, there exist techniques to refine the knowledge graph, in particular to (semi-)automatically complete and correct the knowledge graph~\cite{Paulheim17}, aka \textit{knowledge graph completion} and \textit{knowledge graph correction}, respectively. As distinguished from the creation and enrichment tasks outlined in Section~\ref{sec:create}, refinement typically does not involve applying extraction or mapping techniques over external sources in order to ingest their content into the local knowledge graph. Instead, refinement typically targets improvement of the local knowledge graph as given (but potentially using external sources to verify local content~\cite{Paulheim17}).

\subsection{Completion}\label{ssec:completion}

Knowledge graphs are characterised by incompleteness~\cite{West14}. As such, knowledge graph completion aims at filling in the \textit{missing edges} (aka \textit{missing links}) of a knowledge graph, i.e., edges that are deemed correct but are neither given nor entailed by the knowledge graph. This task is often addressed with \emph{link prediction} techniques proposed in the area of \emph{Statistical Relational Learning}~\cite{Getoor07}, which predict the existence -- or sometimes more generally, predict the probability of correctness -- of missing edges. For instance, one might predict that the edge \gedge[arrin][\thgap]{Moon Valley}{bus}{San Pedro} is a probable missing edge for the graph of Figure~\ref{fig:chileTransport}, given that most bus routes observed are return services (i.e., \gelab{bus} is typically symmetric). Link prediction may target three settings: \textit{general links} involving edges with arbitrary labels, e.g., \gelab{bus}, \gelab{flight}, \gelab{type}, etc.; \textit{type links} involving edges with label \gelab{type}, indicating the type of an entity; and \textit{identity links} involving edges with label \gelab{same as}, indicating that two nodes refer to the same entity (cf.\ Section \ref{sssec:external_identy}). While type and identity links can be addressed using general link prediction techniques, the particular semantics of type and identity links can be addressed with custom techniques. (The related task of generating links across knowledge graphs -- referred to as \textit{link discovery}~\cite{nentwig2017survey} -- will be discussed later in Section~\ref{ssec:principles}.)  

\subsubsection{General link prediction}

Link prediction, in the general case, is often addressed with inductive techniques as discussed in Section~\ref{sec:inductive}, and in particular, knowledge graph embeddings and rule/axiom mining. For example, given Figure~\ref{fig:chileTransport}, using knowledge graph embeddings, we may detect that given an edge of the form \gedge[arrin][\thgap]{$x$}{bus}{$y$}, a (missing) edge \gedge[arrin][\thgap]{$y$}{bus}{$x$} has high plausibility, while using symbol-based approaches, we may learn the high-level rule \begin{tikzpicture}[baseline=-3pt]
	\node[var,compact](x){?x};
	
	\node[var,compact,right=\thgap of x](e){?y}
	edge[arrin] node[lab] {bus} (x);
	\end{tikzpicture} $\Rightarrow$ 
	\begin{tikzpicture}[baseline=-3pt]
	\node[var,compact](y){?y};
	
	\node[var,compact,right=\thgap of y](x){?x}
	edge[arrin] node[lab] {bus} (y);
	\end{tikzpicture}.
Either such approach would help us to predict the missing link \gedge[arrin][\thgap]{Moon Valley}{bus}{San Pedro}.

\subsubsection{Type-link prediction}

Type links are of particular importance to a knowledge graph, where dedicated techniques can be leveraged taking into account the specific semantics of such links. In the case of type prediction, there is only one edge label (\gelab{type}) and typically fewer distinct values (classes) than in other cases, such that the task can be reduced to a traditional classification task~\cite{Paulheim17}, training models to identify each semantic class based on features such as outgoing and/or incoming edge labels on their instances in the knowledge graph~\cite{paulheim2013type,SleemanF13}. For example, assume that in Figure~\ref{fig:chileTransport} we also know that \gnode{Arica}, \gnode{Calama}, \gnode{Puerto Montt}, \gnode{Punta Arenas} and \gnode{Santiago} are of \gelab{type} \gnode{City}. We may then predict that \gnode{Iquique} and \gnode{Easter Island} are also of \gelab{type} \gnode{City} based on the presence of edges labelled \gelab{flight} to/from these nodes, which (we assume) are learnt to be a good feature for prediction of that class (the former prediction is correct, while the latter is incorrect). Graph neural networks (see Section~\ref{ssec:gnns}) can also be used for node classification/type prediction. 

\subsubsection{Identity-link prediction}

\begin{figure}
	\setlength{\vgap}{1cm}
	\setlength{\hgap}{1.2cm}
	\setlength{\sgap}{0.3cm}
	\setlength{\mgap}{0.1cm}
	
	\centering
	\begin{tikzpicture}
	\node[iri,anchor=mid] (ei) {Easter Island};
	
	\node[iri,anchor=mid,below=\vgap of ei] (eip) {7400}
	edge[arrin] node[lab] {population} (ei);
	
	\node[iri,anchor=mid,left=\hgap of ei] (eilat) {27.74}
	edge[arrin] node[lab] {lat} (ei);
	
	\node[iri,anchor=mid,below=\vgap of eilat] (eilon) {109.26}
	edge[arrin] node[lab] {long} (ei);
	
	\node[iri,anchor=mid,right=1.7\hgap of ei] (hr) {Hanga Roa}
	edge[arrin] node[lab] {seat} (ei);
	
	\node[iri,anchor=mid,right=1.7\hgap of hr] (rn) {Rapa Nui}
	edge[arrout] node[lab] {seat} (hr);
	
	\node[iri,anchor=mid,below=\vgap of rn] (rnp) {7500}
	edge[arrin] node[lab] {population} (rn);
	
	\node[iri,anchor=mid,right=\hgap of rn] (rnlat) {27.8}
	edge[arrin] node[lab] {lat} (rn);
		
	\node[iri,anchor=mid,below=\vgap of rnlat] (rnlon) {109.2}
	edge[arrin] node[lab] {long} (rn);
	
	\node[iri,anchor=mid,below=\vgap of hr,xshift=-\hgap] (whs) {World Heritage Site}
	edge[arrin] node[lab] {named} (ei);
	
	\node[iri,anchor=mid,below=\vgap of hr,xshift=\hgap] (p) {Pacific}
		edge[arrin] node[lab] {ocean} (rn);
	\end{tikzpicture}
	\caption{Identity linking example, where Rapa Nui and Easter Island refer to the same island \label{fig:identity}}
\end{figure}
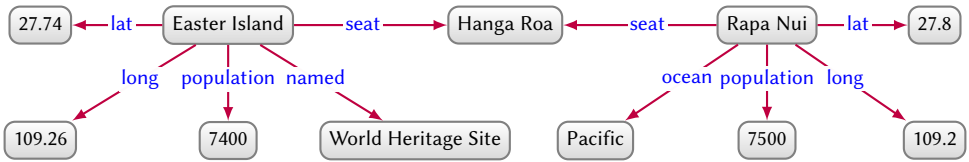

Predicting identity links involves searching for nodes that refer to the same entity; this is analogous to the task of \textit{entity matching} (aka record linkage, deduplication, etc.) considered in more general data integration settings~\cite{KopckeR10}. Such techniques are generally based on two types of \textit{matchers}: \textit{value matchers} determine how similar the values of two entities on a given property are, which may involve similarity metrics on strings, numbers, dates, etc.; while \textit{context matchers} consider the similarity of entities based on various nodes and edges~\cite{KopckeR10}. An illustrative example is given in Figure~\ref{fig:identity}, where value matchers will compute similarity between values such as \gnode{7400} and \gnode{7500}, while context matchers will compute similarity between \gnode{Easter Island} and \gnode{Rapa Nui} based on their surrounding information, such as their having similar latitudes, longitudes, populations, and the same seat (by way of comparison, a value matcher on this pair of nodes would measure string similarity between ``\texttt{Easter Island}'' and ``\texttt{Rapa Nui}''). 

A major challenge in this setting is efficiency, where a pairwise matching would require $O(n^2)$ comparisons for $n$ the number of nodes. To address this issue, \textit{blocking} can be used to group similar entities into (possibly overlapping, possibly disjoint) ``blocks'' based on similarity-preserving keys, with matching performed within each block~\cite{isele2011efficient,KopckeR10,DraisbachN11}; for example, if matching places based on latitude/longitude, blocks may represent geographic regions. An alternative to discrete blocking is to use \textit{windowing} over entities in a similarity-preserving ordering~\cite{DraisbachN11}, or to consider searching for similar entities within \textit{multi-dimensional spaces} (e.g., spacetime~\cite{santipantakis2019stld}, spaces with Minkowski distances \cite{minkowski}, orthodromic spaces~\cite{orchid}, etc.~\cite{SherifN18}). The results can either be pairs of nodes with a computed confidence of them referring to the same entity, or crisp identity links extracted based on a fixed threshold, binary classification, etc.~\cite{KopckeR10}. For confident identity links, the nodes' edges may then be \textit{consolidated}~\cite{HoganZUPD12}; for example, we may select \gnode{Easter Island} as the canonical node and merge the edges of \gnode{Rapa Nui} onto it, enabling us to find, e.g., \textit{World Heritage Sites in the Pacific Ocean} from Figure~\ref{fig:identity} based on the (consolidated) sub-graph \begin{tikzpicture}[baseline=-3pt]
\node[iri,compact](whs){World Heritage Site};

\node[iri,compact,right=1.2cm of whs](ei){Easter Island}
  edge[arrout] node[lab] {named} (whs);
  
\node[iri,compact,right=1.2cm of ei](p){Pacific}
  edge[arrin] node[lab] {ocean} (ei);
\end{tikzpicture}.

\subsection{Correction}

As opposed to completion -- which finds new edges in a knowledge graph -- correction identifies and removes existing incorrect edges in the knowledge graph. We here divide the principal approaches for knowledge graph correction into two main lines: \textit{fact validation}, which assigns a plausibility score to a given edge, typically in reference to external sources; and \textit{inconsistency repairs}, which aim to resolve inconsistencies found in the knowledge graph through ontological axioms.

\subsubsection{Fact validation}

The task of \emph{fact validation} (aka \emph{fact checking})~\cite{gerber2015defacto,syed2018factcheck,yin2008truth, syed2019copaal,EstevesRRL18,shiralkar2017finding,shi2016discriminative,socher2013reasoning,bordes2013translating} involves assigning plausibility or \textit{veracity} scores 
to facts/edges, typically between $0$ and $1$. An ideal fact-checking function assumes a hypothetical reference universe (an ideal knowledge graph) and would return $1$
for the fact \gedge[arrin][\thgap]{Santa Lucía}{city}{Santiago} (being true) while returning $0$ for \gedge[arrin][\thgap]{Sotomayor}{city}{Santiago} (being false). There is a clear relation between fact validation and link prediction -- with both relying on assessing the plausibility of edges/facts/links -- and indeed the same numeric- and symbol-based techniques can be applied for both cases. However, fact validation often considers online assessment of edges given as input, whereas link prediction is often an offline task that generates novel candidate edges to be assessed from the knowledge graph. Furthermore, works on fact validation are characterised by their consideration of external reference sources, which may be \emph{unstructured sources}~\cite{gerber2015defacto,syed2018factcheck,Samadi2016,yin2008truth} or \emph{structured sources} ~\cite{syed2019copaal,shiralkar2017finding,shi2016discriminative,socher2013reasoning,bordes2013translating}.

Approaches based on unstructured sources assume that they are given a \textit{verbalisation function} -- using, for example, rule-based approaches~\cite{ngonga2013sorry,ell2014sparql}, encoder--decoder architectures~\cite{gardent2017webnlg}, etc. -- that is able to translate edges into natural language. Thereafter, approaches for computing the plausibility of facts in natural language -- called \emph{fact finders}~\cite{Pasternack2010,pasternack2011making} -- can be directly employed. Many fact finding algorithms construct an $n$-partite (often bipartite) graph whose nodes are facts and sources, where a source is connected to a fact if the source ``evidences'' the fact, i.e., if it contains a text snippet that matches -- with sufficient confidence -- the verbalisation of the input edge. Two mutually-dependent scores, namely the trustworthiness of sources and the plausibility of facts, are then calculated based on this graph, where fact finders differ on how they compute these scores~\cite{pasternack2011making}. Here we mention three scores proposed by~\citet{Pasternack2010}:
\begin{itemize} 
\item \emph{Sums}~\cite{Pasternack2010} adapts the HITS algorithm~\cite{kleinberg1999hubs} by defining sources as hubs (with 0 authority score) and facts as authorities (with 0 hub score).
\item \emph{Average Log}~\cite{Pasternack2010} extends HITS with a normalisation factor that prevents a single source from receiving a high trustworthiness score by evidencing many facts (that may be false).
\item \emph{Investment}~\cite{Pasternack2010} lets the scores of facts grow with a non-linear function based on ``investments'' coming from the connected sources. The score a source receives from a fact is based on the individual facts in this particular source compared to the other connected sources. 
\end{itemize}
\noindent
\citet{pasternack2011making} then show that these three algorithms can be generalised into a single multi-layered graph-based framework within which (1) a source can support a fact with a weight expressing uncertainty, (2) similar facts can support each other, and (3) sources can be grouped together leading to an implicit support between sources of the same group. Other approaches for fact checking of knowledge graphs later extended this framework~\cite{galland2010,Samadi2016}. Alternative approaches based on classifiers have also emerged, where commonly-used features include trust scores for information sources, co-occurrences of facts in sources, and so forth~\cite{gerber2015defacto,syed2018factcheck}.

Approaches for fact validation based on structured data typically assume external knowledge graphs as reference sources and are based on finding paths that evidence the input edge being validated. Unsupervised approaches search for undirected~\cite{shiralkar2017finding,ciampaglia2015computational} or directed~\cite{syed2019copaal} paths up to a given threshold length that evidence the input edge. The relatedness between input edges and paths is computed using a mutual information function, such as normalized pointwise mutual information~\cite{bouma2009normalized}. Supervised approaches rather extract features for input edges from external knowledge graphs~\cite{sun2011pathsim,zhao2015automatic,lao2010relational} and use these features to train a classification model to label the edges as true or false. An important set of features are  \emph{metapaths}, which encode sequences of predicates that correlate positively with the edge label of the input edge. Amongst such works, PredPath~\cite{shi2016discriminative} automatically extracts metapaths based on type information. Several approaches rather encode the reference nodes and edges using graph embeddings (see Section~\ref{ssec:embeddings}), which are then used to estimate the plausibility of the input edge being validated. 

\subsubsection{Inconsistency repairs} Ontologies can contain axioms -- such as disjointness -- that lead to inconsistencies. While such axioms can be provided by experts, they can also be derived through symbolic learning, as discussed in Section~\ref{ssec:symlearn}. Such axioms can then be used to detect inconsistencies. With respect to correcting a knowledge graph, however, detecting inconsistencies is not enough: techniques are also required to \textit{repair} such inconsistencies, which itself is not a trivial task. In the simplest case, we may have an instance of two disjoint classes, such as that \gnode{Santiago} is of type \gnode{City} and \gnode{Airport}, which are stated or found to be disjoint. To repair the inconsistency, it would be preferable to remove only the ``incorrect'' class, but which should we remove? This is not a trivial question, particularly if we consider that one edge can be involved in many inconsistencies, and one inconsistency can involve many edges. The issue of computing repairs becomes more complex when entailment is considered, where we not only need to remove the stated type, but also all of the ways in which it might be entailed; for example, removing the edge \gedge[arrin][\thgap]{Santiago}{type}{Airport} is insufficient if we further have an edge \gedge[arrin][\thgap]{Arica}{flight}{Santiago} combined with an axiom \gedge[arrin][\thgap]{flight}{range}{Airport}. \citet{TopperKS12} suggest potential repairs for such violations -- remove a domain/range constraint, remove a disjointness constraint, remove a type edge, remove an edge with a domain/range constraint -- where one is chosen manually. In contrast, \citet{BonattiHPS11} propose an automated method to repair inconsistencies based on \textit{minimal hitting sets}~\cite{Reiter87}, where each set is a minimal explanation for an inconsistency. The edges to remove are chosen based on scores of the trustworthiness of their sources and how many minimal hitting sets they are either elements of or help to entail an element of, where the knowledge graph is revised to avoid re-entailment of the removed edges.  Rather than repairing the data, another option is to evaluate queries under inconsistency-aware semantics, such as returning \textit{consistent answers} valid under every possible repair~\cite{LukasiewiczMS13}.

\subsection{Other refinement tasks}

In comparison to the quality clusters discussed in Section~\ref{sec:quality}, the refinement methods discussed here address particular aspects of the accuracy, coverage, and coherency dimensions. Beyond these, one could conceive of further refinement methods to address further quality issues of knowledge graphs, such as succinctness. In general, however, the refinement tasks of \textit{knowledge graph completion} and \textit{knowledge graph correction} have received the majority of attention until now. For further details on knowledge graph refinement, we refer to the survey by~\citet{Paulheim17}.

\section{Publication}\label{sec:publish}

While it may not be desirable to publish, for example, enterprise knowledge graphs that offer a competitive advantage to a company~\cite{NoyGJNPT19}, it may be desirable -- or even required -- to publish other knowledge graphs, such as those produced by volunteers~\cite{VrandecicK14,MahdisoltaniBS15,LehmannIJJKMHMK15}, 
by publicly-funded research~\cite{CallahanCAD13,GrothLGGHP14,uniprot2014}, by governmental organisations~\cite{HendlerHMT12,ShadboltO13}, etc. Publishing refers to making the knowledge graph (or part thereof) accessible to the public, often over the Web. Knowledge graphs published as open data are then called open knowledge graphs (discussed in Section~\ref{sec:openkgs}).

In the following, we first discuss two sets of principles that have been proposed to guide the publication of data on the Web. We next discuss access protocols that constitute the interfaces by which the public can interact with the content of a knowledge graph. Finally, we consider techniques to restrict the access or usage of (parts of) a knowledge graph, as appropriate. 

\subsection{Best Practices}\label{ssec:principles}

We now discuss two key sets of principles for publishing data, namely the FAIR Principles proposed by~\citet{wilkinson2016fair}, and the Linked Data Principles proposed by \citet{ldprinciples}.

\subsubsection{FAIR Principles}\label{ssec:fair}

The FAIR Principles were originally proposed in the context of publishing scientific data~\cite{wilkinson2016fair} -- particularly motivated by maximising the impact of publicly-funded research -- but the principles generally apply to other situations where data are to be published in a manner that facilitates their re-use by external agents, with particular emphasis on machine-readability.

FAIR itself is an acronym for four foundational principles, each with particular goals~\cite{wilkinson2016fair}, that may apply to \textit{data}, \textit{metadata}, or both -- the latter being denoted \textit{(meta)data}.\footnote{Metadata are data about data. The distinction is often important in observational sciences, where in astronomy, for example, data may include raw image data, while metadata may include the celestial coordinates and time of the image.} We now describe the FAIR principles (slightly rephrasing the original wording in some cases for brevity~\cite{wilkinson2016fair}).

\def\labelitemiii{$\circ$}
\begin{itemize}
\item \textit{Findability} refers to the ease with which external agents who might benefit from the dataset can initially locate the dataset. Four sub-goals should be met:
\begin{itemize}
\item F1: (meta)data are assigned a globally unique and persistent identifier.
\item F2: data are described with rich metadata (see R1).
\item F3: metadata clearly and explicitly include the identifier of the data they describe.
\item F4: (meta)data are registered or indexed in a searchable resource.
\end{itemize}

\item \textit{Accessibility} refers to the ease with which external agents (once they have located the dataset) can access the dataset. Two goals are defined, the first with two sub-goals:
\begin{itemize}
\item A1: (meta)data are retrievable by their identifier using a standard protocol.
\begin{itemize}
\item A1.1: the protocol is open, free, and universally implementable.
\item A1.2: the protocol allows for authentication and authorisation, where necessary.
\end{itemize}
\item A2. metadata are accessible, even when the data are no longer available.
\end{itemize}

\item \textit{Interoperability} refers to the ease with which the dataset can be exploited (in conjunction with other datasets) using standard tools. Three goals are defined:
\begin{itemize}
\item I1: (meta)data use an accessible, shared, and general knowledge representation formalism.
\item I2: (meta)data use vocabularies that follow FAIR principles.
\item I3: (meta)data include qualified references to other (meta)data.
\end{itemize}

\item \textit{Reusability} refers to the ease with which the dataset can be re-used in conjunction with other datasets. One goal is defined (with three sub-goals):
\begin{itemize}
\item R1: meta(data) are richly described with a plurality of accurate and relevant attributes.
\begin{itemize}
\item R1.1. (meta)data are released with a clear and accessible data usage license.
\item R1.2. (meta)data are associated with detailed provenance.
\item R1.3. (meta)data meet domain-relevant community standards.
\end{itemize}
\end{itemize}
\end{itemize}

\noindent In the context of knowledge graphs, a variety of vocabularies, tools, and services have been proposed that both directly and indirectly help to satisfy the FAIR principles. In terms of \textit{Findability}, as discussed in Section~\ref{sec:graph}, IRIs are built into the RDF model, providing a general schema for global identifiers. In addition, resources such as the Vocabulary of Interlinked Datasets (VoID)~\cite{AlexanderCHZ09} allow for representing meta-data about graphs, while services such as DataHub~\cite{BhardwajBCDEMP15} provide a central repository of such dataset descriptions. Access protocols that enable \textit{Accessibility} will be discussed in Section~\ref{ssec:access}, while mechanisms for authorisation will be discussed in Section~\ref{ssec:UsageControl}. With respect to \textit{Interoperability}, as discussed in Section~\ref{sec:deductive}, ontologies serve as a general knowledge representation formalism, and can in turn be used to describe vocabularies that follow FAIR principles. Finally, regarding \textit{Reusability}, licensing will be discussed in Section~\ref{ssec:UsageControl}, while the \textit{PROV Data Model}~\cite{prov13} discussed in Section~\ref{sec:knowledge} allows for capturing detailed provenance.

A number of knowledge graphs have been published using FAIR principles, where \citet{wilkinson2016fair} explicitly mention Open PHACTS~\cite{GrothLGGHP14}, a data integration platform for drug discovery, and UniProt~\cite{uniprot2014}, a large collection of protein sequence and annotation data, as conforming to FAIR principles. Both datasets offer graph views of their content through the RDF data model.

\subsubsection{Linked Data Principles}\label{sssec:ld}

\begin{figure}
\setlength{\hgap}{3.2cm}
\setlength{\vgap}{1.4cm}

\colorlet{cng}{orange!15}
\colorlet{cnd}{blue!3}
\colorlet{ex1fill}{orange!15}
\colorlet{ex1text}{orange!30!black}
\colorlet{ex2fill}{green!15}
\colorlet{ex2text}{green!30!black}
\colorlet{ex3fill}{blue!15}
\colorlet{ex3text}{blue!30!black}
\colorlet{ex4fill}{purple!15}
\colorlet{ex4text}{purple!30!black}
\colorlet{ex5fill}{black!15}
\colorlet{ex5text}{black!80}
\colorlet{exbg}{black!03}

\tikzset{
	elab/.style={ 
        lab,
		fill=exbg, 
		opacity=1,
		text opacity=1,
		style={inner sep=0.8,outer sep=0.8}
	}
}

\newcommand{\dtl}[2]{{\def\arraystretch{0.6}\begin{tabular}{@{}c@{}}#1\\{\scriptsize\color{black!65} #2}\end{tabular}}}

\setlength{\hgap}{3.6cm}

\newlength{\hhgap}
\setlength{\hhgap}{\hgap}

\newlength{\vvgap}
\setlength{\vvgap}{0.2\vgap}

\newlength{\bbigheight}
\setlength{\bbigheight}{5.5cm}

\newlength{\bsmallheight}
\setlength{\bsmallheight}{3cm}

\newlength{\bsmallwidth}
\setlength{\bsmallwidth}{2.2\hgap}

\newlength{\bssmallwidth}
\setlength{\bssmallwidth}{1.5\hgap}

\newlength{\bbigwidth}
\setlength{\bbigwidth}{3.83\hgap}

\newcommand{\der}[2]{{\def\arraystretch{0.6}\begin{tabular}{@{}l@{}}#1\\\multicolumn{1}{r}{\color{purple}\ensuremath{\hookrightarrow}\footnotesize \Mundus~#2}\end{tabular}}}

\centering

\begin{tikzpicture}

\node[rectangle,minimum width=\bsmallwidth,minimum
height=\bsmallheight,fill=ex1fill,anchor=mid] (cldlp) at (0,2.3)  {};

\node[rectangle,fill=white!93!black,anchor=south west] (tabone) at (cldlp.north west) {\color{purple}\Mundus~\texttt{cld:LP2018} $\times$};

\node[above right=\vvgap of cldlp.south west,iri,anchor=south west] (cllp)
{\der{cle:LP2018}{cld:LP2018}};

\node[above=\vgap of cllp,iri,anchor=mid] (concert) {\der{clv:Concert}{clv:vocab}}
  edge[arrin] node[lab,fill=ex1fill,yshift=-0.3ex] {\der{rdf:type}{rdf:}}
(cllp);  

\node[right=\hgap of concert,iri,anchor=mid] (wdpj)
{\der{wd:Q142701}{wdd:Q142701}}
  edge[arrin] node[lab,fill=ex1fill,xshift=2ex] {\der{clv:headliner}{clv:vocab}}
(cllp);
  
\node[right=\hgap of cllp,lit,anchor=mid] (cllpname)
{"Lollapalooza 2018"} 
  edge[arrin] node[lab,fill=ex1fill,xshift=-0.5ex] {\der{rdfs:label}{rdfs:}} (cllp);
 
\node[rectangle, minimum width=\bssmallwidth,minimum
height=\bsmallheight,fill=ex2fill,anchor=south west,xshift=0.1\vgap] (wddpj) at
(cldlp.south east)  {}; 

\node[rectangle,fill=white!93!black,anchor=south west] (tabtwo) at (wddpj.north west) {\color{purple}\Mundus~\texttt{wdd:Q142701} $\times$};

  
\node[right=0.77\hgap of wdpj,iri,anchor=mid] (wdpjB)
{\der{wd:Q142701}{wdd:Q142701}};

\node[below=\vgap of wdpjB,lit,anchor=mid,xshift=-0.38\hgap] (dob)
{\dtl{"1990"}{xsd:gYear}}
   edge[arrin] node[lab,fill=ex2fill,xshift=-0.4cm] {\der{wdt:P571}{wdd:P571}} (wdpjB);
   
\node[below=\vgap of wdpjB,iri,anchor=mid,xshift=0.38\hgap] (fpw)
{\der{wd:Q221535}{wdd:Q221535}}
   edge[arrin] node[lab,fill=ex2fill,xshift=0.4cm] {\der{wdt:P527}{wdd:P527}} (wdpjB);
    
\end{tikzpicture}

\caption{Two example Linked Data documents from two websites, each containing an RDF graph, where \texttt{wd:Q142701} refers to Pearl Jam in Wikidata while \texttt{wdd:Q142701} refers to the RDF graph about Pearl Jam, and where \texttt{wd:Q221535} refers to Eddie Vedder while \texttt{wdd:Q221535} refers to the RDF graph about Eddie Vedder; the edge-label \gelab{wdt:571} refers to ``inception'' in Wikidata, while \gelab{wdt:527} refers to ``has part''}\label{fig:ld}
\end{figure}
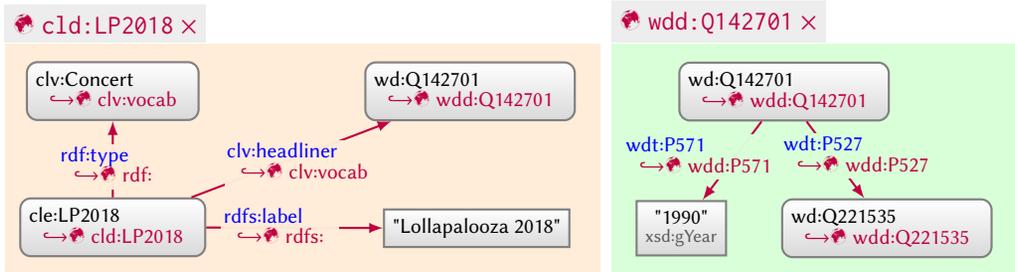

\citet{wilkinson2016fair} state that FAIR Principles ``precede implementation choices'', meaning that the principles do not cover \textit{how} they can or should be achieved. Preceding the FAIR Principles by almost a decade are the Linked Data Principles, proposed by \citet{ldprinciples}, which provide a technical basis for one possible way in which these FAIR Principles can be achieved. Specifically the Linked Data Principles are as follows:

\begin{enumerate}
\item Use IRIs as names for things.
\item Use HTTP IRIs so those names can be looked up.
\item When a HTTP IRI is looked up, provide useful content about the entity that the IRI names using standard data formats.
\item Include links to the IRIs of related entities in the content returned.
\end{enumerate}

\noindent These principles were proposed in a Semantic Web setting, where for principle~(3), the standards based on RDF (including RDFS, OWL, etc.) are currently recommended for use, particularly because they allow for naming entities using HTTP IRIs, which further paves the way for satisfying all four principles. As such, these principles outline a way in which (RDF) graph-structured data can be published on the Web such that these graphs are interlinked to form what \citet{ldprinciples} calls a ``Web of Data'', whose goal is to increase automation on the Web by making content available not only in (HTML) documents intended for human consumption, but also as (RDF) structured data that machines can locate, retrieve, combine, validate, reason over, query over, etc., towards solving tasks automatically. Conceptually, the Web of Data is then composed of graphs of data published on individual web-pages, where one can click on a node or edge-label -- or more precisely perform a HTTP lookup on an IRI of the graph -- to be transported to another graph elsewhere on the Web with relevant content on that node or edge-label, and so on recursively.

In Figure~\ref{fig:ld}, we show a simple example with two Linked Data documents published on the Web, with each containing an RDF graph. As discussed in Section~\ref{sec:identity}, terms such as \texttt{clv:Concert}, \texttt{wd:Q142701}, \texttt{rdfs:label}, etc., are abbreviations for IRIs, where, for example, \texttt{wd:Q142701} expands to \url{http://www.wikidata.org/entity/Q142701}. Prefixes beginning with \texttt{cl} are fictitious prefixes we assume to have been created by the Chilean tourist board. The IRIs prefixed with {\color{purple}$\hookrightarrow$\Mundus} indicate the document returned if the node is looked up. The leftmost document is published by the tourist board and describes Lollapalooza 2018 (identified by the node \gnode{cle:LP2018}), which links to the headlining act Pearl Jam (\gnode{wd:Q142701}) described by an external knowledge graph, namely Wikidata. By looking up the node \gnode{wd:Q142701} in the leftmost graph, the IRI \textit{dereferences} (i.e., returns via HTTP) the document with the RDF graph on the right describing that entity in more detail. From the rightmost document, the node \gnode{wd:Q221535} can be looked up, in turn, to find a graph about Eddie Vedder (not shown in the example). The IRIs for entities and documents are distinguished to ensure that we do not confuse data about the entity and the document; for example, while \texttt{wd:Q221535} refers to Eddie Vedder, the IRI \texttt{\color{purple}{wd\textbf{d}:Q221535}} refers to the document about Eddie Vedder; if we were to assign a last-modified date to the document, we should use \gnode{\color{purple}{wd\textbf{d}:Q221535}} not \gnode{wd:Q221535}. In Figure~\ref{fig:ld}, we can further observe that edge labels (which are also IRIs) and nodes representing classes (e.g., \gnode{clv:Concert}) can also be dereferenced, typically returning semantic definitions of the respective terms. 

A key challenge is posed by the fourth principle -- include links to related entities -- as illustrated in Figure~\ref{fig:ld}, where \gnode{wd:Q221535} in the leftmost graph constitutes a link to related content about Pearl Jam in an external knowledge graph. Specifically, the \textit{link discovery} task considers adding such links from one knowledge graph to another, which may involve inclusion of IRIs that dereference to external graphs (per Figure~\ref{fig:ld}), or links with special semantics such as identity links. In comparison with the link prediction task discussed in Section~\ref{ssec:completion}, which is used to complete links within a knowledge graph, link discovery aims to discover links across knowledge graphs, which involves unique aspects: first, link discovery typically considers disjoint sets of source (local) nodes and target (remote) nodes; second, the knowledge graphs may often use different vocabularies; third, while in link prediction there already exist local examples of the links to predict, in link discovery, there are often no existing links between knowledge graphs to learn from. A common technique is to define manually-crafted linkage rules (aka link specifications) that apply heuristics for defining links that potentially incorporate similarity measures~\cite{NgomoA11,silk}. Link discovery is greatly expedited by the provision of standard identifier schemes within knowledge graphs, such as ISBNs for books, alpha-2 and alpha-3 codes for countries (e.g., \textsc{cl}, \textsc{clp}), or even links to common knowledge graphs such as DBpedia~\cite{LehmannIJJKMHMK15} or Wikidata~\cite{VrandecicK14} (that themselves include standard identifiers). We refer to the survey on link discovery by \citet{nentwig2017survey} for more details. 

Further guidelines have been proposed that provide finer-grained recommendations for publishing Linked Data, relating to how best implement dereferencing, what kinds of links to include, how to publish and interlink vocabularies, amongst other considerations~\cite{ldbook,JanowiczHAKV14}. We refer to the book by \citet{ldbook} for more discussion on how to publish Linked Data on the Web.

\subsection{Access Protocols}\label{ssec:access}

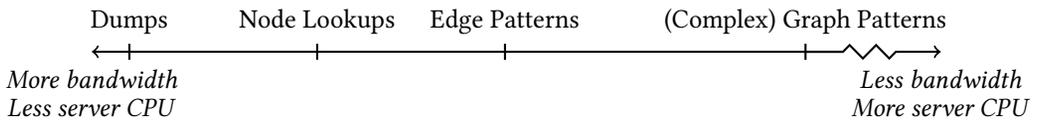
\begin{figure}[tb]
\begin{tikzpicture}[decoration={zigzag, pre length=5mm, post length=5mm}]
    \draw[<-,thick] (-0.5,0) -- (9.5,0);
    \draw[decorate,thick,->] (9,0) -- (10.8,0); 

    \foreach \x in {0,2.5,5,9}
      \draw[thick] (\x cm,3pt) -- (\x cm,-3pt);

    \draw (-0.5,0) node[below=3pt] {\textit{More bandwidth}} node[below=14pt] {\textit{Less server CPU}};
    \draw (0,0) node[above=3pt] {Dumps};
    \draw (2.5,0) node[above=3pt] {Node Lookups};
    \draw (5,0) node[above=3pt] {Edge Patterns};
    \draw (9,0) node[above=3pt] {(Complex) Graph Patterns};
    \draw (10.8,0) node[below=3pt] {\textit{Less bandwidth}} node[below=14pt] {\textit{More server CPU}};
  \end{tikzpicture}
  \caption{Access protocols for knowledge graphs, from simple protocols (left) to more complex protocols (right) \label{fig:access}}
\end{figure}

Publishing involves allowing the public to interact with the knowledge graph, which implies the provision of \textit{access protocols} that define the requests that agents can make and the response that they can expect as a result. Per the \textit{Accessibility} principle of FAIR (specifically A1.1), this protocol should be open, free, and universally implementable. In the context of knowledge graphs, as shown in Figure~\ref{fig:access}, there are a number of access protocols to choose from, varying from simple protocols that allow users to simply download all content, towards protocols that accept and evaluate increasingly complex requests. While simpler protocols require less computation on the server that publishes the data, more complex protocols allow agents to request more specific data, thus reducing bandwidth. A knowledge graph may also offer a variety of access protocols catering to different agents with different requirements~\cite{VerborghSCCMW14}. We now discuss such access protocols.

\subsubsection{Dumps}

A dump is a file or collection of files containing the content of the knowledge graph available for download. The request in this case is for the file(s) and the response is the content of the file(s). In order to publish dumps, first of all, concrete -- and ideally standard -- syntaxes are required to serialise the graph. While for RDF graphs there are various standard syntaxes available based on XML~\cite{rdfxml11}, JSON~\cite{jsonld}, custom syntaxes~\cite{turtle}, and more besides, currently there are only non-standard syntaxes available for property graphs~\cite{TomaszukASLC19}. Second, to reduce bandwidth, compression methods can be applied. While standard compression such as GZIP or BZip2 can be straightforwardly applied on any file, custom compression methods have been proposed for graphs that not only offer better compression ratios than these standard methods, but also offer additional functionalities, such as compact indexes for performing efficient lookups once the file is downloaded~\cite{FernandezMGPA13}. Finally, to further reduce bandwidth, when the knowledge graph is updated, ``diffs'' can be computed and published to obviate the need for agents to download all data from scratch (see~\cite{TummarelloMBE07,PapavasileiouFFKC13,AhnIEZK14}). Still, however, dumps are only suited to certain use-cases, in particular for agents that wish to maintain a full local copy of a knowledge graph. If an agent were rather only interested in, for example, all food festivals in Santiago, downloading the entire dump would require transferring and processing a lot of irrelevant data.

\subsubsection{Node lookups} Protocols for performing node lookups accept a node (id) request (e.g., \gnode{cle:LP2018} in Figure~\ref{fig:ld}) and return a (sub-)graph describing that node (e.g., the document \texttt{\color{purple}cld:LP2018}). Such a protocol is the basis for the Linked Data principles outlined previously, where node lookups are implemented through HTTP dereferencing, which further allows nodes in remote graphs to be referenced from across the Web. Although there are varying definitions on what content should be returned for a node~\cite{cbd}, a common convention is to return a sub-graph containing either all outgoing edges for that node or all incident edges (both outgoing and incoming) for that node~\cite{HoganUHCPD12}. Though simple, mechanisms for answering graph patterns can be implemented on top of a node lookup interface by traversing from node to node according to the particular graph pattern~\cite{HartigBF09}; for example, to find all food festivals in Santiago -- represented by the graph pattern \begin{tikzpicture}[baseline=-3pt]
    \setlength{\hgap}{1cm}
	\node[iri,compact,anchor=center] (e) {Food Festival};
	
	\node[var,compact,anchor=center,right=1.1\hgap of e] (ev) {\textbf{?ff}}
	edge[arrout] node[lab] {type} (e);
	
	\node[iri,compact,anchor=center,right=1.4\hgap of ev] (s) {Santiago}
	edge[arrin] node[lab] {location} (ev);  
\end{tikzpicture} -- we may perform a node lookup for \gnode{Santiago}, subsequently performing a node lookup for each node connected by a \gelab{location} edge to \gnode{Santiago}, returning those nodes declared to be of type \gnode{Food Festival}. However, such an approach may not be feasible if no starting node is declared (e.g., if all nodes are variables), if the node lookup service does not return incoming edges, etc. Furthermore, the client agent may need to request more data than necessary, where the document returned for \gnode{Santiago} may return a lot of irrelevant data, and where nodes with a \gelab{location} in \gnode{Santiago} that do not represent instances of \gnode{Food Festival} still need to be looked up to check their type. On the plus side, node lookups are relatively inexpensive for servers to support.

\subsubsection{Edge patterns}

Edge patterns -- also known as \textit{triple patterns} in the case of directed, edge-labelled graphs -- are singleton graph patterns, i.e., graph patterns with a single edge. Examples of edge patterns are \begin{tikzpicture}[baseline=-3pt]
    \setlength{\hgap}{0.7cm}

	\node[var,compact,anchor=center] (ev) {\textbf{?ff}};
	
	\node[iri,compact,anchor=center,right=1.4\hgap of ev] (s) {Food Festival}
	edge[arrin] node[lab] {type} (ev);  
\end{tikzpicture} or \begin{tikzpicture}[baseline=-3pt]
    \setlength{\hgap}{1cm}

	\node[var,compact,anchor=center] (ev) {\textbf{?ff}};
	
	\node[iri,compact,anchor=center,right=1.4\hgap of ev] (s) {Santiago}
	edge[arrin] node[lab] {location} (ev);  
\end{tikzpicture}, etc., where any term can be a variable or a constant. A protocol for edge patterns accepts such a pattern and returns all solutions for the pattern. Edge patterns provide more flexibility than node lookups, where graph patterns are more readily decomposed into edge patterns than node lookups. With respect to the agent interested in food festivals in Santiago, they can first, for example, request solutions for the edge pattern \begin{tikzpicture}[baseline=-3pt]
    \setlength{\hgap}{1cm}

	\node[var,compact,anchor=center] (ev) {\textbf{?ff}};
	
	\node[iri,compact,anchor=center,right=1.4\hgap of ev] (s) {Santiago}
	edge[arrin] node[lab] {location} (ev);  
\end{tikzpicture} and locally join/intersect these solutions with those of \begin{tikzpicture}[baseline=-3pt]
    \setlength{\hgap}{0.7cm}

	\node[var,compact,anchor=center] (ev) {\textbf{?ff}};
	
	\node[iri,compact,anchor=center,right=1.4\hgap of ev] (s) {Food Festival}
	edge[arrin] node[lab] {type} (ev);  
\end{tikzpicture}. Given that some edge patterns (e.g., \begin{tikzpicture}[baseline=-3pt]
    \setlength{\hgap}{0.5cm}

	\node[var,compact,anchor=center] (ev) {\textbf{?x}};
	
	\node[var,compact,anchor=center,right=1.4\hgap of ev] (s) {\textbf{?z}}
	edge[arrin] node[lab] {\color{black}\textbf{?y}} (ev);  
\end{tikzpicture}) can return many solutions, protocols for edge patterns may offer additional practical features such as iteration or pagination over results~\cite{VerborghSHHVMHC16}. Much like node lookups, the server cost of responding to a request is relatively low and easy to predict. However, the server may often need to transfer irrelevant intermediate results to the client, which in the previous example may involve returning nodes located in Santiago that are not food festivals. This issue is further aggravated if the client does not have access to statistics about the knowledge graph in order to plan how to best perform the join; for example, if there are relatively few food festivals but many things located in Santiago, rather than intersecting the solutions of the two aforementioned edge patterns, it should be more efficient to send a request for each food festival to see if it is in Santiago, but deciding this requires statistics about the knowledge graph. Extensions to the edge-pattern protocol have thus been proposed to allow for more efficient joins~\cite{HartigLP17}, such as allowing batches of solutions to be sent alongside the edge pattern, returning only solutions compatible with the solutions in the request~\cite{HartigA16} (e.g., sending a batch of solutions for \begin{tikzpicture}[baseline=-3pt]
    \setlength{\hgap}{0.7cm}

	\node[var,compact,anchor=center] (ev) {\textbf{?ff}};
	
	\node[iri,compact,anchor=center,right=1.4\hgap of ev] (s) {Food Festival}
	edge[arrin] node[lab] {type} (ev);  
\end{tikzpicture} to join with the solutions for the request \begin{tikzpicture}[baseline=-3pt]
    \setlength{\hgap}{1cm}

	\node[var,compact,anchor=center] (ev) {\textbf{?ff}};
	
	\node[iri,compact,anchor=center,right=1.4\hgap of ev] (s) {Santiago}
	edge[arrin] node[lab] {location} (ev);  
\end{tikzpicture}).

\subsubsection{(Complex) graph patterns} Another alternative is to let client agents make requests based on (complex) graph patterns (see Section~\ref{ssec:querying}), with the server returning (only) the final solutions. In our running example, this involves the client issuing a request for \begin{tikzpicture}[baseline=-3pt]
    \setlength{\hgap}{1cm}
	\node[iri,compact,anchor=center] (e) {Food Festival};
	
	\node[var,compact,anchor=center,right=1\hgap of e] (ev) {\textbf{?ff}}
	edge[arrout] node[lab] {type} (e);
	
	\node[iri,compact,anchor=center,right=1.4\hgap of ev] (s) {Santiago}
	edge[arrin] node[lab] {location} (ev);  
\end{tikzpicture} and directly receiving the relevant results. Compared with the previous protocols, this protocol is much more efficient in terms of bandwidth: it allows clients to make more specific requests and the server to return more specific responses. However, this reduction in bandwidth use comes at the cost of the server having to evaluate much more complex requests, where, furthermore, the costs of a single request are much more difficult to anticipate. While a variety of optimised engines exist for evaluating (complex) graph patterns (e.g.,~\cite{virtuoso,Miller13,ThompsonPC14} amongst many others), the problem of evaluating such queries is known to be intractable~\cite{AnglesABHRV17}. Perhaps for this reason, public services offering such a protocol (most often supporting SPARQL queries~\cite{sparql11}) have been found to often exhibit downtimes, timeouts, partial results, slow performance, etc.~\cite{ArandaHUV13}. Even considering such issues, however, popular services continue to receive -- and successfully evaluate -- millions of requests/queries per day~\cite{malyshev2018getting,SaleemAHMN15}, with difficult (worst-case) instances being rare in practice~\cite{BonifatiMT17}.

\subsubsection{Other protocols} While Figure~\ref{fig:access} makes explicit reference to some of the most commonly-encountered access protocols found for knowledge graphs in practice, one may of course imagine other protocols lying almost anywhere on the spectrum from more simple to more complex interfaces. To the right of (Complex) Graph Patterns, one could consider supporting even more complex requests, such as queries with entailments~\cite{Glimm11}, queries that allow recursion~\cite{ReutterSV15}, federated queries that can join results from remote services~\cite{ArandaACP13}, or even (hypothetically) supporting Turing-complete requests that allow running arbitrary procedural code on a knowledge graph. As mentioned at the outset, a server may also choose to support multiple, complementary protocols~\cite{VerborghSCCMW14}.

\subsection{Usage Control}\label{ssec:UsageControl}

Considering our hypothetical tourism knowledge graph, at first glance, one might assume that the knowledge required to deliver the envisaged services is public and thus can be used both by the tourism board and the tourists. On closer inspection, however, we may see the need for usage control in various forms:
\begin{inparaenum}[(i)]
	\item both the tourist board and its partners should associate an appropriate license with knowledge that they contribute to the knowledge graph, such that the terms of use are clear to all parties; 
	\item a tourist might opt to install an app on their mobile phone that could be used to recommend tourist attractions based on their location, bringing with it potential privacy concerns; 
	\item the tourist board may be required to report criminal activities to the police services and thus may need to encrypt personal information; and 
	\item the tourist board could potentially share information relating to tourism demographics in an anonymous format to allow for improving transport infrastructure on strategic routes.
\end{inparaenum}
Thus in this section, we examine the state of the art in terms of knowledge graph licensing, usage policies, encryption, and anonymisation.

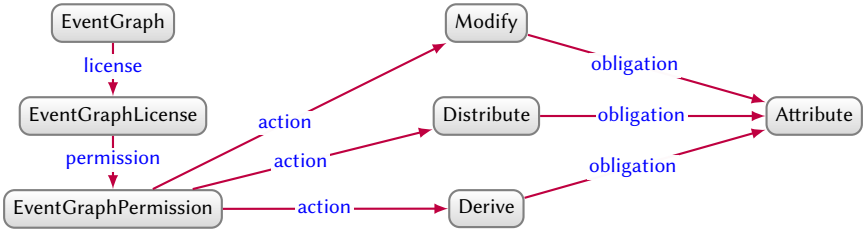
\begin{figure}
	\setlength{\vgap}{0.7cm}
	\setlength{\hgap}{3cm}
	
	\begin{tikzpicture}
	
	\node[iri,anchor=center] (fId) {EventGraph};
	
	\node[iri,anchor=center,below=\vgap of fId] (lic) {EventGraphLicense}
	edge[arrin] node[lab] {license} (fId);   
	
	\node[iri,anchor=center,below=\vgap of lic] (per) {EventGraphPermission}
	edge[arrin] node[lab] {permission} (lic);
	
	\node[iri,anchor=center,right=\hgap of lic] (dis) {Distribute}  
	edge[arrin] node[lab] {action} (per); 
	
	\node[iri,anchor=center,above=\vgap of dis] (mod) {Modify}
	edge[arrin] node[lab] {action} (per);  
	
	\node[iri,anchor=center,below=\vgap of dis] (der) {Derive}  
	edge[arrin] node[lab] {action} (per); 
	
	\node[iri,anchor=center,right=\hgap of dis] (attr) {Attribute}  
	edge[arrin] node[lab] {obligation} (dis)
	edge[arrin] node[lab] {obligation} (mod)
	edge[arrin] node[lab] {obligation} (der);
	
	\end{tikzpicture}
	\caption{Associating licenses with event data, along with permissions, actions, and obligations 
		\label{fig:license}}
\end{figure}

\subsubsection{Licensing} 

When it comes to associating machine readable licenses with knowledge graphs, the W3C Open Digital Rights Language (ODRL)~\cite{odrl} provides an information model and related vocabularies that can be used to specify permissions, duties, and prohibitions with respect to actions relating to assets.
ODRL supports fine-grained descriptions of digital rights that are represented as -- and thus can be embedded within -- graphs.
Figure~\ref{fig:license} illustrates a license granting the assignee the permission to \gnode{Modify}, \gnode{Distribute}, and \gnode{Derive} work from the \gnode{EventGraph} (e.g., Figure~\ref{fig:delg}); however the assignee is obliged to \gnode{Attribute} the copyright holder.
From a modelling perspective, ODRL can be used to model several well known license families, for instance Apache, Creative Commons (CC), and Berkeley Software Distribution (BSD), to name but a few~\cite{CabrioAV14,panasiuk2018modeling}.
Additionally, \citet{CabrioAV14} propose methods to automatically extract machine-readable licenses from unstructured text. 
From a reasoning perspective, license compatibility validation and composition techniques \citep{villata2012licenses,guido_heuristics_2013,DBLP:conf/esws/MoreauSPD19} can be used to combine knowledge graphs that are governed by different licenses. Such techniques are employed by the the Data Licenses Clearance Center (DALICC), which includes a library of standard machine readable licenses, and tools that enable users both to compose arbitrary custom licenses and also to verify the compatibility of different licenses \citep{pellegrini2019DALICC}.

\begin{figure}
	\setlength{\vgap}{0.8cm}
	\setlength{\hgap}{1.8cm}
	
	\begin{tikzpicture}
	
	\node[iri,anchor=center] (pol) {LocationGraphConsentPolicy};
	
	\node[iri,anchor=center,below=\vgap of pol] (proc) {Analyse}
	edge[arrin] node[lab] {processing} (pol);  
	
	\node[iri,anchor=center,left=\hgap of pol] (pur) {EventRecommendations}  
	edge[arrin] node[lab] {purpose} (pol); 
	
	\node[iri,anchor=center,left=\hgap of proc] (data) {LocationGraph}  
	edge[arrin] node[lab] {data} (pol); 
	
	\node[iri,anchor=center,right=\hgap of proc] (rec) {CompanyStaff}  
	edge[arrin] node[lab] {recipients} (pol); 
	
	\node[iri,anchor=center,right=\hgap of pol] (sto) {InternalServers}  
	edge[arrin] node[lab] {storage} (pol);
	
	\end{tikzpicture}
	\caption{A policy for the usage of a sub-graph of location data in the knowledge graph
		\label{fig:usage}}
\end{figure}
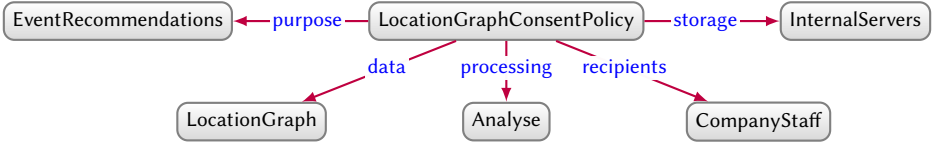

\subsubsection{Usage policies} 

Access control policies based on edge patterns can be used to restrict access to parts of a knowledge graph~\cite{Reddivari2005,Flouris2010,Kirrane2013}.
WebAccessControl (WAC)\footnote{WAC, \url{http://www.w3.org/wiki/WebAccessControl}} is an access control framework for graphs that uses WebID for authentication and provides a vocabulary for specifying access control policies.
Extensions of this WAC vocabulary have been proposed to capture privacy preferences \citep{SaccoP11} and to cater for contextual constraints  \citep{Villata2011,Costabello2012}. 
Although ODRL is primarily used to specify licenses, profiles to specify access policies \citep{steyskal2014} and regulatory obligations \citep{agarwal2018legislative, devos2019ODRL} have also been proposed in recent years, as discussed in the survey by \citet{kirrane2017access}.

As a generalisation of access policies, usage policies specify how data can be used: what kinds of processing can be applied, by whom, for what purpose, etc. The example usage policy presented in Figure~\ref{fig:usage} states that the process \gnode{Analyse} of \gnode{LocationGraph} can be performed on \gnode{InternalServers} by members of \gnode{CompanyStaff} in order to provide \gnode{EventRecommendations}.  Vocabularies for usage policies have been proposed by the SPECIAL H2020 project~\cite{special} and the W3C Data Privacy Vocabularies and Controls Community Group (DPVCG)~\cite{dpv,bonatti2019big}. Once specified, usage policies can then be used to verify that data processing conforms to legal norms and the consent provided by subjects~\citep{DelanauxBRT18,bonatti2019big}.

\begin{figure}
	\setlength{\vgap}{1cm}
	\setlength{\hgap}{1.8cm}
	
	\begin{tikzpicture}
	\node[iri,anchor=center] (xy) {Claimant-XY12SDA};  
	  
	\node[iri,anchor=center,right=1.8\hgap of xy,dashed] (js) {John Smith}
	  edge[arrin,dashed] node[lab] {Claimant-name} (xy);
	
	\node[iri,anchor=center,right=\hgap of js] (cId) {CipherName-XY12SDA}
	edge[arrin,dashed] node[lab] {cipher} (js);  
	
	\node[iri,anchor=center,below=\vgap of js] (ccrypto) {zhk...kjg}
	edge[arrin] node[lab] {crypto} (cId)
	edge[arrin] node[lab] {Claimant-name-enc} (xy);  
	
	\node[iri,anchor=center,below=\vgap  of cId] (ckey) {2048}  
	edge[arrin] node[lab] {keylength} (cId);
	
	\node[iri,anchor=center,right=\hgap of ckey] (calg) {rsa}  
	edge[arrin] node[lab] {algorithm} (cId);
	
	\end{tikzpicture}
	\caption{Directed edge-labelled graph with the name of the claimant encrypted; plaintext elements are dashed and may be omitted from published data (possibly along with encryption details)
		\label{fig:crypto}}
\end{figure}
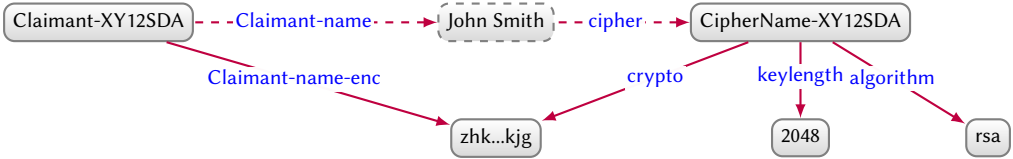

\subsubsection{Encryption} 

Rather than internally controlling usage, the tourist board could use encryption mechanisms on parts of the published knowledge graph, for example relating to reports of crimes, and provide keys to partners who should have access to the plaintext. 
While a straightforward approach is to encrypt the entire graph (or sub-graphs) with one key, more fine-grained encryption can be performed for individual nodes or edge-labels in a graph, potentially providing different clients access to different information through different keys~\cite{giereth2005partial}.
The CryptOntology~\cite{gerbracht2008possibilities} can further be used to embed details about the encryption mechanism used within the knowledge graph. Figure~\ref{fig:crypto} illustrates how this could be used to encrypt the names of claimants from Figure~\ref{fig:direct}, storing the ciphertext \gnode{zhk...kjg}, as well as the key-length and encryption algorithm used. In order to grant access to the plaintext, one approach is to encrypt individual edges with symmetric keys so as to allow specific types of edge patterns to only be executed by clients with the appropriate key~\cite{kasten2013towards}. This approach can be used, for example, to allow clients who know a claimant ID (e.g., \gnode{Claimant-XY12SDA}) and have the appropriate key to find (only) the name of the claimant through an edge pattern \begin{tikzpicture}[baseline=-3pt]
    \setlength{\hgap}{2.3cm}

	\node[iri,compact,anchor=center] (ev) {Claimant-XY12SDA};
	
	\node[var,compact,anchor=center,right=\hgap of ev] (s) {\textbf{?name}}
	edge[arrin] node[lab] {Claimant-name} (ev);  
\end{tikzpicture}. A key limitation of this approach, however, is that it requires attempting to decrypt all edges to find all possible solutions. A more efficient alternative is to combine functional encryption and specialised indexing to retrieve solutions from the encrypted graph without attempting to decrypt all edges~\cite{FernandezKPS17}. 

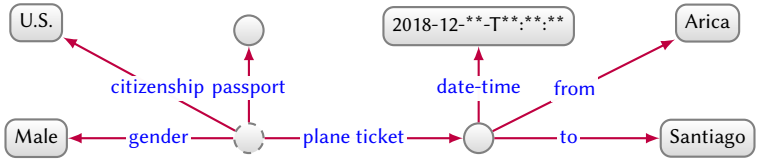
\begin{figure}
	\setlength{\vgap}{1cm}
	\setlength{\hgap}{2.2cm}
	
	\begin{tikzpicture}
	\node[bnode,anchor=center] (r) {};
	
	\node[iri,anchor=center,above=\vgap of r] (d) {2018-12-**-T**:**:**}
	edge[arrin] node[lab] {date-time} (r);

	\node[iri,anchor=center,right=\hgap of r] (sp) {Santiago}
	edge[arrin] node[lab] {to} (r);
	
	\node[iri,anchor=center,above=\vgap of sp] (c) {Arica}
	edge[arrin] node[lab] {from} (r);
	
	\node[bnode,anchor=center,dashed,left=1.2\hgap of r] (p) {}
	edge[arrout] node[lab] {plane ticket} (r);
	
	\node[bnode,anchor=center,above=\vgap of p] (pp) {}
	edge[arrin] node[lab] {passport} (p);
	
	\node[iri,anchor=center,left=\hgap of p] (pg) {Male}
	edge[arrin] node[lab] {gender} (p);	
	
	\node[iri,anchor=center,above=\vgap of pg] (pc) {U.S.}
	edge[arrin] node[lab] {citizenship} (p);		
	\end{tikzpicture}
	
	\caption{Anonymised sample of a directed edge-labelled graph describing a passenger (dashed) of a flight \label{fig:anonymised}}
\end{figure}

\subsubsection{Anonymisation} 

Consider that the tourist board acquires information on transport taken by individuals within the country, which can be used to understand trajectories taken by tourists. However, from a data-protection perspective, it would be advisable to remove any personal data from the knowledge graph to avoid leaks of information about each individual's travel. 

A first approach to anonymisation is to suppress and generalise knowledge in a graph such that individuals cannot be identified, based on $k$-anonymity~\citep{samarati1998protecting}\footnote{$k$-anonymity guarantees that the data of an individual is indistinguishable from at least $k-1$ other individuals.}, $l$-diversity \citep{li2007t}\footnote{$l$-diversity guarantees that sensitive data fields have at least $l$ diverse values within each group of individuals; this avoids leaks such as that all tourists from Austria (a group of individuals) in the data have been pick-pocketed (a sensitive attribute).}, etc. 
Approaches to apply $k$-anonymity on graphs identify and suppress ``quasi-identifiers'' that would allow a given individual to be distinguished from fewer than $k -1$ other individuals~\cite{radulovic2015towards,HeitmannEtAl2017}.
Figure~\ref{fig:anonymised} illustrates a possible result of $k$-anonymisation for a sub-graph describing a flight passenger, where quasi-identifiers (passport, plane ticket) have been converted into blank nodes, ensuring that the passenger (the dashed blank node) cannot be distinguished from $k - 1$ other individuals.
In the context of a graph, however, \textit{neighbourhood attacks}~\cite{ZhouP11} -- using information about neighbours -- can also break $k$-anonymity, where we also suppress the day and time of the flight, which, though not sensitive information per se, could otherwise break $k$-anonymity for passengers (if, for example, a particular flight had fewer than $k$ males from the U.S.\ onboard). 

More complex neighbourhood attacks may rely on more abstract graph patterns, observing that individuals can be deanonymised purely from knowledge of the graph structure, even if all nodes and edge labels are left blank; for example, if we know that a team of $k-1$ players take flights together for a particular number of away games, we could use this information for a neighbourhood attack that reveals the set of players in the graph. Hence a number of guarantees specific to graphs have been proposed, including $k$-degree anonymity~\cite{LiuT08}, which ensures that individuals cannot be deanonymised by attackers with knowledge of the degree of particular individuals. The approach is based on minimally modifying the graph to ensure that each node has at least $k - 1$ other nodes with the same degree. A stronger guarantee, called $k$-isomorphic neighbour anonymity~\cite{ZhouP08}, avoids neighbourhood attacks where an attacker knows how an individual is connected to nodes in their neighbourhood; this is done by modifying the graph to ensure that for each node, there exist at least $k - 1$ nodes with isomorphic (i.e., identically structured) neighbourhoods elsewhere in the graph. Both approaches only protect against attackers with knowledge of bounded neighbourhoods. An even stronger notion is that of $k$-automorphism~\cite{ZouCO09a}, which ensures that for every node, it is structurally indistinguishable from $k -1$ other nodes, thus avoiding any attack based on structural information (as a trivial example, a $k$-clique or a $k$-cycle satisfy $k$-automorphism). Many of these techniques for anonymisation of graph data were originally motivated by social networks~\cite{NarayananS09}, though they can also be applied to knowledge graphs, per the work of \citet{LinT17}, who adapt $k$-automorphism for directed edge-labelled graphs (specifically RDF graphs).

While the aforementioned approaches anonymise data, a second approach is to apply anonymisation when answering queries, such as adding noise to the solutions in a way that preserves privacy. One approach is to apply $\varepsilon$-differential privacy~\citep{Dwork:2006:DP:2097282.2097284}\footnote{$\varepsilon$-differential privacy ensures that the probability of achieving a given result from some process (e.g., query) applied to data, to which random noise is added, differs no more than $e^\varepsilon$ when the data includes or excludes any individual.} for querying graphs~\cite{Silva2017}. Such mechanisms are typically used for aggregate (e.g., count) queries, where noise is added to avoid leaks about individuals. To illustrate, differential privacy may allow for counting the number of passengers of specified nationalities taking specified flights, adding (just enough) random noise to the count to ensure that we cannot tell, within a certain probability (controlled by $\varepsilon$), whether or not a particular individual took a flight, where we would require (proportionally) less noise for common nationalities, but more noise to ``hide'' individuals from more uncommon nationalities.

These approaches require information loss for stronger guarantees of privacy; which to choose is thus heavily application dependent. If the anonymised data are to be published in their entirety as a ``dump'', then an approach based on $k$-anonymity can be used to protect individuals, while $l$-diversity can be used to protect groups. On the other hand, if the data are to be made available, in part, through a query interface, then $\varepsilon$-differential privacy is a more suitable framework.

\section{Knowledge Graphs in Practice}\label{sec:kgs}

In this section, we discuss some of the most prominent knowledge graphs that have emerged in the past years. We begin by discussing open knowledge graphs, which have been published on the Web per the guidelines and protocols described in Section~\ref{sec:publish}. We later discuss enterprise knowledge graphs that have been created by companies for a diverse range of applications.

\subsection{Open Knowledge Graphs}\label{sec:openkgs}

By \textit{open knowledge graphs}, we specifically refer to knowledge graphs published under the Open Data philosophy, namely that ``\textit{open means anyone can freely access, use, modify, and share for any purpose (subject, at most, to requirements that preserve provenance and openness)}''.\footnote{See \url{http://opendefinition.org/}} Many open knowledge graphs have been published in the form of \textit{Linked Open Datasets}~\cite{ldbook}, which are (RDF) graphs published under the Linked Data principles (see Section~\ref{sssec:ld}) following the Open Data philosophy. Many of the most prominent open knowledge graphs -- including DBpedia~\cite{LehmannIJJKMHMK15}, YAGO~\cite{suchanek2007yago}, Freebase~\cite{bollacker2007freebase}, and Wikidata~\cite{VrandecicK14} -- cover multiple domains, representing a broad diversity of entities and relationships; we first discuss these in turn. Later we discuss some of the other (specific) domains for which open knowledge graphs are currently available. Most of the open knowledge graphs we discuss in this section are modelled in RDF, published following Linked Data principles, and offer access to their data through dumps (RDF), node lookups (Linked Data), graph patterns (SPARQL) and, in some cases, edge patterns (Triple Pattern Fragments).

\subsubsection{DBpedia}\label{sssec:dbpedia}

The DBpedia project was developed to extract a graph-structured representation of the semi-structured data embedded in Wikipedia articles~\cite{auer2007dbpedia}, enabling the integration, processing, and querying of these data in a unified manner. The resulting knowledge graph is further enriched by linking to external open resources, including images, webpages, and external datasets such as DailyMed, DrugBank, GeoNames, MusicBrainz, New York Times, and WordNet~\cite{LehmannIJJKMHMK15}. The DBpedia extraction framework consists of several components, corresponding to abstractions of Wikipedia article sources, graph storage and serialisation destinations, wiki-markup extractors, parsers, and extraction managers~\cite{bizer2009dbpedia}. Specific extractors are designed to process labels, abstracts, interlanguage links, images, redirects, disambiguation pages, external links, internal pagelinks, homepages, categories, and geocoordinates. The content in the DBpedia knowledge graph is not only multidomain, but also multilingual: as of 2012, DBpedia contained labels and abstracts in up to 97 different languages \cite{mendes2012dbpedia}. 
Entities within DBpedia are classified using four different schemata in order to address varying application requirements \cite{bizer2009dbpedia}. These schemata include a Simple Knowledge Organization System (SKOS) representation of Wikipedia categories, a Yet Another Great Ontology (YAGO) classification schema (discussed in the following), an Upper Mapping and Binding Exchange Layer (UMBEL) ontology categorisation schema, and a custom schema called the DBpedia ontology with classes such as \texttt{Person}, \texttt{Place}, \texttt{Organisation}, and \texttt{Work}~\cite{LehmannIJJKMHMK15}. DBpedia also supports live synchronisation in order to remain consistent with dynamic Wikipedia articles~\cite{LehmannIJJKMHMK15}.

\subsubsection{Yet Another Great Ontology} YAGO likewise extracts graph-structured data from Wikipedia, which are then unified with the hierarchical structure of WordNet to create a ``\textit{light-weight and extensible ontology with high quality and coverage}''~\cite{suchanek2007yago}. This knowledge graph aims to be applied for various information technology tasks, such as machine translation, word sense disambiguation, query expansion, document classification, data cleaning, information integration, etc. 
While earlier approaches automatically extracted structured knowledge from text using pattern matching, natural language processing (NLP), and statistical learning, the resulting content tended to lack in quality when compared with what was possible through manual construction~\cite{suchanek2007yago}. However, manual construction is costly, making it challenging to achieve broad coverage and keep the data up-to-date. In order to extract data with high coverage and quality, YAGO (like DBpedia) mostly extracts data from Wikipedia infoboxes and category pages, which contain basic entity information and lists of articles for a specific category, respectively; these, in turn, are unified with hierarchical concepts from WordNet~\cite{suchanek2008yago}. A schema -- called the YAGO model -- provides a vocabulary defined in RDFS; this model allows for representing words as entities, capturing synonymy and ambiguity \cite{suchanek2007yago}. The model further supports reification, $n$-ary relations, and data types~\cite{suchanek2008yago}. Refinement mechanisms employed within YAGO include canonicalisation, where each edge and node is mapped to a unique identifier and duplicate elements are removed, and type checking, where nodes that cannot be assigned to a class by deductive or inductive methods are eliminated \cite{suchanek2008yago}. YAGO would be extended in later years to support spatio-temporal context~\cite{YAGO} and multilingual Wikipedias~\cite{MahdisoltaniBS15}.

\subsubsection{Freebase}

Freebase was a general collection of human knowledge that aimed to address some of the large scale information integration problems associated with the decentralised nature of the Semantic Web, such as uneven adoption, implementation challenges, and distributed query performance limitations \cite{bollacker2007platform}. Unlike DBpedia and YAGO -- which are mostly extracted from Wikipedia/WordNet -- Freebase solicited contributions directly from human editors. Included in the Freebase platform were a scalable data store with versioning mechanisms; a large data object store (LOB) for the storage of text, image, and media files; an API that could be queried using the Metaweb Query Language (MQL); a Web user interface; and a lightweight typing system \cite{bollacker2007platform}. The latter typing system was designed to support collaborative processes. Rather than forcing ontological correctness or logical consistency, the system was implemented as a loose collection of structuring mechanisms -- based on datatypes, semantic classes, properties, schema definitions, etc. -- that allowed for incompatible types and properties to coexist simultaneously~\cite{bollacker2007platform}. Content could be added to Freebase interactively through the Web user interface or in an automated way by leveraging the API's write functionality. Freebase had been acquired by Google in 2010, where the content of Freebase formed an important part of the Google Knowledge Graph announced in 2012~\cite{GoogleKG}. When Freebase became read-only as of March 2015, the knowledge graph contained over three billion edges. Much of this content was subsequently migrated to Wikidata~\cite{pellissier2016freebase}. 

\subsubsection{Wikidata}
As exploited by DBpedia and YAGO, Wikipedia contains a wealth of semi-structured data embedded in info-boxes, lists, tables, etc. However, these data have traditionally been curated and updated manually across different articles and languages; for example, a goal scored by a Chilean football player may require manual updates in the player's article, the tournament article, the team article, lists of top scorers, and so forth, across hundreds of language versions. Manual curation has led to a variety of data quality issues, including contradictory data in different articles, languages, etc. The Wikimedia Foundation thus uses Wikidata as a centralised, collaboratively-edited knowledge graph to supply Wikipedia -- and arbitrary other clients -- with data. Under this vision, a fact could be added to Wikidata once, triggering the automatic update of potentially multitudinous articles in Wikipedia across different languages~\cite{VrandecicK14}. Like Wikipedia, Wikidata is also considered a secondary source containing \textit{claims} that should reference primary sources, though claims can also be initially added without reference~\cite{PiscopoKPS17}. Wikidata further allows for different viewpoints in terms of potentially contradictory (referenced) claims~\cite{VrandecicK14}. Wikidata is multilingual, where nodes and edges are assigned language-agnostic \texttt{Qxx} and \texttt{Pxx} codes (see Figure~\ref{fig:ld}) and are subsequently associated with labels, aliases, and descriptions in various languages~\cite{KaffeePVSCP17}, allowing claims to be surfaced in these languages. Collaborative editing is not only permitted on the data level, but also on the schema level, allowing users to add or modify lightweight semantic axioms~\cite{PiscopoS18} -- including sub-classes, sub-properties, inverse properties, etc. -- as well as shapes~\cite{BonevaDFG19}. Wikidata offers various access protocols~\cite{malyshev2018getting} and has received broad adoption, being used by Wikipedia to generate infoboxes in certain domains~\cite{SaezH18}, being supported by Google~\cite{pellissier2016freebase}, and having been used as a data source for prominent applications such as Apple's Siri, amongst others~\cite{malyshev2018getting}.

\subsubsection{Other open cross-domain knowledge graphs}

A number of other cross-domain knowledge graphs have been developed down through the years. BabelNet~\cite{NavigliPonzetto:12} -- in a similar fashion to YAGO -- is based on unifying WordNet and Wikipedia, but with the integration of additional knowledge graphs such as Wikidata, and a focus on creating a knowledge graph of multilingual lexical forms (organised into multilingual synsets) by transforming lexicographic resources such as Wiktionary and OmegaWiki into knowledge graphs. Compared to other knowledge graphs, lexicalised knowledge graphs such as BabelNet bring together the encyclopedic information found in Wikipedia with the lexicographic information usually found in monolingual and bilingual dictionaries. The Cyc project \cite{lenat1995cyc} aims to encode common-sense knowledge in a machine-readable way, where over 900 person-years of effort~\cite{MatuszekCWD06} have, since 1986, gone into the creation of 2.2 million facts and rules. Though Cyc is proprietary, an open subset called OpenCyc has been published, where we refer to the comparison by \citet{FarberBMR18} of DBpedia, Freebase, OpenCyc, and YAGO for further details. The Never Ending Language Learning (NELL) project~\cite{MitchellCHTYBCM18} has, since 2010, extracted a graph of 120 million edges from the text of web pages using OIE methods (see Section~\ref{sec:create}). Each such open knowledge graph applies different combinations of the languages and techniques discussed in this paper over different sources with differing results. 

\subsubsection{Domain-specific open knowledge graphs}

Open knowledge graphs have been published in a variety of specific domains. \citet{SchmachtenbergBP14} identify the most prominent domains in the context of Linked Data as follows: \textit{media}, relating to news, television, radio, etc. (e.g., the BBC World Service Archive~\cite{RaimondFSA14}); \textit{government}, relating to the publication of data for transparency and development (e.g., by the U.S.~\cite{HendlerHMT12} and U.K.~\cite{ShadboltO13} governments); \textit{publications}, relating to academic literature in various disciplines (e.g., OpenCitations~\cite{PeroniSV17}, SciGraph~\cite{IanaJNBHP19}, Microsoft Academic Knowledge Graph~\cite{MAKG}); \textit{geographic}, relating to places and regions of interest (e.g., LinkedGeoData~\cite{StadlerLHA12}); \textit{life sciences}, relating to proteins, genes, drugs, diseases, etc. (e.g., Bio2RDF~\cite{CallahanCAD13}); and \textit{user-generated content}, relating to reviews, open source projects, etc. (e.g., Revyu~\cite{HeathM08a}). Open knowledge graphs have also been published in other domains, including \textit{cultural heritage}~\cite{HyvonenMKAKRSTPKVTPFSPLN09}, \textit{music}~\cite{RaimondSS09}, \textit{law}~\cite{Montiel-Ponsoda17}, \textit{theology}~\cite{SherifN15}, and even \textit{tourism}~\cite{LuLS16,abs-1805-05744,MaturanaALMH18,ZhangCHYAL19}. The envisaged applications for such knowledge graphs are as varied as the domains from which they emanate, but often relate to integration~\cite{RaimondSS09,CallahanCAD13}, recommendation~\cite{RaimondSS09,LuLS16},
transparency~\cite{HendlerHMT12,ShadboltO13}, archiving~\cite{HyvonenMKAKRSTPKVTPFSPLN09,RaimondFSA14}, decentralisation~\cite{HeathM08a}, multilingual support~\cite{SherifN15}, regulatory compliance~\cite{Montiel-Ponsoda17}, etc.

\subsection{Enterprise Knowledge Graphs}

A variety of companies have announced the creation of proprietary ``enterprise knowledge graphs'' with a variety of goals in mind, which include: improving search capabilities~\citep{GoogleKG,BingKG,AmazonKG,AirBnBKG,UberKG}, providing user recommendations~\cite{AirBnBKG,UberKG}, implementing conversational/personal agents~\cite{eBayKG}, enhancing targetted advertising~\cite{LinkedInKG}, empowering business analytics~\cite{LinkedInKG}, connecting users~\cite{LinkedInKG,NoyGJNPT19}, extending multilingual support~\cite{LinkedInKG}, facilitating research and discovery~\cite{AstraZenecaKG}, assessing and mitigating risk~\cite{ThompsonReutersKG,MaanaKG}, tracking news events~\cite{BloombergKG}, and increasing transport automation~\cite{HensonSTK19}, amongst (many) others. Though highly diverse, these enterprise knowledge graphs do follow some high-level trends, as reflected in the discussion by \citet{NoyGJNPT19}: (1) data are typically integrated into the knowledge graph from a variety of both external and internal sources (often involving text); (2) the enterprise knowledge graph is often very large, with millions or even billions of nodes and edges, posing challenges in terms of scalability; (3) refinement of the initial knowledge graph -- adding new links, consolidating duplicate entities, etc.\ -- is important to improve quality; (4) techniques to keep the knowledge graph up-to-date with the domain are often crucial; (5) a mix of ontological and machine learning representations are often combined or used in different situations in order to draw conclusions from the enterprise knowledge graph; (6) the ontologies used tend to be lightweight, often simple taxonomies representing a hierarchy of classes or concepts.

We now discuss the main industries in which enterprise knowledge graphs have been deployed.

\subsubsection{Web search} Web search engines have traditionally focused on matching a query string with sub-strings in web documents. The Google Knowledge Graph~\cite{GoogleKG,NoyGJNPT19} rather promoted a paradigm of ``\textit{things not strings}'' -- analogous to semantic search~\citep{GuhaMM03} -- where the search engine would now try to identify the entities that a particular search may be expressing interest in. The knowledge graph itself describes these entities and how they interrelate. One of the main user-facing applications of the Google Knowledge Graph is the ``Knowledge Panel'', which presents a pane on the right-hand side of (some) search results describing the principal entity that the search appears to be seeking, including some images, attribute--value pairs, and a list of related entities that users also search for. The Google Knowledge Graph was key to popularising the modern usage of the phrase ``knowledge graph'' (see Appendix~\ref{sec:defs}). Other major search engines, such as Microsoft Bing\footnote{Microsoft's Knowledge Graph was previously called ``Satori'' (meaning \textit{understanding} in Japanese).}~\cite{BingKG}, would later announce knowledge graphs along similar lines.

\subsubsection{Commerce} Enterprise knowledge graphs have also been announced by companies that are principally concerned with selling or renting goods and services. A prominent example of such a knowledge graph is that used by Amazon~\cite{AmazonKG,dong2019building}, which describes the products on sale in their online marketplace. One of the main stated goals of this knowledge graph is to enable more advanced (semantic) search features for products, as well as to improve product recommendations to users of its online marketplace. Another knowledge graph for commerce was announced by eBay~\cite{eBayKG}, which encodes product descriptions and shopping behaviour patterns, and is used to power conversational agents that aid users to find relevant products through a natural language interface. Airbnb~\cite{AirBnBKG} have also described a knowledge graph that encodes accommodation for rent, places, events, experiences, neighbourhoods, users, tags, etc., on top of which a taxonomic schema is defined. This knowledge graph is used to offer potential clients recommendations of attractions, events, and activities available in the neighbourhood of a particular home for rent. Uber~\cite{UberKG} have similarly announced a knowledge graph focused on food and restaurants for their ``Uber Eats'' delivery service. The goals are again to offer semantic search features and recommendations to users who are uncertain precisely what kind of food they are looking for.

\subsubsection{Social networks} Enterprise knowledge graphs have also emerged in the context of social networking services. Facebook~\cite{NoyGJNPT19} have gathered together a knowledge graph describing not only social data about users, but also the entities they are interested in, including celebrities, places, movies, music, etc., in order to connect people, understand their interests, and provide recommendations. LinkedIn~\cite{LinkedInKG} announced a knowledge graph containing users, jobs, skills, companies, places, schools, etc., on top of which a taxonomic schema is defined. The knowledge graph is used to provide multilingual translations of important concepts, to improve targetted advertising, to provide advanced features for job search and people search, and likewise to provide recommendations matching jobs to people (and vice versa). Another knowledge graph has been created by Pinterest~\cite{PinterestKG}, describing users and their interests, the latter being organised into a taxonomy. The main use-cases for the knowledge graph are to aid users to more easily find content of interest to them, as well as to enhance revenue through targetted advertisements.

\subsubsection{Finance} The financial sector has also seen deployment of enterprise knowledge graphs. Amongst these, Bloomberg~\cite{BloombergKG} has proposed a knowledge graph that powers financial data analytics, including sentiment analysis for companies based on current news reports and tweets, a question answering service, as well as detecting emerging events that may affect stock values. Thompson Reuters (Refinitiv)~\cite{ThompsonReutersKG} have likewise announced a knowledge graph encoding ``the financial ecosystem'' of people, organisations, equity instruments, industry classifications, joint ventures and alliances, supply chains, etc., using a taxonomic schema to organise these entities. Some of the applications they mention for the knowledge graph include supply chain monitoring, risk assessment, and investment research. Knowledge graphs have also been used for deductive reasoning, with Banca d'Italia~\cite{BellomariniFGS19} using rule-based reasoning to determine, for example, the percentage of ownership of a company by various stakeholders. Other companies exploring financial knowledge graphs include Accenture~\cite{AccentureKG},  Capital One~\cite{CapitalOneKG}, Wells Fargo~\cite{WellsFargoKG}, amongst others.

\subsubsection{Other industries} Enterprises have also been actively developing knowledge graphs to enable novel applications in a variety of other industries, including: \textit{health-care}, where IBM are exploring use-cases for drug discovery~\cite{NoyGJNPT19} and information extraction from package inserts~\cite{GentileGRW19}, while AstraZeneca~\cite{AstraZenecaKG} are using a knowledge graph to advance genomics research and disease understanding; \textit{transport}, where Bosch are exploring a knowledge graph of scenes and locations for driving automation~\cite{HensonSTK19}; \textit{oil \& gas}, where Maana~\cite{MaanaKG} are using knowledge graphs to perform data integration for risk mitigation regarding oil wells and drilling; and more besides. 

\section{Summary and Conclusion}\label{sec:conclude}

We have provided a comprehensive introduction to knowledge graphs, which have been receiving more and more attention in recent years. Under the definition of a knowledge graph as \textit{a graph of data intended to accumulate and convey knowledge of the real world, whose nodes represent entities of interest and whose edges represent relations between these entities}, we have discussed models by which data can be structured as graphs; representations of schema, identity and context; techniques for leveraging deductive and inductive knowledge; methods for the creation, enrichment, quality assessment and refinement of knowledge graphs; principles and standards for publishing knowledge graphs; and finally, the adoption of knowledge graphs in the real world.

\paragraph{Future directions} Research on knowledge graphs can become a confluence of techniques arising from different areas with the common objective of maximising the knowledge -- and thus value -- that can be distilled from diverse sources at large scale using a graph-based data abstraction~\cite{Hogan20}. Pursuing this objective will benefit from expertise on graph databases, knowledge representation, logic, machine learning, graph algorithms and theory, ontology engineering, data quality, natural language processing, information extraction, privacy and security, and more besides. 

While advances in these individual disciplines are sure to continue and to generate further impact, particularly interesting topics arise also from their intersections. In the intersection of data graphs and deductive knowledge, we emphasise emerging topics such as \textit{formal semantics for property graphs}, with languages that can take into account the meaning of labels and property--value pairs on nodes and edges~\cite{Krotzsch0OT18}; and \textit{reasoning and querying over contextual data}, in order to derive conclusions and results valid in a particular setting~\cite{SerafiniH12,zimm-etal-2012-JWS,SchuetzBNSS20}. In the intersection of data graphs and inductive knowledge, we highlight topics such as \textit{similarity-based query relaxation}, allowing to find approximate answers to exact queries based on numerical representations (e.g., embeddings)~\cite{WangWLCZQ18}; \textit{shape induction}, in order to extract and formalise inherent patterns in the knowledge graph as constraints~\cite{Mihindukulasooriya18}; and \textit{contextual knowledge graph embeddings} that provide numeric representations of nodes and edges that vary with time, place, etc.~\cite{KazemiGJKSFP19}. Finally, in the intersection of deductive and inductive knowledge, we mention the topics of \textit{entailment-aware knowledge graph embeddings}~\cite{GuoWWWG16,DemeesterRR16}, that incorporate rules and/or ontologies when computing plausibility; \textit{expressive graph neural networks} proven capable of complex classification analogous to expressive ontology languages~\cite{BarceloKMPRS20}; as well as further advances on \textit{rule and axiom mining}, allowing to extract symbolic, deductive representations from the knowledge graphs~\cite{GalarragaTHS15,BuhmannLW16}.

Aside from specific topics, more general challenges for knowledge graphs include \textit{scalability}, particularly for deductive and inductive reasoning; \textit{quality}, not only in terms of data, but also the models induced from knowledge graphs; \textit{diversity}, such as managing contextual or multi-modal data; \textit{dynamicity}, considering temporal or streaming data; and finally \textit{usability}, which is key to increasing adoption. Though techniques are continuously being proposed to address precisely these challenges, they are unlikely to ever be completely ``solved''; rather they serve as dimensions along which knowledge graphs, and their techniques, tools, etc., will continue to mature.

Given the availability of open knowledge graphs whose quality continue to improve, as well as the growing adoption of enterprise knowledge graphs in various industries, future research on knowledge graphs has the potential to foster key advancements in broad aspects of society. Here we have highlighted just some examples of future research directions of importance to this pursuit.

\subsubsection*{Acknowledgements:} We thank the attendees of the Dagstuhl Seminar on ``Knowledge Graphs'' for discussions that inspired and influenced this paper, and all those that make such seminars possible. We would also like to thank Matteo Palmonari for feedback on Figures~\ref{fig:fsa} and \ref{fig:pg}, as well as Stefan Decker and Carlos Bobed who provided suggestions for the paper. Hogan was supported by Fondecyt Grant No.\ 1181896. Hogan and Gutierrez were funded by ANID -- Millennium Science Initiative Program -- Code ICN17\_002. Cochez did part of the work while employed at Fraunhofer FIT, Germany and was later partially funded by Elsevier's Discovery Lab. Kirrane, Ngonga Ngomo, Polleres and Staab received funding through the project “KnowGraphs” from the European Union’s Horizon programme under the Marie Skłodowska-Curie grant agreement No. 860801. Kirrane and Polleres were supported by the European Union’s Horizon 2020 research and innovation programme under grant 731601. Labra was supported by the Spanish Ministry of Economy and Competitiveness (Society challenges: TIN2017-88877-R). Navigli was supported by the MOUSSE ERC Grant No. 726487 under the European Union's Horizon 2020 research and innovation programme. Rashid was supported by IBM Research AI through the AI Horizons Network. Schmelzeisen was supported by the German Research Foundation (DFG) grant STA 572/18-1.


\bibliographystyle{ACM-Reference-Format}
\bibliography{bibliography}

\appendix

\section{Background}\label{sec:defs}

We now discuss the broader historical context that has paved the way for the modern advent of knowledge graphs, as well as the definitions of the notion of ``knowledge graph'' that have been proposed both before and after the announcement of the Google Knowledge Graph~\cite{GoogleKG}. We remark that the discussion presented here builds upon (but does not subsume) previous discussion by \citet{EhrlingerW16} and by \citet{Bergman19}, which we refer to for further details. Though our goal is to be comprehensive, the list of historical references should not be considered exhaustive.

\subsection{Historical Perspective}

The lineage of knowledge graphs can be traced back to the origins of diagrammatic forms of knowledge representation: a tradition going back at least to Aristotle ($\sim$350 BC), followed by notions such as Euler circles and Venn diagrams that helped humans to reason through visual insights. Later researchers -- particularly \citeH{sylvester}, \citeH{peirce} and \citeH{frege} -- independently devised formal diagrammatic systems that not only help reasoning, but also codify reasoning; in other words, their goal was to use diagrams as formal systems.
  
With the advent of digital computers, programs began to be used to perform formal reasoning and to code representations of knowledge. These developments can be traced back to works such as those of \citeH{ritchens}, \citeH{quillian}, and \citeH{milgram}, which focused on formal representations for natural language, information, and knowledge. These early works faced limitations (at least by modern standards) in terms of the poor computational resources available. From the formal (logical) point of view, a number of influential developments took place in the 70's, including the introduction of \textit{frames} by \citeH{minsky}, the formalisation of \textit{semantic networks} 
by \citeH{Brachman} and \citeH{woods},
and the proposal of \textit{conceptual graphs} by \citeH{sowa}.
These works tried to integrate formal logic with diagrammatic representations of knowledge by giving a (more-or-less) formal semantics to graph representations. But as
Sowa later wrote in the entry ``{\em Semantic networks}'' in the Encyclopedia of Cognitive Science (1987) \cite{sowa2}: ``\textit{Woods (1975) and McDermott (1976) observed, the semantic networks themselves have no well-defined semantics. Standard predicate calculus does have a precisely defined, model theoretic semantics; it is adequate for describing mathematical theories with a closed set of axioms. But the real world is messy, incompletely explored, and full of unexpected surprises.}'' 
  
From this era of exploration and attempts to define programs to simulate the visual and formal reasoning of humans, the following key notions were established that are still of relevance today:

\begin{itemize}
  \item knowledge representation through diagrams (specifically graphs) and visual means;
  \item computational procedures and algorithms to perform formal reasoning; 
  \item combinations of formal (logical) and statistical forms of reasoning; 
  \item relevance of different types of data (e.g., images, sound) as sources of knowledge.
\end{itemize}

These works on conceptual graphs, semantic networks, and frames were direct predecessors of Description Logics, which aimed to give a well-defined semantics to these earlier notions towards building practical reasoning systems for decidable logics. Description Logics stem from the KL-ONE system proposed by \citeH{BrachmanS85}, and the ``\textit{attributive concept descriptions with complements}'' language (aka $\mathcal{ALC}$) proposed by \citeH{Schmidt-SchaussS91}. Description Logics would be further explored in later years (see Section~\ref{sec:dlformal}), and formed the underpinnings of the Web Ontology Language (OWL) standard~\cite{OWL2}. Together with the Resource Description Framework (RDF)~\cite{rdf11}, OWL would become one of the building blocks of the Semantic Web~\cite{berners-lee01}, within which many of the formative ideas and standards underlying knowledge graphs would later be developed, including not only RDF and OWL, but also RDFS~\cite{RDFS}, SPARQL~\cite{sparql11}, Linked Data principles~\cite{ldprinciples}, Shape Expressions~\cite{RDFS,ThorntonSSGMPW19}, and indeed, many of the other concepts, standards and techniques discussed in this paper. Most of the open knowledge graphs discussed in Section~\ref{sec:openkgs} -- including BabelNet~\cite{NavigliPonzetto:12}, DBpedia~\cite{LehmannIJJKMHMK15}, Freebase~\cite{bollacker2007platform}, Wikidata~\cite{VrandecicK14}, YAGO~\cite{suchanek2007yago}, etc. -- have either emerged from the Semantic Web community, or would later adopt its standards.

\subsection{``Knowledge Graphs'': Pre 2012}

Long before the 2012 announcement of the Google Knowledge Graph, various authors had used the phrase ``knowledge graph'' in publications stretching back to the 40's, but with unrelated meaning. To the best of our knowledge, the first reference to a ``knowledge graph'' of relevance to the modern meaning was in a paper by \citeH{Schneider72} in the area of computerised instructional systems for education, where a knowledge graph -- in his case a directed graph whose nodes are units of knowledge (concepts) that a student should acquire, and whose edges denote dependencies between such units of knowledge -- is used to represent and store an instructional course on a computer. An analogous notion of a ``knowledge graph'' was used by \citeH{MarchiM74} to study paths through the knowledge units of an instructional course that yield the highest payoffs for teachers and students in a game-theoretic sense. Around the same time, in a paper on linguistics, \citeH{Kummel73} describes a numerical representation of knowledge, with ``radicals'' -- referring to some symbol with meaning -- forming the nodes of a knowledge graph.

Further authors were to define instantiations of knowledge graphs in the 80's. \citeH{rada1986gradualness} defines a knowledge graph in the context of medical expert systems, where domain knowledge is defined as a weighted graph, over which a ``gradual'' learning process is applied to refine knowledge by making small changes to weights. \citeH{Bakker} defines a knowledge graph with the purpose of cumulatively representing content gleaned from medical and sociological texts, with a focus on causal relationships. Work on knowledge graphs from the same group would continue over the years, with contributions by \citeH{stokman1988structuring} further introducing mereological (\textit{part of}) and instantiation (\textit{is a}) relations to the knowledge graph, and thereafter by \citeT{james}, \citeT{Hoede95}, \citeT{Popping03}, \citeT{zhang}, amongst others, in the decades that followed~\cite{NurdiatiH08}. The notion of knowledge graph used in such works considered a fixed number of relations. Other authors pursued their own parallel notions of knowledge graphs towards the end of the 80's. \citeH{rappaport1988dynamic} describe a user interface for visualising a knowledge-base -- composed of facts and rules --  using a knowledge graph that connects related elements of the knowledge-base. \citeH{SrikanthJ89} use the notion of a knowledge graph to represent the entities and relations involved in projects, particularly software projects, where partitioning techniques are applied to the knowledge graph to modularise the knowledge required in the project.

Continuing to the 90's, the notion of a ``knowledge graph'' would again arise in different, seemingly independent settings. \citeH{de1990hybrid} propose a knowledge graph as a directed graph composed of a taxonomy of instances being related with weighted edges to a taxonomy of classes; they use symbolic learning to extract such knowledge graphs from examples. \citeH{MachadoR90} define a knowledge graph as an acyclic, weighted \textsc{and}--\textsc{or} graph,\footnote{An \textsc{and}--\text{or} graph denotes dependency relations, where \textsc{and} denotes a conjunction of sub-goals on which a goal depends, while \textsc{or} denotes a disjunction of sub-goals.} defining fuzzy dependencies that connect observations to hypotheses through intermediary nodes. These knowledge graphs are elicited from domain experts and can be used to generate neural networks for selecting hypotheses from input observations. Knowledge graphs were again later used by \citeH{DiengGTC92} to represent the results of knowledge acquisition from experts. \citeH{ShimonyDS97} rather define a knowledge graph based on a \textit{Bayesian knowledge base} -- i.e., a Bayesian network that permits directed cycles -- over which Bayesian inference can be applied. This definition was further built upon in a later work by \citeH{JrS99}.

Moving to the 00's, \citeH{Jiang02} introduce the notion of ``plan knowledge graphs'' where nodes represent goals and edges dependencies between goals, further encoding supporting degrees that can change upon further evidence. Search algorithms are then defined on the graph to determine a plan for a particular goal. \citeH{HelmsB05} propose a knowledge graph to represent the flow of knowledge in an organisation, with nodes representing knowledge actors (creators, sharers, users), edges representing knowledge flow from one actor to another, and edge weights indicating the ``velocity'' (delay of flow) and ``viscosity'' (the depth of knowledge transferred). Graph algorithms are then proposed to find bottlenecks in knowledge flow. \citeH{KasneciSIRW08} propose a search engine for knowledge graphs, defined to be weighted directed edge-labelled graphs, where weights denote confidence scores based on the centrality of source documents from which the edge/relation was extracted. From the same group, \citeH{ElbassuoniRSSW09} adopt a similar notion of a knowledge graph, adding edge attributes to include keywords from the source, a count of supporting sources, etc., showing how the graph can be queried. \citeH{CourseyM09} construct a knowledge graph from Wikipedia, where nodes represent Wikipedia articles and categories, while edges represent the proximity of nodes. Subsequently, given an input text, entity linking and centrality measures are applied over the knowledge graph to determine relevant Wikipedia categories for the text. 

Concluding with the 10's (prior to 2012), \citeH{PechsiriP10} use knowledge graphs to capture ``explanation knowledge'' -- the knowledge of why something is the way it is -- by representing events as nodes and causal relationships as edges, claiming that this graphical notation offers intuitive explanations to users; their work focuses on extracting such knowledge graphs from text. \citeH{CorbyF10} use the phrase ``knowledge graph'' in a general way to denote any graph encoding knowledge, proposing an abstract machine for querying such graphs.

Other phrases were used to represent similar notions by other authors, including ``information graphs''~\cite{Kummel73}, ``information networks''~\cite{sun2011pathsim}, ``knowledge networks''~\cite{ciampaglia2015computational}, as well as ``semantic networks''~\cite{Brachman,woods,NavigliPonzetto:12} and ``conceptual graphs''~\cite{sowa}, as mentioned previously. Here we exclusively considered works that (happen to) use the phrase ``knowledge graph'' prior to Google's announcement of their knowledge graph in 2012, where we see that many works had independently coined this phrase for different purposes. Similar to the current practice, all of the works of this period consider a knowledge graph to be formed of a set of nodes denoting entities of interest and a set of edges denoting relations between those entities, with different entities and relations being considered in different works. Some works add extra elements to these knowledge graphs, such as edge weights, edge labels, or other meta-data~\cite{ElbassuoniRSSW09}. Other trends include knowledge acquisition from experts~\cite{rada1986gradualness,MachadoR90,DiengGTC92} and knowledge extraction from text~\cite{Bakker,stokman1988structuring,james,Hoede95}, combinations of symbolic and inductive methods~\cite{MachadoR90,de1990hybrid,ShimonyDS97,JrS99}, as well as the use of rules~\cite{rappaport1988dynamic}, ontologies~\cite{Hoede95}, graph analytics~\cite{SrikanthJ89,HelmsB05,KasneciSIRW08}, learning~\cite{rada1986gradualness,de1990hybrid,ShimonyDS97,JrS99}, and so forth. Later papers (2008--2010) by \citet{KasneciSIRW08}, \citet{ElbassuoniRSSW09}, \citet{CourseyM09} and \citet{CorbyF10} introduce notions of knowledge graph similar to current practice.

However, some trends are not reflected in current practice. Of particular note, quite a lot of the knowledge graphs defined in this period consider edges as denoting a form of dependence or causality, where \begin{tikzpicture}[baseline=-3pt]
    \setlength{\hgap}{0.5cm}
	\node[iri,compact,anchor=center] (x) {$x$};
	
	\node[iri,compact,anchor=center,right=\hgap of x] (y) {$y$}
	edge[arrin] (x);
\end{tikzpicture} may denote that $x$ is a prerequisite for $y$~\cite{Schneider72,MarchiM74,Jiang02} or that $x$ leads to $y$~\cite{rada1986gradualness,Bakker,rappaport1988dynamic,MachadoR90,ShimonyDS97,Jiang02}. In some cases \textsc{and}--\textsc{or} graphs are used to denote conjunctions or disjunctions of such relations~\cite{MachadoR90}, while in other cases edges are weighted to assign a belief to a relation~\cite{MachadoR90,Jiang02,rada1986gradualness}. In addition, papers from 1970--2000 tend to have worked with small graphs, which contrasts with modern practice where knowledge graphs can reach scales of millions or billions of nodes~\cite{NoyGJNPT19}: during this period, computational resources were more limited~\cite{Schneider72}, and fewer sources of structured data were readily available meaning that the knowledge graphs were often sourced solely from human experts~\cite{rada1986gradualness,MachadoR90,DiengGTC92} or from text~\cite{Bakker,stokman1988structuring,james,Hoede95}.

\subsection{``Knowledge Graphs'': 2012 Onwards}

Google Knowledge Graph was announced in 2012~\cite{GoogleKG}. This initial announcement was targeted at a broad audience, mainly motivating the knowledge graph and describing applications that it would enable, where the knowledge graph itself is described as ``\textit{[a graph] that understands real-world entities and their relationships to one another}''~\cite{GoogleKG}. Mentions of ``knowledge graphs'' quickly gained momentum in the research literature from that point. As noted by \citet{Bergman19}, this announcement by Google was a watershed moment in terms of adopting the phrase ``knowledge graph''. However, given the informal nature of the announcement, a technical definition  was lacking~\cite{EhrlingerW16,BonattiDPP18}.

Given that knowledge graphs were gaining more and more attention in the academic literature, formal definitions were becoming a necessity in order to precisely characterise what they were, how they were structured, how they could be used, etc., and more generally to facilitate their study in a precise manner. We can determine four general categories of definitions.

\begin{itemize}
\item[\textit{Category I}:]
The first category simply defines the knowledge graph as a graph where nodes represent entities, and edges represent relationships between those entities. Often a directed edge-labelled graph is assumed (or analogously, a set of binary relations, or a set of triples). This simple and direct definition was popularised by some of the seminal papers on knowledge graph embeddings~\cite{wang2014knowledge,lin2015learning} (2014--2015), being sufficient to represent the data structure upon which these embeddings would operate. As reflected in the survey by \citet{Wang2017KGEmbedding}, the multitude of works that would follow on knowledge graph embeddings have continued to use this definition. Though simple, the \textit{Category I} definition raises some doubts: How is a knowledge graph different from a graph (database)? Where does knowledge come into play?

\item[\textit{Category II}:]
 A second common definition goes as follows: ``\textit{a knowledge graph is a graph-structured knowledge base}'', where, to the best of our knowledge, the earliest usages of this definition in the academic literature were by \citeT{nickel} (2016) and \citeH{SeufertEBKBW16} (interestingly in the formal notation of these initial papers, a knowledge graph is defined analogously to a directed edge-labelled graph). Such a definition raises the question: what, then is a ``knowledge base''? The phrase ``knowledge base'' was popularised in the 70's (possibly earlier) in the context of rule-based expert systems~\cite{BuchananF78}, and later were used in the context of ontologies and other logical formalisms~\cite{BrachmanS85}. The follow-up question then is whether or not one can have a knowledge base (graph-structured or not) without a logical formalism while staying true to the original definitions. Looking in further detail, similar ambiguities have also existed regarding the definition of a ``knowledge base'' (KB). Of note: \citeH{BrachmanL85} -- reporting after a workshop on this issue -- state that ``\textit{if we ask what the KB tells us about the world, we are asking about its Knowledge Level}''. 

\item[\textit{Category III}:] The third category of definitions outline additional, technical characteristics that a ``knowledge graph'' should comply with, where we list some prominent definitions.

\begin{itemize}
\item In an influential survey on knowledge graph refinement, \citet{Paulheim17} lists four criteria that characterise the knowledge graphs considered for the paper. Specifically, that a knowledge graph ``\textit{mainly describes real world entities and their interrelations, organized in a graph; defines possible classes and relations of entities in a schema; allows for potentially interrelating arbitrary entities with each other; covers various topical domains}''; he thus rules out ontologies without instances (e.g., DOLCE) and graphs of word senses (e.g., WordNet) as not meeting the first two criteria, while relational databases do not meet the third criterion (due to schema restrictions), and domain-specific graphs (e.g., Geonames) are considered to not meet the fourth criterion; this leaves graphs such as DBpedia, YAGO, Freebase, etc.
\item \citet{EhrlingerW16} also review definitions of ``knowledge graph'', where they criticise the \textit{Category II} definitions based on the argument that knowledge bases are often synonymous with ontologies\footnote{Prior definitions of an ontology -- such as by~\citet{GuarinoOS9} -- would seem to contradict this conclusion.}, while knowledge graphs are not; they further criticise Google for calling its knowledge graph a ``knowledge base''. After reviewing prior definitions of terms such as ``knowledge base'', ``ontology'', and ``knowledge graph'', they propose their definition: ``\textit{A knowledge graph acquires and integrates information into an ontology and applies a reasoner
to derive new knowledge}''. In the subsequent discussion, they remark that a knowledge graph is distinguished from an ontology (considered synonymous with a knowledge base) by the provision of reasoning capabilities. 
\item One of the most detailed technical definitions for a ``knowledge graph'' is provided by~\citet{BellomariniFGS19}, who state: ``\textit{A knowledge graph is a semi-structured data model characterized by three components: (i) a ground extensional component, that is, a set of relational constructs for schema and data (which can be effectively modeled
as graphs or generalizations thereof); (ii) an intensional component, that is, a set of inference rules over the constructs of the ground extensional component; (iii) a derived extensional component that can be produced as the result of the application of the inference rules over the ground extensional component (with the so-called “reasoning” process).}'' They remark that ontologies and rules represent analogous structures, and that a knowledge graph is then a knowledge base extended with reasoning along similar lines to the definition provided by \citet{EhrlingerW16}.
\end{itemize} 

We refer to \citet{Bergman19} for a list of further definitions that fit this category. While having a specific, technical definition for knowledge graphs provides a more solid grounding for their study, as \citet{Bergman19} remarks, many of these definitions do not seem to fit the current practice of knowledge graphs. For example, it is not clear which of these definitions the Google Knowledge Graph itself -- responsible for popularising the idea -- would meet (if any). Furthermore, many of the criteria proposed by such definitions are orthogonal to the multitude of works in the area of knowledge graph embeddings~\cite{Wang2017KGEmbedding}.

\item[\textit{Category IV}:] While the previous three categories involve (sometimes conflicting) intensional definitions, the fourth category adopts an extensional definition of knowledge graphs, defining them by example. Knowledge graphs are then characterised by examples such as DBpedia, Google's Knowledge Graph, Freebase, YAGO, amongst others~\cite{BonattiDPP18}. Arguably this category sidesteps the issue of defining a knowledge graph, rather than providing such a definition.
\end{itemize}

These categories refer to definitions that have appeared in the academic literature. In terms of enterprise knowledge graphs, an important reference is the paper of \citet{NoyGJNPT19}, which has been co-authored by leaders of knowledge graph projects from eBay, Facebook, Google, IBM, and Microsoft, and thus can be seen as representing a form of consensus amongst these companies on what is a knowledge graph --- a concept these companies have played a key role in popularising. Specifically this paper states that ``\textit{a knowledge graph describes objects of interest and connections between them}'', and goes on to state that ``\textit{many practical implementations impose constraints on the links in knowledge graphs by defining a schema or ontology}''. They later add ``\textit{Knowledge graphs and similar structures usually provide a shared substrate of knowledge within an organization, allowing different products and applications to use similar vocabulary and to reuse definitions and descriptions that others create. Furthermore, they usually provide a compact formal representation that developers can use to infer new facts and build up the knowledge}''. We interpret this definition as corresponding to \textit{Category I}, but further acknowledging that while not a necessary condition for a knowledge graph, ontologies and formal representations \textit{usually} play a key role. The definition we provide at the outset of the paper is largely compatible with that of \citet{NoyGJNPT19}.
\section{Formal Definitions}\label{sec:formal}

In order to keep the discussion as accessible as possible, the body of the paper uses example-driven explanations of the main concepts and techniques associated with knowledge graphs. In this section, we complement the discussion of the paper with formal definitions.

\subsection{Data Graph Models}

We define the graph data models in line with previous conventions (e.g.,~\cite{AnglesABHRV17}). While different types of constants may be used in different models (e.g., RDF allows IRIs and literals), these definitions use a single (countably) infinite set of constants denoted $\con$. (We thus also abstract away from issues that are not exigent for the current introductory discussion, such as the existential semantics of blank nodes in RDF~\cite{HoganAMP14}, $D$-entailment over literals~\cite{rdf11sem}, positional restrictions~\cite{rdf11}, etc.)

\subsubsection{Directed edge-labelled graph} We first provide definitions for a directed edge-labelled graph.

\begin{definition}[Directed edge-labelled graph]\label{def:delg}
A \emph{directed edge-labelled graph} is a tuple $G \coloneqq (V,E,L)$, where $V \subseteq \con$ is a set of nodes, $L \subseteq \con$ is a set of edge labels, and $E \subseteq V \times L \times V$ is a set of edges.
\end{definition}

\begin{example}
In reference to Figure~\ref{fig:delg}, the set of nodes $V$ has 15 elements, including \texttt{Arica}, \texttt{EID16}, etc. The set of edges $E$ has 23 triples, including (\texttt{Arica},\texttt{flight},\texttt{Santiago}). Bidirectional edges are represented with two edges. The set of edge labels $L$ has 8 elements, including \texttt{start}, \texttt{flight}, etc.
\end{example}

Definition~\ref{def:delg} does not state that $V$ and $L$ are disjoint: though not present in the example, a node can also serve as an edge-label. The definition also permits that nodes and edge labels can be present without any associated edge. Either restriction could be explicitly stated -- if necessary -- in a particular application while still conforming to a directed edge-labelled graph.

In some of the definitions that follow, for ease of presentation, we may treat a set of (directed labelled) edges $E \subseteq V \times L \times V$ as a directed edge-labelled graph $(V,E,L)$, in which case we refer to the graph induced by $E$ assuming that $V$ and $L$ contain all and only those nodes and edge labels, respectively, used in $E$. We may similarly apply set operators on directed edge-labelled graphs, which should be interpreted as applying to their sets of edges; for example, given $G_1 = (V_1,E_1,L_1)$ and $G_2 = (V_2,E_2,L_2)$, by $G_1 \cup G_2$ we refer to the directed edge-labelled graph induced by $E_1 \cup E_2$.

\subsubsection{Heterogeneous graph} We next define the notion of a heterogeneous graph.

\begin{definition}[Heterogeneous graph]\label{def:hg}
A \emph{heterogeneous graph} is a tuple $G \coloneqq (V,E,L,l)$, where $V \subseteq \con$ is a set of nodes, $L \subseteq \con$ is a set of edge and node labels, $E \subseteq V \times L \times V$ is a set of edges, and $l : V \rightarrow L$ maps each node to a label.
\end{definition}

\begin{example}
In reference to Figure~\ref{fig:hg}, the set of nodes $V$ has three elements: \texttt{Santiago}, \texttt{Chile}, and \texttt{Perú}. The set of edges $E$ has 3 triples, including (\texttt{Santiago},\texttt{capital},\texttt{Chile}). The set of edge labels $L$ has 4 elements: \texttt{capital}, \texttt{borders}, \texttt{City}, \texttt{Country}. Finally, with respect to the node labels, $l(\texttt{Santiago}) = \texttt{City}$, $l(\texttt{Chile}) = \texttt{Country}$, and $l(\texttt{Perú}) = \texttt{Country}$.
\end{example}

In heterogeneous graphs, edge and node labels are most commonly called \textit{types}. We remark that by defining edges with labels per directed-edge labelled graphs -- rather than labelling edges with $l$ -- we allow two nodes to be related by $n$ edges with $n$ different labels; e.g., we can represent both $(\texttt{Santiago},\texttt{capital},\texttt{Chile})$ and $(\texttt{Santiago},\texttt{country},\texttt{Chile})$.

\subsubsection{Property graph} Finally, we define a property graph.

\begin{definition}[Property graph]\label{def:pg}
A \emph{property graph} is a tuple $G \coloneqq (V,E,L,P,U,e,l,p)$, where $V \subseteq \con$ is a set of node ids, $E \subseteq \con$ is a set of edge ids, $L \subseteq \con$ is a set of labels, $P  \subseteq \con$ is a set of properties, $U \subseteq \con$ is a set of values, $e : E \rightarrow V \times V$ maps an edge id to a pair of node ids, $l : V \cup E \rightarrow 2^L$ maps a node or edge id to a set of labels, and $p : V \cup E \rightarrow 2^{P \times U}$ maps a node or edge id to a set of property--value pairs.
\end{definition}

\begin{example}
Returning to Figure~\ref{fig:pg}:

\begin{itemize} 
\item the set $V$ contains \texttt{Santiago} and \texttt{Arica}; 
\item the set $E$ contains \texttt{LA380} and \texttt{LA381};
\item the set $L$ contains \texttt{Capital City}, \texttt{Port City}, and \texttt{flight}; 
\item the set $P$ contains \texttt{lat}, \texttt{long}, and \texttt{company}; 
\item the set $U$ contains \texttt{$-$33.45}, \texttt{$-$70.66}, \texttt{LATAM}, \texttt{$-$18.48}, and \texttt{$-$70.33}; 
\item the mapping $e$ gives, for example, $e(\texttt{LA380}) = (\texttt{Santiago},\texttt{Arica})$; 
\item the mapping $l$ gives, for example, $l(\texttt{LA380}) =\{ \texttt{flight} \}$ and $l(\texttt{Santiago}) =\{ \texttt{Capital City} \}$; 
\item the mapping $p$ gives, for example, $p(\texttt{Santiago}) =\{ (\texttt{lat},\texttt{$-$33.45}), (\texttt{long},\texttt{$-$70.66}) \}$ and $p(\texttt{LA380}) =\{ (\texttt{company},\texttt{LATAM}) \}$.
\end{itemize}
\end{example}

Definition~\ref{def:pg} does not require that the sets $V$, $E$, $L$, $P$ or $U$ to be (pairwise) disjoint: we allow, for example, that values are also nodes. Unlike some previous definitions~\cite{AnglesABHRV17}, here we allow a node or edge to have several values for a given property. In practice, systems like Neo4j~\cite{Miller13} may rather support this by allowing an array of values. We view such variations as syntactic.

\subsubsection{Graph dataset} Next we define a graph dataset, where one can consider directed-edge labelled graph datasets, heterogeneous graph datasets, property graph datasets, etc.

\begin{definition}[Graph dataset]\label{def:gd}
A \textit{named graph} is a pair $(n,G)$ where $G$ is a graph, and $n \in \con$ is a graph name. A \emph{graph dataset} is a pair $D \coloneqq (G_D,N)$ where $G_D$ is a directed edge-labelled graph called the \emph{default graph} and $N$ is either the empty set, or a set of named graphs $\{ (n_1,G_1), \ldots (n_k,G_k) \}$ ($k > 0$) such that $n_i = n_j$ if and only if $i = j$ ($1 \leq i \leq k$, $1 \leq j \leq k$).
\end{definition}

\begin{example}
Figure~\ref{fig:gd} provides an example of a directed-edge labelled graph dataset $D$ consisting of two named graphs and a default graph. The default graph does not have a name associated with it. The two graph names are \texttt{Events} and \texttt{Routes}; these are also used as nodes in the default graph. 
\end{example}

An RDF dataset is a graph dataset model standardised by the W3C~\cite{rdf11} where each graph is an RDF graph, and graph names can be blank nodes or IRIs.

\subsection{Querying}

Here we formalise foundational notions relating to queries over graphs, starting with graph patterns, to which we later add relational-style operators and path expressions.

\subsubsection{Graph patterns}\label{app:gps}

We formalise the notions of graph patterns first for directed edge-labelled graphs, and subsequently for property graphs~\cite{AnglesABHRV17}. For these definitions, we introduce a countably infinite set of \textit{variables} $\var$ ranging over (but disjoint from: $\con \cap \var = \emptyset$) the set of constants. We refer generically to constants and variables as \textit{terms}, denoted and defined as $\term = \con \cup \var$.

\begin{definition}[Directed edge-labelled graph pattern]\label{def:delgp}
We define a \emph{directed edge-labelled graph pattern} as a tuple $Q = (V,E,L)$, where $V \subseteq \term$ is a set of node terms, $L \subseteq \term$ is a set of edge terms, and $E \subseteq V \times L \times V$ is a set of edges (triple patterns).
\end{definition}

\begin{example}
Returning to the graph pattern of Figure~\ref{fig:gp}:

\begin{itemize} 
\item the set $V$ contains the constant \texttt{Food Festival} and variables \texttt{?event}, \texttt{?ven1} and \texttt{?ven2}; 
\item the set $L$ contains the constants \texttt{type} and \texttt{venue};
\item the set $E$ contains four edges, including $(\texttt{?event},\texttt{type},\texttt{Food Festival})$, etc.
\end{itemize}
\end{example}

A property graph pattern is defined analogously, allowing variables in any position.

\begin{definition}[Property graph pattern]\label{def:pgp}
We define a \emph{property graph pattern} as a tuple $Q = (V,E,L,P,U,e,l,p)$, where $V \subseteq \term$ is a set of node id terms, $E \subseteq \term$ is a set of edge id terms, $L \subseteq \term$ is a set of label terms, $P  \subseteq \term$ is a  set of property terms, $U \subseteq \term$ is a set of value terms, $e : E \rightarrow V \times V$ maps an edge id term to a pair of node id terms, $l : V \cup E \rightarrow 2^{L}$ maps a node or edge id term to a set of label terms, and $p : V \cup E \rightarrow 2^{P \times U}$ maps a node or edge id term to a set of pairs of property--value terms.
\end{definition}

Towards defining the evaluation of a graph pattern, we first define a partial mapping $\mu : \var \rightarrow \con$ from variables to constants, whose \textit{domain} (the set of variables for which it is defined) is denoted by $\dom(\mu)$. Given a graph pattern $Q$, let $\var(Q)$ denote the set of all variables appearing in (some recursively nested element of) $Q$. Abusing notation, we denote by $\mu(Q)$ the image of $Q$ under $\mu$, meaning that any variable $v \in \var(Q) \cap \dom(\mu)$ is replaced in $Q$ by $\mu(v)$. Observe that when $\var(Q) \subseteq \dom(\mu)$, then $\mu(Q)$ is a data graph (in the corresponding model of $Q$). 

Next, we define the notion of containment between data graphs. For two directed edge-labelled graph pattern $G_1 = (V_1,E_1,L_1)$ and $G_2 = (V_2,E_2,L_2)$, we say that $G_1$ is a \textit{sub-graph} of $G_2$, denoted $G_1 \subseteq G_2$, if and only if $V_1 \subseteq V_2$, $E_1 \subseteq E_2$, and $L_1 \subseteq L_2$.\footnote{Given, for example, $G_1 = (\{a\},\{(a,b,a)\},\{b,c\})$ and $G_2 = (\{a,c\},\{(a,b,a)\},\{b\})$, we remark that $G_1 \not\subseteq G_2$ and $G_2 \not\subseteq G_1$: the former has a label not used on an edge while the latter has a node without an incident edge. In concrete data models like RDF where such cases of nodes or labels without edges cannot occur, the sub-graph relation $G_1 \subseteq G_2$ holds if and only if $E_1 \subseteq E_2$ holds.} Conversely, in property graphs, nodes can often be defined without edges. For two property graphs $G_1 = (V_1,E_1,L_1,P_1,U_1,e_1,l_1,p_1)$ and $G_2 = (V_2,E_2,L_2,P_2,U_2,e_2,l_2,p_2)$, we say that $G_1$ is a \textit{sub-graph} of $G_2$, denoted $G_1 \subseteq G_2$, if and only if $V_1 \subseteq V_2$, $E_1 \subseteq E_2$, $L_1 \subseteq L_2$, $P_1 \subseteq P_2$, $U_1 \subseteq U_2$, for all $x \in E_1$ it holds that $e_1(x) = e_2(x)$, and for all $y \in E_1 \cup V_1$ it holds that $l_1(y) \subseteq l_2(y)$ and $p_1(y) \subseteq p_2(y)$.

We are now ready to define the evaluation of a graph pattern.

\begin{definition}[Evaluation of a graph pattern]\label{def:evgp}
Let $Q$ be a graph pattern and let $G$ be a data graph. We then define the \emph{evaluation of graph pattern $Q$ over the data graph $G$}, denoted $Q(G)$, to be the set of mappings $\{ \mu \mid \mu(Q) \subseteq G \text{ and } \dom(\mu) = \var(Q) \}$.
\end{definition}

\begin{example}
Figure~\ref{fig:gp} enumerates all of the mappings given by the evaluation of the depicted graph pattern over the data graph of Figure~\ref{fig:delg}. Each non-header row indicates a mapping $\mu$.
\end{example}

The final results of evaluating a graph pattern may then vary depending on the choice of semantics: the results under \textit{homomorphism-based semantics} are defined as $Q(G)$. Conversely, under \textit{isomorphism-based} semantics, mappings that send two edge variables to the same constant and/or mappings that send two node variables to the same constant may be excluded from the results. Henceforth we assume the more general \textit{homomorphism-based semantics}.

\subsubsection{Complex graph patterns}\label{app:cgps}

We now define complex graph patterns.

\begin{definition}[Complex graph pattern]\label{def:cgp}
\emph{Complex graph patterns} are defined recursively:
\begin{itemize}
    \item If $Q$ is a graph pattern, then $Q$ is a \textit{complex graph pattern}.
    \item If $Q$ is a complex graph pattern, and $\mathcal{V} \subseteq \var(Q)$, then $\pi_\mathcal{V}(Q)$ is a \emph{complex graph pattern}.
    \item If $Q$ is a complex graph pattern, and $R$ is a selection condition with boolean and equality connectives ($\wedge$, $\vee$, $\neg$, $=$) , then $\sigma_R(Q)$ is a \emph{complex graph pattern}.
    \item If $Q_1$ and $Q_2$ are complex graph patterns, then $Q_1 \Join Q_2$, $Q_1 \cup Q_2$, $Q_1 - Q_2$ and  and $Q_1 \rhd Q_2$ are also \emph{complex graph patterns}.
\end{itemize}
\end{definition}

Next we define the evaluation of complex graph patterns. First, given a mapping $\mu$, for a set of variables $\mathcal{V} \subseteq \var$ let $\mu[\mathcal{V}]$ denote the mapping $\mu'$ such that $\dom(\mu') = \dom(\mu) \cap \mathcal{V}$ and $\mu(v) = \mu'(v)$ for all $v \in \dom(\mu')$ (in other words, $\mu[\mathcal{V}]$ projects the variables $\mathcal{V}$ from $\mu$). Furthermore, letting $R$ denote a boolean selection condition and $\mu$ a mapping, by $\mu \models R$ we denote that $\mu$ satisfies the boolean condition. Finally, we define two mappings $\mu_1$ and $\mu_2$ to be \textit{compatible}, denoted $\mu_1 \sim \mu_2$, if and only if $\mu_1(v) = \mu_2(v)$ for all $v \in \dom(\mu_1) \cap \dom(\mu_2)$ (in other words, they map all common variables to the same constant). We are now ready to provide the definition.

\begin{definition}[Complex graph pattern evaluation]\label{def:evalcgp} Given a complex graph pattern $Q$, if $Q$ is a graph pattern, then $Q(G)$ is defined per Definition~\ref{def:evgp}. Otherwise:
\begin{align*}
 \pi_\mathcal{V}(Q)(G) \coloneqq &  \,\{ \mu[\mathcal{V}] \mid \mu \in Q(G) \} \\
 \sigma_R(Q)(G) \coloneqq & \, \{ \mu \mid \mu \in Q(G)\text{ and }\mu \models R\}\\
 Q_1 \Join Q_2(G) \coloneqq &  \,\{ \mu_1 \cup \mu_2 \mid \mu_1 \in Q_2(G)\text{ and } \mu_2 \in Q_1(G)\text{ and }\mu_1 \sim \mu_2 \} \\
 Q_1 \cup Q_2(G) \coloneqq &  \,\{ \mu \mid \mu \in Q_1(G)\text{ or } \mu \in Q_2(G) \} \\
  Q_1 - Q_2(G) \coloneqq & \,\{ \mu \mid \mu \in Q_1(G)\text{ and } \mu \notin Q_2(G) \} \\
 Q_1 \rhd Q_2(G) = & \,\{ \mu \mid \mu \in Q_1(G)\text{ and }\nexists \mu_2 \in Q_2(G)\text{ such that }\mu \sim \mu_2 \} \\  
\end{align*}
\end{definition}

Based on these query operators, we can also define some additional syntactic operators, such as the \textit{left-join} ($\LeftJoin$, aka \textit{optional}):
\begin{align*}
 Q_1 \LeftJoin Q_2(G) \coloneqq & \,(Q_1(G) \Join Q_2(G)) \cup (Q_1(G) \rhd Q_2(G))
\end{align*}
\noindent We call such operators \textit{syntactic} as they do not add expressivity to the query language.

\begin{example}
Figure~\ref{fig:cgp} illustrates a complex graph pattern and its evaluation.
\end{example}

\subsubsection{Navigational graph patterns}\label{app:ngps}

We first define path expressions and regular path queries.

\begin{definition}[Path expression] A constant (edge label) $c$ is a \emph{path expression}. Furthermore:

\begin{itemize}
    \item If $r$ is a path expression, then $r^-$ (\textit{inverse}) and $r^*$ (\textit{Kleene star}) are \emph{path expressions}.
    \item If $r_1$ and $r_2$ are path expressions, then $r_1 \cdot r_2$ (\textit{concatenation}) and $r_1 \mid r_2$ (\textit{disjunction}) are \emph{path expressions}.
\end{itemize}
\end{definition}

We now define the evaluation of a path expression under the SPARQL 1.1-style semantics whereby the endpoints (pairs of start and end nodes) of the path are returned~\cite{sparql11}.

\begin{definition}[Path expression evaluation (directed edge-labelled graph)] Given a directed edge-labelled graph $G = (V,E,L)$ and a path expression $r$, we define the \emph{evaluation of $r$ over $G$}, denoted $r[G]$, as follows:

\begin{align*}
r[G] \coloneqq &\, \{ (u,v) \mid (u,r,v) \in E \} \,(\text{for }r \in \con) \\
r^-[G] \coloneqq &\, \{ (u,v) \mid (v,u) \in r[G] \} \\
r_1 \mid r_2[G] \coloneqq &\, r_1[G] \cup r_2[G] \\
r_1 \cdot r_2[G] \coloneqq &\, \{ (u,v) \mid \exists w \in V : (u,w) \in r_1[G]\text{ and }(w,v) \in r_2[G]\}\\
r^*[G] \coloneqq &\, V \cup \bigcup_{n \in \mathbb{N^+}} r^n[G]
\end{align*}

\noindent where by $r^n$ we denote the $n$\textsuperscript{th}-concatenation of $r$ (e.g., $r^3 = r \cdot r \cdot r$).
\end{definition}

The evaluation of a path expression on a property graph $G = (V,E,L,P,U,e,l,p)$ can be defined analogously by adapting the first definition (in the case that $r \in \con$) as follows:
\[ r[G] \coloneqq \{(u,v) \mid \exists x \in E : e(x) = (u,v)\text{ and }l(e) = r \} \,.\]

\noindent The rest of the definitions then remain unchanged.

Query languages may support additional operators, some of which are syntactic (e.g., $r^+$ is sometimes used for one-or-more, but can be rewritten as $r \cdot r^*$), while others may add expressivity such as the case of SPARQL~\cite{sparql11}, which allows a limited form of negation in expressions (e.g., $!r$, with $r$ being a constant or the inverse of a constant, matching any path not labelled $r$).

Next we define a regular path query and its evaluation.

\begin{definition}[Regular path query] A \emph{regular path query} is a triple $(x,r,y)$ where $x,y \in \con \cup \var$ and $r$ is a path expression.
\end{definition}

\begin{definition}[Regular path query evaluation] Let $G$ denote a directed edge-labelled graph, $c$, $c_1$, $c_2 \in \con$ denote constants and $z$, $z_1$, $z_2 \in \var$ denote variables. Then the \textit{evaluation of a regular path query} is defined as follows:
\begin{align*}
(c_1,r,c_2)(G) \coloneqq & \{ \mu_\emptyset \mid (c_1,c_2) \in r[G] \} \\
(c,r,z)(G) \coloneqq & \{ \mu \mid \dom(\mu) = \{ z \}\text{ and }(c,\mu(z)) \in r[G] \} \\
(z,r,c)(G) \coloneqq & \{ \mu \mid \dom(\mu) = \{ z \}\text{ and }(\mu(z),c) \in r[G] \} \\
(z_1,r,z_2)(G) \coloneqq & \{ \mu \mid \dom(\mu) = \{ z_1, z_2 \}\text{ and }(\mu(z_1),\mu(z_2)) \in r[G] \}
\end{align*}

\noindent where $\mu_\emptyset$ denotes the empty mapping such that $\dom(\mu) = \emptyset$ (the join identity).
\end{definition}

\begin{definition}[Navigational graph pattern] If $Q$ is a graph pattern, then $Q$ is a \emph{navigational graph pattern}. Furthermore, if $Q$ is a navigational graph pattern and $(x,r,y)$ is a regular path query, then $Q \Join (x,r,y)$ is a \emph{navigational graph pattern}.
\end{definition}

The definition of the evaluation of a navigational graph pattern then follows from the previous definition of a join and the corresponding definition of the evaluation of a regular path query (for a directed edge-labelled graph or a property graph, respectively). Likewise, \textit{complex navigational graph patterns} -- and their evaluation -- are defined by extending this definition in the natural way with the same operators from Definition~\ref{def:cgp} following the same semantics seen in Definition~\ref{def:evalcgp}.

\subsection{Schema}

Here we formalise notions relating to schemata for graphs. Though we present definitions for directed edge-labelled graphs -- which allows for more succinct presentation -- the same concepts can be applied to property graphs and other graph models.

\subsubsection{Semantic schema} We provide definitions that generalise semantic schemata in Appendix~\ref{app:deductive}.

\subsubsection{Validating schema}\label{app:shapes} We define shapes following conventions used by~\citet{Labra-Gayo2019}.

\begin{definition}[Shape] A \emph{shape} \shape{} is defined as:

\smallskip
\begin{tabular}{p{1cm}  p{0.5cm} p{3cm} l}
	\shape & ::=  & \shapeTrue & true\\
	& $|$  & $\datatype{N}$ & node belongs to the set of nodes $N$ \\
	& $|$  & $\bcond{\mathrm{cond}}$ & node satisfies the boolean condition $\mathrm{cond}$ \\
	& $|$  & $\shape_1 \sAnd \shape_2$ & conjunction of shape $\shape_1$ and shape $\shape_2$\\
	& $|$  & $\sNot \shape $ & negation of shape $\shape$\\
	& $|$  & \shapeRef{s} & reference to shape with label $s$\\
	& $|$  & $\qualified{\predSpec}{\shape}{min}{max}$ & between $min$ and $max$ outward edges (inclusive)\\ 
	& & &  with label $p$ to nodes satisfying shape \shape \\  
\end{tabular}

\medskip
\noindent
where $min \in \mathbb{N}_{(0)}$, $max \in \mathbb{N}_{(0)} \cup \{ \unbounded \}$, with ``\unbounded''\ indicating unbounded.
\end{definition}

\begin{definition}[Shapes schema]
A \emph{shapes schema} is defined as a tuple $\Sigma \coloneq (\Phi,S,\lambda)$ where $\Phi$ is a set of shapes, $S$ is a set of shape labels, and $\lambda : S \rightarrow \Phi$ is a total function from labels to shapes.
\end{definition}

\begin{example}
The shapes schema from Figure~\ref{fig:shapeExample} can be expressed as:

\smallskip
\begin{tabular}{p{1cm}  p{0.4cm} l}
\shap{Event} & 
 $\mapsto$ & 
 $\qualifiedL{name}{\datatypeL{string}}{1}{\unbounded}\sAnd\qualifiedL{start}{\datatypeL{dateTime}}{1}{1}\sAnd\qualifiedL{end}{\datatypeL{dateTime}}{1}{1}$ \\
& & $\qquad\sAnd\qualifiedL{type}{\shapeTrue}{1}{\unbounded}\sAnd\qualifiedL{venue}{\shapeRefL{Venue}}{1}{\unbounded}$ \\
\shap{Venue} & $\mapsto$ & $\shapeRefL{Place}\:\sAnd\qualifiedL{indoor}{\datatypeL{boolean}}{0}{1}\sAnd\qualifiedL{city}{\shapeRefL{City}}{0}{1}$ \\
\shap{City} & $\mapsto$ & $\shapeRefL{Place}\:\sAnd\qualifiedL{population}{(\datatypeL{int}\sAnd \bcond{>5000})}{0}{1}$ \\
\shap{Place} & $\mapsto$ & $\qualifiedL{lat}{\datatypeL{float}}{0}{1}\sAnd\qualifiedL{long}{\datatypeL{float}}{0}{1}$ \\
& & $\qquad\sAnd\qualifiedL{flight}{\shapeRefL{Place}}{0}{\unbounded}\sAnd\qualifiedL{bus}{\shapeRefL{Place}}{0}{\unbounded}$ \\
\end{tabular}
\end{example}

In a shapes schema, shapes may refer to other shapes, giving rise to a graph that is sometimes known as the \textit{shapes graph}~\cite{SHACLSpec}. Figure~\ref{fig:shapeExample} illustrates a shapes graph of this form.
\medskip

The semantics of a shape $\shape{}$ is defined in terms of the evaluation of \shape{} over the nodes of a graph $G = (V,E,L)$ with respect to a shapes map $\sigma$ associating nodes and shape labels that apply to them.

\begin{definition}[Shapes map]
Given a graph $G = (V,E,L)$ and a schema $\Sigma = (\Phi,S,\lambda)$, a \emph{shapes map} is a (partial) mapping $\sigma: V \times S \rightarrow \{ 0, 1 \}$.
\end{definition} 

The precise semantics of a shape then depends on  whether or not $\sigma$ is a total or partial mapping: whether or not it is defined for every value in $V \times S$. In this paper, we present the semantics for the more straightforward case where $\sigma$ is assumed to be a total shapes map.

\begin{definition}[Shape evaluation] Given a shapes schema $\Sigma \coloneq (\Phi,S,\lambda)$, the semantics of a shape $\shape{} \in \Phi$ is defined in terms of a \emph{shape evaluation function} $\semantics{\shape}{G}{v}{\sigma} \in \{ 0 , 1 \}$, for a graph $G = (V,E,L)$, a node $v \in V$ and a total shapes map $\sigma$, such that:

\smallskip
\begin{tabular}{p{3.3cm} p{0.5cm} l}
\semantics{\shapeTrue}{G}{v}{\sigma} & = & 1 \\
\semantics{\datatype{\ensuremath{N}}}{G}{v}{\sigma} & = & 1 iff $v \in N$\\
\semantics{\bcond{\mathrm{cond}}}{G}{v}{\sigma} & = & 1 iff $\mathrm{cond}(v)$ is true  \\
\semantics{\shape_1 \sAnd \shape_2}{G}{v}{\sigma} & = & $\min\{\semantics{\shape_1}{G}{v}{\sigma}, \semantics{\shape_2}{G}{v}{\sigma}\}$ \\
\semantics{\sNot \shape}{G}{v}{\sigma} & = & $1 - \semantics{\shape}{G}{v}{\sigma}$\\
\semantics{\shapeRef{s}}{G}{v}{\sigma} & = & 1 iff $\sigma(v,s) = 1$ \\  
\semantics{\qualified{\predSpec}{\shape}{min}{max}}{G}{v}{\sigma} & = & 1 iff $min \leq \lvert \{ (v,\predSpec,u)\in E \mid \semantics{\shape}{G}{u}{\sigma}=1 \} \rvert \leq max$\\
\end{tabular}

\medskip
\noindent
If $\semantics{\shape}{G}{v}{\sigma} = 1$, then $v$ is said to \textit{satisfy} $\shape$ in $G$ under $\sigma$.
\end{definition}

Typically for the purposes of validating a graph with respect to a shapes schema, a \textit{target} is defined that requires certain nodes to satisfy certain shapes.

\begin{definition}[Shapes target]
Given a directed edge-labelled graph $G = (V,E,L)$ and a shapes schema $\Sigma = (\Phi,S,\lambda)$, a \emph{shapes target} $T$ is a set of pairs of nodes and shape labels: $T \subseteq V \times S$.
\end{definition}

The nodes that a shape targets can be selected manually, based on the type(s) of the nodes, based on the results of a graph query, etc.~\cite{Corman2018b, Labra-Gayo2019}.
\medskip

Lastly, we can define the notion of a valid graph under a given shapes schema and target.

\begin{definition}[Valid graph] Given a shapes schema $\Sigma = (\Phi,S,\lambda)$, a graph $G = (V,E,L)$, and a shapes target $T$, we say that \emph{$G$ is valid under $\Sigma$ and $T$} if and only if there exists a shapes map $\sigma$ such that, for all $s \in S$ and $v \in V$ it holds that $\sigma(v,s) =  \semantics{\lambda(s)}{G}{v}{\sigma}$, and  $(v,s) \in T$ implies $\sigma(v,s) = 1$.
\end{definition}

\begin{example}
Taking the graph $G$ from Figure~\ref{fig:delg} and the shapes schema $\Sigma$ from Figure~\ref{fig:shapeExample}, first assume an empty shapes target $T = \{\}$. If we consider a shapes map where (for example) $\sigma(\gnode{EID15},\shap{Event}) = 1$, $\sigma(\gnode{Santa Lucía},\shap{Venue}) = 1$, $\sigma(\gnode{Santa Lucía},\shap{Place}) = 1$, etc., but where $\sigma(\gnode{EID16},\shap{Event}) = 0$ (as it does not have the required values for \gelab{start} and \gelab{end}), etc., then we see that $G$ is valid under $\Sigma$ and $T$. However, if we were to define a shapes target $T$ to ensure that the $\shap{Event}$ shape targets $\gnode{EID15}$ and $\gnode{EID16}$ -- i.e., to define $T$ such that $\{ (\gnode{EID15},\shap{Event}), (\gnode{EID16},\shap{Event}) \} \subseteq T$ -- then the graph would no longer be valid under $\Sigma$ and $T$ since $\gnode{EID16}$ does not satisfy $\shap{Event}$.
\end{example}

The semantics we present here assumes that each node in the graph either satisfies or does not satisfy each shape labelled by the schema. More complex semantics -- for example, based on Kleene's three-valued logic~\cite{Corman2018b, Labra-Gayo2019} -- have been proposed that support partial shapes maps, where the satisfaction of some nodes for some shapes can be left undefined. Shapes languages in practice may support other forms of constraints, such as counting on paths~\cite{SHACLSpec}. In terms of implementing validation with respect to shapes, work has been done on translating constraints into sets of graph queries, whose results are input to a SAT solver for recursive cases~\cite{CormanFRS19a}.

\subsubsection{Emergent schema}\label{sec:feschema}

Emergent schemata are often based on the notion of a quotient graph.

\begin{definition}[Quotient graph]\label{def:qg} Given a directed-edge labelled graph $G = (V,E,L)$, a graph $\mathcal{G} = (\mathcal{V},\mathcal{E},L)$ is a \emph{quotient graph} of $G$ if and only if:
	
\begin{itemize}
	\item $\mathcal{V}$ is a partition of $V$ without the empty set, i.e., $\mathcal{V} \subseteq (2^V - \emptyset)$, $V = \bigcup_{U\in \mathcal{V}} U$, and for all $U\in \mathcal{V}$, $W\in \mathcal{V}$, it holds that $U = W$ or $U \cap W = \emptyset$; \emph{and}
	\item $\mathcal{E} = \{ (U,l,W) \mid U \in \mathcal{V}, W \in \mathcal{V}\text{ and there exist } u \in U, w \in W\text{ such that }(u,l,w) \in E \} $.
\end{itemize}
\end{definition}

Intuitively speaking, a quotient graph can merge multiple nodes into one node, where the merged node preserves the edges of its constituent nodes. For an input graph $G = (V,E,L)$, there is an exponential number of potential quotient graphs: as many as there are partitions of the input graphs' nodes. On one extreme, the input graph is a quotient graph of itself (turning nodes like \gnode{u} into singleton nodes like \gnode{\{u\}}). On the other extreme, a single node \gnode{$V$}, with all input nodes, and loops $(V,l,V)$ for each edge-label $l$ used in $E$, the set of input edges, is also a quotient graph. Practical quotient graphs typically fall somewhere in between, where the partition  $\mathcal{V}$ of $V$ is often defined in terms of an \textit{equivalence relation} $\sim$ on the set $V$ such that $\mathcal{V} \coloneqq {\sim}/V$; i.e., $\mathcal{V}$ is defined as the \textit{quotient set} of $V$ with respect to $\sim$; for example, we might define an equivalence relation on nodes such that $u \sim v$ if and only if they have the same set of defined types, where ${\sim}/V$ is then a partition whose parts contain all nodes with the same types. Another way to induce a quotient graph is to define the partition in a way that preserves some of the topology of the input graph. One way to formally define this idea is through \textit{simulation} and \textit{bisimulation}.

\begin{definition}[Simulation]\label{def:sim} Given two directed-edge labelled graph $G \coloneqq (V,E,L)$ and $G' \coloneqq (V',E',L')$, let $R \subseteq V \times V'$ be a relation between the nodes of $G$ and $G'$, respectively.  We call $R$ a \emph{simulation} on $G$ and $G'$ if, for all $(v,v') \in R$, the following holds:

\begin{itemize}
	\item if $(v,p,w) \in E$ then there exists $w'$ such that $(v',p,w') \in E'$ and $(w,w') \in R$.
\end{itemize}

\noindent If a simulation exists on $G$ and $G'$, we say that $G'$ \emph{simulates} $G$, denoted $G \rightsquigarrow G'$.
\end{definition}

\begin{definition}[Bisimulation]\label{def:bisim}
If $R$ is a simulation on $G$ and $G'$, we call it a \emph{bisimulation} if, for all $(v,v') \in R$, the following condition holds:

\begin{itemize}
	\item if $(v'p,w') \in E'$ then there exists $w$ such that $(v,p,w) \in E$ and $(w,w') \in R$.
\end{itemize}

\noindent
If a bisimulation exists on $G$ and $G'$, we say that they are \emph{bisimilar}, denoted $G \approx G'$.
\end{definition}

Bisimulation ($\approx$) is then an equivalence relation on graphs. By defining the (bi)simulation relation $R$ in terms of set membership $\in$, every quotient graph simulates its input graph, but does not necessarily bisimulate its input graph. This gives rise to the notion of \textit{bisimilar quotient graphs}.

\begin{example}
Figures~\ref{fig:emergentSchema} and~\ref{fig:emergentSchema2} exemplify quotient graphs for the graph of Figure~\ref{fig:delg}. Figure~\ref{fig:emergentSchema} simulates but is not bisimilar to the data graph. Figure~\ref{fig:emergentSchema2} is bisimilar to the data graph. Often the goal will be to compute the most concise quotient graph that satisfies a given condition; for example, the nodes without outgoing edges in Figure~\ref{fig:emergentSchema2} could be merged while preserving bisimilarity.
\end{example}

\subsection{Context}

\subsubsection{Annotation domain}\label{sec:annotationDomain}

We define an annotation domain per Zimmermann et al.~\cite{zimm-etal-2012-JWS}.

\begin{definition}[Annotation domain]\label{def:anndom} Let $A$ be a set of \emph{annotation values}. An \emph{annotation domain} is defined as an idempotent, commutative semi-ring $D = \langle A,\oplus,\otimes,\bot,\top \rangle$.
\end{definition} 

This definition can be used to instantiate specific domains of context. Letting $D$ be a semi-ring imposes that, for any values $a, a_1, a_2, a_3$ in $A$, the following hold:

\begin{itemize}
	\item $(a_1 \oplus a_2) \oplus a_3 = a_1 \oplus (a_2 \oplus a_3)$
	\item $(\bot \oplus a) = (a \oplus \bot) = a$
	\item $(a_1 \oplus a_2) = (a_2 \oplus a_1)$
	\item $(a_1 \oplus a_2) = (a_2 \oplus a_1)$
	\item $(a_1 \otimes a_2) \otimes a_3 = a_1 \otimes (a_2 \otimes a_3)$
	\item $(\top \otimes a) = (a \otimes \top) = a$
	\item $a_1 \otimes (a_2 \oplus a_3) = (a_1 \otimes a_2) \oplus (a_1 \otimes a_3)$
	\item $(a_1 \oplus a_2) \otimes a_3 = (a_1 \otimes a_3) \oplus (a_2 \otimes a_3)$
	\item $(\bot \otimes a) = (a \otimes \bot) = \bot$
\end{itemize}

\noindent
The requirement that it be a commutative semi-ring imposes the following constraint:

\begin{itemize}
	\item $(a_1 \otimes a_2) = (a_2 \otimes a_1)$
\end{itemize}

\noindent
Finally, the requirement that it be an idempotent semi-ring imposes the following constraint:

\begin{itemize}
	\item $(a \oplus a) = a$
\end{itemize}

\noindent Idempotence induces a partial order: $a_1 \leq a_2$ if and only if $a_1 \oplus a_2 = a_2$. Imposing these conditions on the annotation domain allow for reasoning and querying to be conducted over the annotation domain in a well-defined manner. Annotated graphs can then be defined in the natural way:

\begin{definition}[Annotated directed-edge labelled graph] Letting $D = \langle A,\oplus,\otimes,\bot,\top \rangle$ denote an idempotent, commutative semi-ring, we define an \emph{annotated directed-edge labelled graph} $G \coloneq (V,E_A,L)$ where $V \subseteq \con$ is a set of nodes, $L \subseteq \con$ is a set of edge labels, and $E_A \subseteq V \times L \times V \times A$ is a set of edges annotated with values from $A$.
\end{definition}

Figure~\ref{fig:time} exemplifies query answering on a graph annotated with days of the year. Formally this domain can be defined as follows: $A \coloneq 2^{\mathbb{N}_{[1,365]}}$, $\oplus \coloneq \cup$, $\otimes \coloneq \cap$, $\top \coloneq \mathbb{N}_{[1,365]}$, $\bot \coloneq \emptyset$, where one may verify that $D = \langle 2^{\mathbb{N}_{[1,365]}}, \cup, \cap, \mathbb{N}_{[1,365]}, \emptyset \rangle$ is indeed an idempotent, commutative semi-ring.

\subsection{Deductive Knowledge}\label{app:deductive}

We provide some formal definitions for concepts relating to deductive knowledge, starting with the notion of an interpretation for a graph. We then describe some logical formalisms by which reasoning can be conducted over graphs, describing rules and Description Logics.

\subsubsection{Graph interpretations}\label{sec:interpretation}

A graph interpretation -- or simply interpretation -- captures the assumptions under which the semantics of a graph can be defined. We define interpretations for directed edge-labelled graphs, though the notion extends naturally to other graph models.

\begin{definition}[Graph interpretation]
A \emph{(graph) interpretation} $I$ is defined as a pair $I \coloneq (\Gamma,\inp{\cdot})$ where $\Gamma = (V_\Gamma,E_\Gamma,L_\Gamma)$ is a (directed edge-labelled) graph called the \emph{domain graph} and $\inp{\cdot} : \con \rightarrow V_\Gamma \cup L_\Gamma$ is a partial mapping from constants to terms in the domain graph. 
\end{definition}

We denote the domain of the mapping $\inp{\cdot}$ by $\textrm{dom}(\inp{\cdot})$. For interpretations under the UNA, the mapping $\inp{\cdot}$ is required to be injective, while  with no UNA (NUNA), no such requirement is necessary. Interpretations that \textit{satisfy} a graph are then said to be \textit{models} of that graph. We first define this notion for a base case that ignores ontological features.

\begin{definition}[Graph models]
Let $G \coloneq (V,E,L)$ be a directed edge-labelled graph. An interpretation $I \coloneq (\Gamma,\inp{\cdot})$ \textit{satisfies} $G$ if and only if the following hold:
\begin{itemize}
\item $V \cup L \subseteq \textrm{dom}(\inp{\cdot})$; 
\item for all $v \in V$, it holds that $\inp{v} \in V_\Gamma$; 
\item for all  $l \in L$, it holds that $\inp{l} \in L_\Gamma$; and 
\item for all $(u,l,v) \in E$, it holds that $(\inp{u},\inp{l},\inp{v}) \in E_\Gamma$. 
\end{itemize}

\noindent
If $I$ \textit{satisfies} $G$ we call $I$ a \emph{(graph) model} of $G$.
\end{definition}

\smallskip
Next we define models under semantics conditions (e.g., of ontology features).

\begin{definition}[Semantic condition] Let $2^G$ denote the set of all (directed edge-labelled) graphs. A \emph{semantic condition} is a mapping $\phi : 2^{G} \rightarrow \{ \text{true}, \text{false} \}$. An interpretation $I \coloneq (\Gamma,\inp{\cdot})$ is a model of $G$ under $\phi$ if and only if $I$ is a model of $G$ and $\phi(\Gamma)$. Given a set of semantic conditions $\Phi$, we say that $I$ is a model of $G$ if and only if $I$ is a model of $G$ and for all $\phi \in \Phi$, $\phi(\Gamma)$ is true.
\end{definition}

We do not restrict the language used to define semantic conditions, but, for example, we can define the \textsc{Has Value} semantic condition of Table~\ref{tab:ontClass} in FOL as follows:
\[ \forall c, p, y \Big( \big(  \Gamma(c,\gielab{prop},p) \wedge \Gamma(c,\gielab{value},y) \big) \leftrightarrow  \forall x \big( \Gamma(x,\gielab{type},c) \leftrightarrow \Gamma(x,p,y)  \big) \Big) \]
\noindent Here we overload $\Gamma$ as a ternary predicate to capture the edges of $\Gamma$. The above FOL formula defines an if-and-only-if version of the semantic condition for \textsc{Has Value}. The other semantic conditions enumerated in Tables~\ref{tab:ontEqIneq}--\ref{tab:ontClass} can be defined in a similar way~\cite{SchneiderS11}.\footnote{Note that although these tables consider axioms originating in the data graph, it suffices to check their image in the domain graph since $I$ only satisfies $G$ if the edges of $G$ defining the axioms are reflected in $I$.} 

Finally, we can define entailment considering such semantic conditions.

\begin{definition}[Graph entailment]\label{def:ent} Letting $G_1$ and $G_2$ denote two (directed edge-labelled) graphs, and $\Phi$ a set of semantic conditions, we say that \emph{$G_1$ entails $G_2$ under $\Phi$} -- denoted $G_1 \models_\Phi G_2$ -- if and only if any model of $G_1$ under $\Phi$ is also a model of $G_2$ under $\Phi$.
\end{definition}

An example of entailment is discussed in Section~\ref{sec:ontSemantics}. Note that in a slight abuse of notation, we may simply write $G \models_\Phi (s,p,o)$ to denote that $G$ entails the edge $(s,p,o)$ under $\Phi$.
\medskip

Under OWA, entailment is as defined as given in Definition~\ref{def:ent}. Under CWA, we make the additional assumption that if $G \not\models_\Phi e$, where $e$ is an edge (strictly speaking, a \textit{positive} edge), then $G \models_\Phi \neg e$; in other words, under CWA we assume that any (positive) edges that $G$ does not entail under $\Phi$ can be assumed false according to $G$ and $\Phi$.\footnote{In FOL, the CWA only applies to positive \textit{facts}, whereas edges in a graph can be used to represent other FOL formulae. If one wished to maintain FOL-compatibility under CWA, additional restrictions on the types of edge $e$ may be needed.}

\subsubsection{Rules}\label{app:rules}

Given a graph pattern $Q$ -- be it a directed edge-labelled graph pattern per Definition~\ref{def:delgp} or a property graph pattern per Definition~\ref{def:pgp} -- recall that $\var(Q)$ denotes the variables appearing in $Q$. We can now define the notion of a rule for graphs.

\begin{definition}[Rule] A \emph{rule} is a pair $R \coloneqq (B,H)$ such that $B$ and $H$ are graph patterns and $\var(H) \subseteq B$. We call $B$ the \textit{body} of the rule while we call $H$ the \textit{head} of the rule. 
\end{definition} 

This definition of a rule applies for directed edge-labelled graphs and property graphs by considering the corresponding type of graph pattern. The head is considered to be a conjunction of edges. Given a graph $G$, a rule is \textit{applied} by computing the mappings from the body to the graph and then using those mappings to substitute the variables in $H$. The restriction $\var(H) \subseteq B$ ensures that the results of this substitution is a graph, with no variables in $H$ left unsubstituted.
\medskip

\begin{definition}[Rule application] Given a rule $R = (B,H)$ and a graph $G$, we define the \emph{application of $R$ over $G$} as the graph $R(G) \coloneqq \bigcup_{\mu \in B(G)} \mu(H)$.
\end{definition} 

Given a set of rules $\mathcal{R} \coloneqq \{ R_1, \ldots, R_n \}$ and a knowledge graph $G$, towards defining the set of inferences given by the rules over the graph, we denote by  $\mathcal{R}(G) \coloneqq \bigcup_{R \in \mathcal{R}} R(G)$ the union of the application of all rules of $\mathcal{R}$ over $G$, and we denote by $\mathcal{R}^+(G) \coloneqq \mathcal{R}(G) \cup G$ the extension of $G$ with respect to the application of $\mathcal{R}$. Finally, we denote by $\mathcal{R}^k(G)$ (for $k \in \mathbb{N^+}$) the recursive application of $\mathcal{R}^+(G)$, where $\mathcal{R}^1(G) \coloneqq \mathcal{R}^+(G)$, and $\mathcal{R}^{i+1}(G) \coloneqq \mathcal{R}^+(\mathcal{R}^{i}(G))$. We are now ready to define the \textit{least model}, which captures the inferences possible for $\mathcal{R}$ over $G$.

\begin{definition}[Least model] The \emph{least model of $\mathcal{R}$ over $G$} is defined as $\mathcal{R}^*(G) \coloneq \bigcup_{k\in \mathbb{N}}(R^k(G))$.
\end{definition} 

At some point $R^{k'}(G) = R^{k'+1}(G)$: the rule applications reach a fixpoint and we have the least model. Once the least model $\mathcal{R}^*(G)$ is computed, the entailed data can be treated as any other data. 
\medskip

Rules can be used to support graph entailments of the form $G_1 \models_\Phi G_2$. We say that a set of rules $\mathcal{R}$ is \textit{correct} for $\Phi$ if, for any graph $G$, $G \models_\Phi \mathcal{R}^*(G)$. We say that $\mathcal{R}$ is \textit{complete} for $\Phi$ if, for any graph $G$, there does not exist an edge $e$ such that $G \models_\Phi e$ and $e \not\in \mathcal{R}^*(G)$. Table~\ref{tab:rulesRdfs} exemplifies a correct (but incomplete) set of rules for the semantic conditions laid out by the RDFS standard~\cite{RDFS}. 

Alternatively, rules can be directly specified in a rule language such as Notation3 (N3)~\cite{n3}, Rule Interchange Format (RIF)~\cite{rif}, Semantic Web Rule Language (SWRL)~\cite{swrl}, or SPARQL Inferencing Notation (SPIN)~\cite{spin}. Languages such as SPIN represent rules as graphs, allowing the rules of a knowledge graph to be embedded in the data graph. Taking advantage of this fact, we can then consider a form of graph entailment $G_1 \cup \gamma(\mathcal{R}) \models_\Phi G_2$, where by $\gamma(\mathcal{R})$ we denote the graph representation of rules $\mathcal{R}$. If the set of rules $\mathcal{R}$ is correct and complete for $\Phi$, we may simply write $G_1 \cup \gamma(\mathcal{R}) \models G_2$, indicating that $\Phi$ captures the same semantics for $\gamma(\mathcal{R})$ as applying the rules in $\mathcal{R}$; formally, $G_1 \cup \gamma(\mathcal{R}) \models \mathcal{R}(G_1 \cup \gamma(\mathcal{R}))$ and there does not exist an edge $e$ such that $G_1 \cup \gamma(\mathcal{R}) \models e$ but $e \not\in \mathcal{R}^*(G_1 \cup \gamma(\mathcal{R}))$. This allows us to view rules as another form of graph entailment.

\subsubsection{Description Logics}\label{sec:dlformal}

Table~\ref{tab:dlsem} provides definitions for all of the constructs typically found in Description Logics. The syntax column denotes how the construct is expressed in DL. A DL knowledge base then consists of an A-Box, a T-Box, and an R-Box. 

\begin{definition}[DL knowledge base]
	A \emph{DL knowledge base} $\mathsf{K}$ is defined as a tuple $(\mathsf{A},\mathsf{T},\mathsf{R})$, where $\mathsf{A}$ is the \emph{A-Box}: a set of assertional axioms; $\mathsf{T}$ is the \emph{T-Box}: a set of class (aka concept/terminological) axioms; and $\mathsf{R}$ is the \emph{R-Box}: a set of relation (aka property/role) axioms.
\end{definition}

\noindent
The semantics column defines the meaning of axioms using \textit{interpretations}. These interpretations are typically defined in a slightly different way to those previously defined for graphs, though the idea is roughly the same.

\begin{definition}[DL interpretation]
A \textit{DL interpretation} $I$ is defined as a pair $(\inpdom,\inp{\cdot})$, where $\inpdom$ is the \textit{interpretation domain}, and $\inp{\cdot}$ is the \textit{interpretation function}. The interpretation domain is a set of individuals. The interpretation function accepts a definition of either an individual $a$, a class $C$, or a relation $R$, mapping them, respectively, to an element of the domain ($\inp{a} \in \inpdom$), a subset of the domain ($\inp{C} \subseteq \inpdom$), or a set of pairs from the domain ($\inp{R} \subseteq \inpdom \times \inpdom$).
\end{definition}

An interpretation $I$ \textit{satisfies} a knowledge-base $\mathsf{K}$ if and only if, for all of the syntactic axioms in $\mathsf{K}$, the corresponding semantic conditions in Table~\ref{tab:dlsem} hold for $I$. In this case, we call $I$ a \textit{model} of $\mathsf{K}$. 
\medskip

As an example, for $\mathsf{K} \coloneq  (\mathsf{A},\mathsf{T},\mathsf{R})$, let:

\begin{itemize}
	\item $\mathsf{A} \coloneq \{ \texttt{City(Arica)}, \texttt{City(Santiago)}, \texttt{flight(Arica,Santiago)} \}$;
	\item $\mathsf{T} \coloneq \{ \texttt{City} \sqsubseteq \texttt{Place}, \exists\texttt{flight}.\top \sqsubseteq \exists\texttt{nearby}.\texttt{Airport} \} $;
	\item $\mathsf{R} \coloneq \{ \texttt{flight} \sqsubseteq \texttt{connectsTo} \} $.
\end{itemize}

\noindent
For $I = (\inpdom,\inp{\cdot})$, let:

\begin{itemize} 
	\item $\inpdom \coloneq \{ \Arica, \Santiago, \AricaAirport \}$;
	\item $\inp{\texttt{Arica}} \coloneq \Arica$, $\inp{\texttt{Santiago}} \coloneq \Santiago$, $\inp{\texttt{AricaAirport}} \coloneq \AricaAirport$;
	\item $\inp{\texttt{City}} \coloneq \{ \Arica, \Santiago \}$, $\inp{\texttt{Airport}} \coloneq \{ \AricaAirport \}$; 
	\item $\inp{\texttt{flight}} \coloneq \{ (\Arica, \Santiago) \}$, $\inp{\texttt{connectsTo}} \coloneq \{ (\Arica, \Santiago) \}$, $\inp{\texttt{sells}} \coloneq \{ (\AricaAirport, \Coffee) \}$.
\end{itemize}

\noindent
The interpretation $I$ is not a model of $\mathsf{K}$ since it does not have that $\Arica$ is \texttt{nearby} some \texttt{Airport}, nor that $\Arica$ and $\Santiago$ are in the class \texttt{Place}. However, if we \textit{extend} $I$ with the following:

\begin{itemize} 
	\item $\inp{\texttt{Place}} \coloneq \{ \Arica, \Santiago \}$; \item $\inp{\texttt{nearby}} \coloneq \{ (\Arica,\AricaAirport) \}$.
\end{itemize}

\noindent

Now $I$ is a model of $\mathsf{K}$. Note that although $\mathsf{K}$ does not imply that $\texttt{sells(Arica,coffee)}$ while $I$ indicates that $\AricaAirport$ \texttt{sells} $\Coffee$, $I$ is still a model of $\mathsf{K}$ since $\mathsf{K}$ is not assumed to be a complete description of the world, as per the Open World Assumption.

Finally, the notion of a model gives rise to the key notion of entailment. 

\begin{definition}
	Given two DL knowledge bases $\mathsf{K}_1$ and $\mathsf{K}_2$, we define that $\mathsf{K}_1$ entails $\mathsf{K}_2$, denoted $\mathsf{K}_1 \models \mathsf{K}_2$, if and only if any model of $\mathsf{K}_1$ is a model of $\mathsf{K}_2$.
\end{definition}

The entailment relation tells us which knowledge bases hold as a logical consequence of which others: if all models of $\mathsf{K}_1$ are also models of $\mathsf{K}_2$ then, intuitively speaking, $\mathsf{K}_2$ says nothing new over $\mathsf{K}_1$. For example, let  $\mathsf{K}_1$ denote the knowledge base $\mathsf{K}$ from the previous example, and define a second knowledge base with one assertion: $\mathsf{K}_2 \coloneq ( \{ \texttt{connectsTo}(\texttt{Arica},\texttt{Santiago}) \}, \{\}, \{\} )$. Though $\mathsf{K}_1$ does not assert this axiom, it does entail $\mathsf{K}_2$: to be a model of $\mathsf{K}_2$, an interpretation must have that $(\inp{\texttt{Arica}},\inp{\texttt{Santiago}}) \in \inp{\texttt{connectsTo}}$, but this must also be the case for any interpretation that satisfies $\mathsf{K}_1$ since it must have that $(\inp{\texttt{Arica}},\inp{\texttt{Santiago}}) \in \inp{\texttt{flight}}$ and $\inp{\texttt{flight}} \subseteq \inp{\texttt{connectsTo}}$.

Unfortunately, the problem of deciding entailment for knowledge bases expressed in the DL composed of the unrestricted use of all of the axioms of Table~\ref{tab:dlsem} combined is undecidable. We could, for example, reduce instances of the Halting Problem to such entailment. Hence DLs in practice restrict use of the features listed in Table~\ref{tab:dlsem}. Different DLs then apply different restrictions, implying different trade-offs for expressivity and the complexity of the entailment problem. Most DLs are founded on one of the following base DLs (we use indentation to denote derivation):

\begin{itemize}
	\item[$\mathcal{ALC}$] (\textit{$\mathcal{A}$ttributive $\mathcal{L}$anguage with $\mathcal{C}$omplement}~\cite{Schmidt-SchaussS91}), supports atomic classes, the top and bottom classes, class intersection, class union, class negation, universal restrictions and existential restrictions. Relation and class assertions are also supported.
	\begin{itemize}
		\item[$\mathcal{S}$] extends $\mathcal{ALC}$ with transitive closure.
	\end{itemize}
\end{itemize}

\noindent
These base languages can be extended as follows:

\begin{itemize}
	\item[$\mathcal{H}$] adds relation inclusion.
	\begin{itemize}
		\item[$\mathcal{R}$] adds (limited) complex relation inclusion, as well as relation reflexivity, relation irreflexivity, relation disjointness and the universal relation.
	\end{itemize}
	\item[$\mathcal{O}$] adds (limited) nomimals.
	\item[$\mathcal{I}$] adds inverse relations.
	\item[$\mathcal{F}$] adds (limited) functional properties.
	\begin{itemize}
	\item[$\mathcal{N}$] adds (limited) number restrictions (subsuming $\mathcal{F}$ given $\top$).
	\begin{itemize}
	\item[$\mathcal{Q}$] adds (limited) qualified number restrictions (subsuming $\mathcal{N}$ given $\top$).
	\end{itemize}
	\end{itemize}
\end{itemize}

\noindent We use ``(limited)'' to indicate that such features are often only allowed under certain restrictions to ensure decidability; for example, complex relations (chains) typically cannot be combined with cardinality restrictions. DLs are then typically named per the following scheme, where $[a|b]$ denotes an alternative between $a$ and $b$ and $[c][d]$ denotes a concatenation $cd$:
\[ [\mathcal{ALC}|\mathcal{S}][\mathcal{H}|\mathcal{R}][\mathcal{O}][\mathcal{I}][\mathcal{F}|\mathcal{N}|\mathcal{Q}] \]
Examples include $\mathcal{ALCO}$, $\mathcal{ALCHI}$, $\mathcal{SHIF}$, $\mathcal{SROIQ}$, etc. These languages often apply additional restrictions on class and property axioms to ensure decidability, which we do not discuss here. For further details on Description Logics, we refer to the recent book by~\citet{BaaderHLS17}.

\begin{table}[t]
	\centering
	\caption{Description Logic semantics (such that $x, y, z, \inp{a}, \inp{a_1}, \ldots \inp{a_n}, \inp{b}$ are in $\inpdom$)}\label{tab:dlsem}
	\resizebox{\textwidth}{!}{
		\begin{tabular}{lll}
			\toprule
			\textbf{Name} & \textbf{Syntax} & \textbf{Semantics} ($\inp{\cdot}$)\\
			\midrule
			\multicolumn{3}{c}{\textsc{Class Definitions}} \\[1ex]
			Atomic Class & $A$ & \inp{A} (a subset of \inpdom) \\
			Top Class & $\top$ & \inpdom \\
			Bottom Class & $\bot$ & $\emptyset$\\
			Class Negation & $\neg C$ & $\inpdom \setminus \inp{C}$ \\
			Class Intersection & $C \sqcap D$ & $\inp{C} \cap \inp{D}$  \\
			Class Union & $C \sqcup D$ & $\inp{C} \cup \inp{D}$  \\
			Nominal & $\{ a_1, ..., a_n \}$ & $\{ \inp{a_1}, ..., \inp{a_n} \}$ \\
			Existential Restriction & $\exists R.C$ & $\{ x \mid \exists y : (x,y) \in \inp{R}\text{ and }y \in \inp{C} \}$ \\
			Universal Restriction & $\forall R.C$ & $\{ x \mid \forall y : (x,y) \in \inp{R}\text{ implies }y \in \inp{C} \}$ \\
			Self Restriction & $\exists R.\textsf{Self}$ & $\{ x \mid (x,x) \in \inp{R} \}$  \\
			Number Restriction & $\card\,n\,R$ (where $\card \in \{\geq, \leq, = \}$) & $\{ x \mid \#\{ y : (x,y) \in \inp{R} \} \card n \}$ \\
			Qualified Number Restriction & $\card\,n\,R.C$ (where $\card \in \{\geq, \leq, = \}$) & $\{ x \mid \#\{ y : (x,y) \in \inp{R}\text{ and }y \in \inp{C} \} \card n \}$ \\
			\midrule
			\multicolumn{3}{c}{\textsc{Class Axioms} (T-Box)} \\[1ex]
			Class Inclusion & $C \sqsubseteq D$ & $\inp{C} \subseteq \inp{D}$ \\
			\midrule
			\multicolumn{3}{c}{\textsc{Relation Definitions}} \\[1ex]
			Relation 	& $R$ & $\inp{R}$ (a subset of $\inpdom \times \inpdom$) \\
			Inverse Relation & $R^{-}$ & $\{ (y,x) \mid (x,y) \in \inp{R} \}$\\
			Universal Relation & $\textsf{U}$ &  $\inpdom \times \inpdom$\\
			\midrule
			\multicolumn{3}{c}{\textsc{Relation Axioms} (R-Box)} \\[1ex]
			Relation Inclusion & $R \sqsubseteq S$ & $\inp{R} \subseteq \inp{S}$  \\
			Complex Relation Inclusion & $R_1 \circ ... \circ R_n \sqsubseteq S$ & $\inp{R_1} \circ ... \circ \inp{R_n}  \subseteq \inp{S}$\\
			Transitive Relations & $\textsf{Trans}(R)$ & $\inp{R} \circ \inp{R}  \subseteq \inp{R}$ \\
			Functional Relations & $\textsf{Func}(R)$ & $\{ (x,y), (x,z) \} \subseteq \inp{R} $implies $y = z$ \\
			Reflexive Relations & $\textsf{Ref}(R)$ & for all $x : (x,x) \in \inp{R}$   \\
			Irreflexive Relations & $\textsf{Irref}(R)$ &  for all $x : (x,x) \not\in \inp{R}$ \\
			Symmetric Relations & $\textsf{Sym}(R)$ & $\inp{R} = \inp{(R^{-})}$  \\
			Asymmetric Relations & $\textsf{Asym}(R)$ & $\inp{R} \cap \inp{(R^{-})} = \emptyset$ \\
			Disjoint Relations & $\textsf{Disj}(R,S)$ & $\inp{R} \cap \inp{S} = \emptyset$  \\
			\midrule
			\multicolumn{3}{c}{\textsc{Assertional Definitions}} \\[1ex]
			Individual & $a$ & $\inp{a}$ \\
			\midrule
			\multicolumn{3}{c}{\textsc{Assertional Axioms} (A-Box)} \\[1ex]
			Relation Assertion & $R(a,b)$ & $(\inp{a},\inp{b}) \in \inp{R}$ \\
			Negative Relation Assertion & $\neg R(a,b)$ & $(\inp{a},\inp{b}) \not\in \inp{R}$ \\
			Class Assertion & $C(a)$ & $\inp{a} \in \inp{C}$ \\
			Equality & $ a = b $ & $\inp{a} = \inp{b}$ \\
			Inequality & $ a \neq b $ & $\inp{a} \neq \inp{b}$ \\
			\bottomrule
		\end{tabular}
	}
\end{table}

As mentioned in the body of the survey, DLs have been very influential in the definition of OWL, where the OWL 2 DL fragment (roughly) corresponds to the DL $\mathcal{SROIQ}$. For example, the axiom \gedge[arrin][\fhgap]{venue}{domain}{Event} in OWL can be translated to $\exists \texttt{venue}.\top \sqsubseteq \texttt{Event}$, meaning that the class of individuals with some value for \texttt{venue} (in any class) is a sub-class of the class \texttt{Event}. We leave other translations from the OWL axioms of Tables~\ref{tab:ontEqIneq}--\ref{tab:ontClass} to DL as an exercise.\footnote{Though not previously mentioned, OWL defines classes \texttt{Thing} and \texttt{Nothing} that correspond to $\top$ and $\bot$, respectively.} Note, however, that axioms like \gedge[arrin][\fhgap]{sub-taxon of}{subp. of}{subc. of} -- which given a graph such as \begin{tikzpicture}[baseline=-3pt]
\node[iri,compact](y){Fred};

\node[iri,compact,right=\thgap of y](hs){Homo sapiens}
edge[arrin] node[lab] {type} (y);

\node[iri,compact,right=1.5\fhgap of hs](h){Hominini}
edge[arrin] node[lab] {sub-taxon of} (hs);
\end{tikzpicture} entails the edge \gedge[arrin][\thgap]{Fred}{type}{Hominini} -- cannot be expressed in DL:  ``$\texttt{subTaxonOf} \sqsubseteq\,\,\sqsubseteq$'' is not syntactically valid. Hence only a subset of graphs can be translated into well-formed DL ontologies; we refer to the OWL standard for details~\cite{OWL2}.

\subsection{Inductive Knowledge}\label{app:inductive}

We provide further discussion and formal definitions relating to graph parallel frameworks, knowledge graph embeddings, and graph neural networks, as discussed in Section~\ref{sec:inductive}.

\subsubsection{Graph parallel frameworks}\label{app:gpfs}

Before defining a graph parallel framework, in the interest of generality, we first define a directed graph labelled with feature vectors, which captures the type of input that such a framework can accept, with vectors assigned to both nodes and edges.

\begin{definition}[Directed vector-labelled graph]\label{def:dvlg}
We define a \emph{directed vector-labelled graph} $G = (V,E,F,\lambda)$, where $V$ is a set of nodes, $E \subseteq V \times V$ is a set of edges, $F$ is a set of feature vectors, and $\lambda : V \cup E \rightarrow F$ labels each node and edge with a feature vector.
\end{definition}

A directed-edge labelled graph or a property graph may be encoded as a directed vector-labelled graph in a number of ways, depending on the application. The type of node and/or a selection of its attributes may be encoded in the node feature vectors, while the label of an edge and/or a selection of its attributes may be encoded in the edge feature vector (including, for example, weights applied to edges). Typically node feature vectors will all have the same dimensionality, as will edge feature vectors. The directed vector-labelled graph can thus be seen as defining the initial state and features that will be used as input for the graph parallel framework.

\begin{example}\label{ex:dvlg}
We define a directed vector-labelled graph in preparation for later computing PageRank using a graph parallel framework. Let $G = (V,E,L)$ denote a directed edge-labelled graph. Let $|E(u)|$ denote the outdegree of node $u \in V$. We then initialise a directed vector-labelled graph $G' = (V,E',F,\lambda)$ such that $E' = \{ (x,z) \mid \exists y : (x,y,z)\in E  \}$, and for all $u \in V$, we define $\lambda(u) \coloneq \begin{bmatrix} \frac{1}{|V|} \\ |E'(u)| \\  |V| \end{bmatrix}$, and $\lambda(u,v) \coloneq \begin{bmatrix} \, \end{bmatrix}$, with $F \coloneq \{ \lambda(u) \mid u \in V \} \cup \{\lambda(u,v) \mid (u,v) \in E'  \}$, assigning each node a vector containing its initial PageRank score, the outdegree of the node, and the number of nodes in the graph. Conversely, edge-vectors are not used in this case. 
\end{example}

We are now ready to define a graph parallel framework operating over a directed vector-labelled graph. In the following we use $\{\!\!\{ \cdot \}\!\!\}$ to denote a multiset (an unordered set preserving duplicates), $2^{S \rightarrow \mathbb{N}}$ to denote the set of all multisets containing (only) elements from the set $S$, and $\mathbb{R}^a$ to denote the set of all vectors of dimension $a$ (i.e., the set of all vectors containing $a$ real-valued elements).

\begin{definition}[Graph parallel framework]\label{def:gpf}
A \emph{graph parallel framework} (\emph{GPF}) is a triple of functions $\mathfrak{G} \coloneq (\textsc{Msg},\textsc{Agg},\textsc{End})$ such that (with $a, b, c \in \mathbb{N}$):

\begin{itemize}
\item $\textsc{Msg}: \mathbb{R}^a \times \mathbb{R}^b \rightarrow \mathbb{R}^c$
\item $\textsc{Agg}: \mathbb{R}^a \times 2^{\mathbb{R}^c  \rightarrow \mathbb{N}} \rightarrow \mathbb{R}^a$
\item $\textsc{End}: 2^{\mathbb{R}^a \rightarrow \mathbb{N}} \rightarrow \{ \mathrm{true}, \mathrm{false} \}$
\end{itemize}
\end{definition}

The function $\textsc{Msg}$ defines what message (i.e., vector) must be passed from a node to a neighbouring node along a particular edge, given the current feature vectors of the node and the edge; the function $\textsc{Agg}$ is used to compute a new feature vector for a node, given its previous feature vector and incoming messages; the function $\textsc{End}$ defines a condition for termination of vector computation. The integers $a$, $b$ and $c$ denote the dimensions of node feature vectors, edge feature vectors, and message vectors, respectively; we assume that $a$ and $b$ correspond with the dimensions of input feature vectors for nodes and edges. Given a GPF $\mathfrak{G} = (\textsc{Msg}, \textsc{Agg}, \textsc{End})$, a directed vector-labelled graph $G = (V, E, F, \lambda)$, and a node $u \in V$, we define the output vector assigned to node $u$ in $G$ by $\mathfrak{G}$  (written $\mathfrak{G}(G, u)$) as follows. First let $\mathbf{n}_u^{(0)} \coloneq \lambda(u)$. For all $i\geq 1$, let: 

\begin{align*}
 M_u^{(i)} & \coloneq  \left\{\!\!\left\{ \textsc{Msg}\left(\mathbf{n}_v^{(i-1)},\lambda(v,u)\right) \bigl\lvert\, (v,u) \in E \right\}\!\!\right\} \\
 \mathbf{n}_{u}^{(i)} & \coloneq \textsc{Agg}\left(\mathbf{n}_u^{(i-1)},M_u^{(i)}\right)
\end{align*}

\noindent
If $j$ is the smallest integer for which $\textsc{End}(\{\!\!\{ \mathbf{n}_u^{(j)} \mid u \in V \}\!\!\})$ is true, then $\mathfrak{G}(G, u) \coloneq \mathbf{n}_u^{(j)}$.

This particular definition assumes that vectors are dynamically computed for nodes, and that messages are passed only to outgoing neighbours, but the definitions can be readily adapted to consider dynamic vectors for edges, or messages being passed to incoming neighbours, etc. We now provide an example instantiating a GPF to compute PageRank over a directed graph.

\begin{example}
We take as input the directed vector labelled graph $G' = (V,E,F,\lambda)$ from Example~\ref{ex:dvlg} for a PageRank GPF. First we define the messages passed from $u$ to $v$:
\[
\textsc{Msg}\left(\mathbf{n}_v,\lambda(v,u)\right) \coloneq \begin{bmatrix}
\frac{d(\mathbf{n}_{v})_1}{(\mathbf{n}_{v})_2}\\
\end{bmatrix}
\]
where $d$ denotes PageRank's constant dampening factor (typically $d \coloneq 0.85$) and $(\mathbf{n}_{v})_k$ denotes the $k$\textsuperscript{th} element of the $\mathbf{n}_{v}$ vector. In other words, $v$ will pass to $u$ its PageRank score multiplied by the dampening factor and divided by its degree (we do not require $\lambda(v,u)$ in this particular example). Next we define the function for $u$ to aggregate the messages it receives from other nodes:
\[
\textsc{Agg}\left(\mathbf{n}_u,M_u\right) \coloneq \begin{bmatrix} \frac{1 - d}{(\mathbf{n}_{u})_3} + \sum_{\mathbf{m} \in M_u}(\mathbf{m})_1 \\ (\mathbf{n}_{u})_2 \\ (\mathbf{n}_{u})_3 \\
\end{bmatrix}
\]
Here, we sum the scores received from other nodes along with its share of rank from the dampening factor, copying over the node's degree and the total number of nodes for future use. Finally, there are a number of ways that we could define the termination condition; here we simply define: 
\[ \textsc{End}(\{\!\!\{ \mathbf{n}_u^{(i)} \mid u \in V \}\!\!\}) \coloneq (i \geq \textsf{z}) \]
where $\textsf{z}$ is a fixed number of iterations, at which point the process stops.
\end{example}

We may note in this example that the total number of nodes is duplicated in the vector for each node of the graph. Part of the benefit of GPFs is that only local information in the neighbourhood of the node is required for each computation step. In practice, such frameworks may allow additional features, such as global computation steps whose results are made available to all nodes~\cite{DBLP:conf/sigmod/MalewiczABDHLC10}, operations that dynamically modify the graph~\cite{DBLP:conf/sigmod/MalewiczABDHLC10}, etc.

\subsubsection{Knowledge graph embeddings}\label{app:gembeddings}

As discussed in Section~\ref{ssec:embeddings}, knowledge graph embeddings represent graphs in a low-dimensional numeric space.\footnote{To the best of our knowledge, the term ``\textit{knowledge graph embedding}'' was coined by \citet{wang2014knowledge} in order to distinguish the case from a ``graph embedding'' that considers a single relation (i.e., an undirected or directed graph). Earlier papers rather used the phrase ``\textit{multi-relational data}''~\cite{nickel2013tensor,bordes2013translating,GlorotBWB13}.} Before defining the key notions, we introduce mathematical objects related to tensor calculus, on which embeddings heavily rely.

\begin{definition}[Vector, matrix, tensor, order, mode]
For any positive integer $a$, a \emph{vector} of dimension $a$ is a family of real numbers indexed by integers in $\{1, \ldots, a\}$. For $a$ and $b$ positive integers, an $(a,b)$-matrix is a family of real numbers indexed by pairs of integers in $\{1, \ldots, a\} \times \{1, \ldots, b\}$. A tensor is a family of real numbers indexed by a finite sequence of integers such that there exist positive numbers $a_1, \ldots, a_n$ such that the indices are all the tuples of numbers in $\{1, \ldots, a_1\} \times \ldots \times \{1, \ldots, a_n\}$. The number $n$ is called the \emph{order} of the tensor, the subindices $i\in \{1, \ldots, n\}$ indicate the \emph{mode} of a tensor, and each $a_i$ defines the dimension of the $i$\textsuperscript{th} mode. A 1-order tensor is a vector and a 2-order tensor is a matrix. We denote the set of all tensors as $\mathbb{T}$.
\end{definition}

For specific dimensions $a_1,\ldots,a_n$ of modes, a tensor is an element of $(\cdots(\mathbb{R}^{a_1})^{\ldots})^{a_n}$ but we write $\mathbb{R}^{a_1,\ldots,a_n}$ to simplify the notation. We use lower-case bold font to denote vectors ($\mathbf{x} \in \mathbb{R}^a$), upper-case bold font to denote matrices ($\mathbf{X} \in \mathbb{R}^{a,b}$) and calligraphic font to denote tensors ($\mathcal{X} \in \mathbb{R}^{a_1,\ldots,a_n}$). 
\medskip

Now we are ready to abstractly define knowledge graph embeddings.

\begin{definition}[Knowledge graph embedding]
Given a directed edge-labelled graph $G = (V,E,L)$, a \emph{knowledge graph embedding of $G$} is a pair of mappings $(\varepsilon,\rho)$ such that $\varepsilon : V \rightarrow \mathbb{T}$ and $\rho : L \rightarrow \mathbb{T}$.
\end{definition}

In the most typical case, $\varepsilon$ and $\rho$ map nodes and edge-labels, respectively, to vectors of fixed dimension. In some cases, however, they may map to matrices. Given this abstract notion of a knowledge graph embedding, we can then define a plausibility score.

\begin{definition}[Plausibility]
A \emph{plausibility scoring function} is a partial function $\phi : \mathbb{T} \times \mathbb{T} \times \mathbb{T} \rightarrow \mathbb{R}$. Given a directed edge-labelled graph $G = (V,E,L)$, an edge $(s,p,o) \in V \times L \times V$, and a knowledge graph embedding $(\varepsilon,\rho)$ of $G$, the plausibility of $(s,p,o)$ is given as $\phi(\varepsilon(s),\rho(p),\varepsilon(o))$.
\end{definition}

Edges with higher scores are considered to be more plausible. Given a graph $G = (V,E,L)$, we assume a set of positive edges $E^+$ and a set of negative edges $E^{-}$. Positive edges are often simply the edges in the graph: $E^+ \coloneq E$. Negative edges use the vocabulary of $G$ (i.e., $E^- \subseteq V \times L \times V$) and typically are defined by taking edges $(s,p,o)$ from $E$ and changing one of the terms of each edge -- most often, but not always, one of the nodes -- such that the edge is no longer in $E$. Given sets of positive and negative edges, and a plausibility scoring function, the objective is then to find the embedding that maximises the plausibility of edges in $E^+$ while minimising the plausibility of edges in $E^{-}$. Specific knowledge graph embeddings then instantiate the type of embedding considered and the plausibility scoring function in (a wide variety of) different ways. 

In Table~\ref{tab:kges}, we define the plausibility scoring function used by different models for knowledge graph embeddings, and further provide details of the types of embeddings considered. To simplify the definitions of embeddings given in Table~\ref{tab:kges}, we will use $\mathbf{e}_x$ to denote $\varepsilon(x)$ when it is a vector, and we will use $\mathbf{r}_y$ to denote $\rho(y)$ when it is a vector and $\mathbf{R}_y$ to denote $\rho(y)$ when it is a matrix. Some models use additional parameters (aka weights) that -- although they do not form part of the entity/relation embeddings -- are learnt to compute the plausibility score from the embeddings. We denote these as $\mathbf{v}$, $\mathbf{V}$, $\mathcal{V}$, $\mathbf{w}$, $\mathbf{W}$ $\mathcal{W}$ (for vectors, matrices or tensors). We use $d_e$ and $d_r$ to denote the dimensionality chosen for entity embeddings and relation embeddings, respectively. Often it is assumed that $d_e = d_r$, in which case we will write $d$. Sometimes weights may have their own dimensionality, which we denote $w$. The embeddings in Table~\ref{tab:kges} use a variety of operators on vectors, matrices and tensors. In the interest of keeping the discussion self-contained, we refer to the latter part of this section for definitions of these operators and other conventions used.

\begin{table}
\caption{Details for selected knowledge graph embeddings, including the plausibility scoring function $\phi(\varepsilon(s),\rho(p),\varepsilon(o))$ for edge \gedge[arrin][0.6cm]{$s$}{$p$}{$o$}, and other conditions applied} \label{tab:kges}
\footnotesize
\setlength{\tabcolsep}{2.5pt}
\begin{tabular}{lll}
\toprule
\multirow{1}{*}{\textbf{Model}} & \multirow{1}{*}{$\phi(\varepsilon(s),\rho(p),\varepsilon(o))$} & \textbf{Conditions} (for all $x \in V$, $y \in L$)  \\
\midrule
\multirow{1}{*}{TransE~\cite{bordes2013translating}} & \multirow{1}{*}{$- \|\mathbf{e}_s + \mathbf{r}_p - \mathbf{e}_o\|_q$} & $\mathbf{e}_x \in \mathbb{R}^{d}$, $\mathbf{r}_y \in \mathbb{R}^{d}$, $q \in \{1,2\}$, $\|\mathbf{e}_x\|_2  = 1$ \\
\midrule

\multirow{2}{*}{TransH~\cite{wang2014knowledge}} & \multirow{2}{*}{$-\|(\mathbf{e}_s - (\T{\mathbf{e}_s}\mathbf{w}_p)\mathbf{w}_p) + \mathbf{r}_p - (\mathbf{e}_o - (\T{\mathbf{e}_o} \mathbf{w}_p)\mathbf{w}_p)\|^{2}_{2}$} 
& $\mathbf{e}_x \in \mathbb{R}^{d}$, $\mathbf{r}_y \in \mathbb{R}^{d}$, $\mathbf{w}_y \in \mathbb{R}^d$, \\
& & $\|\mathbf{w}_y\|_2 = 1$ , $\frac{\T{\mathbf{w}_y} \mathbf{r}_y}{\|\mathbf{r}_y\|_2} \approx 0$, $\|\mathbf{e}_x\|_2  \leq 1$ \\
\midrule

\multirow{2}{*}{TransR~\cite{TransD}}& \multirow{2}{*}{$-\|\mathbf{W}_p\mathbf{e}_s + \mathbf{r}_p - \mathbf{W}_p\mathbf{e}_o\|^{2}_{2}$} & $\mathbf{e}_x \in \mathbb{R}^{d_e}$, $\mathbf{r}_y \in \mathbb{R}^{d_r}$, $\mathbf{W}_y \in \mathbb{R}^{d_r , d_e}$, \\
& & $\|\mathbf{e}_x\|_2 \leq 1$, $\|\mathbf{r}_y\|_2 \leq 1$, $\|\mathbf{W}_y\mathbf{e}_x\|_2 \leq 1$  \\
\midrule

\multirow{2}{*}{TransD~\cite{TransD}} & \multirow{2}{*}{$-\|(\mathbf{w}_p\otimes\mathbf{w}_s + \mathbf{I})\mathbf{e}_s + \mathbf{r}_p - (\mathbf{w}_p\otimes\mathbf{w}_o + \mathbf{I})\mathbf{e}_o\|^{2}_{2}$} & $\mathbf{e}_x \in \mathbb{R}^{d_e}$, $\mathbf{r}_y \in \mathbb{R}^{d_r}$,
$\mathbf{w}_x \in \mathbb{R}^{d_e}$, $\mathbf{w}_y \in \mathbb{R}^{d_r}$, \\
& & $\|\mathbf{e}_x\|_2 \leq 1$, $\|\mathbf{r}_y\|_2 \leq 1$, $\|(\mathbf{w}_y\otimes\mathbf{w}_x + \mathbf{I})\mathbf{e}_x\|_2 \leq 1$  \\
\midrule

\multirow{1}{*}{RotatE~\cite{SunDNT19}} &  \multirow{1}{*}{\multirow{1}{*}{$- \|\mathbf{e}_s \odot \mathbf{r}_p - \mathbf{e}_o\|_2$}} &  $\mathbf{e}_x \in \mathbb{C}^{d}$, $\mathbf{r}_y \in \mathbb{C}^{d}$, $\|\mathbf{r}_y\|_2 = 1$ \\
\midrule

\multirow{1}{*}{RESCAL~\cite{nickel2013tensor}} & \multirow{1}{*}{$\T{\mathbf{e}_s} \mathbf{R}_p \mathbf{e}_o$} & $\mathbf{e}_x \in \mathbb{R}^{d}$, $\mathbf{R}_y \in \mathbb{R}^{d , d}$,  $\|\mathbf{e}_x\|_2 \leq 1$, $\|\mathbf{R}_y\|_{2,2} \leq 1$ \\
\midrule

\multirow{1}{*}{DistMult~\cite{distmult}} & \multirow{1}{*}{$\T{\mathbf{e}_s} \D{\mathbf{r}_p} \mathbf{e}_o$} & $\mathbf{e}_x \in \mathbb{R}^{d}$, $\mathbf{r}_y \in \mathbb{R}^{d}$,  $\|\mathbf{e}_x\|_2 = 1$, $\|\mathbf{r}_y\|_2 \leq 1$ \\
\midrule

\multirow{1}{*}{HolE~\cite{NickelRP16}} & \multirow{1}{*}{$\T{\mathbf{r}_p} (\mathbf{e}_s \star \mathbf{e}_o)$} & $\mathbf{e}_x \in \mathbb{R}^{d}$, $\mathbf{r}_y \in \mathbb{R}^{d}$, $\|\mathbf{e}_x\|_2 \leq 1$, $\|\mathbf{r}_y\|_2 \leq 1$ \\
\midrule

\multirow{1}{*}{ComplEx~\cite{TrouillonWRGB16}} &  \multirow{1}{*}{$\mathrm{Re}(\T{\mathbf{e}_s} \D{\mathbf{r}_p} \overline{\mathbf{e}}_o)$} &  $\mathbf{e}_x \in \mathbb{C}^{d}$, $\mathbf{r}_y \in \mathbb{C}^{d}$, $\|\mathbf{e}_x\|_2 \leq 1$, $\|\mathbf{r}_y\|_2 \leq 1$ \\
\midrule

\multirow{2}{*}{SimplE~\cite{Kazemi018}} & \multirow{2}{*}{$\frac{\T{\mathbf{e}_s} \D{\mathbf{r}_p} \mathbf{w}_o + \T{\mathbf{e}_o} \D{\mathbf{w}_p} \mathbf{w}_s}{2}$} & $\mathbf{e}_x \in \mathbb{R}^{d}$, $\mathbf{r}_y \in \mathbb{R}^{d}$,  $\mathbf{w}_x \in \mathbb{R}^{d}$, $\mathbf{w}_y \in \mathbb{R}^{d}$,\\
& & $\|\mathbf{e}_x\|_2 \leq 1$, $\|\mathbf{w}_x\|_2 \leq 1$, $\|\mathbf{r}_y\|_2 \leq 1, \|\mathbf{w}_y\|_2 \leq 1$  \\
\midrule

\multirow{1}{*}{TuckER~\cite{BalazevicAH19a}} & \multirow{1}{*}{$\mathcal{W} \otimes_1 \T{\mathbf{e}_s} \otimes_2 \T{\mathbf{r}_p} \otimes_3 \T{\mathbf{e}_o}$} & $\mathbf{e}_x \in  \mathbb{R}^{d_e}$, $\mathbf{r}_y \in \mathbb{R}^{d_r}$, $\mathcal{W} \in \mathbb{R}^{d_e , d_r , d_e }$ \\
 \midrule

\multirow{2}{*}{SME Linear~\cite{GlorotBWB13}} & \multirow{2}{*}{$\T{(\mathbf{V}\mathbf{e}_s + \mathbf{V}'\mathbf{r}_p + \mathbf{v})} (\mathbf{W}\mathbf{e}_o + \mathbf{W}'\mathbf{r}_p + \mathbf{w})$} & $\mathbf{e}_x \in  \mathbb{R}^{d}$, $\mathbf{r}_y \in  \mathbb{R}^{d}$,  $\mathbf{v} \in \mathbb{R}^w$, $\mathbf{w} \in \mathbb{R}^w$, $\|\mathbf{e}_x\|_2 = 1$, \\
 & & $\mathbf{V} \in \mathbb{R}^{w , d},\mathbf{V}' \in \mathbb{R}^{w , d}, \mathbf{W} \in \mathbb{R}^{w , d}, \mathbf{W}' \in \mathbb{R}^{w , d}$\\
 \midrule
 
\multirow{2}{*}{SME Bilinear~\cite{GlorotBWB13}} & \multirow{2}{*}{$\T{((\mathcal{V} \otimes_3  \T{\mathbf{r}_p}) \mathbf{e}_s + \mathbf{v})}((\mathcal{W} \otimes_3  \T{\mathbf{r}_p}) \mathbf{e}_o + \mathbf{w})$} &  $\mathbf{e}_x \in  \mathbb{R}^{d}$, $\mathbf{r}_y \in  \mathbb{R}^{d}$, $\mathbf{v} \in \mathbb{R}^w$, $\mathbf{w} \in \mathbb{R}^w$, $\|\mathbf{e}_x\|_2 = 1$,\\
 & & $\mathcal{V} \in \mathbb{R}^{w , d , d}$, $\mathcal{W} \in \mathbb{R}^{w , d , d}$ \\
  \midrule

\multirow{3}{*}{NTN~\cite{socher2013reasoning}} &  \multirow{3}{*}{$\T{\mathbf{r}_p} \psi\left(\T{\mathbf{e}_s} \mathcal{W} \mathbf{e}_o + \mathbf{W} \begin{bmatrix}\mathbf{e}_s\\\mathbf{e}_o\end{bmatrix} + \mathbf{w}\right) $} & $\mathbf{e}_x \in  \mathbb{R}^{d}$, $\mathbf{r}_y \in  \mathbb{R}^{d}$, $\mathbf{w} \in \mathbb{R}^{w}$, $\mathbf{W} \in \mathbb{R}^{w , 2d}$, \\
 & & $\mathcal{W} \in \mathbb{R}^{d , w , d}$, $\|\mathbf{e}_x\|_2 \leq 1$, $\|\mathbf{r}_y\|_2 \leq 1$,\\
 & & $\|\mathbf{w}\|_2  \leq 1$ ,  $\|\mathbf{W}\|_{2,2}  \leq 1$, $\|\mathcal{W}^{[\cdot:i:\cdot]}_{1\leq i \leq w}\|_{2,2}  \leq 1$\\
 \midrule
  
\multirow{3}{*}{MLP~\cite{DongGHHLMSSZ14}} & \multirow{3}{*}{$\T{\mathbf{v}} \psi\left(\mathbf{W} \begin{bmatrix}\mathbf{e}_s\\\mathbf{r}_p\\\mathbf{e}_o\end{bmatrix} + \mathbf{w}\right) $} & \multirow{3}{*}{\begin{tabular}{@{}l@{}}
$\mathbf{e}_x \in  \mathbb{R}^{d}$, $\mathbf{r}_y \in  \mathbb{R}^{d}$, $\mathbf{v} \in \mathbb{R}^{w}$, $\mathbf{w} \in \mathbb{R}^{w}$, $\mathbf{W} \in \mathbb{R}^{w , 3d}$\\
$\|\mathbf{e}_x\|_2 \leq 1$ $\|\mathbf{r}_y\|_2 \leq 1$ 
\end{tabular}}\\
 & & \\
 & & \\
\midrule

\multirow{3}{*}{ConvE~\cite{DettmersMS018}} & \multirow{3}{*}{$\psi\left(\T{\mathrm{vec}\left(\psi\left( \mathcal{W} * \begin{bmatrix}\mathbf{e}_s^{[a, b]}\\\mathbf{r}_p^{[a, b]}\end{bmatrix} \right)\right)} \mathbf{W}\right) \mathbf{e}_o  $} & \multirow{3}{*}{%
\begin{tabular}{@{}l@{}}
$\mathbf{e}_x \in  \mathbb{R}^{d}$, $\mathbf{r}_y \in  \mathbb{R}^{d}$, $d = ab$,\\ $\mathbf{W} \in \mathbb{R}^{w_1(w_2 + 2a - 1)(w_3 + b - 1) , d}$, $\mathcal{W} \in \mathbb{R}^{w_1 , w_2 , w_3}$\\
\end{tabular}
} \\
  & & \\
  & & \\ 
\midrule

\multirow{3}{*}{HypER~\cite{BalazevicAH19b}} & \multirow{3}{*}{$\psi\T{\left(\mathrm{vec}\left( \T{\mathbf{r}_p} \mathcal{W} * \mathbf{e}_s \right)} \mathbf{W} \right) \mathbf{e}_o$} & \multirow{3}{*}{%
\begin{tabular}{@{}l@{}}
$\mathbf{e}_x \in  \mathbb{R}^{d_e}$, $\mathbf{r}_y \in  \mathbb{R}^{d_r}$, $\mathbf{W} \in \mathbb{R}^{w_2(w_1 + d_e - 1) , d_e}$, \\ $\mathcal{W} \in \mathbb{R}^{d_r , w_1 , w_2}$\\
\end{tabular}
} \\
& & \\
& & \\
\bottomrule
\end{tabular}
\end{table}

The embeddings listed in Table~\ref{tab:kges} vary in complexity, ranging from simple models such as TransE~\cite{bordes2013translating} and DistMult~\cite{distmult}, to more complex ones, such as SME Bilinear~\cite{GlorotBWB13} and ConvE~\cite{DettmersMS018}. A trade-off underlies these proposals in terms of the number of parameters used, where more parameters increases computational costs, but increases the expressiveness of the model in terms of the model's capability to capture latent features of the graph. To increase expressivity, many of the models in Table~\ref{tab:kges} use additional parameters beyond the embeddings themselves. A possible formal guarantee of such models is \textit{full expressiveness}, which, given any disjoint sets of positive edges $E^+$ and negative edges $E^{-}$, asserts that the model can always correctly partition those edges. On the one hand, for example, DistMult~\cite{distmult} cannot distinguish an edge $\gedge[arrin][0.6cm]{s}{p}{o}$ from its inverse $\gedge[arrin][0.6cm]{o}{p}{s}$, so by adding an inverse of an edge in $E^+$ to $E^{-}$, we can show that it is \textit{not} fully expressive. On the other hand, models such as ComplEx~\cite{TrouillonWRGB16}, SimplE~\cite{Kazemi018}, and TuckER~\cite{BalazevicAH19a} have been proven to be fully expressive given sufficient dimensionality; for example, TuckER~\cite{BalazevicAH19a} with dimensions $d_r = |L|$ and $d_e = |V|$ trivially satisfies full expressivity since its core tensor $\mathcal{W}$ then has sufficient capacity to store the full one-hot encoding of any graph. This formal property is useful to show that the model does not have built-in limitations for numerically representing a graph, though of course in practice the dimensions needed to reach full expressivity are often impractical/undesirable.

Here we have not discussed language models for embedding~\cite{ristoski2016rdf2vec,cochez2017global}, which are based on a distinct set of principles, or entailment-aware models~\cite{WangWG15,GuoWWWG16,DemeesterRR16}, which add additional scoring constraints on top of the types of models listed in Table~\ref{tab:kges}. For further information on such works, we refer to the survey by \citet{Wang2017KGEmbedding} and/or the corresponding papers.
\medskip

We continue by defining in detail the operators and conventions used in Table~\ref{tab:kges}. We start with the conventions used, thereafter defining the pertinent operators.

\begin{itemize}
\item We use indexed parentheses -- such as $(\mathbf{x})_{i}$, $(\mathbf{X})_{ij}$, or $(\mathcal{X})_{{i_1}\ldots{i_n}}$ -- to denote elements of vectors, matrices, and tensors, respectively. If a vector $\mathbf{x} \in \mathbb{R}^a$ is used in a context that requires a matrix, the vector is interpreted as an $(a, 1)$-matrix (i.e., a column vector) and can be turned into a row vector (i.e., a $(1,a)$-matrix) using the transpose operation $\mathbf{x}^T$. We use $\mathbf{x}^\mathrm{D} \in \mathbb{R}^{a,a}$ to denote the diagonal matrix with the values of the vector $\mathbf{x} \in \mathbb{R}^{a}$ on its diagonal. We denote the identity matrix by $\mathbf{I}$ such that if $j=k$, then $(\mathbf{I})_{jk} = 1$; otherwise $(\mathbf{I})_{jk} = 0$.

\item We denote by $\begin{bmatrix}\mathbf{X}_1\\\vdots\\\mathbf{X_n}\end{bmatrix}$ the vertical stacking of matrices $\mathbf{X}_1, \ldots, \mathbf{X}_n$ with the same number of columns. Given a vector $\mathbf{x} \in \mathbb{R}^{ab}$, we denote by $\mathbf{x}^{[a,b]} \in \mathbb{R}^{a,b}$ the ``reshaping'' of $\mathbf{x}$ into an $(a,b)$-matrix such that $(\mathbf{x}^{[a,b]})_{ij} = (\mathbf{x})_{(i + a(j-1))}$. Conversely, given a matrix $\mathbf{X} \in \mathbb{R}^{a,b}$, we denote by $\mathrm{vec}(\mathbf{X}) \in \mathbb{R}^{ab}$ the \textit{vectorisation} of $\mathbf{X}$ such that $\mathrm{vec}(\mathbf{X})_k = (\mathbf{X})_{ij}$ where $i = ((k-1)\,\mathrm{mod}\,m) + 1$ and $j = \frac{k - i}{m} + 1$ (observe that $\mathrm{vec}(\mathbf{x}^{[a,b]}) = \mathbf{x}$).

\item Given a tensor $\mathcal{X} \in \mathbb{R}^{a,b,c}$, we denote by $\mathcal{X}^{[i:\cdot:\cdot]} \in \mathbb{R}^{b,c}$, the $i$\textsuperscript{th} \textit{slice} of tensor $\mathcal{X}$ along the first mode; for example, given $\mathcal{X} \in \mathbb{R}^{5,2,3}$, then $\mathcal{X}^{[4:\cdot:\cdot]}$ returns the $(2,3)$-matrix consisting of the elements $\begin{bmatrix} (\mathcal{X})_{411} & (\mathcal{X})_{412} & (\mathcal{X})_{413} \\ (\mathcal{X})_{421} & (\mathcal{X})_{422} & (\mathcal{X})_{423} \end{bmatrix}$. Analogously, we use $\mathcal{X}^{[\cdot : i : \cdot]}  \in \mathbb{R}^{a,c}$ and $\mathcal{X}^{[\cdot:\cdot:i]}  \in \mathbb{R}^{b,c}$ to indicate the $i$\textsuperscript{th} slice along the second and third modes of $\mathcal{X}$, respectively.

\item We denote by $\psi(\mathcal{X})$ the element-wise application of a function $\psi$ to the tensor $\mathcal{X}$, such that $(\psi(\mathcal{X}))_{in_1\ldots i_n} = \psi(\mathcal{X}_{i_1\ldots i_n})$. Common choices for $\psi$ include a sigmoid function (e.g., the logistic function $\psi(x) = \frac{1}{1 + e^{-x}}$ or the hyperbolic tangent function $\psi(x) = \mathrm{tanh}\,x = \frac{e^x - e^{-x}}{e^x + e^{-x}}$), the rectifier ($\psi(x) = \mathrm{max}(0,x)$), softplus ($\psi(x) = \mathrm{ln}(1 + e^x)$), etc.
\end{itemize}

\medskip
The first and most elemental operation we consider is that of matrix multiplication.

\begin{definition}[Matrix multiplication]
The \emph{multiplication of matrices} $\mathbf{X} \in \mathbb{R}^{a,b}$ and $\mathbf{Y}  \in \mathbb{R}^{b,c}$ is a matrix $\mathbf{XY} \in \mathbb{R}^{a,c}$ such that ($\mathbf{XY})_{ij} = \sum_{k=1}^b (\mathbf{X})_{ik}(\mathbf{Y})_{kj}$. The matrix multiplication of two tensors $\mathcal{X} \in \mathbb{R}^{a_1,\ldots,a_m,c}$ and $\mathcal{Y} \in \mathbb{R}^{c,b_1,\ldots,b_n}$ 
is a tensor $\mathcal{XY} \in \mathbb{R}^{a_1,\ldots,a_{m},b_{1},\ldots,b_{n}}$ such that ($\mathcal{XY})_{i_1\ldots i_m i_{m+1}\ldots i_{m+n}} = \sum_{k=1}^c (\mathcal{X})_{i_1\ldots i_m k}(\mathcal{Y})_{k i_{m+1}i_{m+n}}$.
\end{definition}

For convenience, we may implicitly add or remove modes with dimension 1 for the purposes of matrix multiplication and other operators; for example, given two vectors $\mathbf{x} \in \mathbb{R}^{a}$ and $\mathbf{y} \in \mathbb{R}^{a}$, we denote by $\T{\mathbf{x}}\mathbf{y}$ (aka the dot or inner product) the multiplication of matrix $\T{\mathbf{x}} \in \mathbb{R}^{1,a}$ with $\mathbf{y} \in \mathbb{R}^{a,1}$ such that $\T{\mathbf{x}}\mathbf{y} \in \mathbb{R}^{1,1}$ (i.e., a scalar in $\mathbb{R}$); conversely, $\mathbf{x}\T{\mathbf{y}} \in \mathbb{R}^{a,a}$ (the outer product).
\medskip

Constraints on embeddings are sometimes given in terms of norms, defined next.

\begin{definition}[$L^p$-norm, $L^{p,q}$-norm]
For $p\in \mathbb{R}$, the \emph{$L^p$-norm} of a vector $\mathbf{x}\in \mathbb{R}^a$ is the scalar $\|\mathbf{x}\|_p \coloneq (|(\mathbf{x})_1|^p + \ldots + |(\mathbf{x})_a|^p)^{\frac{1}{p}}$, where $|(\mathbf{x})_i|$ denotes the absolute value of the $i$\textsuperscript{th} element of $\mathbf{x}$. For $p,q\in \mathbb{R}$, the \emph{$L^{p,q}$-norm} of a matrix $\mathbf{X}\in\mathbb{R}^{a,b}$ is the scalar $\|\mathbf{X}\|_{p,q} \coloneq \left( \sum_{j=1}^b \left( \sum_{i=1}^a |(\mathbf{X})_{ij}|^p \right)^{\frac{q}{p}} \right)^\frac{1}{q}$.
\end{definition}

The $L^1$ norm (i.e., $\|\mathbf{x}\|_1$) is thus simply the sum of the absolute values of $\mathbf{x}$, while the $L^2$ norm (i.e., $\|\mathbf{x}\|_2$) is the (Euclidean) length of the vector. The Frobenius norm of the matrix $\mathbf{X}$ then equates to $\|\mathbf{X}\|_{2,2} = \left( \sum_{j=1}^b \left( \sum_{i=1}^a |(\mathbf{X})_{ij}|^2 \right) \right)^\frac{1}{2}$; i.e., the square root of the sum of the squares of all elements.
\medskip

Another type of product used by embedding techniques is the Hadamard product, which multiplies tensors of the same dimension and computes their product element-wise.

\begin{definition}[Hadamard product]
Given two tensors $\mathcal{X} \in \mathbb{R}^{a_1,\ldots,a_n}$ and $\mathcal{Y} \in \mathbb{R}^{a_1,\ldots,a_n}$, the \emph{Hadamard product} $\mathcal{X} \odot \mathcal{Y}$ is defined as a tensor in $\mathbb{R}^{a_1,\ldots,a_n}$, with each element computed as $(\mathcal{X} \odot \mathcal{Y})_{i_1\ldots i_{n}} \coloneq (\mathcal{X})_{i_1\ldots i_{n}} (\mathcal{Y})_{i_1\ldots i_{n}}$.
\end{definition}

Other embedding techniques -- namely RotatE~\cite{SunDNT19} and ComplEx~\cite{TrouillonWRGB16} -- uses \textit{complex space} based on complex numbers. With a slight abuse of notation, the definitions of vectors, matrices and tensors can be modified by replacing the set of real numbers $\mathbb{R}$ by the set of complex numbers $\mathbb{C}$, giving rise to complex vectors, complex matrices, and complex tensors. In this case, we denote by $\mathrm{Re}(\cdot)$ the real part of a complex number. Given a complex vector $\mathbf{x} \in \mathbb{C}^I$, we denote by $\overline{\mathbf{x}}$ its complex conjugate (swapping the sign of the imaginary part of each element). Complex analogues of the aforementioned operators can then be defined by replacing the multiplication and addition of real numbers with the analogous operators for complex numbers, where RotateE~\cite{SunDNT19} uses the complex Hadamard product, and ComplEx~\cite{TrouillonWRGB16} uses complex matrix multiplication.
\medskip

One embedding technique -- MuRP~\cite{BalazevicAH19} -- uses hyperbolic space, specifically based on the Poincaré ball. As this is the only embedding we cover that uses this space, and the formalisms are lengthy (covering the Poincaré ball, Möbius addition, Möbius matrix--vector multiplication, logarithmic maps, exponential maps, etc.), we rather refer the reader to the paper for further details~\cite{BalazevicAH19}.
\medskip

As discussed in Section~\ref{ssec:embeddings}, tensor decompositions are an important concept for many embeddings, and at the heart of such decompositions is the tensor product.

\begin{definition}[Tensor product]
Given two tensors $\mathcal{X} \in \mathbb{R}^{a_1,\ldots,a_m}$ and $\mathcal{Y} \in \mathbb{R}^{b_1,\ldots,b_n}$, the \emph{tensor product} $\mathcal{X} \otimes \mathcal{Y}$ is defined as  a tensor in $\mathbb{R}^{a_1,\ldots,a_m,b_1,\ldots,b_n}$, with each element computed as $(\mathcal{X} \otimes \mathcal{Y})_{i_1\ldots i_{m}j_1\ldots j_n} \coloneq (\mathcal{X})_{i_1 \ldots i_m} (\mathcal{Y})_{j_1 \ldots j_n}$.\footnote{Please note that ``$\otimes$'' is used here in an unrelated sense to its use in Definition~\ref{def:anndom}.}
\end{definition}

To illustrate the tensor product, assume that $\mathcal{X} \in \mathbb{R}^{2,3}$ and $\mathcal{Y} \in \mathbb{R}^{3,4,5}$. The result of $\mathcal{X} \otimes \mathcal{Y}$ will be a tensor in $\mathbb{R}^{2,3,3,4,5}$. Element $(\mathcal{X} \otimes \mathcal{Y})_{12345}$ will be computed by multiplying $(\mathcal{X})_{12}$ and $(\mathcal{Y})_{345}$.
\medskip

An $n$-mode product is used by other embeddings to transform elements along a mode of a tensor.

\begin{definition}[$n$-mode product]
For a positive integer $n$, a tensor $\mathcal{X} \in \mathbb{R}^{a_1,\ldots,a_{n-1},a_n,a_{n+1},\ldots,a_m}$ and matrix $\mathbf{Y} \in \mathbb{R}^{b,a_n}$, the \emph{$n$-mode product} of $\mathcal{X}$ and $\mathbf{Y}$ is the tensor $\mathcal{X} \otimes_n \mathbf{Y} \in \mathbb{R}^{a_1,\ldots,a_{n-1},b,a_{n+1},\ldots,a_m}$ such that $(\mathcal{X} \otimes_n \mathbf{Y})_{i_1\ldots i_{n-1}ji_{n+1}\ldots i_m} \coloneq \sum_{k=1}^{a_n} (\mathcal{X})_{i_1 \ldots i_{n-1}ki_{n+1} \ldots i_m} (\mathbf{Y})_{jk}$.
\end{definition}

To illustrate, let us assume that $\mathcal{X} \in \mathbb{R}^{2,3,4}$ and $\mathbf{Y} \in \mathbb{R}^{5,3}$. The result of $\mathcal{X} \otimes_2 \mathbf{Y}$ will be a tensor in $\mathbb{R}^{2,5,4}$, where, for example, $(\mathcal{X} \otimes_2 \mathbf{Y})_{142}$ will be given as $(\mathcal{X})_{112}(\mathbf{Y})_{41} + (\mathcal{X})_{122}(\mathbf{Y})_{42} + (\mathcal{X})_{132}(\mathbf{Y})_{43}$. Observe that if $\mathbf{y} \in \mathbb{R}^{a_n}$ -- i.e., if $\mathbf{y}$ is a (column) vector -- then the $n$-mode tensor product $\mathcal{X} \otimes_n \T{\mathbf{y}}$ ``flattens'' the $n$\textsuperscript{th} mode of $\mathcal{X}$ to one dimension, effectively reducing the order of $\mathcal{X}$ by one.
\medskip

One embedding technique -- HolE~\cite{NickelRP16} -- uses a circular correlation operator.

\begin{definition}[Circular correlation]
The \emph{circular correlation} of vector $\mathbf{x} \in \mathbb{R}^a$ with $\mathbf{y} \in \mathbb{R}^a$ is the vector $\mathbf{x} \star \mathbf{y} \in \mathbb{R}^{a}$ such that $(\mathbf{x} \star \mathbf{y})_k \coloneq \sum_{i=1}^a (\mathbf{x})_i (\mathbf{y})_{(((k+i-2) \,\mathrm{mod}\,a)+1)}$. 
\end{definition}

Each element of $\mathbf{x} \star \mathbf{y}$ is the sum of $a$ elements along a diagonal of the outer product $\mathbf{x} \otimes \mathbf{y}$ that ``wraps'' if not the primary diagonal. Assuming $a = 5$, then $(\mathbf{x} \star \mathbf{y})_1 = (\mathbf{x})_1(\mathbf{y})_1 + (\mathbf{x})_2(\mathbf{y})_2 + (\mathbf{x})_3(\mathbf{y})_3 + (\mathbf{x})_4(\mathbf{y})_4 + (\mathbf{x})_5(\mathbf{y})_5$, or a case that wraps: $(\mathbf{x} \star \mathbf{y})_4 = (\mathbf{x})_1(\mathbf{y})_4 + (\mathbf{x})_2(\mathbf{y})_5 + (\mathbf{x})_3(\mathbf{y})_1 + (\mathbf{x})_4(\mathbf{y})_2 + (\mathbf{x})_5(\mathbf{y})_3$.
\medskip

Finally, a couple of neural models that we include -- namely ConvE~\cite{DettmersMS018} and HypER~\cite{BalazevicAH19b} -- are based on convolutional architectures using the convolution operator.

\begin{definition}[Convolution]
Given two matrices $\mathbf{X} \in \mathbb{R}^{a,b}$ and $\mathbf{Y} \in \mathbb{R}^{e,f}$, the \emph{convolution} of $\mathbf{X}$ and $\mathbf{Y}$ is the matrix $\mathbf{X} * \mathbf{Y} \in \mathbb{R}^{(a + e - 1),(b + f - 1)}$ such that $(\mathbf{X} * \mathbf{Y})_{ij} = \sum_{k=1}^a \sum_{l=1}^b (\mathbf{X})_{kl} (\mathbf{Y})_{(i+k-a)(j+l-b)}$.\footnote{We define the convolution operator per the convention for convolutional neural networks. Strictly speaking, the operator should be called \textit{cross-correlation}, where traditional convolution requires the matrix $\mathbf{X}$ to be initially ``rotated'' by 180°. Since in our settings the matrix $\mathbf{X}$ is learnt, rather than given, the rotation is redundant.} In cases where $(i+k-a) < 1$, $(j+l-b) < 1$, $(i+k-a) > e$ or $(j+l-b) > f$ (i.e., where $(\mathbf{Y})_{(i+k-a)(j+l-b)}$ lies outside the bounds of $\mathbf{Y}$), we say that $(\mathbf{Y})_{(i+k-a)(j+l-b)} = 0$.
\end{definition}

Intuitively speaking, the convolution operator overlays $\mathbf{X}$ in every possible way over $\mathbf{Y}$ such that at least one pair of elements $(\mathbf{X})_{ij},(\mathbf{Y})_{lk}$ overlaps, summing the products of pairs of overlapping elements to generate an element of the result. Elements of $\mathbf{X}$ extending beyond $\mathbf{Y}$ are ignored (equivalently we can consider $\mathbf{Y}$ to be ``zero-padded'' outside its borders). To illustrate, given $\mathbf{X} \in \mathbb{R}^{3,3}$ and $\mathbf{Y} \in \mathbb{R}^{4,5}$, then $\mathbf{X} * \mathbf{Y} \in \mathbb{R}^{6,7}$, where, for example, $(\mathbf{X} * \mathbf{Y})_{11} = (\mathbf{X})_{33}(\mathbf{Y})_{11}$ (with the bottom right corner of $\mathbf{X}$ overlapping the top left corner of $\mathbf{Y}$), while $(\mathbf{X} * \mathbf{Y})_{34} = (\mathbf{X})_{11}(\mathbf{Y})_{12} + (\mathbf{X})_{12}(\mathbf{Y})_{13} + (\mathbf{X})_{13}(\mathbf{Y})_{14} + (\mathbf{X})_{21}(\mathbf{Y})_{22} + (\mathbf{X})_{22}(\mathbf{Y})_{23} + (\mathbf{X})_{23}(\mathbf{Y})_{24} + (\mathbf{X})_{31}(\mathbf{Y})_{32} + (\mathbf{X})_{32}(\mathbf{Y})_{33} + (\mathbf{X})_{33}(\mathbf{Y})_{34}$ (with $(\mathbf{X})_{22}$ -- the centre of $\mathbf{X}$ -- overlapping $(\mathbf{Y})_{23}$).\footnote{Models applying convolutions may differ regarding how edge cases are handled, or on the ``stride'' of the convolution applied, where, for example, a stride of 3 for $(\mathbf{X} * \mathbf{Y})$ would see the kernel $\mathbf{X}$ centred only on elements $(\mathbf{Y})_{ij}$ such that $i\,\mathrm{mod}\,3 = 0$ and $j\,\mathrm{mod}\,3 = 0$, reducing the number of output elements by a factor of 9. We do not consider such details here.} In a convolution $\mathbf{X} * \mathbf{Y}$, the matrix $\mathbf{X}$ is often called the ``kernel'' (or ``filter''). Often several kernels are used in order to apply multiple convolutions. Given a tensor $\mathcal{X} \in \mathbb{R}^{c,a,b}$ (representing $c$ $(a,b)$-kernels) and a matrix $\mathbf{Y} \in \mathbb{R}^{e,f}$, we denote by $\mathcal{X} * \mathbf{Y} \in \mathbb{R}^{c,(a + e - 1),(b + f - 1)}$ the result of the convolutions of the $c$ first-mode slices of $\mathcal{X}$ over $\mathbf{Y}$ such that $(\mathcal{X} * \mathbf{Y})^{[i:\cdot:\cdot]} = \mathcal{X}^{[i:\cdot:\cdot]} * \mathbf{Y}$ for $1 \leq i \leq c$, yielding a tensor of results for $c$ convolutions.

\subsubsection{Graph neural networks}\label{app:gnns}

We now provide high-level definitions for graph neural networks (GNNs) inspired by (for example) the definitions provided by~\citet{XuHLJ19}. We assume that the GNN accepts a directed vector-labelled graph as input (see Definition~\ref{def:dvlg}).
\medskip

We first abstractly define a recursive graph neural network.

\begin{definition}[Recursive graph neural network]
A \emph{recursive graph neural network} (\emph{RecGNN}) is a pair of functions $\mathfrak{R} \coloneq ( \mathrm{\textsc{Agg}},\mathrm{\textsc{Out}} )$, such that (with $a, b, c \in \mathbb{N}$):

\begin{itemize}
\item $\textsc{Agg}: \mathbb{R}^a \times 2^{(\mathbb{R}^a \times \mathbb{R}^b)  \rightarrow \mathbb{N}} \rightarrow \mathbb{R}^a$
\item $\textsc{Out}: \mathbb{R}^a  \rightarrow \mathbb{R}^c$
\end{itemize}
\end{definition}

The function $\textsc{Agg}$ computes a new feature vector for a node, given its previous feature vector and the feature vectors of the nodes and edges forming its neighbourhood; the function $\textsc{Out}$ transforms the final feature vector computed by $\textsc{Agg}$ for a node to the output vector for that node. We assume that $a$ and $b$ correspond to the dimensions of the input node and edge vectors, respectively, while $c$ denotes the dimension of the output vector for each node. Given a RecGNN  $\mathfrak{R} = ( \mathrm{\textsc{Agg}},\mathrm{\textsc{Out}} )$, a directed vector-labelled graph $G = (V,E,F,\lambda)$, and a node $u \in V$, we define the output vector assigned to node $u$ in $G$ by $\mathfrak{R}$ (written $\mathfrak{R}(G,u)$) as follows. First let $\mathbf{n}_u^{(0)} \coloneq \lambda(u)$. For all $i \geq 1$, let:
\[ \mathbf{n}_u^{(i)} \coloneq \mathrm{\textsc{Agg}} \left( \mathbf{n}_u^{(i-1)}, \{\!\!\{ (\mathbf{n}_v^{(i-1)},\lambda(v,u)) \mid (v,u) \in E  \}\!\!\}  \right) \]
If $j \geq 1$ is an integer such that $\mathbf{n}_u^{(j)} = \mathbf{n}_u^{(j-1)}$ for all $u \in V$, then $\mathfrak{R}(G,u) \coloneq \textsc{Out}(\mathbf{n}_u^{(j)})$.

In a RecGNN, the same aggregation function ($\mathrm{\textsc{Agg}}$) is applied recursively until a fixpoint is reached, at which point an output function ($\mathrm{\textsc{Out}}$) creates the final output vector for each node. While in practice RecGNNs will often consider a static feature vector and a dynamic state vector~\cite{ScarselliGTHM09}, we can more concisely encode this as one vector, where part may remain static throughout the aggregation process representing input features, and part may be dynamically computed representing the state. In practice, $\textsc{Agg}$ and $\textsc{Out}$ are often based on parametric combinations of vectors, with the parameters learnt based on a sample of output vectors for labelled nodes. 

\begin{example} The aggregation function for the GNN of~\citet{ScarselliGTHM09} is given as:
\[\mathrm{\textsc{Agg}}(\mathbf{n}_u,N) \coloneq \sum_{(\mathbf{n}_v,\mathbf{a}_{vu})\in N}f_{\mathbf{w}}(\mathbf{n}_u,\mathbf{n}_v,\mathbf{a}_{vu})\] 
where $f_{\mathbf{w}}(\cdot)$ is a contraction function with parameters $\mathbf{w}$. The output function is defined as:
\[ \mathrm{\textsc{Out}} \left( \mathbf{n}_u \right) \coloneq g_{\textbf{w}'}(\mathbf{n}_u)\]
where again $g_{\mathbf{w}'}(\cdot)$ is a function with parameters $\mathbf{w'}$. Given a set of nodes labelled with their expected output vectors, the parameters $\mathbf{w}$ and $\mathbf{w}'$ are learnt.
\end{example}

There are notable similarities between graph parallel frameworks (GPFs; see Definition~\ref{def:gpf}) and RecGNNs. While we defined GPFs using separate \textsc{Msg} and \textsc{Agg} functions, this is not essential: conceptually they could be defined in a similar way to RecGNN, with a single \textsc{Agg} function that ``pulls'' information from its neighbours (we maintain \textsc{Msg} to more closely reflect how GPFs are defined/implemented in practice). The key difference between GPFs and GNNs is that in the former, the functions are defined by the user, while in the latter, the functions are generally learnt from labelled examples. Another difference arises from the termination condition present in GPFs, though often the GPF's termination condition will -- like in RecGNNs -- reflect convergence to a fixpoint.

\medskip
Next we abstractly define a non-recursive graph neural network.

\begin{definition}[Non-recursive graph neural network]
A \emph{non-recursive graph neural network} (NRecGNN) with $l$ layers is an $l$-tuple of functions $\mathfrak{N} \coloneq ( \mathrm{\textsc{Agg}}^{(1)},\ldots,\mathrm{\textsc{Agg}}^{(l)} )$, such that, for $1 \leq k \leq l$ (with $a_0, \ldots a_l, b \in \mathbb{N}$), $\textsc{Agg}^{(k)}: \mathbb{R}^{a_{k-1}} \times 2^{(\mathbb{R}^{a_{k-1}} \times \mathbb{R}^b)  \rightarrow \mathbb{N}} \rightarrow \mathbb{R}^{a_{k}}$.
\end{definition}

Each function $\textsc{Agg}^{(k)}$ (as before) computes a new feature vector for a node, given its previous feature vector and the feature vectors of the nodes and edges forming its neighbourhood. We assume that $a_0$ and $b$ correspond to the dimensions of the input node and edge vectors, respectively, where each function $\textsc{Agg}^{(k)}$ for $2 \leq k \leq l$ accepts as input node vectors of the same dimension as the output of the function $\textsc{Agg}^{(k-1)}$. Given an NRecGNN  $\mathfrak{N} = ( \mathrm{\textsc{Agg}}^{(1)},\ldots,\mathrm{\textsc{Agg}}^{(l)} )$, a directed vector-labelled graph $G = (V,E,F,\lambda)$, and a node $u \in V$, we define the output vector assigned to node $u$ in $G$ by $\mathfrak{N}$ (written $\mathfrak{N}(G,u)$) as follows. First let $\mathbf{n}_u^{(0)} \coloneq \lambda(u)$. For all $i \geq 1$, let:
\[ \mathbf{n}_u^{(i)} \coloneq \mathrm{\textsc{Agg}}^{(i)} \left( \mathbf{n}_u^{(i-1)}, \{\!\!\{ (\mathbf{n}_v^{(i-1)},\lambda(v,u)) \mid (v,u) \in E  \}\!\!\}  \right) \]
Then $\mathfrak{N}(G,u) \coloneq \mathbf{n}_u^{(l)}$.

In an $l$-layer NRecGNN, a different aggregation function can be applied at each step (i.e., in each layer), up to a fixed number of steps $l$. We do not consider a separate $\mathrm{\textsc{Out}}$ function as it can be combined with the final aggregation function $\mathrm{\textsc{Agg}}^{(l)}$. When the aggregation functions are based on a convolutional operator, we call the result a \textit{convolutional graph neural network} (\textit{ConvGNN}). We refer to the survey by \citet{abs-1901-00596} for discussion of ConvGNNs proposed in the literature.
\medskip

We have considered GNNs that define the neighbourhood of a node based on its incoming edges. However, these definitions can be adapted to also consider outgoing neighbours by either adding inverse edges to the directed vector-labelled graph in pre-processing, or by adding outgoing neighbours as arguments to the $\mathrm{\textsc{Agg}}(\cdot)$ function. More generally, GNNs (and indeed GPFs) relying solely on the neighbourhood of each node have limited expressivity in terms of their ability to distinguish nodes and graphs~\cite{XuHLJ19}; for example, \citet{BarceloKMPRS20} show that such NRecGNNs have a similar expressiveness for classifying nodes as the $\mathcal{ALCQ}$ Description Logic discussed in Section~\ref{sec:dlformal}. More expressive GNN variants have been proposed that allow the aggregation functions to access and update a globally shared vector~\cite{BarceloKMPRS20}. We refer to the papers by \citeT{XuHLJ19} and \citeT{BarceloKMPRS20} for further discussion on the expressivity of GNNs. 

\subsubsection{Symbolic learning}\label{app:sym}

We provide some abstract formal definitions for the tasks of \textit{rule mining} and \textit{axiom mining}  over graphs, which we generically call \textit{hypothesis mining}. First we introduce \textit{hypothesis induction}: a task that captures a more abstract (ideal) case for hypothesis mining. 

\begin{definition}[Hypothesis induction] The task of \emph{hypothesis induction} assumes a particular graph entailment relation $\models_\Phi$ (see Definition~\ref{def:ent}; hereafter simply $\models$). Given \textit{background knowledge} in the form of a knowledge graph $G$ (a directed edge-labelled graph, possibly extended with rules or ontologies), a set of \textit{positive edges} $E^{+}$ such that $G$ does not entail any edge in $E^{+}$ (i.e., for all $e^{+} \in E^{+}$, $G \not\models e^{+}$) and $E^{+}$ does not contradict $G$ (i.e., there is a model of $G \cup E^{+}$), and a set of \textit{negative edges} $E^{-}$ such that $G$ does not entail any edge in $E^-$ (i.e., for all $e^{-} \in E^{-}$, $G \not\models e^{-}$), the task is to find a set of \textit{hypotheses} (i.e., a set of directed edge-labelled graphs) $\Psi$ such that:

\begin{itemize}
\item $G \not\models \psi$ for all $\psi \in \Psi$ (the background knowledge does not entail any hypothesis)
\item $G \cup \Psi^* \models E^{+}$ (the background knowledge and hypotheses entail all positive edges);
\item for all $e^{-} \in E^{-}$,  $G \cup \Psi^* \not\models e^{-}$ (the background knowledge and hypotheses do not entail any negative edge);
\item $G \cup \Psi^* \cup E^{+}$ has a model (the background knowledge, hypotheses and positive edges taken together do not contain a contradiction);
\item for all $e^{+} \in E^{+}$,  $\Psi^* \not\models e^{+}$ (the hypotheses alone do not entail a positive edge).
\end{itemize}

\noindent where by $\Psi^* \coloneq \cup_{\psi \in \Psi} \psi$ we denote the union of all graphs in $\Psi$.
\end{definition}

\begin{example}
Let us assume ontological entailment $\models$ with semantic conditions $\Phi$ as defined in Tables~\ref{tab:ontEqIneq}--\ref{tab:ontClass}. Given the graph of Figure~\ref{fig:airports} as the background knowledge $G$, along with

\begin{itemize}
\item a set of positive edges $E^{+} = \{ \gedge{SCL}{flight}{ARI}, \gedge[arrin][2.2cm]{SCL}{domestic flight}{ARI} \}$, and
\item a set of negative edges $E^{-} = \{ \gedge{ARI}{flight}{LIM}, \gedge[arrin][2.2cm]{SCL}{domestic flight}{LIM} \}$,
\end{itemize}

\noindent
then a set of hypotheses $\Psi = \{ \gedge{flight}{type}{Symmetric}, \gedge{domestic flight}{type}{Symmetric} \}$ would entail all positive edges in $E^{+}$ and no negative edges in $E^{-}$ when combined with $G$.
\end{example}

This task represents a somewhat idealised case. Often there is no set of positive edges distinct from the background knowledge itself. Furthermore, hypotheses not entailing a few positive edges, or entailing a few negative edges, may still be useful. The task of \textit{hypothesis mining} rather accepts as input the background knowledge $G$ and a set of negative edges $E^{-}$ (such that for all $e^{-} \in E^{-}$, $G \not\models e^{-}$), and attempts to \textit{score} individual hypotheses $\psi$ (such that $G \not\models \psi$) in terms of their ability to ``explain'' $G$ while minimising the number of elements of $E^{-}$ entailed by $G$ and $\psi$. 

We can now abstractly define the task of hypothesis mining.

\begin{definition}[Hypothesis mining]
Given a knowledge graph $G$, a set of negative edges $E^{-}$, a scoring function $\sigma$, and a threshold $\textsf{min}_{\sigma}$, the goal of \emph{hypothesis mining} is to identify a set of hypotheses $\{ \psi \mid G \not\models \psi\text{ and }\sigma(\psi,G,E^{-}) \geq \textsf{min}{\sigma} \}$.
\end{definition}

There are two main scoring functions used for $\sigma$ in the literature: \textit{support} and \textit{confidence}.

\begin{definition}[Hypothesis support and confidence]
Given a knowledge graph $G = (V,E,L)$ and a hypothesis $\psi$, the \emph{positive support} of $\psi$ is defined as follows:
\[ \sigma^{+}(\psi,G) \coloneq |\{ e \in E \mid G' \not\models e \text{ and }G' \cup \psi \models e \}| \] 
where $G'$ denotes $G$ with the edge $e$ removed. Further given a set of negative edges $E^{-}$, the \emph{negative support} of $\psi$ is defined as follows:
\[ \sigma^{-}(\psi,G,E^{-}) \coloneq |\{ e^{-} \in E^{-} \mid G \cup \psi \models e^{-} \}| \] 
Finally, the \emph{confidence} of $\psi$ is defined as $\sigma^\pm(\psi,G,E^{-}) \coloneq \frac{\sigma^{+}(\psi,G)}{\sigma^{+}(\psi,G) + \sigma^{-}(\psi,G,E^{-})}$.
\end{definition}

We have yet to define how the set of negative edges are defined, which, in the context of a knowledge graph $G$, depends on which assumption is applied:

\begin{itemize}
\item \textit{Closed world assumption (CWA)}: For any (positive) edge $e$, $G \not\models e$ if and only if $G \models \neg e$. Under CWA, any edge $e$ not entailed by $G$ can be considered a negative edge.
\item \textit{Open world assumption}: For a (positive) edge $e$, $G \not\models e$ does not necessarily imply $G \models \neg e$. Under OWA, the negation of an edge must be entailed by $G$ for it to be considered negative.
\item \textit{Partial completeness assumption (PCA)}: If there exists $(s,p,o)$ such that $G \models (s,p,o)$, then for all $o'$ such that $G \not\models (s,p,o')$, it holds that $G \models \neg(s,p,o')$. Under PCA, if $G$ entails some outgoing edge(s) labelled $p$ from a node $s$, then such edges are assumed to be complete, and any edge $(s,p,o)$ not entailed by $G$ can be considered a negative edge.
\end{itemize}

Knowledge graphs are generally incomplete -- in fact, one of the main applications of hypothesis mining is to try to improve the completeness of the knowledge graph -- and thus it would appear unwise to assume that any edge that is not currently entailed is false/negative. We can thus rule out CWA. Conversely, under OWA, potentially few (or no) negative edges might be entailed by the given ontologies/rules, and thus hypotheses may end up having low negative support despite entailing many edges that do not make sense in practice. Hence the PCA can be adopted as a heuristic to increase the number of negative edges and apply more sensible scoring of hypotheses.

Different implementations of hypothesis mining may consider different logical languages. Rule mining, for example, mines hypotheses expressed either as monotonic rules (with positive edges) or non-monotonic edges (possibly with negated edges). On the other hand, axiom mining considers hypotheses expressed in a logical language such as Description Logics. Particular implementations may, for practical reasons, impose further syntactic restrictions on the hypotheses generated, such as to impose thresholds on their length, on the symbols they use, or on other structural properties (such as ``closed rules'' in the case of the AMIE rule mining system~\cite{GalarragaTHS13}; see Section~\ref{ssec:symlearn}). Systems may further implement different search strategies for hypotheses. Systems such as AMIE~\cite{GalarragaTHS13}, RuLES~\cite{HoSGKW18}, CARL~\cite{TanonSRMW17}, DL-Learner~\cite{BuhmannLW16}, etc., propose \textit{discrete mining} that recursively generates candidate formulae through refinement/genetic operators that are then scored and checked for threshold criteria, thus navigating a branching search space. On the other hand, systems such as NeuralLP~\cite{YangYC17} and DRUM~\cite{SadeghianADW19} apply \textit{differentiable mining} that allows for learning (path-like) rules and their scores in a more continuous fashion (e.g., using gradient descent). We refer to Section~\ref{ssec:symlearn} for further discussion and examples of such techniques for mining hypotheses.

\end{document}